\documentclass[11pt]{article} 
\usepackage{tikz}
\usepackage{pgfplots}

\usepackage{anyfontsize}
\usepackage{eqnarray}
\usepackage{epsf}
\usepackage{multirow}

\usepackage{hhline}
\usepackage{caption}
\usepackage{graphicx}
\usepackage{subfigure}
\usepackage{bm} %
\usepackage{epsfig}
\usepackage{amsmath}
\usepackage{amsthm}  %
\usepackage{amsfonts}
\usepackage{mathrsfs}
\usepackage{bbm}
\usepackage{extarrows}  %
\usepackage{amssymb} %
\usepackage{verbatim}%
\usepackage{longtable} %
\usepackage{framed} %
\usepackage{color}
\usepackage{hyperref} %
\usepackage{float}

\newcommand{\e}{\varepsilon}

\newcommand{\mbR}{\mathbb{R}}
\newcommand{\idf}{\mathbbm{1}}  %
\newcommand{\mc}{\mathcal}

\newcommand{\ea}{ \end{aligned}}
\newcommand{\ba}{ \begin{aligned}}

\newcommand{\eeq}{\end{aligned}\end{equation}}
\newcommand{\beq}{\begin{equation}\begin{aligned}}
\newcommand{\mb}{\mathbb}
\newcommand{\mbe}{\mb{E}}
\newcommand{\mr}{\mathrm}

\newcommand{\me}{\mr{e}}

\newcommand{\abs}[1]{\left|#1\right|}
\newcommand{\Bigabs}[1]{\Big|#1\Big|}
\newcommand{\bigabs}[1]{\big|#1\big|}
\newcommand{\ceil}[1]{\left \lceil#1\right\rceil} %
\newcommand{\flr}[1]{\left \lfloor#1\right\rfloor} %
\newcommand{\norm}[1]{\left\|#1\right\|}
\newcommand{\ykh}[1]{\left(#1\right)}
\newcommand{\zkh}[1]{\left[#1\right]}
\newcommand{\hkh}[1]{\left\{#1\right\}}
\newcommand{\bigykh}[1]{\big(#1\big)}
\newcommand{\Bigykh}[1]{\Big(#1\Big)}

\newcommand{\qd}{\vee}  %
\newcommand{\qx}{\wedge} %

\newcommand{\setl}[2]{\hkh{\left.#1\right|#2}}
\newcommand{\setr}[2]{\hkh{#1\left|#2\right.}}

\newcommand{\set}[2]{\left\{\left.\vphantom{#2}#1\right.\right|\left.#2\vphantom{#1}\right\}}

\newcommand{\setm}[3]{\left\{\left.#2\vphantom{#3}\vphantom{\raisebox{#1ex}{}}\right.\right|\left.#3\vphantom{#2}\vphantom{\raisebox{#1ex}{}}\right\}}

\newcommand{\setma}[3]{\left\{\vphantom{\raisebox{#1ex}{}}\right.#2\left.\vphantom{\raisebox{#1ex}{}}\right|\left.\vphantom{\raisebox{#1ex}{}}\right.#3\left.\vphantom{\raisebox{#1ex}{}}\right\}}

\definecolor{darkgreen}{rgb}{0,0.5,0.1}

\newcommand{\betikz}{\begin{tikzpicture}}
\newcommand{\eetikz}{\end{tikzpicture}}

\def\[#1\]{\begin{align*}#1\end{align*}}

\newcommand{\tbe}{\bm{E}}

\newcommand{\rp}{\mc{R}_P}

\newcommand{\ep}{\mc{E}_P}

\newcommand{\dom}{\mathbf{dom}}
\newcommand{\ran}{\mathbf{ran}}
\newcommand{\iffun}{{\mathtt{i}\kern-0.295em \mathtt{f}}}

\usepackage{relsize}

\usepackage{bbding}
\usepackage{pifont}

\usepackage{xspace}

\usepackage{enumitem}

\newtheorem{theorem}{Theorem}
\newtheorem{lemma}{Lemma}

\usepackage{geometry}

\allowdisplaybreaks[4]

\begin{document}
\title{Optimal Convergence Rates of Deep  Neural Network Classifiers$^\dag$\footnotetext{\dag~The work of Lei Shi is supported by National Natural Science Foundation of China [Project No. 12171039]. The work of Ding-Xuan Zhou is partially supported by the Australian Research Council under project DP240101919 and partially supported by InnoHK initiative, the Government of the HKSAR, China, and the Laboratory for AI-Powered Financial Technologies, Australia. Email addresses: zihan.zhang@sydney.edu.au (Z. Zhang), leishi@fudan.edu.cn (L. Shi), dingxuan.zhou@sydney.edu.au (D-X. Zhou). The corresponding author is Lei Shi.}}

\author{Zihan Zhang$^1$, Lei Shi$^2$ and Ding-Xuan Zhou$^1$\\
	\small $^1$  School of Mathematics and Statistics, 
		The University of Sydney, Sydney NSW 2006, Australia\\
	\small $^2$ School of Mathematical Sciences and Shanghai Key Laboratory for Contemporary\\
	\small Applied Mathematics, Fudan University, Shanghai 200433, China}
\date{}

\maketitle

\begin{abstract}
In this paper, we study the binary classification problem on $[0,1]^d$ under the Tsybakov noise condition (with exponent $s \in [0,\infty]$) and the compositional assumption. This assumption requires the conditional class probability function of the data distribution to be the composition of $q+1$ vector-valued multivariate functions, where each component function is either a maximum value function or a H\"{o}lder-$\beta$ smooth function that depends only on $d_*$ of its input variables. Notably, $d_*$ can be significantly smaller than the input dimension $d$. We prove that, under these conditions, the optimal convergence rate for the excess 0-1 risk of classifiers is  
$ \left( \frac{1}{n} \right)^{\frac{\beta\cdot(1\wedge\beta)^q}{{\frac{d_*}{s+1}+(1+\frac{1}{s+1})\cdot\beta\cdot(1\wedge\beta)^q}}} $,   
which is independent of the input dimension $d$. Additionally, we demonstrate that ReLU deep neural networks (DNNs) trained with hinge loss can achieve this optimal convergence rate up to a logarithmic factor. This result provides theoretical justification for the excellent performance of ReLU DNNs in practical classification tasks, particularly in high-dimensional settings. %
The generalized approach is of independent interest.  
\end{abstract}


\maketitle

\newcommand{\cyi}{c_6}   %
\newcommand{\cling}{c_0}
\newcommand{\cer}{c_7}
\newcommand{\csi}{c_{8}}
\newcommand{\cwu}{c_1}  %
\newcommand{\csan}{c_9}
\newcommand{\cliu}{c_2}
\newcommand{\cqi}{c_{11}}
\newcommand{\cba}{c_4}
\newcommand{\cjiu}{c_{5}}
\newcommand{\cshi}{c_{10}}
\newcommand{\cshiyi}{c_{13}}
\newcommand{\cshier}{c_{12}}
\newcommand{\cshisan}{c_{3}}

\newcommand{\zaverage}{expected\xspace}

\newcommand{\fdnn}{\mc{F}^{\mathbf{NN}}}

\newcommand{\FNN}{\mathbf{NN}}

\section{Introduction}\label{23111501}

In this paper, we study the binary classification problem on a finite-dimensional input space $[0,1]^d$ using deep neural networks (DNNs) with the rectified linear unit (ReLU) activation function. For simplicity, throughout this paper, the term ``classification" refers to ``binary classification" unless otherwise specified.

We first describe the setting of classification. Throughout this paper, we consider $[0,1]^d$ as the input space and $\hkh{+1, -1}$ as the output space, where $+1$ and $-1$ denote the two classes. Note that choosing $[0,1]^d$ to be the input space does not entail any loss of     generality, since any distribution supported on a bounded subset of $\mbR^d$ can be transformed to a distribution on $[0,1]^d$ via scaling and translation. Let $P$ be a probability measure on $[0,1]^d \times \{1,-1\}$, regarded as the joint distribution of the input and output data. For any function $f: [0,1]^d \to \mbR$, the map $\mr{sgn} \circ f$ takes inputs from $[0,1]^d$ to outputs in $\hkh{1, -1}$, where $\mr{sgn}(\cdot)$ is the sign function defined as $\mr{sgn}(t)=1$ if $t\geq 0$ and $\mr{sgn}(t)=-1$ otherwise. Thus, a natural way to evaluate the classification performance of a deterministic function $f$ is by measuring the probability that the induced map $\mr{sgn} \circ f$ makes an incorrect prediction. This leads to the definition of the 0-1 risk (or misclassification error) of $f$:
\[
 \mc{R}_P(f):=P\ykh{\setl{(x,y)\in[0,1]^d\times \{1,-1\}}{\mr{sgn}(f(x))\neq y}}.
 \] The classification performance of $f$ can also be characterized by its excess 0-1 risk (or excess misclassification error), defined as
  $\mc{E}_P(f):=\rp(f)-\inf\limits_{g\in\mc F_d}\rp(g)$,  where $\mc F_d$ is the set of all Borel  measurable functions from $[0,1]^d$ to $\mb R$. The advantage of using $\mc{E}_P(\cdot)$ is that its minimum value is always zero, whereas the minimum of $\mc{R}_P(\cdot)$ depends on the distribution $P$. Therefore, $\mc{E}_P(\cdot)$ provides a unified criterion for assessing the classification capability of $f$ as the data distribution $P$ varies within the statistical model $\mc{H}$, that is, the family of   distributions $P$ that satisfy the assumptions we impose.
 
 The goal of classification problem is to construct an estimator $\hat{f}_n$ taking values in $\mc F_d$, based on an i.i.d. sample $\hkh{(X_i,Y_i)}_{i=1}^n$ drawn from the distribution $P$, such that the expected excess 0-1 risk $\tbe_{P^{\otimes n}}\zkh{\ep(\hat f_n)}$ remains uniformly small whenever $P$ satisfies our assumptions. Specifically, we aim to make the quantity $\sup\limits_{P\in\mc{H}}\tbe_{P^{\otimes n}}\zkh{\ep(\hat f_n)}$ as small as possible, where $\tbe_{P^{\otimes n}}$ represents the expectation with respect to the sample  $\hkh{(X_i,Y_i)}_{i=1}^n$. Here, the subscript $P^{\otimes n}$ indicates the joint distribution of  $\hkh{(X_i,Y_i)}_{i=1}^n$, i.e., $P^{\otimes n}:=\underbrace{P\times \cdots\times  P}_{n}$. To demonstrate that a sequence of estimators $\hkh{\hat{f}_n}_{n=1}^\infty$  produced by a specific approach performs well asymptotically as the sample size $n \to \infty$, we need to compare the convergence rate of  $\sup\limits_{P\in\mc{H}}\tbe_{P^{\otimes n}}\zkh{\ep(\hat f_n)}$ with the minimax rate $\inf\limits_{\hat{g}_n} \sup\limits_{P\in\mc{H}}\tbe_{P^{\otimes n}}\zkh{\ep(\hat g_n)}$,
  where the infimum is taken over all estimators $\hat{g}_n: [0,1]^d \to \mbR$. If $ \sup\limits_{P\in\mc{H}}\tbe_{P^{\otimes n}}\zkh{\ep(\hat f_n)}$ converges to zero at the same rate as $\inf\limits_{\hat{g}_n} \sup\limits_{P\in\mc{H}}\tbe_{P^{\otimes n}}\zkh{\ep(\hat g_n)}$, that is,  
  $\sup\limits_{P\in\mc{H}}\tbe_{P^{\otimes n}}\zkh{\ep(\hat f_n)}
   \asymp
   \inf\limits_{\hat{g}_n} \sup\limits_{P\in\mc{H}}\tbe_{P^{\otimes n}}\zkh{\ep(\hat g_n)}$, 
    then we say that  the estimators $\hkh{\hat{f}_n}_{n=1}^\infty$ achieve an optimal convergence rate under the assumption $P \in \mc{H}$. Furthermore, if 
   $\sup\limits_{P\in\mc{H}}\tbe_{P^{\otimes n}}\zkh{\ep(\hat f_n)}\lesssim \ykh{\log n}^{\theta}\cdot \inf\limits_{\hat{g}_n} \sup\limits_{P\in\mc{H}}\tbe_{P^{\otimes n}}\zkh{\ep(\hat g_n)}$
    for some $\theta\in[0,\infty)$, we say that the estimators $\hkh{\hat{f}_n}_{n=1}^\infty$ achieve an optimal convergence rate up to a logarithmic factor.  One commonly employed approach for producing  $\hat{f}_n\in\mc{F}$ based on the sample $\hkh{(X_i,Y_i)}_{i=1}^n$ is empirical risk minimization (ERM) with respect to a given loss function $\phi:\mbR\to[0,\infty)$, defined by 
    \beq\label{20241005003201}
    \hat{f}_n\in\mathop{\arg\min}_{f\in\mc{F}}\frac{1}{n}\sum_{i=1}^n\phi(Y_i\cdot f(X_i)). 
    \eeq An estimator $\hat{f}_n$ satisfying \eqref{20241005003201} is referred to as an empirical $\phi$-risk minimizer (or $\phi$-ERM) over $\mc{F}$ with respect to the sample $\hkh{(X_i,Y_i)}_{i=1}^n$.

In this paper, we derive explicit minimax convergence rates
$
\inf\limits_{\hat{g}_n} \sup\limits_{P \in \mc{H}} \tbe_{P^{\otimes n}} \zkh{\ep(\hat g_n)}
$
for a class of statistical models $\mc{H}$, which consist of distributions satisfying practical and reasonable assumptions. Furthermore, we show that under these assumptions, ReLU DNN estimators can achieve optimal convergence rates up to a logarithmic factor. These results provide a theoretical justification for the well-known empirical success of ReLU DNNs in classification tasks. The specific formulation of these assumptions can be found in Section \ref{240119033824}. Here, we give a brief overview. To this end, we introduce some notations. Let $P_X$ denote the marginal distribution of $P$ on $[0,1]^d$. For $x \in [0,1]^d$, we use $P(\cdot \,|\, x)$ to denote the regular conditional distribution of $P$ on $\hkh{-1, 1}$ given $x$, and define the conditional class probability (CCP) function as $\eta_P(x) := P(\hkh{1} \,|\, x)$. Given $\beta \in (0, \infty)$ and integers $q \geq 0$, $0 < d_* \leq d$, we assume throughout this paper that the CCP function $\eta_P$ of the data distribution $P$ is the composition of $q + 1$ vector-valued multivariate functions, each having a relatively simple structure: their component functions are either H\"older-$\beta$ smooth functions defined essentially on a lower-dimensional space $[0,1]^{d_*}$, or functions that compute the maximum of some input components. The compositional assumption is reasonable and significant, motivated by the following observation: ``in the natural sciences-physics, chemistry, biology-many phenomena seem to be described well by processes that take place at a sequence of increasing scales and are local at each scale, in the sense that they can be described well by neighbor-to-neighbor interactions'' (Appendix 2 of \cite{poggio2016and}). In addition, we  also assume the well-known Tsybakov noise condition 
 \beq\label{23080401}
 	P_X\ykh{\setl{x\in[0,1]^d }{ \big|2\cdot P(\hkh{1}|x)-1\big|\leq t}} \leq \alpha\cdot t^{s}, \quad \forall\;t\in(0,\infty)\cap (0, \tau], 
 \eeq where  $\alpha\in(0,\infty)$, $\tau\in(0,\infty]$,  and $s\in[0,\infty]$.  This condition, first introduced in \cite{tsybakov2004optimal}, plays a key role in generalization analysis of binary classification, as it enables the derivation of fast convergence rates (cf. \cite{kim2021fast, steinwart2007fast}). The noise condition \eqref{23080401} describes the amount of the input data $x$  that makes the conditional probability $P(\hkh{1}|x)$ close to $1/2$. From a practical standpoint, even the strongest noise condition (i.e., \eqref{23080401} with $s=\infty$) is reasonable. In many real-world classification problems—particularly in image classification—the CCPs of most inputs are either very close to $0$ or very close to $1$, with only a small fraction near $1/2$ (see Section 4 of \cite{kim2021fast} and the discussion following Assumption 2 in Section 3 of \cite{KOHLER2025106188}). It is worth noting that when $\tau \geq 1$ and $s = \infty$, the noise condition \eqref{23080401} reduces to the case where the CCP function $P(\hkh{1}|\cdot)$ is almost surely equal to $0$ or $1$—the so-called noise-free scenario.
 
Many existing works have adopted a similar compositional assumption on the CCP function of the data distribution $P$, with or without the Tsybakov noise condition \eqref{23080401}, to establish convergence rates for ReLU DNN estimators in classification tasks. For example, in \cite{KOHLER2025106188}, the authors establish convergence rates for the expected excess 0-1 risk of ReLU deep convolutional neural network (CNN) estimators (with a specific structure), trained using logistic loss for classification. Their analysis assumes that the CCP function $\eta_P$ satisfies the so-called hierarchical max-pooling model, which requires $\eta_P$ to be the maximum of several functions $f_i$. Each $f_i$ is defined as a composition of some H\"older smooth functions of four input variables. In \cite{kohler2022rate}, the authors study a model that generalizes the one considered in \cite{KOHLER2025106188}. In their framework,  $\eta_P$ is assumed to take the form $g(f_1,f_2,\ldots,f_{d_*})$, where each $f_i$ is a composition of some H\"older-$p_1$ smooth functions of four input variables, and $g$ is a H\"older-$p_2$ smooth function. Under this model, they establish convergence rates for the \zaverage excess 0-1 risk of ReLU deep CNN estimators trained with the least square loss. In our recent work \cite{zhangzihan2023classification}, under the same compositional assumption described in this paper, but without imposing the noise condition \eqref{23080401}, we derive the optimal (up to a logarithmic factor) convergence rate ${{\ykh{\frac{(\log n)^5}{n}}^{{\frac{\beta\cdot(1\qx\beta)^q}{{d_*+\beta\cdot(1\qx\beta)^q}}}}}}$ for the expected excess logistic risk (see \eqref{eq3}) of ReLU DNN estimators. We also establish a (sub-optimal) convergence rate ${{{\ykh{\frac{(\log n)^5}{n}}^{{\frac{\beta\cdot(1\qx\beta)^q}{{2d_*+2\beta\cdot(1\qx\beta)^q}}}}}}}$ for their \zaverage excess 0-1 risk. Furthermore, we show that the noise condition \eqref{23080401} generally cannot improve the convergence rate of the \zaverage excess logistic risk (see the discussion following Theorem 2.6 on page 35 of \cite{zhangzihan2023classification}, as well as the discussion in  Appendix  \ref{20250919034415}{} of this paper for more details). In \cite{bos2022convergence}, the authors study multiclass classification using ReLU DNNs, where all the conditional class probability (CCP) functions $\mb{P}(Y_1=\text{the $k$-th class}|X_1=x)$ are assumed to be H\"older-$\beta$ smooth and satisfy the so-called small value bound condition (see Definition 3.1 in \cite{bos2022convergence}). This condition introduces a noise assumption tailored to the multiclass setting; however, it is not equivalent to the Tsybakov noise condition \eqref{23080401} in the binary classification case. Therefore, in the following discussion, we focus on results that do not involve noise conditions. It follows from Theorem 3.3 of \cite{bos2022convergence}, with $L \asymp \log n \asymp B$, that the expected Kullback-Leibler divergence from the learned CCP to the true CCP (see the last equation on page 2727 of \cite{bos2022convergence} for the detailed definitions of $L$, $B$, and the \zaverage Kullback-Leibler divergence) achieves the convergence rate  ${K^{\frac{\beta+3d}{\beta+d}}{n}^{-\frac{\beta}{d+\beta}}\cdot {\log^4n}}$, where $K\geq 2$ denotes the number of classes. This rate matches (up to a logarithmic factor) the convergence rate of the \zaverage excess logistic risk established in our previous work \cite{zhangzihan2023classification} under the setting without the compositional assumption, that is, by setting $d_* = d$ and $q = 1$ in the rate ${{\ykh{\frac{(\log n)^5}{n}}^{{\frac{\beta\cdot(1\qx\beta)^q}{{d_*+\beta\cdot(1\qx\beta)^q}}}}}}$ derived by Theorem 2.3 of \cite{zhangzihan2023classification}. Since the \zaverage Kullback-Leibler divergence is indeed equivalent to the \zaverage excess logistic risk in the binary classification case, we conclude that, without considering the compositional structure or noise conditions, Theorem 3.3 of \cite{bos2022convergence} and Theorem 2.3 of \cite{zhangzihan2023classification} actually establish the same convergence rate for binary classification. The work \cite{bos2022convergence} also discusses the compositional structure of the CCP functions. The compositional assumption they introduce can be regarded as a special case of the compositional function space considered in this paper and in our previous work \cite{zhangzihan2023classification} (see Section \ref{240119033824} for details). In Section 3 of \cite{bos2022convergence}, the authors introduce the compositional assumption and claim that it leads to an improved convergence rate, although they do not provide a proof. It is worth noting that the convergence rate they claim under the compositional structure is consistent with the results in our work \cite{zhangzihan2023classification}, while our results further achieve sharper estimates for the power index with respect to $\log n$. However, our derivation in \cite{zhangzihan2023classification} demonstrates that extending the convergence rates from the setting without the compositional assumption to the compositional case is highly non-trivial.  Moreover, \cite{bos2022convergence} does not provide any lower bound to demonstrate that the established convergence rate is optimal. Besides, \cite{bos2022convergence} does not derive convergence rates for the (excess) misclassification error, which is exactly the aforementioned (excess) 0-1 risk in the binary classification setting and is also a fundamental metric for evaluating the performance of classifiers. In other learning tasks such as regression  (see,  e.g.,  \cite{schmidt2020nonparametric,kohler2021rate}), the composition assumption has also been considered. Thanks to the inherently lower-dimensional structure of the CCP function, the convergence rates established in the aforementioned works are independent of the input dimension $d$. These results help explain why ReLU DNNs can overcome the curse of dimensionality. However, none of the results in \cite{kohler2022rate,KOHLER2025106188,zhangzihan2023classification} provides corresponding optimal convergence rates of the excess 0-1 risk (i.e., the excess misclassification error) of estimators in classification. As a result,  we cannot deduce from these works how fast ReLU DNN estimators converge relative to the optimal rates. In this work, we fill the gap by deriving explicit convergence rates under the aforementioned compositional  assumption and noise condition (see Theorem \ref{20231204212113}). Furthermore, we justify the optimality of these rates by establishing corresponding minimax lower bounds (see Theorem \ref{23102601} and Theorem \ref{23102602}). Our lower bound results serve as a benchmark for assessing the sharpness of the convergence rates established in previous works, such as \cite{kohler2022rate,KOHLER2025106188,zhangzihan2023classification}, and play an important role in characterizing the asymptotic optimality of DNN classifiers. Moreover, we show that ReLU DNNs trained with the hinge loss $\phi_{\mathbf{h}}(t)=\max\hkh{0,1-t}$ can achieve the optimal convergence rates up to logarithmic factors (see Theorem \ref{20231125000604} and Theorem \ref{231215231016}), thereby demonstrating that DNN classifiers based on the hinge loss are theoretically optimal. In particular, we find that this theoretical optimality is not easily achieved by other loss functions. For example, according to \cite{zhangzihan2023classification}, ReLU DNNs trained with logistic loss may be suboptimal (cf. the discussion on page \pageref{241214215551} following Theorem \ref{231215231016} for more details). Our results and the related discussions indicate that, as a classical classification loss function, the hinge loss still offers significant advantages in training DNN classifiers.

To obtain the optimal convergence of ReLU DNN classifiers, we generalize the novel oracle inequality proposed in our recent work \cite{zhangzihan2023classification} (see Theorem 2.1 of \cite{zhangzihan2023classification}, and Theorem \ref{231123013855} in this paper). This inequality establishes the framework of generalization analysis for estimators trained by ERM with respect to any Lipschitz continuous loss function $\phi$. Specifically, we replace the original condition (see (2.5) of \cite{zhangzihan2023classification}) with a more general one (see \eqref{231123022412}). Our result preserves the key advantage of the original inequality: it enables the derivation of sharp generalization error bounds without relying on the explicit form of the target function associated with the $\phi$-risk. As such, this result is of independent interest and can be useful for establishing generalization bounds for estimators trained with other loss functions $\phi$—particularly in cases where the corresponding target function is less favorable. For instance, this occurs when the target function is unbounded, which can happen if the loss function $\phi$ is strictly positive. For more details, we refer the reader to Theorem \ref{231123013855} and the accompanying discussion.

In conclusion, our main contributions in this paper are summarized as follows.

\begin{itemize}
\item In Theorem \ref{231123013855}, we establish a novel oracle inequality that generalizes Theorem 2.1 of \cite{zhangzihan2023classification} and provides framework of generalization analysis for estimators trained by ERM with respect to an arbitrary Lipschitz loss function $\phi$. This inequality exhibits a broad applicability and remains effective even in cases where the target function—when training with $\phi$-risk—is not well-behaved, such as when it is unbounded. This is achieved by employing a surrogate function $\psi$ instead of relying directly on the specific form of the target function. Although in this paper we focus on applying this result to the hinge loss $\phi_{\mathbf{h}}$, this novel inequality is still of independent interest.
    
\item We consider the compositional assumption, which requires the CCP function $\eta_P(x)$ to be expressed as the composition $h_q\circ h_{q-1}\circ\cdots h_0$ of several functions $h_i:[0,1]^{d_i^{\mathbf{in}}}\to\mbR^{d_i^{\mathbf{out}}}$. Each component function of $h_i$ is either a H\"older-$\beta$ smooth function with respect to at most $d_*\leq {d_i^{\mathbf{in}}}$ of its input variables, or a function computes the maximum of some of its input variables. Under this compositional assumption, together with the Tsybakov noise condition \eqref{23080401}, we show that the \zaverage excess 0-1 risk of ReLU DNN estimators trained with hinge loss $\phi_{\mathbf{h}}$ can achieve the convergence rates 
   \begin{equation}\label{eq1}
     \ykh{\frac{(\log n)^3}{n}}^{{\frac{\beta\cdot(1\qx\beta)^q}{{\frac{d_*}{s+1}+(1+\frac{1}{s+1})\cdot\beta\cdot(1\qx\beta)^q}}}},
   \end{equation} which is optimal up to a logarithmic factor. See   Theorem \ref{20231125000604} and Theorem  \ref{231215231016}. Notably, the rate is independent of the input dimension $d$, meaning that this result avoids the curse of dimensionality. Therefore, it offers a theoretical explanation for the efficiency of ReLU DNNs in practical high-dimensional classification problems.

\item We justify the optimality of the  convergence rate \eqref{eq1} mentioned above by showing that the minimax rate $\asymp{\ykh{\frac{1}{n}}^{{\frac{\beta\cdot(1\qx\beta)^q}{{\frac{d_*}{s+1}+(1+\frac{1}{s+1})\cdot\beta\cdot(1\qx\beta)^q}}}}}$. Specifically, we establish minimax lower bounds in Theorem \ref{23102601} and  Theorem \ref{23102602}, which imply that the minimax rate $\gtrsim \ykh{\frac{1}{n}}^{{\frac{\beta\cdot(1\qx\beta)^q}{{\frac{d_*}{s+1}+(1+\frac{1}{s+1})\cdot\beta\cdot(1\qx\beta)^q}}}}$. Besides, in Theorem \ref{20231204212113}, we construct a specific series  of estimators  which can achieve the convergence rate $\ykh{\frac{1}{n}}^{{\frac{\beta\cdot(1\qx\beta)^q}{{\frac{d_*}{s+1}+(1+\frac{1}{s+1})\cdot\beta\cdot(1\qx\beta)^q}}}},$ indicating that the minimax rate  $\lesssim \ykh{\frac{1}{n}}^{{\frac{\beta\cdot(1\qx\beta)^q}{{\frac{d_*}{s+1}+(1+\frac{1}{s+1})\cdot\beta\cdot(1\qx\beta)^q}}}}$. 

\end{itemize}

The rest of this paper is organized as follows. In the remainder of this section, we introduce conventions and notations used through this paper.  The next three sections present our main results along with related discussions. Specifically, we introduce the novel oracle inequality in Section \ref{241217205730}, and provide the results on the generalization upper bounds and minimax lower rates in Section \ref{240119033824}  and Section \ref{2412172103}, respectively. Finally,  Section \ref{2412172107} summarizes the entire paper outlines potential directions for future research. Besides the main body of this paper, three appendices are included.  Appendix  \ref{250914042717} provides rigorous and detailed versions of some definitions that are only briefly described in the main body of this paper, 
Appendix   \ref{2412172110} contains all  the detailed proofs of the  main results of this paper,  
and Appendix 
\ref{20250928194706}
provides some additional  discussions.

\subsection{Conventions and Notations}

Throughout this paper, we adopt the following conventions: $0^0:=1$, $1^\infty:=1$, $\frac{z}{0}:=\infty=:\infty^{c}$, $\log(\infty):=\infty$,  $\log 0:=-\infty$, $0\cdot w:=0=:w\cdot 0$ and $\frac{a}{\infty}:=0=:b^{\infty}$, for any $a\in\mbR, b\in[0,1), c\in(0,\infty)$, $z\in[0,\infty]$, and $w\in[-\infty,\infty]$. Here, $\log$ denotes the natural logarithm function (i.e.,  the logarithm with base $\me$). We use $\mb N$ to denote the set of all positive integers, $\mb Z$ to denote the set of all integers, and $\mbR$ to denote the set of all real numbers. Additionally, we define $\mb N_0:=\mb{N}\cup\hkh{0}$.  
To express functions that are defined piecewisely in a more compact way, we use the following notation: For any  statement $\mr{P}$ and any mathematical expressions $a$ and $b$, define 
\[\iffun(\mr{P};a;b):=\begin{cases}
a,&\;\;\text{if the statement $\mr{P}$ is true};\\
b,&\;\;\text{if the statement $\mr{P}$ is false}. 
\end{cases} \]
 Moreover, we use $\idf$ to represent the indicator function of a set $A$ or a  statement $\mr{P}$. Specifically, 
 $\idf_A(x):=\iffun(x\in A;1;0)$ and  $\idf_{\hkh{\mr{P}}}:=\iffun(\mr{P};1;0)$. For any function $f$, let $\dom(f)$ denote its domain and $\ran(f)$ denote its range.  
 If $A \subset \dom(f)$ and $\setm{2}{f(x)}{x \in A} \subset [-\infty, \infty]$, then  we denote  ${\norm{f}_A := \sup \setm{2}{\abs{f(x)}}{x \in A}}$. For each finite-dimensional vector $x$, we use  $\mathbf{dim}(x)$ to denote its dimension, and use $(x)_i$ to denote its $i$-th component.  Moreover, we use  $(x)_I$ to denote the subvector $((x)_{i_1}, (x)_{i_2},\ldots,(x)_{i_k})$, where $1\leq i_1<i_2<\cdots < i_k\leq\mathbf{dim}(x)$ and $I=\hkh{i_1,\ldots,i_k}$. Then, for any $p\in [1,\infty]$, we use $\norm{x}_p$ to denote the $\ell^p$-norm of the vector $x$, and $\norm{x}_0$ to denote the number of nonzero entries in $x$, that is, $\norm{x}_0:=\#\ykh{\set{i\in\mb{N}}{1\leq i\leq\mathbf{dim}(x), (x)_i\neq 0}}$, where $\#(\cdot)$ gives the number of elements of a set. For any  real matrix
 ${\bm A}$ and $t\in[1,\infty]\cup\hkh{0}$, we define $\norm{\bm A}_t:=\norm{a}_t$, where $a$  is the  vector formed by stacking  the entries of $\bm A$. 

For $d\in\mb N$ and $n\in\mb N$, let   $\mc H_0^d$ be the set of all Borel probability measures on $[0,1]^d\times\hkh{-1,1}$, $\mc F_{d}$ be the set of all Borel measurable functions from $[0,1]^d$ to $\mbR$, and $\mc{X}_d^n:=\ykh{[0,1]^d\times\hkh{-1,1}}^n$. Then, for 
 $z\in[0,\infty]$, define  
$\mc F_{d,z}:=\setr{f\in\mc{F}_d}{\norm{f}_{[0,1]^d}\leq z}$.  For $F\in(0,\infty)$, define the truncation function 
\begin{equation}\label{eq2}
\bm{T}_F:[-\infty,\infty]\to[-F,F], \ t\mapsto \iffun(t\in[-F,F];t;\mr{sgn}(t)\cdot F), 
\end{equation} where recall that $\mr{sgn}(t):=\iffun(t\geq 0;1;-1)$ is the sign of $t$. We use the symbols $\qd$ and $\qx$ to denote the maximum and minimum operators  respectively. That is, $a_1\qd a_2\qd\cdots\qd  a_k:=\max\hkh{a_1,a_2,\ldots,a_k}$, and $a_1\qx a_2\qx\cdots \qx  a_k:=\min\hkh{a_1,a_2,\ldots,a_k}$.  Let  $m\in\mb N$,  $\mr{r}\in(0,\infty)$, $\Omega$ be a set, and $\mc{F}$ be a set of  functions from $\Omega$ to $\mbR^m$. Define $\mc{N}(\mc{F},\mr{r}):=\inf\set{\#(\mc{A})}{\begin{minipage}{240.84pt}$\mc{A}\subset\mc{F}$, and for any function  $f\in\mc{F}$ there exists a function ${g}\in\mc{A}$ such that $\sup\limits_{x\in\Omega}\norm{f(x)-g(x)\vphantom{2^{2^2}}}_\infty\leq \mr{r}$\end{minipage}}$.  
 In other words, $\mc{N}(\mc{F},\mr{r})$ is the covering number of $\mc{F}$ with radius $\mr{r}$ in the uniform norm.

\section{A Novel Oracle Inequality} 
\label{241217205730}

In this section, we present a novel oracle inequality which serves as the fundamental tool for deriving the generalization bounds provided in Section \ref{240119033824}. 

Recall the definition of the  truncation function $\bm{T}_F$ given  in \eqref{eq2}. 
For any Borel measurable function $f:[0,1]^d\to\mbR$ and any probability distribution $P$ on $[0,1]^d\times\hkh{-1,1}$, we define the $\phi$-risk of $f$ with respect to $P$ by
 $\mc{R}^\phi_P(f):=\int_{[0,1]^d\times\hkh{-1,1}}\phi(y\cdot f(x))\mr{d}P(x,y)$, 
 and define the excess $\phi$-risk as
 \begin{equation}\label{eq3}
 \mc{E}_P^\phi(f):=\mc{R}_P^{\phi}(f)-\inf_{g\in\mc{F}_d}\mc{R}_P^\phi(g). 
 \end{equation}

\begin{theorem} \label{231123013855}Let $n\in\mb N$ and $d\in\mb N$. Consider i.i.d. sample  $\{(X_i,Y_i)\}_{i=1}^n$ drawn from a  distribution $P$ on $[0,1]^d\times\{-1,1\}$ and a nonempty  class $\mc{F}$ of  Borel measurable functions from $[0,1]^d$ to $\mbR$. Let $F\in(0,\infty)$, $J\in(0,\infty)$ and $\phi:\mbR\to[0,\infty)$ be a measurable function with $\sup\limits_{-F\leq z<t\leq F}\abs{\frac{\phi(t)-\phi(z)}{t-z}}\leq J$.   Denote $\hat{f}_n^\dagger$ as an empirical $\phi$-risk minimizer over $\mc{F}$, i.e., 
	\beq\label{231123022329}
	\hat{f}_n^\dagger\in\mathop{{\arg\min}}_{f\in\mc{F}}\frac{1}{n}\sum_{i=1}^n\phi\ykh{Y_if(X_i)}.
	\eeq Suppose that there exist a measurable function $\psi:[0,1]^d\times\{-1,1\}\to\mbR$ and constants  $M\in(0,\infty)$, $H\in\mbR$,  $\Gamma\in(0,\infty)$, $\gamma\in(0,\infty)$, $\theta\in[0,1]$  such that  
	$W:=\max\hkh{3,\;\mc{N}\ykh{\mc{F},\gamma}}<\infty$,
	\beq\label{231208074141}
	\sup\setm{2.3}{\phi(y\cdot \bm{T}_F(f(x)))-\phi(y\cdot f(x))}{(x,y)\in[0,1]^d\times\hkh{-1,1},\;f\in\mc{F}}\leq H, 
	\eeq\beq\label{231123022357}
			  \sup\hkh{\abs{\psi(x,y)}\qd\phi(y\cdot \bm T_F(f(x)))\Big|(x,y)\in [0,1]^d\times\{-1,1\},\;f\in\mc{F}} \leq M, \eeq
	\beq\label{231123022412}
			\Psi:=\int_{[0,1]^d\times\{-1,1\}}{\psi(x,y)}\mr{d}P(x,y)\leq \inf_{f\in\mc{F}}\int_{[0,1]^d\times\{-1,1\}}{\phi(y\cdot \bm T_F(f(x)))}\mr{d}P(x,y), 
			\eeq and
			\beq\label{231123022427}
			&\int_{[0,1]^d\times\{-1,1\}}{\ykh{\phi(y\cdot \bm{T}_F(f(x)))-\psi\ykh{x,y}}^2}\mr{d}P(x,y)\\ &\leq\Gamma \cdot \abs{{\int_{[0,1]^d\times\{-1,1\}}\ykh{\phi(y\cdot \bm T_F(f(x)))-\psi(x,y)}\mr{d}P(x,y)}}^\theta,\;\forall\;f\in\mc{F}. 
	\eeq Then for any $\e\in[0,\infty)$, there holds
	\beq\label{24011917:30:09}
	&\mbe\zkh{\mc{R}_P^{\phi}\big(\bm T_F\circ\hat{f}_n^\dagger\big)-\Psi}\leq(1+\e)\cdot\inf_{f\in\mc{F}}\Big(\mc{R}_P^\phi(f)-\Psi\Big)\\
	&\;\;\;\;\;\;\;\;\;\;\;+\abs{2+\e}\cdot J\cdot\gamma+\frac{8\cdot M\cdot(1+\e)}{n}\cdot\log W+8\cdot\abs{\frac{\Gamma\cdot(1+\e)^2}{n\cdot\e^\theta}\cdot\log W}^{\frac{1}{2-\theta}}+(1+\e)\cdot H.
	\eeq
\end{theorem}

The proof of Theorem \ref{231123013855} is given in Appendix  \ref{20251011003902}. Although  Theorem \ref{231123013855} is stated under the assumption that the input space is $[0,1]^d$,  its proof remains valid when the input space is an arbitrary measurable space. In particular,  replacing $[0,1]^d$ with $\mathbb{R}^d$ enables this theorem to be applied to data generated by Gaussian mixture models. In this context, the generalization analysis of ReLU DNNs classifiers trained with the hinge loss has been investigated by \cite{tianyi2024classification}.

In our following main results, i.e., Theorem \ref{20231125000604}, Theorem \ref{231215231016}, and  Theorem \ref{20231204212113}, we will use Theorem \ref{231123013855} to derive tight upper bounds for the quantity $\mbe\zkh{\ep(\hat f_n)}$. Inequality \eqref{24011917:30:09} leads to  upper bounds for the \zaverage excess 0-1 risk $\mbe\zkh{\ep(\hat f_n)}$ because, for a broad class of loss functions  $\phi$ that satisfy very mild conditions, the so-called comparison inequality holds. Specifically, this inequality takes the form 
\beq\label{241005145754}
\mc E_P(f)\leq c\cdot \ykh{\mc E_P^\phi(f)}^{s_1}\text{ for any } f\in \mc F_d\text{ and} \text{ any  $P\in\mc H_0^d$}, 
\eeq where $c\in(0,\infty)$ and $s_1\in[0,1]$ are constants  independent of $f$ and $P$ (see, e.g., Theorem 2.1 of \cite{zhang2004statistical}; Theorem 1 of \cite{bartlett2006convexity}; Theorem 8.29 of \cite{steinwart2008support}; Theorem 1.1 of \cite{xiang2011classification}; and Subsection 2.1 of  \cite{benabid2021comparison}). Indeed, \eqref{241005145754} implies that 
\[
&
\mbe\zkh{\ep(\hat f_n)}
=
\mbe\zkh{\ep(\bm{T}_F\circ\hat f_n)}
\leq 
\mbe\zkh{c\cdot \ykh{\mc E_P^\phi(\bm{T}_F\circ\hat f_n)}^{s_1}}
\\&
\leq
c\cdot\abs{ \vphantom{\raisebox{3ex}{}}\mbe\zkh{{\mc E_P^\phi(\bm{T}_F\circ\hat f_n)}}}^{s_1}
=
c\cdot\abs{ \vphantom{\raisebox{3ex}{}}\mbe\zkh{{\mc R_P^\phi(\bm{T}_F\circ\hat f_n)}-\Psi}+\Psi-\inf_{g\in\mc{F}_d}\mc{R}_P^\phi(g)}^{s_1}, 
\] which, together with \eqref{24011917:30:09}, yields  upper bounds for $\mbe\zkh{\ep(\hat f_n)}$. 

In this paper,  we only focus on a special case of Theorem \ref{231123013855} when $\phi$ is the hinge loss $\phi_{\mathbf{h}}(t)=\max\hkh{1-t,0}$. We formulate this case as Theorem \ref{231126053801} in Appendix  \ref{20250918073254}, and use it to derive generalization bounds given in Theorem \ref{20231125000604}, Theorem \ref{231215231016}, and  Theorem \ref{20231204212113}. To this end, it is necessary to construct a surrogate function $\psi$ and establish inequalities of the form \eqref{231123022427}. In this work, we take $\psi(x,y)=\phi(y\cdot f^*_{P}(x))$, where $f^*_P(x)=\mr{sgn}(2\eta_P(x)-1)=\mr{sgn}(2\cdot P(\hkh{1}|x)-1)$ is the target function for both the hinge risk and the 0-1 risk. We then employ a classical result, which asserts that
\beq\label{241005183901}
&
\int_{}\abs{\phi_{\mathbf{h}}(yf(x))-\psi(x,y)}^2\mr{d}P(x,y)
= \int_{}\abs{\phi_{\mathbf{h}}(yf(x))-\phi_{\mathbf{h}}(y\cdot\eta_P(x))}^2\mr{d}P(x,y)
\\&\leq  
\Gamma\cdot\ykh{\mc E_P^{\phi_{\mathbf{h}}}(f)}^{\frac{s}{s+1}}
=
\Gamma\cdot\ykh{\mc R_P^{\phi_{\mathbf{h}}}(f)-\int\psi(x,y)\mr{d}P(x,y)}^{\frac{s}{s+1}}
,
\;\forall\;f\in\mc F_{d,1}
\eeq for any $P$ satisfying Tsybakov noise condition  \eqref{23080401} with noise exponent $s$. This inequality has been used in, e.g., \cite{steinwart2007fast,kim2021fast}, to establish upper bounds for excess hinge risks of estimators (cf. Lemma 6.1 of \cite{steinwart2007fast}, Lemma A.2 of \cite{kim2021fast},  and Lemma \ref{23081001} in {Appendix }  \ref{20250918073254}{}). 
From this, we can infer that the bounds given in this special case of Theorem \ref{231123013855} might also be derived using similar arguments to those employed in existing works such as \cite{steinwart2007fast,kim2021fast}, where inequalities of the form \eqref{241005183901} are exploited. However, we emphasize that, although the application of Theorem \ref{231123013855} here is quite specific, the theorem itself has a much broader potential scope, which remains largely unexplored. It is, in fact,  a very general result requiring only that $\phi$ be Lipschitz continuous. In particular, Theorem \ref{231123013855} is applicable in situations where the target function of the $\phi$-risk—that is, the function minimizing the $\phi$-risk $\mc{R}_P^\phi(\cdot)$—exhibits poor properties, such as unboundedness. This is because the theorem does not directly rely on the specific form of the target function for  $\phi$, but instead employ a surrogate function $\psi$. An appropriately constructed $\psi$ can help overcome difficulties arising form the undesirable properties of the target function.  The idea of using a surrogate function $\psi$  was first proposed in our previous paper \cite{zhangzihan2023classification}, in which we establish an oracle inequality analogous to \eqref{24011917:30:09} for the logistic loss $\phi_{\mathbf{l}}(t)=\log(1+\me^{-t})$. In this case, the corresponding target function $f^*_{\phi_{\mathbf{l}},P}(x)=\log\frac{\eta_P(x)}{1-\eta_P(x)}$ may be unbounded, but through a careful construction of $\psi$, we successfully established minimax optimal convergence rates for classifiers trained with $\phi_{\mathbf{l}}$ (see Theorem 2.1 in \cite{zhangzihan2023classification} and the discussion therein). Indeed, Theorem \ref{231123013855} directly implies Theorem 2.1 of \cite{zhangzihan2023classification}. To see this, one simply takes $H=0$, $\theta=1$, $\mc{F}$ to be contained in $\mc{F}_{d,F}=\set{f:[0,1]^d\to[-F,F]}{\text{$f$ is measurable}}$, and $\phi$ to be the logistic loss $\phi_{\mathbf{l}}$ in Theorem \ref{231123013855}. Then, Theorem 2.1 of \cite{zhangzihan2023classification} follows immediately, which is merely an exceedingly particular application of Theorem \ref{231123013855}. Investigating how to apply Theorem \ref{231123013855} to derive tight generalization bounds for other loss functions—particularly for well-known strictly positive loss functions such as the exponential loss and the LUM (Large-Margin Unified Machine; cf. \cite{liu2011hard,benabid2021comparison}) loss, for which the corresponding target functions are unbounded—is a topic worth further study. We will explore this direction in our forthcoming work.

In addition to its generality, Theorem \ref{231123013855} can also be sharp, as the bound \eqref{24011917:30:09} in Theorem \ref{231123013855}  can yield optimal convergence rates. For example, in this work, we successfully apply the bound \eqref{24011917:30:09} to derive optimal convergence rates for the \zaverage excess 0-1 risk of estimators (see Theorem \ref{20231125000604}, Theorem \ref{231215231016}, and Theorem \ref{20231204212113}). Moreover,  in our previous work \cite{zhangzihan2023classification},  we also succeeded in using Theorem 2.1 therein—which, as previously mentioned, is a direct corollary of Theorem \ref{231123013855}—to obtain optimal convergence rates (up to a logarithmic factor) for the \zaverage excess logistic risk of estimators (see Theorem 2.3 of \cite{zhangzihan2023classification}). The key to effectively applying Theorem \ref{231123013855} lies in constructing a suitable function $\psi$ such that the conditions required by Theorem \ref{231123013855}—namely, inequalities \eqref{231123022357}, \eqref{231123022412}, and \eqref{231123022427}—are satisfied,  with the constants $\Gamma$ and $M$ kept sufficiently small. Among these inequalities, the boundedness conditions  \eqref{231123022357} and \eqref{231123022412} are typically easy to fulfill. However, establishing \eqref{231123022427} generally requires considerably more effort, as it is closely tied to the specific construction of $\psi$. For some loss function $\phi$, and $\psi$ constructed depended on $\phi$, threre are many existing tight inequalities which take the form of \eqref{231123022427} and lead to optimal convergence rates. For example,  in our previous work \cite{zhangzihan2023classification}, we establish an inequality of the form \eqref{231123022427} with $\phi$ being the logistic loss $\phi_{\mathbf{l}}$, $\theta=1$, and $\psi$ carefully constructed. See Lemma C.10  of  \cite{zhangzihan2023classification} and its proof for the inequality and the detailed construction of  $\psi$.   Using this result, together with Theorem 2.1 of \cite{zhangzihan2023classification}—which is a special case of Theorem \ref{231123013855} presented here—we prove that the \zaverage excess $\phi_{\mathbf{l}}$-risk of ReLU  DNN estimators can achieve optimal convergence rates (up to logarithmic factors) under the compositional assumption on the CCP function $\eta_P$ of the data distribution $P$. In \cite{farrell2021}, the authors establish an inequality implying  \eqref{231123022427} with  $\phi=\phi_{\mathbf{l}}$, $\theta=1$, and $\psi(x,y)=\phi(y\cdot f^*_{\phi_{\mathbf{l}},P}(x))$, where  $f^*_{\phi_{\mathbf{l}},P}(x)=\log\frac{\eta_P(x)}{1-\eta_P(x)}$ is the target function of $\phi_{\mathbf{l}}$-risk, under the assumption that $f^*_{\phi_{\mathbf{l}},P}$ lies in the unit ball of some H\"older  space (and is therefore bounded). Based on this inequality, they show that the $L^2_{P_X}$ distance from suitable ReLU DNN estimators to the target function  $f^*_{\phi_{\mathbf{l}},P}$  achieves optimal convergence rates (up to logarithmic factors).  See Lemma 8 in \cite{farrell2021} for the inequality and Corollary 1  in \cite{farrell2021} for the result regarding the optimal rates. In this paper, as previously mentioned, we use \eqref{231123022427} with $\psi(x,y)=\phi_{\mathbf{h}}\ykh{y\cdot\mr{sgn}(2\eta_P(x)-1)}$ and  $\gamma=\frac{s}{s+1}$  to establish upper bounds for the \zaverage excess 0-1 risk of estimators. These results are obtained under the compositional assumption on the CCP function $\eta_P$ and the Tsybakov noise condition \eqref{23080401} with exponent $s$, as shown in Theorem \ref{20231125000604}, Theorem \ref{231215231016}, and Theorem \ref{20231204212113} (see \eqref{241005183901} and the discussions therein), which can lead to optimal convergence rates. 

Nonetheless,  if $\psi$ is chosen inappropriately, then Theorem \ref{231123013855} may give loose bounds and fail to yield optimal rates. To illustrate this point, let us take one of the results from the paper \cite{KOHLER2025106188} as an example. In \cite{KOHLER2025106188}, the authors establish convergence rates for the \zaverage excess 0-1 risk and the excess $\phi_{\mathbf{l}}$-risk of certain ReLU convolutional neural network estimators $\hat{f}_n^{\mathbf{CNN}}$ under the so-called  $(\beta, C)$-smooth hierarchical max-pooling model of level $q$. In this model, the CCP function $\eta_P$ is assumed to be the maximum of several functions $f_i$, where each $f_i$ is a composition of $q$ multivariate functions $h_{i,q}, h_{i,q-1}, \ldots, h_{i,1}$ (i.e., $\eta_P = \max_i f_i = \max_i h_{i,q} \circ h_{i,q-1} \circ \cdots \circ h_{i,1}$). Each component function of $h_{i,j}$ is required to be H\"older-$\beta$ smooth with respect to four of its input variables. Under this model, for $\beta\geq 1$, the authors in  \cite{KOHLER2025106188} prove that $\mbe\zkh{\mc{E}^{\phi_{\mathbf{l}}}_P\big(\hat{f}_n^{\mathbf{CNN}}\big)}$ and $\mbe\zkh{\mc{E}_P\big(\hat{f}_n^{\mathbf{CNN}}\big)}$ can achieve the convergence rate ${(\log n)^2\cdot \abs{\frac{1}{n}}^{\frac{\beta}{2\beta+4}\qx\frac{1}{4}}}$ and  ${(\log n)\cdot \abs{\frac{1}{n}}^{\frac{\beta}{4\beta+8}\qx\frac{1}{8}}}$, respectively (see  Theorem 1 of \cite{KOHLER2025106188} and its proof in Section A.2 of the supplementary material of \cite{KOHLER2025106188}). The convergence rate ${(\log n)\cdot \abs{\frac{1}{n}}^{\frac{\beta}{4\beta+8}\qx\frac{1}{8}}}$ of $\mbe\zkh{\mc{E}_P\big(\hat{f}_n^{\mathbf{CNN}}\big)}$ is directly obtained from the convergence rate ${(\log n)^2\cdot \abs{\frac{1}{n}}^{\frac{\beta}{2\beta+4}\qx\frac{1}{4}}}$ of $\mbe\zkh{\mc{E}^{\phi_{\mathbf{l}}}_P\big(\hat{f}_n^{\mathbf{CNN}}\big)}$  by applying the comparison inequality 
\beq\label{241107191558}
\mc E_P(f)\leq \sqrt{2}\cdot \ykh{\mc E_P^\phi(f)}^{1/2},\;\forall\; f\in \mc F_d,\;\forall\;P\in\mc H_0^d \eeq
(see Lemma 1 of \cite{KOHLER2025106188}, Theorem 8.29 of \cite{steinwart2008support}, and the discussion following (8.27) in \cite{steinwart2008support}). Note that the convergence rate ${(\log n)\cdot \abs{\frac{1}{n}}^{\frac{\beta}{4\beta+8}\qx\frac{1}{8}}}$ of $\mbe\zkh{\mc{E}_P\big(\hat{f}_n^{\mathbf{CNN}}\big)}$  is not so satisfactory. In fact, the convergence must be slower than $n^{-1/8}$. As we will show in Theorem \ref{20231204212113} below, under the above model, the optimal convergence rate of the \zaverage excess 0-1 risk of estimators is ${{\ykh{\frac{1}{{n}}}^{\frac{\beta}{{{4}+2\cdot\beta}}}}}$. This is because the model described above is actually covered by our compositional model $\mc G_d^{\mathbf{CHOM}}(q, K,d_\star, d_*, \beta,r)$, which is first proposed in \cite{zhangzihan2023classification} and will also be considered in this paper (see Appendix  \ref{20250916114433}  for its detailed definition), with $d_*=4$. By setting $d_*=4$, $s=0$, and using $\beta\geq 1$ in Theorem \ref{20231204212113}, we immediately obtain the rate ${{\ykh{\frac{1}{{n}}}^{\frac{\beta\cdot(1\qx\beta)^q}{{{d_*}+2\cdot\beta\cdot(1\qx\beta)^q}}}}}= {{\ykh{\frac{1}{{n}}}^{\frac{\beta}{{{4}+2\cdot\beta}}}}}$. Obviously, the optimal rate ${{\ykh{\frac{1}{{n}}}^{\frac{\beta}{{{4}+2\cdot\beta}}}}}$ is much better than the derived rate ${(\log n)\cdot \abs{\frac{1}{n}}^{\frac{\beta}{4\beta+8}\qx\frac{1}{8}}}$ in \cite{KOHLER2025106188}. Note that, as shown in \cite{zhangzihan2023classification}, the comparison inequality \eqref{241107191558} is optimal in the sense that the exponent $1/2$ cannot be increased (see the discussion below (2.21) in \cite{zhangzihan2023classification}). This indicates that the fundamental reason for the slow convergence with rate ${(\log n)\cdot \abs{\frac{1}{n}}^{\frac{\beta}{4\beta+8}\qx\frac{1}{8}}}$ of $\mbe\zkh{\mc{E}_P\big(\hat{f}_n^{\mathbf{CNN}}\big)}$ lies in the unsatisfactory convergence rate ${(\log n)^2\cdot \abs{\frac{1}{n}}^{\frac{\beta}{2\beta+4}\qx\frac{1}{4}}}$ for $\mbe\zkh{\mc{E}^{\phi_{\mathbf{l}}}_P\big(\hat{f}_n^{\mathbf{CNN}}\big)}$ derived in \cite{KOHLER2025106188}, which is even worse than $n^{-1/4}$. In contrast, by setting $d_*=4$ in Theorem 2.3 of \cite{zhangzihan2023classification}, we can show that the expected excess $\phi_{\mathbf{l}}$-risk of ReLU DNN estimators achieves the convergence rate ${\ykh{\frac{\log^5n}{n}}}^{\frac{\beta}{4+\beta}}$, which approaches $\frac1n$ as $\beta \to +\infty$.

To explain why the authors in \cite{KOHLER2025106188} are unable to establish a convergence rate better than $\frac{1}{\sqrt{n}}$ for the \zaverage excess $\phi_{\mathbf{l}}$-risk of estimators, we point out that they essentially derive their convergence rates by applying our Theorem \ref{231123013855} in the special case where $\psi\equiv 0=\e=H$. Indeed,  we prove Theorem \ref{231123013855} via the error decomposition \eqref{20231124115324}  in  Appendix  \ref{20251011003902}{}, which states  that
  \beq\label{241018133828}
  &\mbe\zkh{\mc{E}^\phi_P\big(\bm{T}_F\circ\hat{f}_n^\dagger\big)}
  \leq S_\e+(1+\e)\cdot H+\e\cdot\ykh{\inf_{g\in\mc{F}_d}\mc{R}^\phi_P(g)-\Psi}+(1+\e)\cdot\inf_{f\in\mc{F}}\mc{E}^\phi_P(f)
  \eeq with 
  \beq\label{241105110626}
  S_\e:= \mbe\zkh{\mc{R}^\phi_P\big(\bm{T}_F\circ\hat{f}_n^\dagger\big)-\Psi-\frac{1+\e}{n}\sum_{i=1}^n\ykh{\phi(Y_i\cdot\bm{T}_F(\hat{f}_n^\dagger(X_i)))-\psi(X_i,Y_i)}}. 
  \eeq In \cite{KOHLER2025106188}, the authors choose the hypothesis space $\mc{F}$ to consist of ReLU DNNs whose maximum norms are uniformly bounded by some constant  $\beta_n$. Taking $F=\beta_n$, we have that $\bm{T}_F(f(x))=f(x)$ for any $x\in[0,1]^d$ and any $f\in\mc{F}, 
  $ meaning that $\bm{T}_F\circ\hat{f}_n^\dagger=\hat{f}_n^\dagger$, and 
  \[
  \setm{2.3}{\phi(y\cdot \bm{T}_F(f(x)))-\phi(y\cdot f(x))}{(x,y)\in[0,1]^d\times\hkh{-1,1},\;f\in\mc{F}}=\hkh{0}. 
  \] Thus, \eqref{231208074141} holds for $H=0$. Plugging $\bm{T}_F\circ\hat{f}_n^\dagger=\hat{f}_n^\dagger$ and $\psi\equiv 0=\e=H$ into \eqref{241018133828}, we obtain 
    \beq\label{241018141056}
    \mbe\zkh{\mc{E}^\phi_P\big(\hat{f}_n^\dagger\big)}
     & \leq \mbe\zkh{\mc{R}^\phi_P\big(\hat{f}_n^\dagger\big)-\frac{1}{n}\sum_{i=1}^n\phi(Y_i\cdot\hat{f}_n^\dagger(X_i))}+\inf_{f\in\mc{F}}\mc{E}^\phi_P(f)
     \\
     &\leq
     \mbe\zkh{\sup_{f\in\mc{F}}\abs{\mc{R}^\phi_P\big(f\big)-\frac{1}{n}\sum_{i=1}^n\phi(Y_i\cdot f(X_i))}}+\inf_{f\in\mc{F}}\mc{E}^\phi_P(f). 
    \eeq 
    This is exactly the error decomposition used in \cite{KOHLER2025106188} (see Lemma 2 therein). To further bound the right-hand side of \eqref{241018141056}, the authors of \cite{KOHLER2025106188} use the identity 
    \beq\label{2411051059}
    \mbe\zkh{\sup_{f\in\mc{F}}\abs{\mc{R}^\phi_P\big(f\big)-\frac{1}{n}\sum_{i=1}^n\phi(Y_i\cdot f(X_i))}}=\int_0^\infty\mb{P}\ykh{{\sup_{f\in\mc{F}}\abs{\mc{R}^\phi_P\big(f\big)-\frac{1}{n}\sum_{i=1}^n\phi(Y_i\cdot f(X_i))}}>t}\mr{d}t, 
    \eeq and apply the (uniform) Hoeffding's inequality (see Theorem 9.1 in \cite{gyorfi2002distribution}) to bound the probability 
     \[{\mb{P}\ykh{{\sup\limits_{f\in\mc{F}}\abs{\mc{R}^\phi_P\big(f\big)-\frac{1}{n}\sum_{i=1}^n\phi(Y_i\cdot f(X_i))}}>t}}\]
     (see Section 5 of \cite{KOHLER2025106188}). They obtain the bound 
     \[
     \mb{P}\ykh{{\sup\limits_{f\in\mc{F}}\abs{\mc{R}^\phi_P\big(f\big)-\frac{1}{n}\sum_{i=1}^n\phi(Y_i\cdot f(X_i))}}>t}\leq 
     \min\hkh{W_t\cdot\exp\ykh{-\frac{n\cdot t^2}{a_n}},1}, 
     \] where $a_n\geq 1$,  and $W_t\geq 1$ is the covering number of the hypothesis space $\mc{F}$ (up to some constant factor). Therefore, the derived  upper bound for $\mbe\zkh{\mc{E}^\phi_P\big(\hat{f}_n^\dagger\big)}$ by the approach described above satisfies 
     \[
     \int_0^\infty   \min\hkh{W_t\cdot\exp\ykh{-\frac{n\cdot t^2}{a_n}}\mr{d}t, 1}
     \geq 
       \int_0^\infty   \exp\ykh{-\frac{n\cdot t^2}{a_n}}\mr{d}t = 
       \frac{\sqrt{\pi}}{2}\cdot \sqrt{\frac{a_n}{n}}>\frac{1}{\sqrt{n}}. 
     \] This explains why the derived convergence rates of the \zaverage excess $\phi_{\mathbf{l}}$-risk in \cite{KOHLER2025106188} cannot be better than $\frac{1}{\sqrt n}$.

In contrast, in \cite{zhangzihan2023classification}, we essentially apply Theorem \ref{231123013855}  with $\phi$ being the logistic loss $\phi_{\mathbf{l}}$ and $\psi$ carefully constructed. This careful construction of $\psi$ enables us to fully exploit the information contained in the variance of random samples, rather than merely relying on their ranges as in the case wherer $\psi\equiv 0$ when applying the uniform Hoeffding's inequality. In particular, we prove 
 \[&\int_{[0,1]^d\times\{-1,1\}}{\ykh{\phi(y f(x))-\psi\ykh{x,y}}^2}\mr{d}P(x,y)\\ &\leq12500\cdot\abs{\log\ykh{1+\me^F}}^2\cdot {\int_{[0,1]^d\times\{-1,1\}}\ykh{\phi(y\cdot f(x))-\psi(x,y)}\mr{d}P(x,y)}
\] for some measurable  $\psi:[0,1]^d\times\hkh{-1,1}\to \zkh{0,\log\ykh{\ykh{1+\me^F}\cdot 10\cdot\log\ykh{1+\me^F}}}$ and all measurable  $f:[0,1]^d\to[-F,F]$,  which yields \eqref{231123022427} with $\theta=1$. As a result, we can establish the probability bound  
  \[
&\mb{P}\ykh{\mc{R}^\phi_P\big(\bm{T}_F\circ\hat{f}_n^\dagger\big)-\Psi-\frac{1+\e}{n}\sum_{i=1}^n\ykh{\phi(Y_i\cdot\bm{T}_F(\hat{f}_n^\dagger(X_i)))-\psi(X_i,Y_i)}>t}
\\&\leq
\mb{P}\ykh{\sup_{f\in\mc{F}}\ykh{\mc{R}^\phi_P\big(\bm{T}_F\circ f\big)-\Psi-\frac{1+\e}{n}\sum_{i=1}^n\ykh{\phi(Y_i\cdot\bm{T}_F(f(X_i)))-\psi(X_i,Y_i)}}>t}\\&\leq \min\hkh{W_t'\cdot\exp\ykh{-\frac{n\cdot t}{a_n'}},1},
  \] where $a_n'\geq 1$ depends only on $F$ and $\e$ (noting that $F$ may itself depend on $n$)  and $W_t'\geq 3$ is the covering number of the hypothesis space $\mc{F}$ (see  \eqref{241021093747} and \eqref{20231124104043} with $\theta=1$ in  Appendix  \ref{20251011003902}). This leads to a bound for $S_\e$ defined in \eqref{241105110626}:
     \[
     S_\e&\leq \int_0^\infty\mb{P}\ykh{{\mc{R}^\phi_P\big(\bm{T}_F\circ\hat{f}_n^\dagger\big)-\Psi-\frac{1+\e}{n}\sum_{i=1}^n\ykh{\phi(Y_i\cdot\bm{T}_F(\hat{f}_n^\dagger(X_i)))-\psi(X_i,Y_i)}}>t}\mr{d}t
     \\&\leq \int_0^\infty \min\hkh{W_t'\cdot\exp\ykh{-\frac{n\cdot t}{a_n'}},1}\mr{d}t\leq {t_*}+\int_{t_*}^\infty W_t'\cdot\exp\ykh{-\frac{n\cdot t}{a_n'}}\mr{d}t
     \\&\leq 
    {t_*}+\int_{t_*}^\infty W_{t_*}'\cdot\exp\ykh{-\frac{n\cdot t}{a_n'}}\mr{d}t=t_*+W_{t_*}'\cdot \frac{a_n'}{n}\cdot\exp\ykh{-\frac{n\cdot t_*}{a_n'}}=t_*+\frac{a_n'}{n}\cdot \frac{W_{t_*}'}{W_{a_n'/n}}
    \\&\leq 
    t_*+\frac{a_n'}{n}\leq \frac{a_n'}{n}\cdot2\cdot \log\abs{W'_{a_n'/n}}, 
     \] where $t_*:=\frac{a_n'}{n}\cdot\log\abs{W'_{a_n'/n}}$. Plugging the above inequality into  \eqref{241018133828}, we deduce that 
      \[
       &\mbe\zkh{\mc{E}^\phi_P\big(\bm{T}_F\circ\hat{f}_n^\dagger\big)}
       \leq
        S_\e+(1+\e)\cdot H+\e\cdot\ykh{\inf_{g\in\mc{F}_d}\mc{R}^\phi_P(g)-\Psi}+(1+\e)\cdot\inf_{f\in\mc{F}}\mc{E}^\phi_P(f)
       \\&\leq 
        \frac{a_n'}{n}\cdot2\cdot \log\abs{W'_{a_n'/n}}+(1+\e)\cdot H+\e\cdot\ykh{\inf_{g\in\mc{F}_d}\mc{R}^\phi_P(g)-\Psi}+(1+\e)\cdot\inf_{f\in\mc{F}}\mc{E}^\phi_P(f). 
       \] Therefore, the convergence rate of $\mbe\zkh{\mc{E}^\phi_P\big(\bm{T}_F\circ\hat{f}_n^\dagger\big)}$ can approach $\frac{1}{n}$ provided that the other terms on the above inequality, namely, $a_n'$, $\log\abs{W'_{a_n'/n}}$, $H$, ${\inf\limits_{g\in\mc{F}_d}\mc{R}^\phi_P(g)-\Psi}$, and $\inf\limits_{f\in\mc{F}}\mc{E}^\phi_P(f)$, are well controlled. This aligns with what we have observed from the convergence rate ${\ykh{\frac{\log^5n}{n}}^{{\frac{\beta}{4+\beta}}}}$  derived from taking $d_*=4$ in Theorem 2.3 of \cite{zhangzihan2023classification}: it can be arbitrary close to $\frac1n$ when  $\beta$ is sufficiently large, since $\lim\limits_{\beta\to+\infty}\frac{\beta}{4+\beta}=1$.

Finally, we would like to point out that the condition \eqref{231208074141} in Theorem \ref{231123013855}, which, roughly speaking,  measures the error induced by taking the truncation operation $\bm{T}_F$ on $\hat{f}_n^\dagger$,  is usually easy to verify.  For example,
\eqref{231208074141} holds with $H=0$ if $\mc{F}$ is contained in $\mc{F}_{d,F}=\set{f:[0,1]^d\to[-F,F]}{f\text{ is measurable}}$. Besides, 
 \eqref{231208074141}  holds with $H=\phi(F)$ if $\phi$ is convex or decreasing. In this case, $H$ can be small and will not affect the convergence rate of the derived bound, i.e., the right-hand side of \eqref{24011917:30:09}, provided $F$ is chosen to be sufficiently large. For example, when $\phi$ is the logistic loss $\phi_{\mathbf{l}}(t)=\log(1+\me^{-t})$, a typical choice of $F$ is $\log n$. This implies that $H$ can be taken as $\phi_{\mathbf{l}}(\log n)\leq \frac{1}{n}$, which is sufficiently small and usually decays faster than the other terms on the right-hand side of \eqref{24011917:30:09}.

\section{Generalization Upper Bounds}\label{240119033824} 

In this section, we will apply Theorem \ref{231123013855} to derive several upper bounds for the excess 0-1 risk (i.e., the excess misclassification error) of estimators under the noise condition \eqref{23080401}, together with the compositional assumption (mentioned earlier and to be rigorously defined shortly)  on the conditional class probability function $\eta_P(x):=P(\hkh{1}|x)$ of the data distribution $P$. These results lead to sharp convergence rates.  Specifically,  we establish optimal convergence rates (up to logarithmic factors) for ReLU neural networks  trained with hinge loss in Theorem \ref{20231125000604} and Theorem \ref{231215231016}. Furthermore, in Theorem \ref{20231204212113}, we construct a particular sequence of estimators $\hkh{\hat{f}_n^{\lozenge}}_{n=1}^\infty$ which can rigorously achieve the optimal convergence rates.  

Before stating these results, we first introduce mathematical definitions of ReLU neural networks and the aforementioned compositional assumption.  These definitions are essentially the same as those in our previous work \cite{zhangzihan2023classification}. Therefore, we will adopt some notations and definitions from that work, with slight modifications.  In particular, we consider the space of compositional functions $\mc G_d^{\mathbf{CHOM}}(q, K,d_\star, d_*, \beta,r)$ defined in equation (2.32) of \cite{zhangzihan2023classification}, which  consists of  compositional functions of the form $h_q\circ \cdots\circ h_1\circ h_0$, where each component function \[\iffun\ykh{i>1;[0,1]^K;[0,1]^d}\ni x\mapsto(h_i(x))_j\in \iffun\ykh{i<q;[0,1];\mbR}\] of $h_i$ is either a  function that computes the maximum value of $d_\star$ of its input variables or a H\"older-$\beta$ smooth function whose output value only depends on $d_*$ of its input variables  and H\"older-$\beta$ norm is not greater than $r$. Here we additionally assume  that \beq\label{231231083620}
d_*\leq d\text{ and }d_*\leq K 
\eeq because the essential input dimension $d_*$ of $x\mapsto (h_i(x))_j$ should not be greater than its  apparent input dimension $\iffun(i>1;K;d)$. 
 We will also use the symbol $\fdnn_d(G,N,S,B,F)$ to  define the space  consisting of ReLU neural networks of which the depth, width, number of nonzero parameters,  maximum absolute value of parameters, and uniform norm on $[0,1]^d$ are less than or equal to $G$, $N$, $S$, $B$, and $F$  respectively 
as in \cite{zhangzihan2023classification} (cf. equation (1.15) therein). For the convenience of readers, we  provide the detailed definitions of the spaces  $\fdnn_d(G,N,S,B,F)$ and $\mc G_d^{\mathbf{CHOM}}(q, K,d_\star, d_*, \beta,r)$   in Appendix  \ref{20250916114234} and  Appendix  \ref{20250916114433}  respectively,   
 instead of merely citing the symbols.

We now define the class of probability measures to which we assume the data distribution $P$ belongs. Recall the noise condition \eqref{23080401}. For $(d,s,\alpha,\tau)\in\mb N\times[0,\infty]\times(0,\infty)\times(0,\infty]$, define $\mc{T}^{d,s}_{\alpha,\tau}$ as the set of all Borel probability measures $P$ on $[0,1]^d\times\hkh{-1,1}$ satisfying \eqref{23080401}.  
 Note that, for $\alpha\geq 1$,
\beq\label{230105003839}
\mc{T}^{d,0}_{\alpha,\tau}=\hkh{\text{all Borel probability measures on $[0,1]^d\times\hkh{-1,1}$ }}.
\eeq Moreover,  we have  
$1=P_X\ykh{\setl{x\in[0,1]^d }{ \big|2\cdot P(\hkh{1}|x)-1\big|\leq 1}} \leq \alpha\cdot 1^{s}=\alpha$ for any $P\in \mc{T}^{d,s}_{\alpha,\tau}$ with $\tau\geq 1$. Therefore, 
\beq \label{23081002}
\mc{T}^{d,s}_{\alpha,\tau}=\varnothing\text{ if } \idf_{[1,\infty]}(\tau)\cdot\idf_{(0,1)}(\alpha)= 1. 
\eeq Consequently, in our following main theorems where the class $\mc{T}^{d,s}_{\alpha,\tau}$ is considered, we will always assume $\idf_{[1,\infty]}(\tau)\cdot\idf_{(0,1)}(\alpha)\neq 1$ in order to avoid the trivial case where the set is empty. For $(d,q,K,d_\star,d_*,\beta,r)\in\mb N\times(\mb N\cup\hkh{0})\times\mb N\times\mb N\times\mb N\times(0,\infty)\times(0,\infty)$ with $d_*\leq d$ and $d_*\leq K$, define \beq\label{241115172717}
\mc{H}^{d,\beta,r}_{q,K,d_*,d_\star}:=\setr{P}{\begin{minipage}{212pt}  $P$ is a Borel probability measure on $[0,1]^d\times\hkh{-1,1}$ such that  $P_X\ykh{\set{x\in[0,1]^d}{P(\hkh{1}|x)=f(x)}}=1$ for some function  $f$ in $\mc G_d^{\mathbf{CHOM}}(q, K,d_\star, d_*, \beta,r)$\end{minipage}}.
\eeq

In our upcoming Theorem \ref{20231125000604},  Theorem \ref{231215231016}, and Theorem \ref{20231204212113}, we asume the data distribution $P$ belongs to ${\mc{H}^{d,\beta,r}_{q,K,d_*,d_\star}\cap\mc{T}_{\alpha,\tau}^{d,s}}$. In other words, we establish upper bounds for the \zaverage excess 0-1 risk of estimators  when the CCP function $\eta_P(x)= P(\hkh{1}|x)$ lies in the compositional function space $\mc G_d^{\mathbf{CHOM}}(q, K,d_\star, d_*, \beta,r)$ under the noise condition \eqref{23080401}. Recall that  we have already discussed and justified the reasonableness of this model. In addition, we emphasize that this assumption allows maximum value functions to appear in the compositional structure of the CCP function $\eta_P$, since, as stated on page 35 of  \cite{zhangzihan2023classification},  ``taking the maximum value is an important operation to pass key information from lower scale levels to higher ones''. For a more detailed explanation of this point, refer to pages 35-36 of \cite{zhangzihan2023classification}.

Now we now ready to present Theorem \ref{20231125000604}, which establishes upper bounds for the excess 0-1 risk of ReLU DNN estimators trained with hinge loss under the assumption that $P\in{\mc{H}^{d,\beta,r}_{q,K,d_*,d_\star}\cap\mc{T}_{\alpha,\tau}^{d,s}}$ with $s<\infty$, achieving optimal convergence rates up to logarithmic factors.

\begin{theorem}\label{20231125000604}Suppose   $\phi_{\mathbf{h}}(t)=\max\hkh{1-t,0}$ is  the hinge loss,  $n\in\mb N$, $d\in\mb  N$, $q\in\mb N\cup\hkh{0}$, $K\in\mb N$,  $d_\star\in\mb N$, $d_*\in\mb N$, $\beta\in(0,\infty)$,  $r\in(0,\infty)$, $\alpha\in(0,\infty)$,  $\tau\in(0,\infty]$, $s\in[0,\infty)$,  $\idf_{(0,1)}(\alpha)\cdot\idf_{[1,\infty]}(\tau)\neq 1$,  $d_*\leq d$, and $d_*\leq K$.  
Then for suitable  choices of $G,N,S,B$,  the $\phi_{\mathbf{h}}$-ERM $\hat{f}_n^{\FNN}$ over the neural network space $\fdnn_d(G,N,S,B,\infty)$  with respect to the i.i.d. sample $\hkh{(X_i,Y_i)}_{i=1}^n$ drawn from the distribution $P$ on $[0,1]^d\times\hkh{-1,1}$ satisfies 
 \beq\label{231220061650}
\sup\setm{3}{{\bm E}_{P^{\otimes n}}\zkh{\mc{E}_P(\hat{f}_n^{\FNN})}}{P\in\mc{H}^{d,\beta,r}_{q,K,d_*,d_\star}\cap \mc{T}^{d,s}_{\alpha,\tau}}\lesssim \ykh{\frac{(\log n)^3}{{n}}}^{\frac{\beta\cdot(1\qx\beta)^q}{{\frac{d_*}{s+1}+(1+\frac{1}{s+1})\cdot\beta\cdot(1\qx\beta)^q}}}. 
\eeq 
\end{theorem}

The proof of Theorem \ref{20231125000604} is given in Appendix   \ref{20250918073254}, where we will specify the choices of $G,N,S,B$ (which may depend on the sample size $n$) as well (see \eqref{231220060903} therein).

The convergence  rate $\ykh{\frac{(\log n)^3}{{n}}}^{\frac{\beta\cdot(1\qx\beta)^q}{{\frac{d_*}{s+1}+(1+\frac{1}{s+1})\cdot\beta\cdot(1\qx\beta)^q}}}$ established in \eqref{231220061650} 
 is optimal up to the logarithmic factor $\Bigabs{\log n}^{{\frac{3\cdot\beta\cdot(1\qx\beta)^q}{{\frac{d_*}{s+1}+(1+\frac{1}{s+1})\cdot\beta\cdot(1\qx\beta)^q}}}}$,   according to the lower bounds established in Theorem \ref{23102601} (cf. \eqref{25eq240118175544} and the comments therein). Another advantage of this rate is its independence from the input dimension $d$, which means that it avoids the curse of dimensionality. As previously discussed, the assumption that the data distribution $P\in\mc{H}^{d,\beta,r}_{q,K,d_*,d_\star}\cap \mc{T}^{d,s}_{\alpha,\tau}$ is reasonable. Therefore, the result of Theorem \ref{20231125000604} provides a theoretical explanation for the efficient performance of deep neural networks in high-dimensional classification problems.

Note that Theorem \ref{20231125000604} excludes the case $s=\infty$. However, as mentioned right after \eqref{23080401},  the noise condition \eqref{23080401} with $s=\infty$ is still quite reasonable and is likely to be satisfied in practice. Since the  exponent of the derived rate 
 $\abs{\vphantom{\raisebox{3ex}{}}\frac{(\log n)^3}{n}}^{{\frac{\beta\cdot(1\qx\beta)^q}{{\frac{d_*}{s+1}+(1+\frac{1}{s+1})\cdot\beta\cdot(1\qx\beta)^q}}}}$  
in Theorem \ref{20231125000604} satisfies 
\[
&\lim_{s\to+\infty}{\frac{\beta\cdot(1\qx\beta)^q}{{\frac{d_*}{s+1}+(1+\frac{1}{s+1})\cdot\beta\cdot(1\qx\beta)^q}}}=1={\frac{\beta\cdot(1\qx\beta)^q}{{0+(1+0)\cdot\beta\cdot(1\qx\beta)^q}}}={\frac{\beta\cdot(1\qx\beta)^q}{{\frac{d_*}{\infty+1}+(1+\frac{1}{\infty+1})\cdot\beta\cdot(1\qx\beta)^q}}}, \] 
one may conjecture that in the case $s=\infty$, the estimator $\hat{f}_n^{\FNN}$ can achieve the convergence rate 
\beq\label{240104155820}
\sup\setm{3}{{\bm E}_{P^{\otimes n}}\zkh{\mc{E}_P(\hat{f}_n^{\FNN})}}{P\in\mc{H}^{d,\beta,r}_{q,K,d_*,d_\star}\cap \mc{T}^{d,s}_{\alpha,\tau}}\lesssim \ykh{\frac{(\log n)^3}{n}}^{{\frac{\beta\cdot(1\qx\beta)^q}{{\frac{d_*}{\infty+1}+(1+\frac{1}{\infty+1})\cdot\beta\cdot(1\qx\beta)^q}}}}=\frac{(\log n)^3}{n}. 
\eeq 
Indeed, the rate $\frac{(\log n)^3}{n}$ shown in \eqref{240104155820} can even be refined. In the next theorem, we obtain a slightly faster rate $\frac{\log n}{n}$. 

 \begin{theorem}\label{231215231016}Suppose   $\phi_{\mathbf{h}}(t)=\max\hkh{1-t,0}$ is the hinge loss,  $n\in\mb N$, $d\in\mb  N$, $q\in\mb N\cup\hkh{0}$, $K\in\mb N$,  $d_\star\in\mb N$, $d_*\in\mb N$, $\beta\in(0,\infty)$,  $r\in(0,\infty)$, $\alpha\in(0,\infty)$,  $\tau\in(0,\infty]$,  $\idf_{(0,1)}(\alpha)\cdot\idf_{[1,\infty]}(\tau)\neq 1$,    $d_*\leq d$ and $d_*\leq K$. 
 Then for  suitable  choices of $G,N,S,B$,  the $\phi_{\mathbf{h}}$-ERM $\hat{f}_n^{\FNN}$ over the neural network space $\fdnn_d(G,N,S,B,\infty)$  with respect to the i.i.d. sample $\hkh{(X_i,Y_i)}_{i=1}^n$ drawn from the distribution $P$ on $[0,1]^d\times\hkh{-1,1}$ satisfies 
   \beq\label{231222054619}
  \sup\setm{3}{{\bm E}_{P^{\otimes n}}\zkh{\mc{E}_P(\hat{f}_n^{\FNN})}}{P\in\mc{H}^{d,\beta,r}_{q,K,d_*,d_\star}\cap \mc{T}^{d,\infty}_{\alpha,\tau}}\lesssim\frac{\log n}{n}. 
  \eeq

\end{theorem}

The proof of Theorem \ref{231215231016} is given in  Appendix  \ref{20250918113923} of this paper, where we will specify the choices of $G,N,S,B$ (which may depend on the sample size $n$) as well (see \eqref{231222054545} therein). 
As in Theorem \ref{20231125000604}, we can deduce from Theorem \ref{23102601} (with $s=\infty$) that  the convergence rate $\frac{\log n}{n}$ derived in Theorem  \ref{231215231016} is optimal up to the logarithmic factor $\log n$.  Remarkably, this fast, dimension-free rate $\frac{\log n}{n}$  can be achieved by neural networks of fixed size, in the sense that their depth and width do not need to grow with $n$ as  $n\to\infty$ (see the lower bounds of $G$ and $N$ given in \eqref{231222054545} in Appendix  \ref{20250918113923}). In contrast, the depth and width of networks which achieve the convergence rates stated in Theorem \ref{20231125000604} when the noise condition \eqref{23080401} with exponent $s<\infty$ is satisified must increase as $n \to \infty$. This marks a fundamental distinction between the case $s=\infty$ and the case $s<\infty$. Therefore, compared to Theorem \ref{20231125000604}, Theorem \ref{231215231016} offers further insight into the efficiency of deep neural networks.

The results of Theorem \ref{20231125000604} and Theorem \ref{231215231016} can be  briefly summarized as follows. Under the assumption that  the data distribution $P\in\mc{H}^{d,\beta,r}_{q,K,d_*,d_\star}\cap \mc{T}^{d,s}_{\alpha,\tau}$, a neural network estimator $\hat{f}_n^{\FNN}$ trained with hinge loss can achieve the convergence rate 
\beq\label{23eq240118233153}
\sup\setm{3}{{\bm E}_{P^{\otimes n}}\zkh{\mc{E}_P(\hat{f}_n^{\FNN})}}{P\in\mc{H}^{d,\beta,r}_{q,K,d_*,d_\star}\cap \mc{T}^{d,s}_{\alpha,\tau}}\lesssim \ykh{\frac{(\log n)^{1+2\cdot\idf_{\mbR{}}(s)}}{n}}^{{\frac{\beta\cdot(1\qx\beta)^q}{{\frac{d_*}{s+1}+(1+\frac{1}{s+1})\cdot\beta\cdot(1\qx\beta)^q}}}}
\eeq
for  $d\in\mb  N$, $q\in\mb N\cup\hkh{0}$, $K\in\mb N$,  $d_\star\in\mb N$, $d_*\in\mb N$, $\beta\in(0,\infty)$,  $r\in(0,\infty)$, $\alpha\in(0,\infty)$,  $\tau\in(0,\infty]$, and $s\in[0,\infty]$. Here, the rate  ${\ykh{\frac{(\log n)^{1+2\cdot\idf_{\mbR{}}(s)}}{n}}^{{\frac{\beta\cdot(1\qx\beta)^q}{{\frac{d_*}{s+1}+(1+\frac{1}{s+1})\cdot\beta\cdot(1\qx\beta)^q}}}}}$ is optimal up to a logarithmic factor.  By taking $q=0$ and ${d_*=d_\star=K=d}$ in \eqref{23eq240118233153}, we obtain 
\beq\label{241115184352}
\sup\setm{3}{{\bm E}_{P^{\otimes n}}\zkh{\mc{E}_P(\hat{f}_n^{\FNN})}}{P\in\mc{H}^{d,\beta,r}_{0,d,d,d}\cap \mc{T}^{d,s}_{\alpha,\tau}}\lesssim \ykh{\frac{(\log n)^{1+2\cdot\idf_{\mbR{}}(s)}}{n}}^{{\frac{\beta}{{\frac{d}{s+1}+(1+\frac{1}{s+1})\cdot\beta}}}}
\eeq
Since $\mc{H}^{d,\beta,r}_{0,d,d,d}$ contains all probability measures $P$ on $[0,1]^d\times\hkh{-1,1}$ with its conditional class probability function $P(\hkh{1}|\;\cdot\;)\in \mc{B}^\beta_r([0,1]^d)$,   \eqref{241115184352} actually yields  Theorem 3.3 in \cite{kim2021fast}. Specifically, Theorem 3.3 in \cite{kim2021fast}  states that  estimators $\hat{f}_n^{\FNN}$ trained by empirical hinge risk minimization over an appropriately chosen ReLU DNN hypothesis space $\fdnn_d(G,N,S,B,1)$, under the assumption that the CCP function $\eta_P$ is H\"older-$\beta$ smooth on $[0,1]^d$ and satisfies the Tsybakov noise condition \eqref{23080401}, can achieve the convergence rate ${(\frac{\log^3n}{n})^{\frac{\beta}{\frac{d}{1+s}+(1+1/(s+1))\cdot\beta}}}$. This coincides with our rate $\ykh{\frac{(\log n)^{1+2\cdot\idf_{\mbR{}}(s)}}{n}}^{{\frac{\beta}{{\frac{d}{s+1}+(1+\frac{1}{s+1})\cdot\beta}}}}$ in \eqref{241115184352} when $s<\infty$, but is strictly slower than ours when $s=\infty$. Moreover, it is worth noting that our hypothesis space used in Theorem \ref{20231125000604} and Theorem \ref{231215231016} is of the form $\fdnn_d(G,N,S,B,\infty)$. Compared with Theorem 3.3 of \cite{kim2021fast}, we do not require the  unform norms of DNNs in the hypothe space to be bounded by $1$. In practical training procedures, the uniform norms of DNNs are generally not controlled. Therefore, our choice of hypothesis space is more reasonable than that in Theorem 3.3 of \cite{kim2021fast}. 

Another noteworthy   result concerning  convergence rates of neural network classification is \cite{meyer2023optimal}, where optimal convergence    rates are obtained under the assumption that the decision boundary of $P$ (that is, the boundary of the Bayes decision set   $\set{x\in[0,1]^d}{\eta_P(x)\geq\frac12}$ in $[0,1]^d$) is piecewise smooth. 
However, it is not meaningful to make a direct comparison between the rates in \cite{meyer2023optimal}  and those in this paper because the setting in \cite{meyer2023optimal}  is  entirely different from ours, where we require  $\eta_P$ to be a compositional H\"older smooth function. Indeed, even when $\eta_P$ is  compositional H\"older smooth,  the decision boundary of $P$ may not be piecewise smooth;  Conversely, even when $P$ has a piecewise smooth decision boundary, $\eta_P$ may be non-smooth. In Appendix  \ref{20250919034321} of this paper,  we provide a detailed  discussion on the difference between the setting in \cite{meyer2023optimal} and that in this paper  so as   to promote a better understanding of the contribution of this paper and how it relates to  \cite{meyer2023optimal}.

In Theorem 2.3 \label{241214215551}  of our previous work \cite{zhangzihan2023classification}, we establish  convergence rates  for the \zaverage excess logistic risk of ReLU DNN estimators under the compositional model $\mc{H}^{d,\beta,r}_{q,K,d_*,d_\star}$, but without the noise condition \eqref{23080401}. Specifically, Theorem 2.3 of \cite{zhangzihan2023classification} states that 
\beq\label{241121010851}
\sup\setm{3}{{\bm E}_{P^{\otimes n}}\zkh{\mc{E}_P^{\phi_{\mathbf{l}}}(\tilde{f}_n^{\FNN})}}{P\in\mc{H}^{d,\beta,r}_{q,K,d_*,d_\star}}
\lesssim \ykh{\frac{(\log n)^5}{n}}^{{\frac{\beta\cdot(1\qx\beta)^q}{{d_*+\beta\cdot(1\qx\beta)^q}}}}
\eeq 
for  estimators $\tilde{f}_n^{\FNN}$ trained with the logistic loss $\phi_{\mathbf{l}}(t)=\log(1+\me^{-t})$ over suitable ReLU DNN spaces. Using the comparison inequality \eqref{241107191558}, we then deduce from \eqref{241121010851} that 
 \beq\label{241121012657}
 \sup\setm{3}{{\bm E}_{P^{\otimes n}}\zkh{\mc{E}_P^{}(\tilde{f}_n^{\FNN})}}{P\in\mc{H}^{d,\beta,r}_{q,K,d_*,d_\star}}\lesssim \sqrt{\ykh{\frac{(\log n)^5}{n}}^{{\frac{\beta\cdot(1\qx\beta)^q}{{d_*+\beta\cdot(1\qx\beta)^q}}}}}=\ykh{\frac{(\log n)^5}{n}}^{{\frac{\beta\cdot(1\qx\beta)^q}{{2d_*+2\beta\cdot(1\qx\beta)^q}}}}. 
 \eeq
On the other hand, taking $s=0$ and $\alpha=\tau=2$ in \eqref{23eq240118233153} yields
\beq \label{241121012631}
&\sup\setm{3}{{\bm E}_{P^{\otimes n}}\zkh{\mc{E}_P(\hat{f}_n^{\FNN})}}{P\in\mc{H}^{d,\beta,r}_{q,K,d_*,d_\star}}
\\&
\xlongequal{\because\eqref{230105003839}}
\sup\setm{3}{{\bm E}_{P^{\otimes n}}\zkh{\mc{E}_P(\hat{f}_n^{\FNN})}}{P\in\mc{H}^{d,\beta,r}_{q,K,d_*,d_\star}\cap \mc{T}^{d,0}_{2,2}}
\lesssim \ykh{\frac{(\log n)^{3}}{n}}^{{\frac{\beta\cdot(1\qx\beta)^q}{{{d_*}+2\cdot\beta\cdot(1\qx\beta)^q}}}}. 
\eeq
Obviously, the rate ${\ykh{\frac{(\log n)^{3}}{n}}^{{\frac{\beta\cdot(1\qx\beta)^q}{{{d_*}+2\cdot\beta\cdot(1\qx\beta)^q}}}}}$ in  \eqref{241121012631}  is strictly better than the rate ${\ykh{\frac{(\log n)^5}{n}}^{{\frac{\beta\cdot(1\qx\beta)^q}{{2d_*+2\beta\cdot(1\qx\beta)^q}}}}}$ in  \eqref{241121012657},  indicating that ReLU DNNs trained with  logistic loss under the assumption $P\in \mc{H}^{d,\beta,r}_{q,K,d_*,d_\star}$ are sub-optimal for minimizing the excess 0-1 risk. This observation has particularly insightful explanation: as discussed on page 5 of \cite{zhangzihan2023classification}, estimators trained with the logistic loss $\phi_{\mathbf{l}}$ capture more information about the exact value of the CCP function $\eta_P(x)$ than we actually need to minimize the excess 0-1 risk $\mc{E}_P(\cdot)$, leading to inefficiency in minimizing the 0-1 risk. Indeed, the sign of $2\eta_P(x)-1$ suffices for minimizing $\mc{E}_P(\cdot)$, as $\mr{sgn}(2\eta_P(x)-1)$ minimizes the excess 0-1 risk. However, estimators trained with $\phi_{\mathbf{l}}$ also encode the exact values of $\eta_P(x)$, since they effectively learn the minimizer of the excess $\phi_{\mathbf{l}}$-risk, i.e., $\log\frac{\eta_P(x)}{1-\eta_P(x)}$, from which $\eta_P(x)$ can be recovered. We actually demonstrated that ReLU DNN estimators $\tilde{f}_n^{\FNN}$ trained with $\phi_{\mathbf{l}}$ can be highly efficient in minimizing the excess $\phi_{\mathbf{l}}$-risk. In fact, Theorem 2.6 of \cite{zhangzihan2023classification} shows that the convergence rate ${\ykh{\frac{(\log n)^5}{n}}^{{\frac{\beta\cdot(1\qx\beta)^q}{{d_*+\beta\cdot(1\qx\beta)^q}}}}}$ is optimal up to logarithmic factors (see also the discussion on page 35 of \cite{zhangzihan2023classification}). This implies that the estimators $\tilde{f}_n^{\FNN}$ indeed carry substantial information about the values of $\eta_P(x)$—information not required for minimizing the excess 0-1 risk—thereby causing inefficiency in this context. In contrast, ReLU DNNs trained with hinge loss, unlike those trained with logistic loss, can achieve the optimal convergence rate (up to a logarithmic factor) for minimizing the excess 0-1 risk, as established in our Theorems \ref{20231125000604} and Theorem \ref{231215231016}. This is somewhat due to the fact that the minimizer of the excess 0-1 risk given by $\mr{sgn}(2\eta_P(x)-1)$ also minimizes the excess hinge risk $\mc{E}_P^{\phi_{\mathbf{h}}}(\cdot)$. Therefore, training with hinge loss—which essentially learns the minimizer of the excess hinge risk—avoids capturing redundant information not necessary for minimizing the excess 0-1 risk.

One may \label{241217172519}  think that  adding the noise  condition can improve the convergence rate of the \zaverage excess $\phi_{\mathbf{l}}$-risk given in \eqref{241121010851}. Unfortunately, this is not generally true. As mentioned on page 35 of  \cite{zhangzihan2023classification},  ``the noise condition \eqref{23080401} does little to help improve the convergence rate of the excess $\phi_{\mathbf{l}}$-risk in classification''.  
We give a more detailed discussion of this in  Appendix  \ref{20250919034415} of this paper, 
 where we provide both an intuitive explanation and a rigorous mathematical proof of the fact that  the convergence rates of the expected excess logistic risk of estimators cannot be improved even under the Tsybakov noise condition $P\in \mc{T}^{d,s}_{\alpha,\tau}$ for a  wide range of choices of the parameters $\alpha$, $\tau$, $s$ (see \eqref{241202002615} therein), and demonstrate   how the remaining cases are different and more subtle.

In the next Theorem, we construct estimators which rigorously achieve the optimal convergence rate under the assumption that $P\in\mc{H}^{d,\beta,r}_{q,K,d_*,d_\star}\cap \mc{T}^{d,s}_{\alpha,\tau}$.

\begin{theorem}\label{20231204212113}Let $n\in\mb N$, $d\in\mb  N$, $q\in\mb N\cup\hkh{0}$, $K\in\mb N$,  $d_\star\in\mb N$, $d_*\in\mb N$, $\beta\in(0,\infty)$,  $r\in(0,\infty)$, $\alpha\in(0,\infty)$,  $\tau\in(0,\infty]$, and $s\in[0,\infty]$. Suppose $d_*\leq d$, $d_*\leq K$, and $\idf_{(0,1)}(\alpha)\cdot\idf_{[1,\infty]}(\tau)\neq 1$. Then there exist a constant $\cwu\in(0,\infty)$ depending only on $(d,q,K,d_\star,d_*,\beta,r,\tau,\alpha)$ and an estimator ${\hat f_n^{\lozenge}}:[0,1]^d\to\mbR$ based upon the i.i.d. sample $\hkh{(X_i,Y_i)}_{i=1}^n$ drawn from $P$ on $[0,1]^d\times\hkh{-1,1}$ such that \beq\label{231208034119}
	\sup\setm{3}{{\bm E}_{P^{\otimes n}}\zkh{\mc{E}_P(\hat{f}_n^{\lozenge})}}{P\in\mc{H}^{d,\beta,r}_{q,K,d_*,d_\star}\cap \mc{T}^{d,s}_{\alpha,\tau}}\leq\cwu\cdot\ykh{\frac{1}{{n}}}^{\frac{\beta\cdot(1\qx\beta)^q}{{\frac{d_*}{s+1}+(1+\frac{1}{s+1})\cdot\beta\cdot(1\qx\beta)^q}}}. 
	\eeq
	\end{theorem}
The proof of Theorem \ref{20231204212113} is given in Appendix  \ref{20251011004301} of this paper. According to the lower bound established in Theorem \ref{23102601} and equation \eqref{25eq240118175544}, we deduce that the estimator $\hat{f}_n^{\lozenge}$  rigorously achieves  the minimax optimal convergence rate    
 $\inf\limits_{\hat{f}_n}\sup\setl{{\bm E}_{P^{\otimes n}}\zkh{\mc{E}_P(\hat{f}_n)}\vphantom{\Bigg|}}{P\in\mc{H}^{d,\beta,r}_{q,K,d_*,d_\star}\cap\mc{T}^{d,s}_{\alpha,\tau}}\asymp{\ykh{\frac{1}{{n}}}^{\frac{\beta\cdot(1\qx\beta)^q}{{\frac{d_*}{s+1}+(1+\frac{1}{s+1})\cdot\beta\cdot(1\qx\beta)^q}}}}$. 
The idea in the proof of Theorem \ref{20231204212113} is to apply Theorem \ref{231123013855} with $\phi$ being the hinge loss and a carefully constructed finite set $\mc{F}$ (see  \eqref{75eq240119025342} in Appendix   \ref{20251011004301} of this paper for the detailed construction).  Therefore, the optimality of the derived rate in Theorem \ref{20231204212113} justifies the tightness of our oracle inequality established in Theorem \ref{231123013855}.

\section{Minimax Lower Rates}\label{2412172103}

In this subsection, we provide minimax lower rates for the excess 0-1 risk (i.e., the excess misclassification error) of estimators, which justify the optimality of the convergence rates  established in Theorem \ref{20231125000604}, Theorem \ref{231215231016}, and Theorem \ref{20231204212113}, where the data distribution $P$ is assumed to belong to the class $\mc{H}^{d,\beta,r}_{q,K,d_*,d_\star}\cap \mc{T}^{d,s}_{\alpha,\tau}$. To this end, we first introduce a class of probabilities $\mc{H}^{d,\beta,r,\Lambda}_{q,K,d_*}$, which is a strict subset of $\mc{H}^{d,\beta,r}_{q,K,d_*,d_\star}$. We then establish minimax lower rates under the assumption that the data distribution $P\in \mc{H}^{d,\beta,r,\Lambda}_{q,K,d_*} \cap \mc{T}^{d,s}_{\alpha,\tau}$, thereby obtaining lower bounds for the case $P\in\mc{H}^{d,\beta,r}_{q,K,d_*,d_\star}\cap \mc{T}^{d,s}_{\alpha,\tau}$. 
 The space  $\mc{H}^{d,\beta,r,\Lambda}_{q,K,d_*}$, roughly speaking, consists of distributions $P$ on $[0,1]^d\times\hkh{-1,1}$ of which the marginal distribution $P_X$ on $[0,1]^d$ has a probability density function bounded by $\Lambda$, and the  CCP function $\eta_P$ can be expressed as  $h_q\circ \cdots \circ h_1\circ h_0$, where each component function $x\mapsto (h_i(x))_j$ is H\"older-$\beta$ smooth and depends only on  $d_*$ of its input variables.  One may refer to Appendix  \ref{20250915195346}  for the  detailed definition of $\mc{H}^{d,\beta,r,\Lambda}_{q,K,d_*}$.

\begin{theorem}\label{23102601}
	Let $q\in\mb N\cup\hkh{0}$,  $d\in\mb N$, $d_*\in\mb N$, $K\in\mb N$, $\alpha\in(0,\infty)$,  $\tau\in(0,\infty]$, $s\in[0,\infty]$, $\beta\in(0,\infty)$,  and  $\Lambda\in(1,\infty)$. Suppose that $d_*\leq K$, $d_*\leq d$,  and  $\idf_{(0,1)}(\alpha)\cdot\idf_{[1,\infty]}(\tau)\neq 1$.   Then there exist  constants $\cliu\in(0,\infty)$  depending only on $(\beta,\Lambda)$,  
	$\cshisan\in(0,\infty)$ depending only on $(d_*,\beta,s,q,\alpha,\Lambda)$,  and $\cba\in(0,\infty)$ depending only on $(d_*,\beta,s,q,\Lambda)$ such that 
	\[
	&\inf_{\hat{f}_n}\sup\setl{{\bm E}_{P^{\otimes n}}\zkh{\mc{E}_P(\hat{f}_n)}\vphantom{\Bigg|}}{P\in\mc{H}^{d,\beta,r,\Lambda}_{q,K,d_*}\cap\mc{T}^{d,s}_{\alpha,\tau}}\geq \max\hkh{\frac{1}{32}\cdot\idf_{\hkh{\infty}}(s),{\cshisan}}\cdot \ykh{\frac{1}{{n}}}^{\frac{\beta\cdot(1\qx\beta)^q}{{\frac{d_*}{s+1}+(1+\frac{1}{s+1})\cdot\beta\cdot(1\qx\beta)^q}}}
	\] for any $r\in[\cliu,\infty)$ and any  $n\in\mb N\cap(\cba,\infty)$,   where the infimum  is taken over all estimators ${\hat f_n:[0,1]^d\to\mbR}$  based upon the i.i.d. sample $\hkh{(X_i,Y_i)}_{i=1}^n$ drawn from ${[0,1]^d\times\hkh{-1,1}}$. 
\end{theorem}

The proof of Theorem \ref{23102601} is given in  Appendix  \ref{20251011004523}. By applying Theorem \ref{20231204212113},  Theorem \ref{23102601}, and the fact that
 $\mc{H}^{d,\beta,r,\Lambda}_{q,K,d_*}\cap\mc{T}_{\alpha,\tau}^{d,s}\subset \mc{H}^{d,\beta,r}_{q,K,d_*,d_\star}\cap\mc{T}_{\alpha,\tau}^{d,s}$,  
we obtain 
\[
&\ykh{\frac{1}{{n}}}^{\frac{\beta\cdot(1\qx\beta)^q}{{\frac{d_*}{s+1}+(1+\frac{1}{s+1})\cdot\beta\cdot(1\qx\beta)^q}}} \lesssim \inf_{\hat{f}_n}\sup\setl{{\bm E}_{P^{\otimes n}}\zkh{\mc{E}_P(\hat{f}_n)}\vphantom{\Bigg|}}{P\in\mc{H}^{d,\beta,r,\Lambda}_{q,K,d_*}\cap\mc{T}^{d,s}_{\alpha,\tau}}\\
&\leq \inf_{\hat{f}_n}\sup\setl{{\bm E}_{P^{\otimes n}}\zkh{\mc{E}_P(\hat{f}_n)}\vphantom{\Bigg|}}{P\in\mc{H}^{d,\beta,r}_{q,K,d_*,d_\star}\cap\mc{T}^{d,s}_{\alpha,\tau}}\\&\leq \sup\setm{3}{{\bm E}_{P^{\otimes n}}\zkh{\mc{E}_P(\hat{f}_n^{\lozenge})}}{P\in\mc{H}^{d,\beta,r}_{q,K,d_*,d_\star}\cap \mc{T}^{d,s}_{\alpha,\tau}}\lesssim \ykh{\frac{1}{{n}}}^{\frac{\beta\cdot(1\qx\beta)^q}{{\frac{d_*}{s+1}+(1+\frac{1}{s+1})\cdot\beta\cdot(1\qx\beta)^q}}}, 
\] 
which implies that 
\beq\label{25eq240118175544}
&\inf_{\hat{f}_n}\sup\setl{{\bm E}_{P^{\otimes n}}\zkh{\mc{E}_P(\hat{f}_n)}\vphantom{\Bigg|}}{P\in\mc{H}^{d,\beta,r,\Lambda}_{q,K,d_*}\cap\mc{T}^{d,s}_{\alpha,\tau}}\\
&\asymp
\inf_{\hat{f}_n}\sup\setl{{\bm E}_{P^{\otimes n}}\zkh{\mc{E}_P(\hat{f}_n)}\vphantom{\Bigg|}}{P\in\mc{H}^{d,\beta,r}_{q,K,d_*,d_\star}\cap\mc{T}^{d,s}_{\alpha,\tau}}
\asymp
 \ykh{\frac{1}{{n}}}^{\frac{\beta\cdot(1\qx\beta)^q}{{\frac{d_*}{s+1}+(1+\frac{1}{s+1})\cdot\beta\cdot(1\qx\beta)^q}}}
\eeq 
for $q\in\mb N\cup\hkh{0}$,  $d\in\mb N$, $d_*\in\mb N$, $K\in\mb N$, $\alpha\in(0,\infty)$,  $\tau\in(0,\infty]$, $s\in[0,\infty]$, $\beta\in(0,\infty)$,   $\Lambda\in(1,\infty)$, and $r\in[\cliu,\infty)$ with $d_*\leq K$, $d_*\leq d$,  and  $\idf_{(0,1)}(\alpha)\cdot\idf_{[1,\infty]}(\tau)\neq 1$. Thus, the estimator $\hat{f}_n^{\lozenge}$ in Theorem \ref{20231204212113} achieves the optimal convergence rate, and the neural network estimator $\hat{f}_n^{\FNN}$ in Theorem \ref{20231125000604} and Theorem \ref{231215231016} achieves the optimal convergence rates up to some logarithmic factors. Moreover, since $\mc{H}^{d,\beta,r,\Lambda}_{0,d,d}$ consists of all probability measures on $[0,1]^d\times \hkh{-1,1}$ with its CCP  function $\eta_P$ contained in the H\"older ball $\mc{B}^\beta_r([0,1]^d)$ and its marginal distribution $P_X$ having a $[0,\Lambda]$-valued density function (with respect to the Lebesgue measure) on $[0,1]^d$, equation \eqref{25eq240118175544} with $q=0$ and $d_*=K=d$ yields  the minimax rate 
\beq\label{241212030944}
&\inf_{\hat{f}_n}\sup\setl{{\bm E}_{P^{\otimes n}}\zkh{\mc{E}_P(\hat{f}_n)}\vphantom{\Bigg|}}{P\in\mc{H}^{d,\beta,r,\Lambda}_{0,d,d}\cap\mc{T}^{d,s}_{\alpha,\tau}}
\asymp \ykh{\frac{1}{{n}}}^{\frac{\beta}{{\frac{d}{s+1}+(1+\frac{1}{s+1})\cdot\beta}}}, 
\eeq
which coincides with the minimax rate without the compositional assumption established in Section 4 of \cite{audibert2007fast}. Specifically, the authors of \cite{audibert2007fast} consider the following noise condition 
\beq\label{241210043327}
P_X\ykh{\setl{x\in[0,1]^d }{ 0<\big|2\cdot \eta_P(x)-1\big|\leq t}} \leq \alpha\cdot t^{s}, \quad \forall\;0<t
	\leq \tau, 
\eeq which is quite similar to, yet slightly weaker than, condition \eqref{23080401}. Condition \eqref{241210043327} has been adopted to derive improved convergence rates in binary classification (see, e.g., \cite{feng2021generalization}) as well. Besides,  \cite{audibert2007fast} assumes that the CCP function $\eta_P$ is H\"older-$\beta$ smooth for some $\beta\in(0,\infty)$, and that the marginal distribution $P_X$ on $[0,1]^d$ has a density function bounded from above  (i.e., satisfies the mild density assumption, cf. Definition 2.1 therein). Under these assumptions, the authors of \cite{audibert2007fast} verify the  minimax optimal convergence rate for the \zaverage  excess 0-1 risk to be  ${{\ykh{\frac{1}{{n}}}^{\frac{\beta}{{\frac{d}{s+1}+(1+\frac{1}{s+1})\cdot\beta}}}}}$ (see Theorem 4.1 and Theorem 4.3 of \cite{audibert2007fast}). Specifically, they prove that 
\beq\label{241212030945}
&\inf_{\hat{f}_n}\sup\setl{{\bm E}_{P^{\otimes n}}\zkh{\mc{E}_P(\hat{f}_n)}\vphantom{\Bigg|}}{P\in\mc{H}^{d,\beta,r,\Lambda}_{0,d,d}\cap\overline{\mc{T}}^{d,s}_{\alpha,\tau}}
\asymp \ykh{\frac{1}{{n}}}^{\frac{\beta}{{\frac{d}{s+1}+(1+\frac{1}{s+1})\cdot\beta}}}, 
\eeq
 where \[
{\overline{\mc{T}}}_{\alpha,\tau}^{d,s}:=\hkh{\text{Borel probability measures $P$ on $[0,1]^d\times\hkh{-1,1}$ satisfying \eqref{241210043327}}}. 
\] Although the results \eqref{241212030944} and \eqref{241212030945} seem very similar, they may not be equivalent. In fact, since $\mc{T}^{d,s}_{\alpha,\tau}\subsetneqq \overline{\mc{T}}^{d,s}_{\alpha,\tau}$, the lower rate part of our result \eqref{241212030944}, namely, 
\[
&\inf_{\hat{f}_n}\sup\setl{{\bm E}_{P^{\otimes n}}\zkh{\mc{E}_P(\hat{f}_n)}\vphantom{\Bigg|}}{P\in\mc{H}^{d,\beta,r,\Lambda}_{0,d,d}\cap\mc{T}^{d,s}_{\alpha,\tau}}
\gtrsim \ykh{\frac{1}{{n}}}^{\frac{\beta}{{\frac{d}{s+1}+(1+\frac{1}{s+1})\cdot\beta}}}, 
\] 
is stronger than the lower bound of \eqref{241212030945}, which is given by
\[&\inf_{\hat{f}_n}\sup\setl{{\bm E}_{P^{\otimes n}}\zkh{\mc{E}_P(\hat{f}_n)}\vphantom{\Bigg|}}{P\in\mc{H}^{d,\beta,r,\Lambda}_{0,d,d}\cap\overline{\mc{T}}^{d,s}_{\alpha,\tau}}
\gtrsim  \ykh{\frac{1}{{n}}}^{\frac{\beta}{{\frac{d}{s+1}+(1+\frac{1}{s+1})\cdot\beta}}}. \] 
Conversely, the upper rate part of \eqref{241212030944} 
\beq\label{24121220151}
&\inf_{\hat{f}_n}\sup\setl{{\bm E}_{P^{\otimes n}}\zkh{\mc{E}_P(\hat{f}_n)}\vphantom{\Bigg|}}{P\in\mc{H}^{d,\beta,r,\Lambda}_{0,d,d}\cap\mc{T}^{d,s}_{\alpha,\tau}}
\lesssim  \ykh{\frac{1}{{n}}}^{\frac{\beta}{{\frac{d}{s+1}+(1+\frac{1}{s+1})\cdot\beta}}} 
\eeq 
is weaker than that of \eqref{241212030945}, which is given by  
\beq \label{24121220152} &\inf_{\hat{f}_n}\sup\setl{{\bm E}_{P^{\otimes n}}\zkh{\mc{E}_P(\hat{f}_n)}\vphantom{\Bigg|}}{P\in\mc{H}^{d,\beta,r,\Lambda}_{0,d,d}\cap\overline{\mc{T}}^{d,s}_{\alpha,\tau}}
\lesssim \ykh{\frac{1}{{n}}}^{\frac{\beta}{{\frac{d}{s+1}+(1+\frac{1}{s+1})\cdot\beta}}}. 
\eeq
Unfortunately, our proof of  \eqref{24121220151} cannot be directly used to derive \eqref{24121220152}. This is  because \eqref{24121220151} is proved by applying our Theorem \ref{231123013855}, which relies on the inequality \eqref{23080803} in the proof of Lemma \ref{23081001} in Appendix  \ref{20250918073254}. This inequality takes the form of the condition \eqref{231123022427} of Theorem \ref{231123013855} and requires $P\in \mc{T}^{d,s}_{\alpha,\tau}$. Under the weaker condition $P\in \overline{\mc{T}}^{d,s}_{\alpha,\tau}$, the arguments which originally lead  to  \eqref{23080803} become invalid. Naturally, one may consider adopting the arguments in the proof of  \eqref{24121220152} in \cite{audibert2007fast} (see Section 6.3 therein) to obtain a result that generalizes \eqref{24121220152} to our compositional assumption model. Such a reslut is analogous to the upper rate part of our result \eqref{25eq240118175544}. To be specific, we expect to show 
\[
&\inf_{\hat{f}_n}\sup\setl{{\bm E}_{P^{\otimes n}}\zkh{\mc{E}_P(\hat{f}_n)}\vphantom{\Bigg|}}{P\in\mc{H}^{d,\beta,r,\Lambda}_{q,K,d_*}\cap\overline{\mc{T}}^{d,s}_{\alpha,\tau}}
\\
&\leq \inf_{\hat{f}_n}\sup\setl{{\bm E}_{P^{\otimes n}}\zkh{\mc{E}_P(\hat{f}_n)}\vphantom{\Bigg|}}{P\in\mc{H}^{d,\beta,r}_{q,K,d_*,d_\star}\cap\overline{\mc{T}}^{d,s}_{\alpha,\tau}}
 \lesssim
  \ykh{\frac{1}{{n}}}^{\frac{\beta\cdot(1\qx\beta)^q}{{\frac{d_*}{s+1}+(1+\frac{1}{s+1})\cdot\beta\cdot(1\qx\beta)^q}}}. 
\]
However, we are unable to prove the inequalites above in this paper, as we believe that certain details in the proof of  \eqref{24121220152} in \cite{audibert2007fast} require further clarification. Specifically, in page 25 of \cite{audibert2007fast},  the authors claim that 
\beq\label{241213073501}
&\mb{P}\ykh{\min_{\bar{\eta}\in\mc{N}_n^*}\zkh{R_n(f_{\bar{\eta}})-R_n(f_{{\eta_n}})}\leq 0}
=
\mb{P}\ykh{\min_{\bar{\eta}\in\mc{N}_n^*}\zkh{Z_n(f_{\bar{\eta}})-Z_n(f_{{\eta}_n})+d(f_{\bar{\eta}})-d(f_{{\eta}_n})}\leq 0}
\\&\xlongequal{\text{by definition}}
\mb{P}\ykh{\min_{\bar{\eta}\in\mc{N}_n^*}\zkh{R_n(f_{\bar{\eta}})-R_n(f_{{\eta_n}})+\ykh{R_n(f^*_{\eta_n})-R_n(f^*_{\bar{\eta}})}}\leq 0}, 
\eeq
 which constitutes a key step in the proof of  \eqref{24121220152} in \cite{audibert2007fast}. Here, the notations $R_n(\cdot)$, $Z_n(\cdot)$, $d(\cdot)$, $f_{\bar{\eta}}$, $f_{\eta_n}$, $f_{\bar{\eta}}^*$, $f_{\eta_n}^*$, and $\mc{N}_n^*$ follow the defnitions in that paper. It is not clear why removing the term $R_n(f^*_{\eta_n})-R_n(f^*_{\bar{\eta}})$  from the right hand side of \eqref{241213073501} does not affect the value of that probability, since it is possible that $\mb{P}\ykh{{R_n(f^*_{\eta_n})-R_n(f^*_{\bar{\eta}})}>0}$ is positive.  Indeed, the term $R_n(f^*_{\eta_n})-R_n(f^*_{\bar{\eta}})$ can be expressed as
 \beq\label{241213082944}
 \frac{1}{n}\sum_{i=1}^n\ykh{\idf_{\hkh{\mr{sgn}(2\eta_n(X_i)-1)\neq Y_i}}-\idf_{\hkh{\mr{sgn}({2\bar{\eta}(X_i)}-1)\neq Y_i}}}\cdot\idf_{\hkh{\eta_P(X_i)=1/2}}. 
 \eeq 
However, even under the noise condition \eqref{241210043327}, the probability that ${{\eta_P(X_i)=1/2}}$ may be positive, which implies that the quantity in \eqref{241213082944} may also be positive with positive probability.

In \cite{audibert2007fast}, the authors also consider the so-called strong density assumption on the data distribution $P$ (see Definition 2.2 therein), which requires that the density function of the marginal distribution $P_X$ on $[0,1]^d$ (with respect to the Lebesgue measure) is both bounded from above and bounded away from zero, i.e., 
\beq\label{241213121730}
0<\lambda\leq\frac{\mr{d}P_X(x)}{\mr{d}x}\leq\Lambda,\;\forall\;\text{a.s.}\;x\in[0,1]^d
\eeq 
for some constant $0<\lambda<\Lambda<\infty$. This is a stronger assumption than the mild density assumption adopted in the model $\mc{H}^{d,\beta,r,\Lambda}_{q,K,d_*}$, where the density function $\frac{\mr{d}P_X(x)}{\mr{d}x}$ is only required to be bounded from above by $\Lambda$. In \cite{audibert2007fast}, the authors show that a faster convergence rate than ${\ykh{\frac{1}{{n}}}^{\frac{\beta}{{\frac{d}{s+1}+(1+\frac{1}{s+1})\cdot\beta}}}}$, as established in  \eqref{241212030945}, can be achieved when the mild density assumption is replaced by the strong density assumption. Specifically, in Theorem 3.3 of \cite{audibert2007fast}, they prove that 
\beq\label{241213185805}
\inf_{\hat{f}_n}\sup\setl{{\bm E}_{P^{\otimes n}}\zkh{\mc{E}_P(\hat{f}_n)}\vphantom{\Bigg|}}{P\in{\underline{\mc{H}}^{d,\beta,r,\Lambda,\lambda}_{0,d,d}}\cap\overline{\mc{T}}^{d,s}_{\alpha,\tau}}
 \lesssim
  \ykh{\frac{1}{{n}}}^{\frac{\beta\cdot (1+s)}{2\beta+d}}, 
\eeq 
where 
$\underline{\mc{H}}^{d,\beta,r,\Lambda,\lambda}_{q,K,d_*}:=\set{P\in{\mc{H}}^{d,\beta,r,\Lambda}_{q,K,d_*}}{\text{$P$ satisfies \eqref{241213121730}}}$. 
The rate ${n^{-\frac{\beta\cdot (1+s)}{2\beta+d}}}$ in \eqref{241213185805} is achieved by the plug-in classifier $\hat{f}^*_n=\mr{sgn}(2\hat{\eta}_n^*-1)$, where $\hat{\eta}_n^*$ is obtained by making minor modifications to the so-called local polynomial estimator $\hat{\eta}_n^{\text{LP}}$ of the CCP function $\eta_P$, as defined in Definition 2.3 of \cite{audibert2007fast}.  The estimator $\hat{\eta}_n^{\text{LP}}$ is defined pointwise: for each $x\in[0,1]^d$, the value $\hat{\eta}_n^{\text{LP}}(x)$ is estimated from the data points $(X_i,Y_i)$ such that $X_i$ is close to $x$. Under the mild density assumption
\beq\label{241213214730}
\frac{\mr{d}P_X(x)}{\mr{d}x}\leq \Lambda,\;\forall\;\text{a.s. } x\in[0,1]^d
\eeq 
the density function $\frac{\mr{d}P_X(x)}{\mr{d}x}$ is not required to be bounded away from zero. As a result, there may be many points $x\in[0,1]^d$ at which the value of the density function  $\frac{\mr{d}P_X(x)}{\mr{d}x}$ is small,  meaning  that there are few data points $(X_i,Y_i)$ with $X_i$ near $x$, which leads to inaccurate estimation of $\eta_P(x)$ by $\hat{\eta}^{\text{LP}}_n(x)$. This explains why the convergence rate ${\ykh{\frac{1}{{n}}}^{\frac{\beta}{{\frac{d}{s+1}+(1+\frac{1}{s+1})\cdot\beta}}}}$, derived in \eqref{241212030944} under the mild density assumption \eqref{241213214730}, is slower than the rate ${n^{-\frac{\beta\cdot (1+s)}{2\beta+d}}}$ derived in \eqref{241213185805} under the strong  density assumption \eqref{241213121730}. In contrast, the estimator ${\hat f_n^{\lozenge}}$ that we construct to derive the optimal convergence rate ${\ykh{\frac{1}{{n}}}^{\frac{\beta\cdot(1\qx\beta)^q}{{\frac{d_*}{s+1}+(1+\frac{1}{s+1})\cdot\beta\cdot(1\qx\beta)^q}}}}$ in Theorem \ref{20231204212113} is not based on pointwise estimation, but rather from the empirical hinge risk minimization,  which determines the estimator ${\hat f_n^{\lozenge}}$ in a global manner. Indeed, in the proof of Theorem \ref{20231204212113}, we do not utilize any information about the marginal distribution $P_X$. Consequently, the approach we use to prove Theorem \ref{20231204212113} cannot be directly employed to derive improved convergence rates like that in \eqref{241213185805}, even under the strong density assumption \eqref{241213121730}. Nevertheless, one may expect that some estimator similar to $\hat{f}^*_n=\mr{sgn}(2\hat{\eta}_n^*-1)$, as defined in \cite{audibert2007fast}, could achieve a convergence rate like that in \eqref{241213185805}, and better than the rate ${\ykh{\frac{1}{{n}}}^{\frac{\beta\cdot(1\qx\beta)^q}{{\frac{d_*}{s+1}+(1+\frac{1}{s+1})\cdot\beta\cdot(1\qx\beta)^q}}}}$ in \eqref{25eq240118175544}, provided that the strong density assumption \eqref{241213121730} is additionally imposed. Determining the precise minimax rate in this case, namely 
\[
\inf_{\hat{f}_n}\sup\setl{{\bm E}_{P^{\otimes n}}\zkh{\mc{E}_P(\hat{f}_n)}\vphantom{\Bigg|}}{P\in{\underline{\mc{H}}^{d,\beta,r,\Lambda,\lambda}_{q,K,d_*}}\cap{\mc{T}}^{d,s}_{\alpha,\tau}} 
\] 
or 
\[
\inf_{\hat{f}_n}\sup\setl{{\bm E}_{P^{\otimes n}}\zkh{\mc{E}_P(\hat{f}_n)}\vphantom{\Bigg|}}{P\in{\underline{\mc{H}}^{d,\beta,r,\Lambda,\lambda}_{q,K,d_*}}\cap\overline{\mc{T}}^{d,s}_{\alpha,\tau}}, 
\] merits further study.

The condition $r\geq \cliu$ in Theorem \ref{23102601} is a technical requirement. The proof of Theorem \ref{23102601} is based on an argument akin to Fano's method or Le Cam's method (see Lemma \ref{23102003}, Lemma \ref{23091519}, Lemma \ref{23102201} in Appendix    \ref{20251011004523}   and the comments therein), where we need to construct several probabilities $P_0,\ldots P_{\texttt M}$ in  $\mc{H}^{d,\beta,r}_{q,K,d_*,d_\star}\cap\mc{T}^{d,s}_{\alpha,\tau}$ such that they differ from each other to a sufficient extent. This ensures that the desired minimax lower bound holds over the set $\hkh{P_0, P_1, \ldots, P_{\texttt{M}}}$, i.e.,
\beq\label{241218185128}
&\inf_{\hat{f}_n}\sup\setl{{\bm E}_{P^{\otimes n}}\zkh{\mc{E}_P(\hat{f}_n)}\vphantom{\Bigg|}}{P\in\hkh{P_0,P_1,\ldots,P_{\texttt{M}}}}
	\geq \text{ the desired bound}, 
\eeq 
 which  immediately yields the desired inequality,   since  $\inf\limits_{\hat{f}_n}\sup\setl{{\bm E}_{P^{\otimes n}}\zkh{\mc{E}_P(\hat{f}_n)}\vphantom{\bigg|}}{P\in\mc{H}^{d,\beta,r,\Lambda}_{q,K,d_*}\cap\mc{T}^{d,s}_{\alpha,\tau}}$ is   at least  
	$\inf\limits_{\hat{f}_n}\sup\setl{{\bm E}_{P^{\otimes n}}\zkh{\mc{E}_P(\hat{f}_n)}\vphantom{\bigg|}}{P\in\hkh{P_0,P_1,\ldots,P_{\texttt{M}}}}$. 
If $r$ were too small, the constructed measures $P_0,\ldots,P_{\texttt M}$ would fall outside the space $\mc{H}^{d,\beta,r}_{q,K,d_*,d_\star}\cap\mc{T}^{d,s}_{\alpha,\tau}$, rendering the original argument to be invalid.  A smaller $r$ leads to a less expressive model $\mc{H}^{d,\beta,r}_{q,K,d_*,d_\star}\cap\mc{T}^{d,s}_{\alpha,\tau}$, which may fail to sustain the minimax lower rate ${\ykh{\frac{1}{{n}}}^{\frac{\beta\cdot(1\qx\beta)^q}{{\frac{d_*}{s+1}+(1+\frac{1}{s+1})\cdot\beta\cdot(1\qx\beta)^q}}}}$ established in Theorem \ref{23102601}. Therefore, the lower bound requirement on $r$ is quite reasonable. Of curse it is interesting to further study how to refine the original condition $r\geq \cliu$—specifically, to identify a smaller constant $c'$ such that for $r \geq c'$ (or $r > c'$), the minimax lower rate can still be achieved.  In the next Theorem \ref{23102602}, we address this problem for the special case when the noise condition is not assumed.

\begin{theorem}\label{23102602}
	Suppose  $q\in\mb N\cup\hkh{0}$,  $d\in\mb N$, $d_*\in\mb N$, $K\in\mb N$,     $\beta\in(0,\infty)$, $r\in(\frac{1}{2},\infty)$,   $\Lambda\in[1,\infty)$, $d_*\leq K$ and $d_*\leq d$.   Then there exists a   constant  $\cjiu\in(0,\infty)$ depending only on $(d_*,\beta,q,r)$  such that 
	\[
	&\inf_{\hat{f}_n}\sup\setl{{\bm E}_{P^{\otimes n}}\zkh{\mc{E}_P(\hat{f}_n)}\vphantom{\Bigg|}}{P\in\mc{H}^{d,\beta,r,\Lambda}_{q,K,d_*}}
	\geq \cjiu\cdot \ykh{\frac{1}{n}}^{\frac{\beta\cdot(1\qx\beta)^q}{d_*+2\beta\cdot(1\qx\beta)^q}}
	\] for any  $n\in\mb N$,  where the infimum is taken over all estimators $\hat f_n:[0,1]^d\to\mbR$  based upon the i.i.d. sample $\hkh{(X_i,Y_i)}_{i=1}^n$ drawn from ${[0,1]^d\times\hkh{-1,1}}$. 
\end{theorem}

The proof of Theorem \ref{23102602} is given in Appendix   \ref{20251011010150}. 

Combining Theorem \ref{20231204212113}  with $s=0, \alpha=1, \tau=1$, and Theorem \ref{23102602}, we deduce that
\beq\label{231118001813}
&
\ykh{\frac{1}{n}}^{\frac{\beta\cdot(1\qx\beta)^q}{d_*+2\beta\cdot(1\qx\beta)^q}}\gtrsim\inf_{\hat{f}_n}\sup\setl{{\bm E}_{P^{\otimes n}}\zkh{\mc{E}_P(\hat{f}_n)}\vphantom{\Bigg|}}{P\in\mc{H}^{d,\beta,r}_{q,K,d_*,d_\star}\cap \mc{T}^{d,0}_{1,1}}
\\&\xlongequal{\because\eqref{230105003839}}
\inf_{\hat{f}_n}\sup\setl{{\bm E}_{P^{\otimes n}}\zkh{\mc{E}_P(\hat{f}_n)}\vphantom{\Bigg|}}{P\in\mc{H}^{d,\beta,r}_{q,K,d_*,d_\star}}
\\&\geq\inf_{\hat{f}_n}\sup\setl{{\bm E}_{P^{\otimes n}}\zkh{\mc{E}_P(\hat{f}_n)}\vphantom{\Bigg|}}{P\in\mc{H}^{d,\beta,r,\Lambda}_{q,K,d_*}}\gtrsim \ykh{\frac{1}{n}}^{\frac{\beta\cdot(1\qx\beta)^q}{d_*+2\beta\cdot(1\qx\beta)^q}}\;\text{as }n\to\infty,
\eeq  
provided that $r\in(\frac{1}{2},\infty)$. This shows that the lower rate given by Theorem \ref{23102602} is optimal. However, Theorem \ref{23102602} fails if the condition $r\in(\frac{1}{2},\infty)$ is replaced by $r\in(0,\frac12]$. Indeed, we have  
\beq\label{231118003959}
\inf_{\hat{f}_n}\sup\setl{{\bm E}_{P^{\otimes n}}\zkh{\mc{E}_P(\hat{f}_n)}\vphantom{\Bigg|}}{P\in\mc{H}^{d,\beta,r,\Lambda}_{q,K,d_*}}=0,\;\forall\;n\in\mb N,\;\forall\;r\in(0,{\textstyle\frac{1}{2}}], 
\eeq 
of which the proof is given in Appendix  \ref{20251011004753} of this paper. Therefore, the condition $r\in(\frac{1}{2},\infty)$ in Theorem \ref{23102602} is tight and cannot be improved. 

\section{Conclusion and Future Work}\label{2412172107}

In this work, we establish an oracle inequality in Theorem \ref{231123013855}, which provides the generalization analysis framework to derive tight upper bounds for the \zaverage excess $\phi$-risk of $\phi$-ERMs. This inequality is highly general and subsumes the oracle inequality presented in Theorem 2.1 of our previous paper \cite{zhangzihan2023classification}. By applying Theorem \ref{231123013855}, we derive the optimal convergence rates for the \zaverage excess 0-1 risk of estimators under the assumption that the data distribution $P\in \mc{H}^{d,\beta,r}_{q,K,d_*,d_\star}\cap \mc{T}^{d,s}_{\alpha,\tau}$—that is, under the compositional assumption on the CCP function $\eta_P$ along with the noise condition \eqref{23080401}—as shown in Theorem \ref{20231204212113}.  Furthermore, we demonstrate that ReLU DNNs trained with hinge loss can achieve these optimal rates (up to logarithmic factors) in Theorem \ref{20231125000604} and Theorem \ref{231215231016}. The optimality of these rates is justified by establishing the corresponding minimax lower rates in Theorem \ref{23102601} and Theorem \ref{23102602}.

As mentioned in Section \ref{241217205730}, the oracle inequality in Theorem  \ref{231123013855} is very powerful, and we can consider applying it to establish excess risk bounds for estimators trained with other losses functions, particularly unbounded ones such as the LUM loss. Theorem  \ref{231123013855} remains true  when the input space $[0,1]^d$ is replaced by any  measurable space. Therefore, we can employ Theorem  \ref{231123013855}  to study binary classification  of data distributed in the whole $\mbR^d$, such as in the case of Gaussian mixture models \cite{tianyi2024classification}, or even in infinite-dimensional spaces such as $\ell^p$ spaces and more general Banach spaces.

Furthermore, we can extend our study of optimal convergence in binary classification under assumptions beyond $P \in \mc{H}^{d,\beta,r}_{q,K,d_*,d_\star} \cap \mc{T}^{d,s}_{\alpha,\tau}$. For example, as mentioned in Section \ref{2412172103}, we can investigate the asymptotic behaviors of optimal convergence if the strong density assumption \eqref{241213121730} is imposed, or if the noise condition \eqref{23080401} is replaced by the weaker one \eqref{241210043327}. In addition, we can explore optimal convergence rates in multiclass classification. In particular, we expect that our construction of $P_0, P_1, \ldots, P_{\texttt{M}}$, which satisfies \eqref{241218185128} in the proof of Theorem \ref{23102601} and Theorem \ref{23102602}, can be useful for deriving minimax lower bounds in multiclass classification.

\appendix

\renewcommand{\theequation}{\thesection.\arabic{equation}}

\renewcommand{\thefigure}{\thesection.\arabic{figure}}

\numberwithin{theorem}{section}

\numberwithin{lemma}{section} 

\numberwithin{equation}{section}

\section{Detailed Definitions of  Some  Spaces}
\hypertarget{250914042717}{} \label{250914042717}

\setcounter{figure}{0}

 In  Appendix \ref{250914042717}, we provide  detailed definitions of some spaces  mentioned in Section \ref{240119033824}  and Section \ref{2412172103}. 

\subsection{Definition of $\fdnn_d(G,N,S,B,F)$}\hypertarget{20250916114234}{}\label{20250916114234}

In this part, we define the ReLU neural network space $\fdnn_d(G,N,S,B,F)$.  Recall that the ReLU activation function is given by $\sigma:\mbR\to\mbR,\;t\mapsto\max\hkh{0,t}$. For any positive integer $m$ and $v\in\mbR^m$, define $\sigma_v:\mbR^m\to\mbR^m,\;x\mapsto (\sigma((x-v)_1),\sigma((x-v)_2),\ldots,\sigma((x-v)_m))^\top$. We say that a function $f:\mbR^d\to\mbR$ is a ReLU neural network if there exist $L\in\mb N\cup\hkh{0}$, $\mr m_i\in \mb N$ ($0\leq i\leq L+1$) with $\mr{m}_0=d$ and $\mr m_{L+1}=1$, $v_i\in\mbR^{\mr m_i}$ ($1\leq i\leq L$) and $\bm{W}_i\in\mbR^{\mr{m}_{i+1}\times\mr{m}_i}$ ($1\leq i\leq L$), such that \beq\label{231229041606}
f(x)=\bm W_L\sigma_{v_L}\bm W_{L-1}\sigma_{v_{L-1}}\cdots\bm W_2\sigma_{v_2}\bm W_1\sigma_{v_1}\bm W_0x,\;\forall\;x\in\mbR^d. 
\eeq Then for  $(G,N,S,B,F)\in (0,\infty)\times[1,\infty)\times [0,\infty]\times[0,\infty]\times[0,\infty]$, we define the space of ReLU neural networks as
\[
&\fdnn_d(G,N,S,B,F)\\&:=\set{f:\mbR^d\to\mbR}{\begin{minipage}{297.84pt}there exist  $L\in \mb N\cup\hkh{0}$, $\mr{m}_i\in\mb N$, $\bm W_i\in\mbR^{\mr{m}_{i+1}\times\mr{m}_i}$, $v_i\in\mbR^{\mr m_i}$ ($i\in\mb N\cup\hkh{0}$)  such that $L\leq G$,  $\sup\set{\mr{m}_i}{i\in\mb N}\leq N$,  $\sum\limits_{i=0}^L\Bigabs{\norm{\bm W_i}_0+\norm{v_i}_0}\leq S$, $\sup\limits_{i\in\mb N}{{\norm{v_i}_\infty}}\leq B$, $\sup\limits_{{i\in\mb N\cup\hkh{0}}}{{\norm{\bm W_i}_\infty}}\leq B$,  $\sup\limits_{x\in[0,1]^d}\abs{f(x)}\leq F$, $\mr{m}_0=d$, $\mr{m}_{L+1}=1$, and  \eqref{231229041606} holds \end{minipage}}. 
\] 

\subsection{Definition of $\mc G_d^{\mathbf{CHOM}}(q, K,d_\star, d_*, \beta,r)$}
\hypertarget{20250916114433}{}\label{20250916114433}
In this part, we define the compositional H\"older smooth function space $\mc G_d^{\mathbf{CHOM}}(q, K,d_\star, d_*, \beta,r)$. To this end, we must first clarify the concept of H\"older smoothness. 
Let $\beta\in(0,\infty)$, $m\in\mb{N}$, and $\Omega\subset \mbR^m$. For any real number $b$, define   
$\mb N^{m,b}_{0}:=\setm{2}{x\in \mbR^m}{\norm{x}_1<b\text{ and }(x)_i\in\mb{N}\cup\hkh{0}\text{ for all $i\in\mb{Z}\cap (0,m]$}}$. 
Roughly speaking, for a $\flr{\beta}$-times continuously differentiable function $f:\Omega\to\mbR$, the H\"older-$\beta$ norm of $f$ on $\Omega$ is defined as \beq\label{20240823014747}
\norm{f}_{\mathbf H^{\beta}(\Omega)}:=\max_{\bm u\in\mb N_0^{m,\beta}}\|\mr D^{\bm u} f\|_{{\Omega}}+\max_{\bm u\in \mb N_0^{m,\beta}\setminus \mb N_0^{m,\beta-1}}\abs{\sup_{x\neq z\in\Omega}\frac{\abs{\mr D^{\bm u} f(x)-\mr D^{\bm u}f(z)}}{\norm{x-z}_2^{\beta+1-\ceil{\beta}}}}, 
\eeq 
where we use the symbol $\mr{D}^{\bm u}$  to denote  the partial derivative operator corresponding to the multi-index $\bm u$, that is, ${{\mr D^{\bm u}f(x):=\frac{\partial ^{\norm{\bm u}_1}f(x_1,x_2,\ldots x_m)}{\partial x_1^{(\bm u)_1}\partial x_2^{(\bm u)_2}\cdots \partial x_m^{(\bm u)_m}}}}$. 
However, since the partial derivatives of $f$ at boundary points of $\Omega$ are, in general, not well-defined, a more careful treatment is required. For any set $E\subset\mbR^m$, we use $E^\circ$ to denote the interior of $E$, defined as
\[E^\circ:=\setl{x\in\mbR^m}{\mathscr{C}(x,r)\subset E\text{ for some $r\in(0,1)$}}, \] where
\beq\label{20250918033935}
\mathscr C(x,r):=\setm{2}{y\in\mb R^m}{\norm{y-x}_\infty\leq r}  \eeq is the  infinity-norm ball in $\mbR^m$. 
We say that a set $E\subset\mbR^m$ is a closed domain in $\mbR^m$ if $E$ is the closure of $E^\circ$.  For any $k\in\mb N\cup\hkh{0}$ and any nonempty open set $U$ in $\mbR^m$, we use $\mc C^{k}(U)$ to denote the set of all  $k$-times continuously differentiable functions from $U$ to $\mbR$, i.e., 
 \[
 \mc{C}^k(U):=\hkh{f:U\to\mbR\Big|\text{$f$ is $k$-times continuously differentiable on $U$}},
 \] and define  \[\overline{\mc C}^{k}(U):=\setr{f\in\mc C^k(U)}{\mr{D}^\alpha f\text{ is uniformly continuous on $U$, $\forall\;\alpha\in\mb N_0^{m,k+1}$}}. \]  Then, for any nonempty bounded closed domain $E$ in $\mbR^m$, we define 
 \[
 \mc{C}^k(E):=\setr{f:E\to\mbR}{\text{$f$ is continuous on $E$, and }f|_{E^\circ}\in\overline{\mc{C}}^k(E^\circ)}. 
 \] Suppose $\Omega$ is a nonempty open set or  a nonempty bounded  closed domain  in $\mbR^m$.  Then, for any  $\beta\in(0,\infty)$ and any   $f\in\mc C^{\ceil{\beta-1}}(\Omega)$, we define the H\"older-$\beta$ norm of $f$ on $\Omega$  as 
\[
\norm{f}_{\mathbf H^{\beta}(\Omega)}:=\max_{\bm u\in\mb N_0^{m,\beta}}\|\mr D^{\bm u} f\|_{{\Omega^\circ}}+\max_{\bm u\in \mb N_0^{m,\beta}\setminus \mb N_0^{m,\beta-1}}\abs{\sup_{ x\neq z\in\Omega^\circ}\frac{\abs{\mr D^{\bm u} f(x)-\mr D^{\bm u}f(z)}}{\norm{x-z}_2^{\beta+1-\ceil{\beta}}}}.
\] Note that if we define 
\beq\mr{D}^{\bm u} f(x):=\lim_{\substack{y\to x\\y\in\Omega^\circ}}\mr{D}^{\bm u} f(y), \text{ for $x\in\Omega\setminus\Omega^\circ$,}\eeq 
where the limit exists because in this case  $\mr{D}^{\bm u}f$ is uniformly continuous on $\Omega^{\circ}$, then \eqref{20240823014747} follows.  Thus, the definition of the H\"older-$\beta$ norm in this paper is exactly the same as that in \cite{zhangzihan2023classification} (cf. (2.12) therein).  For $0\leq r\leq\infty$, we define the H\"older ball 
\[
\mc B^\beta_r(\Omega):=\setl{f\in\mc C^{\ceil{\beta-1}}(\Omega)}{\norm{f}_{\mathbf H^\beta(\Omega)}\leq r}.  
\] For any real-valued function $h$ with $\Omega {\subsetneqq} \mathbf{dom}(h)$ and $h|_\Omega\in \mc C^{\ceil{\beta-1}}(\Omega)$, we write $\norm{h}_{\mathbf H^{\beta}(\Omega)}$ for $\norm{h|_\Omega}_{\mathbf H^{\beta}(\Omega)}$.  If $\norm{h}_{\mathbf{H}^\beta(\Omega)}<\infty$, then we say that the function $h$ is H\"older-$\beta$ smooth on $\Omega$. Moreover, we say that a function $f$ is H\"older-$\beta$ smooth if $f$ is H\"older-$\beta$ smooth on $\mathbf{dom}(f)$. 

With the precise definition of H\"older smoothness  in hand,  we are now ready to define the space  of compositional H\"older smooth functions. 
 For any $(d,d_*,\beta,r)\in\mb N\times\mb N\times(0,\infty)\times(0,\infty)$, define 
\beq\label{231231073133}
&\mc G_d^{\mathbf{H}}(d_*, \beta,r):=\setr{f:[0,1]^d\to\mbR}{\begin{minipage}{211.6pt}
		$\exists\;I\subset \zkh{0,d}\cap\mb N$ and $g\in \mc{B}^{\beta}_r\ykh{[0,1]^{d_*}}$ such that  $\abs{I}=d_*$ and $f(x)=g\ykh{(x)_I}$ for $x\in[0,1]^d$
\end{minipage}}, 
\eeq which represents the set of all functions $f:[0,1]^d\to\mbR$ that are essentially H\"older-$\beta$ smooth functions of $d_*$ variables. Note that $\mc G_d^{\mathbf{H}}(d_*, \beta,r)=\varnothing$ if $d<d_*$. For any $(d,d_\star)\in\mb N\times\mb N$, define 
\[
\mc G_d^{\mathbf{M}}(d_\star):=\setr{f:[0,1]^d\to\mbR}{\begin{minipage}{205.44pt}
		$\exists\; I\subset \hkh{1,2,\ldots,d}$ such that $1\leq \#(I)\leq d_\star$ and $ f(x)=\max\setr{(x)_i}{i\in I} \mbox{ for }x\in[0,1]^d$ 
\end{minipage}},
\] which represents the set of all functions $f:[0,1]^d\to\mbR$ that compute the maximum value over a subset of at most $d_\star$ components $(x)_i$ of the input vector $x$. Then, for any $(d,q,K,d_\star,d_*,\beta,r)$ in $\mb N\times(\mb N\cup\hkh{0})\times\mb N\times\mb N\times\mb N\times(0,\infty)\times(0,\infty)$ that satisfies  \eqref{231231083620}, 
 we define 
\label{241107014836} \[
&\mc G_d^{\mathbf{CHOM}}(q, K,d_\star, d_*, \beta,r)\\&:=\setr{h_q\circ\cdots \circ h_1\circ h_0}{\begin{minipage}{389.28pt}
		$h_0,h_1,\ldots, h_{q-1}, h_q$ are functions satisfying the following conditions: 
		\begin{enumerate}[label={{ {\hfil(\roman*)\hfil}} }]
			\item   ${\mathbf{{dom}}}(h_i)=[0,1]^K$ for $0<i\leq q$ and ${\mathbf{{dom}}}(h_0)=[0,1]^d$;
			\item $\mathbf{ran}(h_i)\subset[0,1]^K$ for $0 \leq i<q$ and $\mathbf{ran}(h_q)\subset\mbR$;
			\item For $1\leq j\leq K+(1-K)\cdot\idf_{\hkh{q}}(0)$, the $j$-th component function of $h_0$ given by $\mathbf{dom}(h_0)=[0,1]^d\ni x\mapsto (h_0(x))_j\in\mbR$ belongs to $\mc G_d^{\mathbf{H}}(d_*, \beta,r)\cup \mc G_d^{\mathbf{M}}(d_\star)$;
			\item For $0 < i\leq q$ and $1\leq j\leq K+(1-K)\cdot\idf_{\hkh{q}}(i)$, the $j$-th component function  of $h_i$ given by  $\mathbf{dom}(h_i)=[0,1]^K\ni x\mapsto (h_i(x))_j\in\mbR$ belongs to $\mc G_K^{\mathbf{H}}(d_*, \beta,r)\cup \mc G_K^{\mathbf{M}}(d_\star)$.
\end{enumerate}\end{minipage}}.
\] The definitions of $\mc G_d^{\mathbf{M}}(d_\star)$, $\mc G_d^{\mathbf{H}}(d_*, \beta,r)$ and  $\mc G_d^{\mathbf{CHOM}}(q, K,d_\star, d_*, \beta,r)$ coincide with those in \cite{zhangzihan2023classification} (cf. (2.27), (2.28) and (2.32) therein). Note that \eqref{231231083620} is a very natural requirement: it simply ensures that the essential input dimension $d_*$ of the H\"older-$\beta$ smooth component function $(h_i(x))_j$ does not exceed its (apparent) input dimension, namely, $d$ (if $i=0$) or $K$ (if $0<i\leq q$). Otherwise (i.e., if $d_*>d$ or $d_*>K$), the spaces $\mc G_d^{\mathbf{H}}(d_*, \beta,r)$ or $\mc G_K^{\mathbf{H}}(d_*, \beta,r)$, which appear in the definition above, would be empty, spoiling the richness of $\mc G_d^{\mathbf{CHOM}}(q, K,d_\star, d_*, \beta,r)$. In particular, if $d_*>d\qd K$, then  $\mc G_d^{\mathbf{CHOM}}(q, K,d_\star, d_*, \beta,r)$, as defined above, collapses into a trivial space containing only maximum value functions. Therefore, to maintain the richness of $\mc G_d^{\mathbf{CHOM}}(q, K,d_\star, d_*, \beta,r)$  and avoid the trivial case,  we require  the condition \eqref{231231083620}. We also point out that requiring the domain of $h_i$ to be $[0,1]^K$ in the definition of $\mc G_d^{\mathbf{CHOM}}(q, K,d_\star, d_*, \beta,r)$—rather than an arbitrary cube $[a,b]^K$—entails no loss of generality: by scaling, any compositional function $\tilde h_q\circ\tilde h_{q-1}\circ\cdots\circ\tilde h_1\circ\tilde h_0$ with $\mathbf{dom}(\tilde h_i)=[a,b]^K$ ($0<i\leq q$) can be rewritten as $h_q\circ\cdots \circ h_1\circ h_0$ with $\mathbf{dom}(h_i)=[0,1]^K$ ($0<i\leq q$). To see this, note that $\tilde h_q\circ \tilde h_{q-1}\circ \tilde h_{q-2}\circ\cdots\circ\tilde h_2\circ\tilde h_1\circ\tilde h_0$ can also be expressed as  \[(\tilde h_q\circ A)\circ (A^{-1}\circ \tilde h_{q-1}\circ A)\circ (A^{-1}\circ\tilde h_{q-2}\circ A)\circ\cdots\circ (A^{-1}\circ \tilde h_2\circ A)\circ (A^{-1}\circ \tilde h_1\circ A)\circ  (A^{-1}\circ \tilde h_0),
\] where $A:[0,1]^K\to[a,b]^K,\;\;x\mapsto \bigykh{(b-a)\cdot (x)_1+a,(b-a)\cdot (x)_2+a,\cdots,(b-a)\cdot (x)_K+a}^\top$.

\subsection{Definition of  $\mc{H}^{d,\beta,r,\Lambda}_{q,K,d_*}$}\hypertarget{20250915195346}{}  \label{20250915195346}

In this part, we provide the explicit definition of the space  $\mc{H}^{d,\beta,r,\Lambda}_{q,K,d_*}$.

For any  $t\in[0,1]$, let $\mathscr{M}_{t}$ be the probability measure on $\hkh{-1,1}$ satisfying $\mathscr{M}_{t}(\hkh{1})=a$. For any Borel measurable  function  $\eta:[0,1]^d\to[0,1]$ and any Borel probability measure $\mathscr Q$  on $[0,1]^d$, define 
\begin{align*}
	&P_{\eta,\mathscr Q}:\hkh{\text{all Borel subsets of $[0,1]^d\times\hkh{-1,1}$}}\to [0,1],\\
	&\;\;\;\;\;\;S\mapsto\int_{[0,1]^d}\int_{\hkh{-1,1}}\idf_{S}(x,y)\mr{d}\mathscr M_{\eta(x)}(y)\mr{d}\mathscr Q(x). 
\end{align*}  
For any $\Lambda \in [1,\infty)$ and $d \in \mb N$, we use $\mc{M}_{\Lambda,d}$ to denote the set of all Borel probability measures on $[0,1]^d$ that admit a $[0,\Lambda]$-valued probability density function with respect to the Lebesgue measure, that is,
\beq
\mc M_{\Lambda,d}:=\setr{\mathscr Q}{\begin{minipage}{275.72pt}$\mathscr Q$ is a Borel probability measure on $[0,1]^d$, and there exists a Borel measurable function $f:[0,1]^d\to[0,\Lambda]$ such that $\mathscr Q(A)=\int_{[0,1]^d}f(x)\cdot\idf_A(x)\mr{d}x$ for any Borel set $A\subset[0,1]^d$ \end{minipage}}.
\eeq 
Obviously, $\mc M_{1,d}$ is the trivial singleton set whose unique element is the Lebesgue measure on $[0,1]^d$. 
For any  $(d,q,K,d_*,\beta,r)$ in $\mb N\times(\mb N\cup\hkh{0})\times\mb N\times\mb N\times(0,\infty)\times(0,\infty)$ with \eqref{231231083620}, define
\beq \label{241129192804}
&\mc G_d^{\mathbf{CH}}(q, K, d_*, \beta,r)\\&:=\setr{h_q\circ\cdots \circ h_1\circ h_0}{\begin{minipage}{293.768pt}
		$h_0,h_1,\ldots, h_q$ are functions satisfying the following conditions: 
		\begin{enumerate}[label={{ {\hfil(\roman*)\hfil}} }]
			\item   ${\mathbf{{dom}}}(h_i)=[0,1]^K$ for $0<i\leq q$,  and ${\mathbf{{dom}}}(h_0)=[0,1]^d$;
			\item $\mathbf{ran}(h_i)\subset[0,1]^K$ for $0 \leq i<q$, and $\mathbf{ran}(h_q)\subset\mbR$;
			\item For $1\leq j\leq K+(1-K)\cdot\idf_{\hkh{q}}(0)$, the $j$-th component function of $h_0$ given by $\mathbf{dom}(h_0)=[0,1]^d\ni x\mapsto (h_0(x))_j\in\mbR$ belongs to $h_q\in \mc G_d^{\mathbf{H}}(d_*, \beta,r)$;
			\item  For $0 < i\leq q$ and $1\leq j\leq K+(1-K)\cdot\idf_{\hkh{q}}(i)$, the $j$-th component function  of $h_i$ given by  $\mathbf{dom}(h_i)=[0,1]^K\ni x\mapsto (h_i(x))_j\in\mbR$ belongs to $\mc G_K^{\mathbf{H}}(d_*, \beta,r)$.
\end{enumerate}\end{minipage}},  
\eeq which consists of compositional functions of the form $h_q\circ\cdots\circ h_0$, where each component function $x \mapsto (h_i(x))_j$ is essentially a H\"older-$\beta$ smooth function of $d_*$ variables. This definition of $\mc G_d^{\mathbf{CH}}(q, K, d_*, \beta,r)$ coincides with that given in (2.31) of \cite{zhangzihan2023classification}. Note that $\mc G_d^{\mathbf{CH}}(q, K, d_*, \beta,r)$ is slightly smaller than the space $\mc G_d^{\mathbf{CHOM}}(q, K,d_\star, d_*, \beta,r)$. In contrast to $\mc G_d^{\mathbf{CHOM}}(q, K,d_\star, d_*, \beta,r)$, the component function $x \mapsto (h_i(x))_j$ is not permitted to be a maximum value function. 
Finally, we define 
\[
\mc{H}^{d,\beta,r,\Lambda}_{q,K,d_*}:=\setr{P_{\eta,\mathscr Q}}{\begin{minipage}{224.4pt}  $\eta\in\mc G_d^{\mathbf{CH}}(q, K, d_*, \beta,r)$, $\mathbf{ran}(\eta)\subset[0,1]$, and $\mathscr Q\in\mc M_{\Lambda,d}$ \end{minipage}}. 
\]  
Therefore, $\mc{H}^{d,\beta,r,\Lambda}_{q,K,d_*}$ consists of all probability measures $P$ on $[0,1]^d\times\hkh{-1,1}$ such that the conditional class probability function $P(\hkh{1}|\cdot)$ belongs to $\mc G_d^{\mathbf{CH}}(q, K, d_*, \beta,r)$ and the marginal distribution $P_X$ on $[0,1]^d$ (i.e., the input distribution) has a  density function (with respect to the Lebesgue measure) bounded from above by $\Lambda$. Probabilities $P$ with $P_X$ admiting uniformly bounded density functions  $\frac{\mr{d}P_X(x)}{\mr{d}x}$ are said to satisfy the mild density assumption given in \cite{audibert2007fast} (cf. Definition 2.1 therein).

Obviously, for any $d_\star\in\mb N$, we have 
  $\mc{H}^{d,\beta,r,\Lambda}_{q,K,d_*}\subset \mc{H}^{d,\beta,r}_{q,K,d_*,d_\star}$. 
Since the marginal distributions $P_X$ of distributions $P$ in $\mc{H}^{d,\beta,r}_{q,K,d_*,d_\star}$ are not subject to any restrictions, the space $\mc{H}^{d,\beta,r,\Lambda}_{q,K,d_*}$ is actually much smaller than $\mc{H}^{d,\beta,r}_{q,K,d_*,d_\star}$. This implies that the assumption $P \in \mc{H}^{d,\beta,r,\Lambda}_{q,K,d_*} \cap \mc{T}^{d,s}_{\alpha,\tau}$ is significantly more restrictive than the assumption $P \in \mc{H}^{d,\beta,r}_{q,K,d_*,d_\star} \cap \mc{T}^{d,s}_{\alpha,\tau}$ made in Theorem \ref{20231125000604}, Theorem \ref{231215231016}, and Theorem \ref{20231204212113}. Nevertheless, as shown in Theorem \ref{23102601}, this difference does not affect the minimax lower rates.

\section{Proofs of Main Results}\hypertarget{2412172110}{}\label{2412172110}
\setcounter{figure}{0}

  In  Appendix \ref{2412172110}, we provide proofs of main results of this paper. Specifically, the proof of Theorem \ref{231123013855} is given  in Appendix   \ref{20251011003902}; the proof of Theorem \ref{20231125000604} is given in Appendix   \ref{20250918073254}; the proof of Theorem \ref{231215231016} is given in Appendix   \ref{20250918113923}; the proof of Theorem \ref{20231204212113} is given in Appendix    \ref{20251011004301}; the proof of Theorem \ref{23102601} is given in Appendix   \ref{20251011004523}; the proof of Theorem \ref{23102602} is given in Appendix    \ref{20251011010150}; and the proof of \eqref{231118003959} is given in Appendix    \ref{20251011004753}.

\subsection{Proof of Theorem \ref{231123013855}}\hypertarget{20251011003902}{} \label{20251011003902}

\begin{lemma}\label{20231124104210}Let $A\in[3,\infty)$, $a\in(0,\infty)$, and $b\in[1,2]$. Then 
\[
\int_0^\infty\min\big\{A\cdot \exp\ykh{-a\cdot t^b},1\big\}\mr{d}t\leq 2\cdot \abs{\frac{\log A}{a}}^{\frac{1}{b}}.
\]\end{lemma}
\begin{proof} Take $u=\abs{\frac{\log A}{a}}^{\frac{1}{b}}$. Then we have that \[
&\int_0^\infty\min\big\{A\cdot \exp\ykh{-a\cdot t^b},1\big\}\mr{d}t\leq \int_0^u1\mr{d}t+\int_u^\infty A\cdot \exp(-a\cdot t^b)\mr{d}t\\
&=u+\int_{a\cdot u^b}^\infty A\cdot \exp(-z)\cdot \abs{\frac{1}{a}}^{\frac{1}{b}}\cdot \frac{1}{b}\cdot z^{\frac{1}{b}-1}\mr{d}z
=u+A\cdot\abs{\frac{1}{a}}^{\frac{1}{b}}\cdot \frac{1}{b}\cdot\int_{\log A}^\infty  \exp(-z)\cdot  z^{\frac{1}{b}-1}\mr{d}z
\\&\leq
u+A\cdot\abs{\frac{1}{a}}^{\frac{1}{b}}\cdot \frac{1}{b}\cdot\int_{\log A}^\infty  \exp(-z)\cdot  \Bigabs{\log A}^{\frac{1}{b}-1}\mr{d}z
=
u+\abs{\frac{1}{a}}^{\frac{1}{b}}\cdot \frac{1}{b}\cdot \Bigabs{\log A}^{\frac{1}{b}-1}
\\&=
\abs{\frac{\log A}{a}}^{\frac{1}{b}}+\abs{\frac{1}{a}}^{\frac{1}{b}}\cdot \frac{1}{b}\cdot \Bigabs{\log A}^{\frac{1}{b}-1}
\leq
\abs{\frac{\log A}{a}}^{\frac{1}{b}}+\abs{\frac{1}{a}}^{\frac{1}{b}}\cdot \Bigabs{\log A}^{\frac{1}{b}}=2\cdot \abs{\frac{\log A}{a}}^{\frac{1}{b}}.
\] This proves the desired result. \end{proof}

\begin{proof}[Proof of Theorem \ref{231123013855}] 
It follows from \eqref{231208074141} that
\[
&\mbe\zkh{\mc{R}^\phi_P\big(\bm{T}_F\circ \hat{f}_n^\dagger\big)-\Psi}-\mbe\zkh{\mc{R}^\phi_P\big(\bm{T}_F\circ\hat{f}_n^\dagger\big)-\Psi-\frac{1+\e}{n}\sum_{i=1}^n\ykh{\phi(Y_i\cdot\bm{T}_F(\hat{f}_n^\dagger(X_i)))-\psi(X_i,Y_i)}}\\&=(1+\e)\cdot\mbe\zkh{\frac{1}{n}\sum_{i=1}^n\phi(Y_i\cdot\bm{T}_F(\hat{f}_n^\dagger(X_i)))}-(1+\e)\cdot\mbe\zkh{\frac{1}{n}\sum_{i=1}^n{\psi(X_i,Y_i)}}\\
&\leq(1+\e)\cdot\mbe\zkh{\frac{1}{n}\sum_{i=1}^n\Bigykh{H+\phi(Y_i\cdot\hat{f}_n^\dagger(X_i))}}-(1+\e)\cdot\mbe\zkh{\frac{1}{n}\sum_{i=1}^n{\psi(X_i,Y_i)}}
\\&= (1+\e)\cdot\mbe\zkh{\frac{1}{n}\sum_{i=1}^n\phi(Y_i\cdot\hat{f}_n^\dagger(X_i))}+(1+\e)\cdot H-(1+\e)\cdot\Psi
\\&\leq (1+\e)\cdot\mbe\zkh{\frac{1}{n}\sum_{i=1}^n\phi(Y_i\cdot{f}(X_i))}-(1+\e)\cdot\Psi+(1+\e)\cdot H
\\&=(1+\e)\cdot(\mc{R}_P^\phi(f)-\Psi+H),\;\forall\;f\in\mc{F},\;\forall\;\e\in(-1,\infty), 
\] which means that
\beq\label{20231124115324}
& \mbe\zkh{\mc{R}^\phi_P\big(\bm{T}_F\circ\hat{f}_n^\dagger\big)-\Psi-\frac{1+\e}{n}\sum_{i=1}^n\ykh{\phi(Y_i\cdot\bm{T}_F(\hat{f}_n^\dagger(X_i)))-\psi(X_i,Y_i)}}+(1+\e)\cdot\inf_{f\in\mc{F}}\Big(\mc{R}_P^\phi(f)-\Psi+H\Big)\\
&\geq \mbe\zkh{\mc{R}^\phi_P\big(\bm{T}_F\circ\hat{f}_n^\dagger\big)-\Psi},\;\forall\;\e\in(-1,\infty). 
\eeq Recall that $\mc{N}(\mc{F},\gamma)\leq W<\infty$, meaning that there exist functions $f_1,\ldots,f_W$ in $\mc{F}$ such that 
\[
\mc{F}\subset\bigcup_{j=1}^W\set{g\in\mc{F}}{\sup_{x\in[0,1]^d}\Bigabs{g(x)-f_j(x)}\leq\gamma}.
\] Thus we can well define 
\[
j_*:\mc{F}\to\hkh{1,2,\ldots,W},\;g\mapsto\min\set{j\in\hkh{1,2,\ldots,W}}{\sup_{x\in[0,1]^d}\Bigabs{{g}(x)-f_j(x)}\leq\gamma}.
\] Obviously, 
\beq
\sup_{x\in[0,1]^d}\Bigabs{\bm{T}_F(g(x))-\bm{T}_F(f_{j_*(g)}(x))}\leq\sup_{x\in[0,1]^d}\Bigabs{g(x)-f_{j_*(g)}(x)}\leq\gamma,\;\forall\;g\in\mc{F}.
\eeq Therefore, 
\[
&\Bigg|{\mbe\zkh{\mc{R}^\phi_P\big(\bm{T}_F\circ\hat{f}_n^\dagger\big)}-\mbe\zkh{\mc{R}^\phi_P\big(\bm{T}_F\circ f_{j_*(\hat{f}_n^\dagger)}\big)}}\Bigg|\\&=\Bigg|{\mbe\bigg[\int_{[0,1]^d\times\hkh{-1,1}}\ykh{\phi(y\cdot\bm{T}_F\big(\hat{f}_n^\dagger(x)\big))-\phi(y\cdot\bm{T}_F\big({f}_{j_*(\hat{f}_n^\dagger)}(x)\big))}\mr{d}P(x,y)\bigg]}\Bigg|\\
&\leq
{\mbe\bigg[\int_{[0,1]^d\times\hkh{-1,1}}\abs{\phi(y\cdot\bm{T}_F\big(\hat{f}_n^\dagger(x)\big))-\phi(y\cdot\bm{T}_F\big({f}_{j_*(\hat{f}_n^\dagger)}(x)\big))}\mr{d}P(x,y)\bigg]}
\\&\leq
{\mbe\bigg[\int_{[0,1]^d\times\hkh{-1,1}}J\cdot\abs{y\cdot\bm{T}_F\big(\hat{f}_n^\dagger(x)\big)-y\cdot\bm{T}_F\big({f}_{j_*(\hat{f}_n^\dagger)}(x)\big)}\mr{d}P(x,y)\bigg]}\\&=
{\mbe\bigg[\int_{[0,1]^d\times\hkh{-1,1}}J\cdot\abs{\bm{T}_F\big(\hat{f}_n^\dagger(x)\big)-\bm{T}_F\big({f}_{j_*(\hat{f}_n^\dagger)}(x)\big)}\mr{d}P(x,y)\bigg]}
\\&\leq 
{\mbe\bigg[\int_{[0,1]^d\times\hkh{-1,1}}J\cdot \gamma\mr{d}P(x,y)\bigg]}=J\cdot\gamma,
\] and 
\[
&\Bigg|\mbe\bigg[{\frac{1}{n}\sum_{i=1}^n\phi(Y_i\cdot\bm{T}_F\big(\hat{f}_n^\dagger(X_i)\big))}\bigg]-\mbe\bigg[{\frac{1}{n}\sum_{i=1}^n\phi(Y_i\cdot\bm{T}_F\big({f}_{j_*(\hat{f}_n^\dagger)}(X_i)\big))}\bigg]\Bigg|
\\&=
\Bigg|\mbe\bigg[\frac{1}{n}\sum_{i=1}^n\ykh{\phi(Y_i\cdot\bm{T}_F\big(\hat{f}_n^\dagger(X_i)\big))-\phi(Y_i\cdot\bm{T}_F\big({f}_{j_*(\hat{f}_n^\dagger)}(X_i)\big))}\bigg]\Bigg|
\\&\leq
\mbe\bigg[\frac{1}{n}\sum_{i=1}^n\abs{\phi(Y_i\cdot\bm{T}_F\big(\hat{f}_n^\dagger(X_i)\big))-\phi(Y_i\cdot\bm{T}_F\big({f}_{j_*(\hat{f}_n^\dagger)}(X_i)\big))}\bigg]
\\&\leq 
\mbe\bigg[\frac{1}{n}\sum_{i=1}^nJ\cdot\abs{Y_i\cdot\bm{T}_F\big(\hat{f}_n^\dagger(X_i)\big)-Y_i\cdot\bm{T}_F\big({f}_{j_*(\hat{f}_n^\dagger)}(X_i)\big)}\bigg]
\\&=
\mbe\bigg[\frac{1}{n}\sum_{i=1}^nJ\cdot\abs{\bm{T}_F\big(\hat{f}_n^\dagger(X_i)\big)-\bm{T}_F\big({f}_{j_*(\hat{f}_n^\dagger)}(X_i)\big)}\bigg]
\leq
\mbe\bigg[\frac{1}{n}\sum_{i=1}^nJ\cdot\gamma\bigg]=J\cdot\gamma,
\] meaning that
\beq\label{20231124103953}
&\int_0^\infty\mb{P}\ykh{{\max\setl{\mc{R}^\phi_P\big(\bm{T}_F\circ{f}_{j}\big)-\Psi-\frac{1+\e}{n}\sum_{i=1}^n\ykh{\phi(Y_i\cdot\bm{T}_F\big({f}_{j}(X_i)\big))-\psi(X_i,Y_i)}}{{j\in\hkh{1,\ldots,W}}}}>t}\mr{d}t%
\\&\geq
\mbe\zkh{\max\setl{\mc{R}^\phi_P\big(\bm{T}_F\circ{f}_{j}\big)-\Psi-\frac{1+\e}{n}\sum_{i=1}^n\ykh{\phi(Y_i\cdot\bm{T}_F\big({f}_{j}(X_i)\big))-\psi(X_i,Y_i)}}{{j\in\hkh{1,\ldots,W}}}}
\\&\geq
\mbe\zkh{\mc{R}^\phi_P\big(\bm{T}_F\circ{f}_{j_*(\hat{f}_n^\dagger)}\big)-\Psi-\frac{1+\e}{n}\sum_{i=1}^n\ykh{\phi(Y_i\cdot\bm{T}_F\big({f}_{j_*(\hat{f}_n^\dagger)}(X_i)\big))-\psi(X_i,Y_i)}}
\\&\geq\mbe\zkh{\mc{R}^\phi_P\big(\bm{T}_F\circ\hat{f}_n^\dagger\big)-\Psi-\frac{1+\e}{n}\sum_{i=1}^n\ykh{\phi(Y_i\cdot\bm{T}_F\big(\hat{f}_n^\dagger(X_i)\big))-\psi(X_i,Y_i)}}-J\cdot\gamma-\abs{1+\e}\cdot J\cdot\gamma,\forall\;\e\in\mbR.
\eeq Next, we  use Bernstein's inequality to establish upper bounds for the probability \[\mb{P}\ykh{{\max\setl{\mc{R}^\phi_P\big(\bm{T}_F\circ{f}_{j}\big)-\Psi-\frac{1+\e}{n}\sum_{i=1}^n\ykh{\phi(Y_i\cdot\bm{T}_F\big({f}_{j}(X_i)\big))-\psi(X_i,Y_i)}}{{j\in\hkh{1,\ldots,W}}}}>t}.\] To this end, we first note that
\[
-M=0-M\leq\phi(y\cdot \bm{T}_F\big(f_j(x)\big))-\psi(x,y)\leq  2M,\;\forall\;(x,y)\in[0,1]^d\times\hkh{-1,1},\;\forall\;j\in\hkh{1,2,\ldots,W}, 
\] and 
\[
&0\leq \int_{[0,1]^d\times\hkh{-1,1}}\phi(y\cdot\bm{T}_F\big( f_j(x)\big))\mr{d}P(x,y)-\int_{[0,1]^d\times\hkh{-1,1}}\psi(x,y)\mr{d}P(x,y)\\&=\mbe\Big[{\phi(Y_1\cdot \bm{T}_F\big(f_j(X_1)\big))-\psi(X_1,Y_1)}\Big]=\mc{R}_P^\phi(\bm{T}_F\circ f_j)-\Psi,\;\forall\;j\in\hkh{1,2,\ldots,W}. 
\] Then it follows from Bernstein's inequality  that 
\[
&\mb{P}\ykh{{{\mc{R}^\phi_P(\bm{T}_F\circ {f}_{j})-\Psi-\frac{1+\e}{n}\sum_{i=1}^n\ykh{\phi(Y_i\cdot\bm{T}_F\big({f}_{j}(X_i)\big))-\psi(X_i,Y_i)}}}>t}\\
&=\mb{P}\ykh{{{\mc{R}^\phi_P\big(\bm{T}_F\circ{f}_{j}\big)-\Psi-\frac{1}{n}\sum_{i=1}^n\ykh{\phi(Y_i\cdot\bm{T}_F\big({f}_{j}(X_i)\big))-\psi(X_i,Y_i)}}}>\frac{t+\e\cdot (\mc{R}^\phi_P(\bm{T}_F\circ{f}_{j})-\Psi)}{1+\e}}
\\&\leq \exp\ykh{\frac{-n\cdot\abs{\frac{t+\e\cdot (\mc{R}^\phi_P(\bm{T}_F\circ{f}_{j})-\Psi)}{1+\e}}^2}{2\cdot\mbe\Big[\abs{\phi(Y_1\cdot\bm{T}_F\big({f}_{j}(X_1)\big))-\psi(X_1,Y_1)}^2\Big]+2\cdot M\cdot\frac{t+\e\cdot (\mc{R}^\phi_P(\bm{T}_F\circ{f}_{j})-\Psi)}{1+\e}}}
\\&\leq \exp\ykh{\frac{-n\cdot\abs{\frac{t+\e\cdot (\mc{R}^\phi_P(\bm{T}_F\circ{f}_{j})-\Psi)}{1+\e}}^2}{2\cdot\Gamma\cdot\Big|\mbe\big[\bigykh{\phi(Y_1\cdot\bm{T}_F\big({f}_{j}(X_1)\big))-\psi(X_1,Y_1)}\big]\Big|^\theta+2\cdot M\cdot\frac{t+\e\cdot (\mc{R}^\phi_P(\bm{T}_F\circ{f}_{j})-\Psi)}{1+\e}}}
\\&= \exp\ykh{\frac{-n\cdot\abs{\frac{t+\e\cdot (\mc{R}^\phi_P(\bm{T}_F\circ{f}_{j})-\Psi)}{1+\e}}^2}{2\cdot\Gamma\cdot\abs{\frac{1}{\e}}^\theta\cdot\big(\e\cdot(\mc{R}^\phi_P(\bm{T}_F\circ f_j)-\Psi)\big)^\theta+2\cdot M\cdot\frac{t+\e\cdot (\mc{R}^\phi_P(\bm{T}_F\circ{f}_{j})-\Psi)}{1+\e}}}
\\&\leq\exp\ykh{\frac{-n\cdot\abs{\frac{t+\e\cdot (\mc{R}^\phi_P(\bm{T}_F\circ{f}_{j})-\Psi)}{1+\e}}^2}{2\cdot\Gamma\cdot\abs{\frac{1}{\e}}^\theta\cdot\big(\e\cdot(\mc{R}^\phi_P(\bm{T}_F\circ f_j)-\Psi)+t\big)^\theta+2\cdot M\cdot\frac{t+\e\cdot (\mc{R}^\phi_P(\bm{T}_F\circ {f}_{j})-\Psi)}{1+\e}}}
\\&\leq\exp\ykh{\frac{-n\cdot\abs{\frac{t}{1+\e}}^2}{2\cdot\Gamma\cdot\abs{\frac{1}{\e}}^\theta\cdot t^\theta+2\cdot M\cdot\frac{t}{1+\e}}}
\leq\exp\ykh{\frac{-\frac{1}{2}\cdot n\cdot\abs{\frac{t}{1+\e}}^2}{\max\hkh{2\cdot\Gamma\cdot\abs{\frac{1}{\e}}^\theta\cdot t^\theta,\;2\cdot M\cdot\frac{t}{1+\e}}}}
\\&\leq
\exp\ykh{\frac{-\frac{1}{2}\cdot n\cdot\abs{\frac{t}{1+\e}}^2}{{2\cdot\Gamma\cdot\abs{\frac{1}{\e}}^\theta\cdot t^\theta}}}+\exp\ykh{\frac{-\frac{1}{2}\cdot n\cdot\abs{\frac{t}{1+\e}}^2}{{2\cdot M\cdot\frac{t}{1+\e}}}}
\\&=
\exp\ykh{\frac{-n\cdot\e^\theta\cdot t^{2-\theta}}{{4\cdot\Gamma\cdot\big|1+\e\big|^2}}}+\exp\ykh{\frac{- n\cdot{{t}}}{{4\cdot M\cdot\abs{1+\e}}}},\;
{\;\forall\;t\in(0,\infty),\;\forall\;j\in\hkh{1,2,\ldots,W},\;\forall\;\e\in(0,\infty)}.
\]
Consequently, 
\beq\label{241021093747}
&\mb{P}\ykh{{{\mc{R}^\phi_P(\bm{T}_F\circ {f}_{j})-\Psi-\frac{1+\e}{n}\sum_{i=1}^n\ykh{\phi(Y_i\cdot\bm{T}_F\big({f}_{j}(X_i)\big))-\psi(X_i,Y_i)}}}>t}
\\&\leq
\exp\ykh{\frac{-n\cdot\e^\theta\cdot t^{2-\theta}}{{4\cdot\Gamma\cdot\big|1+\e\big|^2}}}+\exp\ykh{\frac{- n\cdot{{t}}}{{4\cdot M\cdot\abs{1+\e}}}},\;
{\;\forall\;t\in(0,\infty),\;\forall\;j\in\hkh{1,2,\ldots,W},\;\forall\;\e\in(0,\infty)}.
\eeq
Therefore, 
\beq\label{20231124104043}
&\mb{P}\ykh{{\max\setma{3ex}{\mc{R}^\phi_P\big(\bm{T}_F\circ{f}_{j}\big)-\Psi-\frac{1+\e}{n}\sum_{i=1}^n\ykh{\phi(Y_i\cdot\bm{T}_F\big({f}_{j}(X_i)\big))-\psi(X_i,Y_i)}}{{j\in\hkh{1,\ldots,W}}}}>t}\\
&\leq\sum_{j=1}^W\mb{P}\ykh{{{\mc{R}^\phi_P\big(\bm{T}_F\circ{f}_{j}\big)-\Psi-\frac{1+\e}{n}\sum_{i=1}^n\ykh{\phi(Y_i\cdot\bm{T}_F\big({f}_{j}(X_i)\big))-\psi(X_i,Y_i)}}}>t}
\\&\leq W\cdot \exp\ykh{\frac{-n\cdot\e^\theta\cdot t^{2-\theta}}{{4\cdot\Gamma\cdot\big|1+\e\big|^2}}}+W\cdot\exp\ykh{\frac{- n\cdot{{t}}}{{4\cdot M\cdot\abs{1+\e}}}},\;\forall\;t\in(0,\infty),\;\forall\;\e\in(0,\infty). 
\eeq Combining \eqref{20231124103953}, \eqref{20231124104043}, and Lemma \ref{20231124104210}, we deduce that \[
&\mbe\zkh{\mc{R}^\phi_P\big(\bm{T}_F\circ\hat{f}_n^\dagger\big)-\Psi-\frac{1+\e}{n}\sum_{i=1}^n\ykh{\phi(Y_i\cdot\bm{T}_F\big(\hat{f}_n^\dagger(X_i)\big))-\psi(X_i,Y_i)}}-J\cdot\gamma-\abs{1+\e}\cdot J\cdot\gamma\\
&\leq \int_0^\infty\mb{P}\ykh{{\max_{j\in\hkh{1,\ldots,W}}\hkh{\mc{R}^\phi_P\big(\bm{T}_F\circ{f}_{j}\big)-\Psi-\frac{1+\e}{n}\sum_{i=1}^n\ykh{\phi(Y_i\cdot\bm{T}_F\big({f}_{j}(X_i)\big))-\psi(X_i,Y_i)}}{{}}}>t}\mr{d}t
\\&\leq\int_0^\infty \min\hkh{W\cdot \exp\ykh{\frac{-n\cdot\e^\theta\cdot t^{2-\theta}}{{4\cdot\Gamma\cdot\big|1+\e\big|^2}}}+W\cdot\exp\ykh{\frac{- n\cdot{{t}}}{{4\cdot M\cdot\abs{1+\e}}}},\;1}\mr{d}t
\\&\leq
\int_0^\infty \min\hkh{W\cdot\exp\ykh{\frac{- n\cdot{{t}}}{{4\cdot M\cdot\abs{1+\e}}}},\;1}\mr{d}t+\int_0^\infty \min\hkh{W\cdot \exp\ykh{\frac{-n\cdot\e^\theta\cdot t^{2-\theta}}{{4\cdot\Gamma\cdot\big|1+\e\big|^2}}},\;1}\mr{d}t
\\&\leq \frac{8\cdot M\cdot(1+\e)}{n}\cdot\log W+2\cdot\abs{\frac{4\cdot\Gamma\cdot(1+\e)^2}{n\cdot\e^\theta}\cdot\log W}^{\frac{1}{2-\theta}},\;\forall\;\e\in(0,\infty),
\] which yields
\beq\label{20231124115250}
&\mbe\zkh{\mc{R}^\phi_P\big(\bm{T}_F\circ\hat{f}_n^\dagger\big)-\Psi-\frac{1+\e}{n}\sum_{i=1}^n\ykh{\phi(Y_i\cdot\bm{T}_F\big(\hat{f}_n^\dagger(X_i)\big))-\psi(X_i,Y_i)}}
\\&\leq
\abs{2+\e}\cdot J\cdot\gamma+\frac{8\cdot M\cdot(1+\e)}{n}\cdot\log W+2\cdot\abs{\frac{4\cdot\Gamma\cdot(1+\e)^2}{n\cdot\e^\theta}\cdot\log W}^{\frac{1}{2-\theta}}\\&
\leq \abs{2+\e}\cdot J\cdot\gamma+\frac{8\cdot M\cdot(1+\e)}{n}\cdot\log W+8\cdot\abs{\frac{\Gamma\cdot(1+\e)^2}{n\cdot\e^\theta}\cdot\log W}^{\frac{1}{2-\theta}},\;\forall\;\e\in(0,\infty). 
\eeq Finally, it follows from \eqref{20231124115324} and \eqref{20231124115250} that \[
&\mbe\zkh{\mc{R}^\phi_P\big(\bm{T}_F\mkern-4mu\circ\mkern-4mu\hat{f}_n^\dagger\big)-\Psi}
\\&\leq \mbe\zkh{\mc{R}^\phi_P\big(\bm{T}_F\mkern-4mu\circ\mkern-4mu\hat{f}_n^\dagger\big)-\Psi-\frac{1+\e}{n}\sum_{i=1}^n\ykh{\phi(Y_i\cdot\bm{T}_F\big(\hat{f}_n^\dagger(X_i)\big))-\psi(X_i,Y_i)}}+(1+\e)\mkern-3mu\cdot\mkern-5mu\inf_{f\in\mc{F}}\Big(\mc{R}_P^\phi(f)-\Psi+H\Big) 
\\&\leq \abs{2+\e}\cdot J\cdot\gamma+\frac{8 M\cdot(1+\e)}{n}\cdot\log W+8\cdot\abs{\frac{\Gamma\cdot(1+\e)^2}{n\cdot\e^\theta}\cdot\log W}^{\frac{1}{2-\theta}}+(1+\e)\mkern-1mu\cdot\mkern-1mu\inf_{f\in\mc{F}}\Big(\mc{R}_P^\phi(f)-\Psi+H\Big)
\] for any $\e\in(0,\infty)$. Letting  $\e\to 0+$ gives the $\e=0$ case  of the above inequality.   This completes the proof of Theorem \ref{231123013855}.   \end{proof}

\subsection{Proof of Theorem \ref{20231125000604}}\label{240104170523}\hypertarget{20250918073254}{}  \label{20250918073254}

Let $a\in(0,\infty)$, $b\in(0,\infty)$, and  $\nu\in[0,\infty)$. We will prove that   there exist constants 
$\cyi\in[3,\infty)$ depending only on $(d_\star,d_*,\beta,r,q,K,\tau,s,a)$, $\cer\in(0,\infty)$ depending only on $(d,d_\star,d_*,\beta,r,q,K,\tau,s,\alpha,\nu,a,b)$, and $\csi\in(0,\infty)$  depending only on $(d_\star,d_*,\beta,r,q)$, such that:  \beq\label{231220061649}
\sup\setm{3}{{\bm E}_{P^{\otimes n}}\zkh{\mc{E}_P(\hat{f}_n^{\FNN})}}{P\in\mc{H}^{d,\beta,r}_{q,K,d_*,d_\star}\cap \mc{T}^{d,s}_{\alpha,\tau}}\leq\cer\cdot \ykh{\frac{(\log n)^3}{{n}}}^{\frac{\beta\cdot(1\qx\beta)^q}{{\frac{d_*}{s+1}+(1+\frac{1}{s+1})\cdot\beta\cdot(1\qx\beta)^q}}}
\eeq 
provided that 
\beq\label{231220060903}
& \cyi\leq n,\;\;\csi\log n\leq G\leq b \log n, \;\;a \abs{\frac{n}{(\log n)^3}}^{\frac{d_*}{d_*+(s+2)\cdot\beta\cdot(1\qx\beta)^q}}\leq N\leq b \abs{\frac{n}{(\log n)^3}}^{\frac{d_*}{d_*+(s+2)\cdot\beta\cdot(1\qx\beta)^q}},\\
&1\leq B\leq n^\nu b,\;\;a\cdot (\log n)\cdot \abs{\frac{n}{(\log n)^3}}^{\frac{d_*}{d_*+(s+2)\cdot\beta\cdot(1\qx\beta)^q}}\leq S\leq b\cdot (\log n)\cdot \abs{\frac{n}{(\log n)^3}}^{\frac{d_*}{d_*+(s+2)\cdot\beta\cdot(1\qx\beta)^q}}. 
\eeq 
 This immediately leads to Theorem \ref{20231125000604}. The proof is  based on Theorem \ref{231123013855}. 
To apply Theorem \ref{231123013855}, we need an inequality of the form \eqref{231123022427} for the hinge loss $\phi_{\mathbf{h}}(t)=\max\hkh{0,1-t}$, which is stated in Lemma \ref{23081001} below. The proof of Lemma \ref{23081001} is essentially the same as that of Lemma 6.1 in \cite{steinwart2007fast}. The advantage of  Lemma \ref{23081001} here over Lemma 6.1 in  \cite{steinwart2007fast} is that  Lemma \ref{23081001} provides the explicit expression of the prefactor in the bound, rather than relying on the Lorentz quasi-norm 
\[
\norm{\frac{1}{{2\eta_P(\cdot)-1}}}_{L^{q,\infty}(P_X)}:=\sup_{t\in(0,\infty)}\abs{\vphantom{\scalebox{2.5}{1}}t^{1/q}\cdot\inf\set{\lambda\in[0,\infty)}{P_X\ykh{\setm{3}{x\in[0,1]^d}{\textstyle\Bigabs{\frac{1}{2\eta_P(x)-1}}>\lambda}}\leq t}}
\] of the function $\frac{1}{2\eta_P(\cdot)-1}$ (cf.  Definition 4.1 of \cite{bennett1988interpolation}). 

\begin{lemma}\label{23081001} Let $\alpha\in(0,\infty)$, $\tau\in(0,\infty)$, $s\in[0,\infty]$, $d\in\mb N$, $P\in \mc{T}^{d,s}_{\alpha,\tau}$, and $\phi_{\mathbf{h}}(t)=\max\hkh{0,1-t}$ be the hinge loss.  Suppose $g\in\mc F_{d,\infty}$ and 
\beq\label{23080802}
P_X\ykh{\setl{x\in[0,1]^d}{g(x)=\mr{sgn}(2\cdot P(\hkh{1}|x)-1)}}=1. 
\eeq Then there holds 
\beq\label{23080801}
\inf_{f\in\mc F_{d,\infty}}\mc R^{\phi_{\mathbf{h}}}_P(f)=\mc R^{\phi_{\mathbf{h}}}_P(g)
\eeq and 
\beq\label{23080803}
&\int_{[0,1]^d\times\hkh{-1,1}}\abs{\phi_{\mathbf{h}}(yf(x))-\phi_{\mathbf{h}}(yg(x))}^2\mr{d}P(x,y)= \int_{[0,1]^d}\abs{f(x)-g(x)}^2\mr{d}P_X(x)\\&\leq  6\cdot\ykh{\mc E_P^{\phi_{\mathbf{h}}}(f)}^{\frac{s}{s+1}}\cdot \max\hkh{1,\alpha,\frac{1}{\tau}},
\;\forall\;f\in\mc F_{d,1},
\eeq  where $\frac{s}{s+1}:=1$ if $s=\infty$. 
	
\end{lemma}
\begin{proof}\let\tempersanyiyierbayi\phi
\renewcommand{\phi}{{\tempersanyiyierbayi_{\mathbf{h}}}}

We first prove \eqref{23080801}.  Define $u:\mbR\to\mbR, t\mapsto((-1)\qd t)\qx 1$. Then it is easy to verify that 
	\[
	\phi(y\cdot t)\geq \phi(y \cdot u(t))=1-y\cdot u(t),\;\forall\;y\in\hkh{-1,1},\;t\in\mbR. 
	\] Therefore, for any $f\in\mc F_{d,\infty}$,
\[
&\mc R^\phi_P(f)=\int_{[0,1]^d\times\hkh{-1,1}}\phi(y\cdot f(x))\mr{d}P(x,y)\geq \int_{[0,1]^d\times\hkh{-1,1}}\phi(y\cdot u(f(x)))\mr{d}P(x,y)\\
&=\int_{[0,1]^d}\int_{\hkh{-1,1}}\ykh{1-y\cdot u(f(x))}\mr{d}P(y|x)\mr{d}P_X(x)=1-\int_{[0,1]^d}\int_{\hkh{-1,1}}{y\cdot u(f(x))}\mr{d}P(y|x)\mr{d}P_X(x)\\
&=1-\int_{[0,1]^d}\Bigykh{P(\hkh{1}|x)\cdot u(f(x))-P(\hkh{-1}|x)\cdot u(f(x))}\mr{d}P_X(x)\\&=1-\int_{[0,1]^d}\Bigykh{2\cdot P(\hkh{1}|x)-1}\cdot u(f(x))\mr{d}P_X(x)\geq 1-\int_{[0,1]^d}\abs{2\cdot P(\hkh{1}|x)-1}\cdot \abs{u(f(x))}\mr{d}P_X(x)\\
&\geq 1-\int_{[0,1]^d}\abs{2\cdot P(\hkh{1}|x)-1}\mr{d}P_X(x)\\&=1-\int_{[0,1]^d}\Bigykh{2\cdot P(\hkh{1}|x)-1}\cdot\mr{sgn}(2\cdot P(\hkh{1}|x)-1)\mr{d}P_X(x)\\
&=1-\int_{[0,1]^d}\Bigykh{ P(\hkh{1}|x)\cdot\mr{sgn}(2\cdot P(\hkh{1}|x)-1)-P(\hkh{-1}|x)\cdot\mr{sgn}(2\cdot P(\hkh{1}|x)-1)}\mr{d}P_X(x)\\
&=1-\int_{[0,1]^d}\int_{\hkh{-1,1}}{ y\cdot\mr{sgn}(2\cdot P(\hkh{1}|x)-1)}\mr{d}P(y|x)\mr{d}P_X(x)\\
&=\int_{[0,1]^d}\int_{\hkh{-1,1}}\Bigykh{1- y\cdot\mr{sgn}(2\cdot P(\hkh{1}|x)-1)}\mr{d}P(y|x)\mr{d}P_X(x)\\&=\int_{[0,1]^d}\int_{\hkh{-1,1}}\phi\bigykh{ y\cdot\mr{sgn}(2\cdot P(\hkh{1}|x)-1)}\mr{d}P(y|x)\mr{d}P_X(x)\\&\xlongequal{\because\text{\eqref{23080802}}}\int_{[0,1]^d}\int_{\hkh{-1,1}}\phi\bigykh{ y\cdot g(x)}\mr{d}P(y|x)\mr{d}P_X(x)=\mc{R}_P^\phi(g). 
\] Consequently, 
\[
\mc R_P^\phi(g)\leq \inf_{f\in\mc F_{d,\infty}}\mc R^\phi_P(f)\leq \mc R_P^\phi(g),
\] leading to \eqref{23080801}. 

Now it remains to show \eqref{23080803}. To this end, we first note from  \eqref{23080802} that 
\beq\label{231211153720}
P\ykh{\set{(x,y)\in[0,1]^d\times\hkh{-1,1}}{y\cdot g(x)\in\hkh{-1,1}}}=1, 
\eeq meaning together with \eqref{231211153720} that 
\beq\label{231211155129}
&\int_{[0,1]^d\times\hkh{-1,1}}\abs{\phi(yf(x))-\phi(yg(x))}^2\mr{d}P(x,y)\\&=\int_{[0,1]^d\times\hkh{-1,1}}\abs{(1-yf(x))-(1-yg(x))}^2\mr{d}P(x,y)\\
&=\int_{[0,1]^d\times\hkh{-1,1}}\abs{f(x)-g(x)}^2\mr{d}P(x,y)=\int_{[0,1]^d}\abs{f(x)-g(x)}^2\mr{d}P_X(x),\;\forall\;f\in\mc{F}_{d,1}. 
\eeq If $s=0$, then we have from \eqref{231211155129} that 
\[
&\int_{[0,1]^d\times\hkh{-1,1}}\abs{\phi(yf(x))-\phi(yg(x))}^2\mr{d}P(x,y)= \int_{[0,1]^d}\abs{f(x)-g(x)}^2\mr{d}P_X(x)\\&\leq \int_{[0,1]^d}\abs{1+1}^2\mr{d}P_X(x)=4=4\cdot \ykh{\mc E^\phi_P(f)}^{\frac{s}{s+1}} \leq  6\cdot\ykh{\mc E_P^\phi(f)}^{\frac{s}{s+1}}\cdot \max\hkh{1,\alpha,\frac{1}{\tau}} ,
\;\forall\;f\in\mc F_{d,1}, 
\] which yields \eqref{23080803}. In the rest of the proof, we assume $0<s\leq\infty$. It follows from $P\in\mc T^{d,s}_{\alpha,\tau}$ that 
\[
&P_X\ykh{\setl{x\in[0,1]^d}{P(\hkh{1}|x)=1/2}}\leq \varlimsup_{t\to 0+} P_X\ykh{\setl{x\in[0,1]^d}{\abs{2P(\hkh{1}|x)-1}\leq t}}\\
&\leq \varlimsup_{t\to 0+} \alpha\cdot t^{s}=0,
\] meaning that 
\beq
P_X\ykh{\setl{x\in[0,1]^d}{P(\hkh{1}|x)={1}/{2}}}=0.
\eeq Let $f$ be an arbitrary function in $\mc F_{d,1}$.  Define  
\[
h_1(x)=\begin{cases}
	(1-f(x))\cdot\bigabs{2P(\hkh{1}|x)-1} ,&\text{ if }P(\hkh{1}|x)\geq1/2,\\
(1+f(x))\cdot \bigabs{2P(\hkh{1}|x)-1},&\text{ if }P(\hkh{1}|x)<1/2, 
\end{cases}
\] and 
\[
h_2(x)=\begin{cases}
	(1-f(x))^2,&\text{ if }P(\hkh{1}|x)\geq1/2,\\
	(1+f(x))^2,&\text{ if }P(\hkh{1}|x)<1/2.
\end{cases}
\] It is easy to verify by $f\in\mc F_{d,1}$  that 
\[
&\mc E_P^\phi(f)=\mc R_P^\phi(f)-\inf_{v\in\mc F_{d,\infty}}\mc R^\phi_P(v)=\mc R_P^\phi(f)-\mc R_P^\phi(g)\\&=\int_{[0,1]^d}\int_{\hkh{-1,1}}\Bigykh{\phi(yf(x))-\phi(yg(x))}\mr{d}P(y|x)\mr{d}P_X(x)\\
&=\int_{[0,1]^d}\int_{\hkh{-1,1}}\Bigykh{\phi(yf(x))-\phi(y\cdot \mr{sgn}(2P(\hkh{1}|x)-1))}\mr{d}P(y|x)\mr{d}P_X(x)\\
&=\int_{[0,1]^d}\int_{\hkh{-1,1}}y\cdot\Bigykh{\mr{sgn}(2P(\hkh{1}|x)-1)-f(x)}\mr{d}P(y|x)\mr{d}P_X(x)\\
&=\int_{[0,1]^d}\Bigykh{2P(\hkh{1}|x)-1}\cdot\Bigykh{\mr{sgn}(2P(\hkh{1}|x)-1)-f(x)}\mr{d}P_X(x)=\int_{[0,1]^d}h_1(x)\mr{d}P_X(x),
\] and 
\[
h_2(x)\leq\frac{2}{\bigabs{2P(\hkh{1}|x)-1}}\cdot h_1(x),\;\forall\;x\in\setl{z\in[0,1]^d}{P(\hkh{1}|z)\neq 1/2}. 
\] Thus we have that
\beq\label{23080901}
&\int_{[0,1]^d\times\hkh{-1,1}}\abs{\phi(yf(x))-\phi(yg(x))}^2\mr{d}P(x,y)\xlongequal{\because\eqref{231211155129}} \int_{[0,1]^d}\abs{f(x)-g(x)}^2\mr{d}P_X(x)\\
&=\int_{[0,1]^d}\abs{f(x)-\mr{sgn}(2P(\hkh{1}|x)-1)}^2\mr{d}P_X(x)=\int_{[0,1]^d}h_2(x)\mr{d}P_X(x)\\
&=\int_{[0,1]^d}h_2(x)\cdot\idf_{[0,t]}(\bigabs{2P(\hkh{1}|x)-1})\mr{d}P_X(x)+\int_{[0,1]^d}h_2(x)\cdot\idf_{(t,\infty)}(\bigabs{2P(\hkh{1}|x)-1})\mr{d}P_X(x)\\
&\leq \int_{[0,1]^d}4\cdot\idf_{[0,t]}(\bigabs{2P(\hkh{1}|x)-1})\mr{d}P_X(x)\\&\;\;\;\;\;\;\;\;\;\;\;\;\;\;\;\;\;\;\;\;+\int_{[0,1]^d}\frac{2\cdot h_1(x)}{\bigabs{2P(\hkh{1}|x)-1}}\cdot\idf_{(t,\infty)}(\bigabs{2P(\hkh{1}|x)-1})\mr{d}P_X(x)\\
&\leq 4\cdot\max\hkh{\idf_{(\tau,\infty)}(t),\alpha\cdot t^s}+\int_{[0,1]^d}\frac{2\cdot h_1(x)}{t}\mr{d}P_X(x)\\&=4\cdot\max\hkh{\idf_{(\tau,\infty)}(t),\alpha\cdot t^s}+\frac{2\mc E_P^\phi(f)}{t},\;\forall\;t\in(0,\infty),
\eeq where we have used the inequality \eqref{23080401}. If $s=\infty$, then it follows from \eqref{23080901} that 
\[
&\int_{[0,1]^d\times\hkh{-1,1}}\abs{\phi(yf(x))-\phi(yg(x))}^2\mr{d}P(x,y)= \int_{[0,1]^d}\abs{f(x)-g(x)}^2\mr{d}P_X(x)\\
&\leq \varlimsup_{t\to \tau\qx 1-}\ykh{4\cdot\max\hkh{\idf_{(\tau,\infty)}(t),\alpha\cdot t^s}+\frac{2\mc E_P^\phi(f)}{t}}=\frac{2\mc E_P^\phi(f)}{\tau\qx 1}\leq  6\cdot\ykh{\mc E_P^\phi(f)}^{\frac{s}{s+1}}\cdot \max\hkh{1,\alpha,\frac{1}{\tau}},
\]  which implies \eqref{23080803}. If $0<s<\infty$, then we take $t_0:=\ykh{\frac{\mc E_P^\phi(f)}{2\cdot s\cdot\max\hkh{\alpha,\frac{1}{\tau^s}}}}^{\frac{1}{s+1}}$ and deduce from \eqref{23080901} that \[
&\int_{[0,1]^d\times\hkh{-1,1}}\abs{\phi(yf(x))-\phi(yg(x))}^2\mr{d}P(x,y)= \int_{[0,1]^d}\abs{f(x)-g(x)}^2\mr{d}P_X(x)\\
&\leq \varlimsup_{t\to t_0+}\ykh{4\cdot\max\hkh{\idf_{(\tau,\infty)}(t),\alpha\cdot t^s}+\frac{2\mc E_P^\phi(f)}{t}}\\
&\leq \varlimsup_{t\to t_0+}\ykh{4\cdot\max\hkh{\alpha,\frac{1}{\tau^s}}\cdot t^s+\frac{2\mc E_P^\phi(f)}{t}}\\&=4\cdot\max\hkh{\alpha,\frac{1}{\tau^s}}\cdot \ykh{\frac{\mc E_P^\phi(f)}{2\cdot s\cdot\max\hkh{\alpha,\frac{1}{\tau^s}}}}^{\frac{s}{s+1}}+2\cdot \ykh{\mc E_P^\phi(f)}^{\frac{s}{s+1}}\cdot\ykh{{2\cdot s\cdot\max\hkh{\alpha,\frac{1}{\tau^s}}}}^{\frac{1}{s+1}}\\
&=\ykh{\mc E_P^\phi(f)}^{\frac{s}{s+1}}\cdot \ykh{\max\hkh{\alpha^{\frac{1}{s}},\frac{1}{\tau}}}^{\frac{s}{s+1}}\cdot\abs{4\cdot\ykh{\frac{1}{2s}}^{\frac{s}{s+1}}+2\cdot\bigykh{2s}^{\frac{1}{s+1}}}\\
&=\ykh{\mc E_P^\phi(f)}^{\frac{s}{s+1}}\cdot \ykh{\max\hkh{\alpha^{\frac{1}{s}},\frac{1}{\tau}}}^{\frac{s}{s+1}}\cdot\abs{\inf_{z\in(0,\infty)}\ykh{4\cdot z^s+\frac{2}{z}}}\\
&\leq\ykh{\mc E_P^\phi(f)}^{\frac{s}{s+1}}\cdot \ykh{\max\hkh{\alpha^{\frac{1}{s}},\frac{1}{\tau}}}^{\frac{s}{s+1}}\cdot\abs{{4\cdot 1^s+\frac{2}{1}}}=6\cdot\ykh{\mc E_P^\phi(f)}^{\frac{s}{s+1}}\cdot \max\hkh{\alpha^{\frac{1}{s+1}},\ykh{\frac{1}{\tau}}^{\frac{s}{s+1}}}\\
&\leq 6\cdot\ykh{\mc E_P^\phi(f)}^{\frac{s}{s+1}}\cdot \max\hkh{1,\alpha,\frac{1}{\tau}}, 
\] which yields \eqref{23080803} again. This completes the proof of this lemma. \end{proof} 

Combining Theorem \ref{231123013855} and Lemma \ref{23081001}, we immediately obtain the following result.

\begin{theorem}\label{231126053801} Let $d\in\mb N$, $\alpha\in(0,\infty)$,  $\tau\in(0,\infty]$, $s\in[0,\infty]$, $P\in\mc{T}^{d,s}_{\alpha,\tau}$, $\gamma\in(0,\infty)$, and let $\mc{F}$ be a nonempty class of measurable functions from $[0,1]^d$ to $\mbR$. For $n\in\mb N$, consider an i.i.d. sample $\hkh{(X_i,Y_i)}_{i=1}^n$ drawn from $P$ on $[0,1]^d\times\hkh{-1,1}$, and let $\phi_{\mathbf{h}}(t)=\max\hkh{0,1-t}$ be the hinge loss. Define $\hat{f}_n^\dagger$ as the  empirical $\phi_{\mathbf{h}}$-risk minimizer  over $\mc{F}$, i.e., 
	\beq
	\hat{f}_n^{\dagger}\in\mathop{{\arg\min}}_{f\in\mc{F}}\frac{1}{n}\sum_{i=1}^n\phi_{\mathbf{h}}\ykh{Y_if(X_i)}.
	\eeq Suppose 
	\[W:=\max\hkh{3,\;\mc{N}\ykh{\mc{F},\gamma}}<\infty.\]Then 
 \[
 &\abs{2+\e}\cdot\gamma+\frac{16 \cdot(1+\e)}{n}\cdot\log W+8\cdot\abs{\frac{6\cdot\abs{\alpha\qd\frac{1}{\tau}}\cdot|1+\e|^2}{n\cdot\e^{1-\frac{1}{s+1}}}\cdot\log W}^{1-\frac{1}{s+2}}+(1+\e)\cdot\inf_{f\in\mc{F}}\mc{E}_P^{\phi_{\mathbf{h}}}(f)\\&\geq \mbe\zkh{\mc{E}_P^{{\phi_{\mathbf{h}}}}\big(\bm{T}_1\circ\hat{f}_n^\dagger\big)}\geq
 \mbe\zkh{\mc{E}_P\big(\hat{f}_n^\dagger\big)}, \;\forall\;\e\in(0,\infty). 
 \] \end{theorem}

\begin{proof} We first observe from  \eqref{23081002} that
\beq\label{20231128174409}
\max\hkh{\alpha,\frac{1}{\tau}}\geq 1, 
\eeq and an elementary calculation gives
\[
&\sup\setm{2.3}{\phi_{\mathbf{h}}(y\cdot \bm{T}_1(f(x)))-\phi_{\mathbf{h}}(y\cdot f(x))}{(x,y)\in[0,1]^d\times\hkh{-1,1},\;f\in\mc{F}}\\
&\leq\sup\setm{2.3}{\phi_{\mathbf{h}}(y\cdot \bm{T}_1(t))-\phi_{\mathbf{h}}(y\cdot t)}{y\in\hkh{-1,1},\;t\in\mbR}=0. 
\]
Let $g:[0,1]^d\to\hkh{-1,1}$ be an arbitrary measurable function satisfying 
\[
P_X\ykh{\set{x\in[0,1]^d}{g(x)=\mr{sgn}\big(2\cdot P(\{1\}|x)-1\big)}}=1. 
\] Define
\begin{align*}
&\psi:[0,1]^d\times\hkh{-1,1}\to[0,2],\;(x,y)\mapsto\phi_{\mathbf{h}}(y\cdot g(x)). 
\end{align*} According to Lemma \ref{23081001}, 
\beq\label{231128180231}
\Psi&:=\int_{[0,1]^d\times\hkh{-1,1}}\psi(x,y)\mr{d}P(x,y)=\int_{[0,1]^d\times\hkh{-1,1}}\phi_{\mathbf{h}}(y\cdot g(x))\mr{d}P(x,y)=\mc{R}_P^{\phi_{\mathbf{h}}}(g)\\&=\inf_{u\in\mc{F}_d}\mc{R}_P^{\phi_{\mathbf{h}}}(u)\leq \inf_{u\in\mc{F}}\mc{R}_P^{\phi_{\mathbf{h}}}(\bm{T}_1\circ u)=\inf_{u\in\mc{F}}\int_{[0,1]^d\times\hkh{-1,1}}\phi(y\cdot\bm{T}_1(u(x)))\mr{d}P(x,y),
\eeq and 
\[
&\int_{[0,1]^d\times\hkh{-1,1}}\abs{\phi_{\mathbf{h}}(y\cdot \bm{T}_1(f(x)))-\psi(x,y)}^2\mr{d}P(x,y)\\&=\int_{[0,1]^d\times\hkh{-1,1}}\abs{\phi_{\mathbf{h}}(y\cdot \bm{T}_1(f(x)))-\phi_{\mathbf{h}}(y\cdot g(x))}^2\mr{d}P(x,y)\leq  6\cdot\abs{\mc E_P^{\phi_{\mathbf{h}}}(\bm{T}_1\circ f)}^{1-\frac{1}{s+1}}\cdot \max\hkh{1,\alpha,\frac{1}{\tau}}
\\&\xlongequal{\because \eqref{20231128174409}\text{ and } \eqref{231128180231}}
6\cdot\abs{\mc R_P^{\phi_{\mathbf{h}}}(\bm{T}_1\circ f)-\mc{R}_P^{\phi_{\mathbf{h}}}(g)}^{1-\frac{1}{s+1}}\cdot \max\hkh{\alpha,\frac{1}{\tau}}
\\&\xlongequal{\because\eqref{231128180231}}
6\cdot\abs{\int_{[0,1]^d\times\hkh{-1,1}}\Bigykh{\phi_{\mathbf{h}}(y\cdot \bm{T}_1(f(x)))-\psi(x,y)}\mr dP(x,y)}^{1-\frac{1}{s+1}}\cdot \max\hkh{\alpha,\frac{1}{\tau}}
,\;\forall\;f\in\mc{F}.
\]  We then apply Theorem \ref{231123013855} with $\phi=\phi_{\mathbf{h}}$, $F=1$, $J=1$, $M=2$, $\Gamma=6\cdot\max\hkh{\alpha,\frac{1}{\tau}}$, $\theta=1-\frac{1}{s+1}$, $H=0$ to obtain that 
\beq\label{231128182930}
&\abs{2+\e}\cdot\gamma+\frac{16 \cdot(1+\e)}{n}\cdot\log W+8\cdot\abs{\frac{6\cdot\abs{\alpha\qd\frac{1}{\tau}}\cdot|1+\e|^2}{n\cdot\e^{1-\frac{1}{s+1}}}\cdot\log W}^{1-\frac{1}{s+2}}+(1+\e)\cdot\inf_{f\in\mc{F}}\mc{E}_P^{\phi_{\mathbf{h}}}(f)	
\\&\xlongequal{\because\eqref{231128180231}}
\abs{2+\e} J\gamma+\frac{8 M\cdot(1+\e)}{n}\cdot\log W+8\abs{\frac{\Gamma\cdot|1+\e|^2}{n\cdot\e^\theta}\cdot\log W}^{\frac{1}{2-\theta}}+|1+\e|\mkern-1mu\cdot\mkern-3mu\inf_{f\in\mc{F}}\big(\mc{R}_P^{\phi_{\mathbf{h}}}(f)-\Psi+H\big)
	\\&\geq
	\mbe\zkh{\mc{R}_P^{\phi_{\mathbf{h}}}\big(\bm{T}_1\circ\hat{f}_n^\dagger\big)-\Psi}
	\xlongequal{\because\eqref{231128180231}}
	\mbe\zkh{\mc{E}_P^{{\phi_{\mathbf{h}}}}\big(\bm{T}_1\circ\hat{f}_n^\dagger\big)}
	,\;\forall\;\e\in(0,\infty).
\eeq Moreover, it follows from 
\[
\mr{sgn}(f(x))=\mr{sgn}(\bm{T}_1(f(x))),\;\forall\;x\in[0,1]^d,\;\forall\;f\in\mc{F}_d
\] that 
\[
\mbe\zkh{\mc{E}_P\big(\hat{f}_n^\dagger\big)}
=
\mbe\zkh{\mc{E}_P\big(\bm{T}_1\circ\hat{f}_n^\dagger\big)},
\] which, together with Theorem 2.31 of \cite{steinwart2008support}, yields
\beq\label{231128182907}
\mbe\zkh{\mc{E}_P\big(\hat{f}_n^\dagger\big)}
=
\mbe\zkh{\mc{E}_P\big(\bm{T}_1\circ\hat{f}_n^\dagger\big)}\leq\mbe\zkh{\mc{E}_P^{{\phi_{\mathbf{h}}}}\big(\bm{T}_1\circ\hat{f}_n^\dagger\big)}. 
\eeq Combining \eqref{231128182930} and \eqref{231128182907} proves the desired inequality. \end{proof}

The next lemma estimates the approximation error. Recall that the ReLU  function $\sigma$ is given by $\sigma(t)=\max\hkh{0,t}$. 
\begin{lemma}\label{231215142342}Let $d\in\mb  N$, $q\in\mb N\cup\hkh{0}$, $K\in\mb N$,  $d_\star\in\mb N$, $d_*\in\mb N$, $\beta\in(0,\infty)$,  $r\in(0,\infty)$, $\alpha\in(0,\infty)$,  $\tau\in(0,\infty]$, $s\in[0,\infty]$,   $P\in\mc{H}^{d,\beta,r}_{q,K,d_*,d_\star}\cap \mc{T}^{d,s}_{\alpha,\tau}$, and $\phi_{\mathbf{h}}(t)=\max\hkh{1-t,0}$ be the hinge loss. Suppose $d_*\leq d$, $d_*\leq K$, and $\idf_{(0,1)}(\alpha)\cdot\idf_{[1,\infty]}(\tau)\neq 1$. Then there exists  a constant $\mr{c}_{13}\in(2,\infty)$ only depending on $(d_\star,d_*,\beta,r,q)$ such that 
\beq\label{231215142548}
&\inf\set{\mc{E}_P^{\phi_{\mathbf{h}}}({f})}{f\in\fdnn_d\ykh{\mr{c}_{13}\cdot\log\frac{1}{\delta},K\cdot\mr{c}_{13}\cdot\abs{\frac{1}{\delta}}^{\frac{d_*}{\beta\cdot(1\qx\beta)^q}},K\cdot\mr{c}_{13}\cdot\abs{\frac{1}{\delta}}^{\frac{d_*}{\beta\cdot(1\qx\beta)^q}}\cdot\log\frac{1}{\delta},1,1}}
\\&\leq
2\alpha\cdot\delta^{s+1},\;\forall\;\delta\in \left(0,{\textstyle\frac{1}{2}}\qx\tau\right]. 
\eeq
\end{lemma}
\begin{proof}
According to Lemma C.14 of \cite{zhangzihan2023classification}, there exists a constant $\mr{c}_{12}\in(0,\infty)$ only depending on $(d_\star, d_*, \beta,r,q)$ such that
\beq\label{231214024516}
&\inf\set{\sup_{x\in[0,1]^d}\abs{f(x)-\tilde{f}(x)}}{\tilde{f}\in\fdnn_d\ykh{\mr{c}_{12}\log\frac{1}{\e},K\mr{c}_{12}\abs{\frac{1}{\e}}^{\frac{d_*}{\beta\cdot(1\qx\beta)^q}},K\mr{c}_{12}\abs{\frac{1}{\e}}^{\frac{d_*}{\beta\cdot(1\qx\beta)^q}}\cdot\log\frac{1}{\e},1,\infty}}\\
&\leq \frac{\e}{8},\;\forall\;f\in \mc G_d^{\mathbf{CHOM}}(q, K,d_\star, d_*, \beta,r),\;\forall\;\e\in(0,1/2].
\eeq Take
\[
\mr{c}_{13}:=\mr{c}_{12}+110. 
\]Then $\mr{c}_{13}$ only depends on $(d_\star,d_*,\beta,r,q)$ and belongs to $(2,\infty)$. We then show that this $\mr{c}_{13}$ has the desired property. Since $P\in\mc{H}^{d,\beta,r}_{q,K,d_*,d_\star}\cap \mc{T}^{d,s}_{\alpha,\tau}$, we have that there exists a function $g\in \mc G_d^{\mathbf{CHOM}}(q, K,d_\star, d_*, \beta,r)$ such that 
\beq\label{231215060654}
P_X\ykh{\set{x\in[0,1]^d}{g(x)=P(\hkh{1}|x)}}=1.
\eeq Let $\delta$ be an arbitrary number in $\left(0,{\textstyle\frac{1}{2}}\qx\tau\right]$. Then it follows from \eqref{231214024516} that there exists a function \beq\label{231215025509}
\tilde g\in\fdnn_d\ykh{\mr{c}_{12}\cdot\log\frac{1}{\delta},K\cdot\mr{c}_{12}\cdot\abs{\frac{1}{\delta}}^{\frac{d_*}{\beta\cdot(1\qx\beta)^q}},K\cdot\mr{c}_{12}\cdot\abs{\frac{1}{\delta}}^{\frac{d_*}{\beta\cdot(1\qx\beta)^q}}\cdot\log\frac{1}{\delta},1,\infty}\eeq
such that
\beq\label{231214221144}
\sup_{x\in[0,1]^d}\bigabs{\tilde{g}(x)-g(x)}\leq \frac{\delta}{7}. 
\eeq Take \beq\label{231215034849}
k:=\ceil{\frac{-\log\delta}{\log 2}}+1\in[2,5\log\frac{1}{\delta})\cap\mb{N},\eeq  
\[
\tilde l:\mbR\to\mbR,\;t\mapsto2\cdot\sigma\ykh{2^k\cdot\sigma\ykh{2\cdot\sigma({t})-1}}-2\cdot\sigma\ykh{2^k\cdot\sigma\ykh{2\cdot \sigma(t)-1}-1}-1,  
\] and $\tilde\eta:=\tilde l\circ\tilde g$. Obviously, 
\beq\label{231214224930}
\tilde l(z)=-1\leq\tilde{l}(t)\leq 1,\;\forall\;t\in\mbR, \;\forall\;z\in{\textstyle(-\infty,\frac{1}{2}]}, 
\eeq and it is easy to verify from $2^k\cdot(2\cdot t-1)\geq 2^k\cdot\frac{5\cdot\delta}{7}\geq\frac{10}{7}$ that
\beq
&\tilde l(t)=2\cdot\sigma\ykh{2^k\cdot\ykh{2\cdot{t}-1}}-2\cdot\sigma\ykh{2^k\cdot\ykh{2\cdot t-1}-1}-1
\\&=2\cdot\ykh{2^k\cdot\ykh{2\cdot{t}-1}}-2\cdot\ykh{2^k\cdot\ykh{2\cdot t-1}-1}-1=1,
\;\forall\;t\in{\textstyle\left[\frac{1}{2}+\frac{5\cdot\delta}{14},\infty\right)},
\eeq which, together with \eqref{231214221144}, yields
\beq\label{231215061726}
\set{x\in[0,1]^d}{2\cdot g(x)-1>\delta}&\subset \set{x\in[0,1]^d}{ \tilde g(x)>\frac{1}{2}+\frac{5\cdot\delta}{14}\text{ and }\mr{sgn}(2\cdot g(x)-1)=1}
\\&\subset
\set{x\in[0,1]^d}{ \tilde l(\tilde g(x))=1=\mr{sgn}(2\cdot g(x)-1)}, 
\eeq and
\beq\label{231215061738}
\set{x\in[0,1]^d}{2\cdot g(x)-1<-\delta}&\subset \set{x\in[0,1]^d}{ \tilde g(x)<\frac{1}{2}\text{ and }\mr{sgn}(2\cdot g(x)-1)=-1}
\\&\subset
\set{x\in[0,1]^d}{ \tilde l(\tilde g(x))=-1=\mr{sgn}(2\cdot g(x)-1)}. 
\eeq Moreover, it follows from   Lemma C.5 of \cite{zhangzihan2023classification}, \eqref{231215025509}, \eqref{231215034849} and \eqref{231214224930} that
\[
\tilde{\eta}=\tilde l\circ\tilde g&\in \fdnn_d\ykh{k+5+\mr{c}_{12}\cdot\log\frac{1}{\delta},5+K\cdot\mr{c}_{12}\cdot\abs{\frac{1}{\delta}}^{\frac{d_*}{\beta\cdot(1\qx\beta)^q}},32+6k+K\cdot\mr{c}_{12}\cdot\abs{\frac{1}{\delta}}^{\frac{d_*}{\beta\cdot(1\qx\beta)^q}}\cdot\log\frac{1}{\delta},1,1}\\
&\subset 
\fdnn_d\ykh{4k+\mr{c}_{12}\cdot\log\frac{1}{\delta},K\cdot(5+\mr{c}_{12})\cdot\abs{\frac{1}{\delta}}^{\frac{d_*}{\beta\cdot(1\qx\beta)^q}},22k+K\cdot\mr{c}_{12}\cdot\abs{\frac{1}{\delta}}^{\frac{d_*}{\beta\cdot(1\qx\beta)^q}}\cdot\log\frac{1}{\delta},1,1}
\\&\subset
\fdnn_d\ykh{(20+\mr{c}_{12})\cdot\log\frac{1}{\delta},K\cdot(5+\mr{c}_{12})\cdot\abs{\frac{1}{\delta}}^{\frac{d_*}{\beta\cdot(1\qx\beta)^q}},K\cdot(\mr{c}_{12}+110)\cdot\abs{\frac{1}{\delta}}^{\frac{d_*}{\beta\cdot(1\qx\beta)^q}}\cdot\log\frac{1}{\delta},1,1}
\\& \subset
\fdnn_d\ykh{\mr{c}_{13}\cdot\log\frac{1}{\delta},K\cdot\mr{c}_{13}\cdot\abs{\frac{1}{\delta}}^{\frac{d_*}{\beta\cdot(1\qx\beta)^q}},K\cdot\mr{c}_{13}\cdot\abs{\frac{1}{\delta}}^{\frac{d_*}{\beta\cdot(1\qx\beta)^q}}\cdot\log\frac{1}{\delta},1,1} 
\] {(cf. figure \ref{fig241017035901})}.

	\begin{figure}[H]
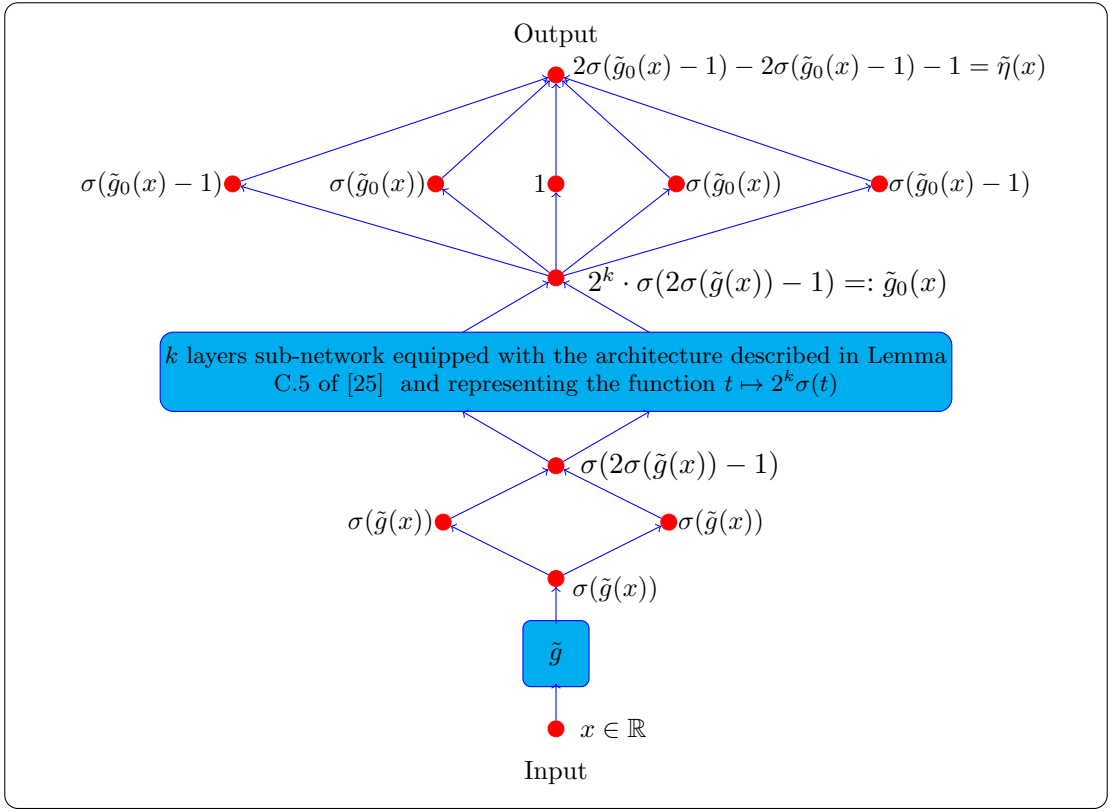

		\centering
		\betikz
			\tikzset{ %
					mypt/.style ={
						circle, %
						minimum width =0pt, %
						minimum height =0pt, %
						inner sep=0pt, %
						draw=blue, %
					}
				}
		\tikzset{
			mybox/.style ={
				rectangle, %
				rounded corners =5pt, %
				minimum width =10pt, %
				minimum height =70pt, %
				inner sep=0.6pt, %
				draw=blue, %
				fill=cyan
		}}
		\tikzset{
					myboxxx/.style ={
						rectangle, %
						rounded corners =5pt, %
						minimum width =10pt, %
						minimum height =30pt, %
						inner sep=0.6pt, %
						draw=blue, %
						fill=cyan
				}}
		\tikzset{
			ttinybox/.style ={
				rectangle, %
				rounded corners =3pt, %
				minimum width =25pt, %
				minimum height =25pt, %
				inner sep=2pt, %
				draw=blue, %
				fill=cyan
		}}
		\tikzset{
			tinybox/.style ={
				rectangle, %
				rounded corners =5pt, %
				minimum width =20pt, %
				minimum height =40pt, %
				inner sep=5pt, %
				draw=blue, %
				fill=cyan
		}}
		\tikzset{
			bigbox/.style ={
				rectangle, %
				rounded corners =5pt, %
				minimum width =367pt, %
				minimum height =305pt, %
				inner sep=0pt, %
				draw=black, %
				fill=none
		}}
		\tikzset{
			tinycircle/.style ={
				circle, %
				minimum width =6pt, %
				minimum height =6pt, %
				inner sep=0pt, %
				draw=red, %
				fill=red %
			}
		}
		\node[tinycircle] (1) at (0,0) {};
		\node[right] at (0,0){$\;\;x\in\mbR$};
		\node[below] at (-0.0,-0.3) {Input};
		
		\node[ttinybox] at (0,1) {$\tilde{g}$};
		\node[above] (11) at (0,1.15) {};
		\node[below] (x11) at (0,0.85) {};
		\node[tinycircle] (2) at (0,2) {};
		\filldraw[->,blue] (11)--(2);
		\filldraw[->,blue] (1)--(x11);
		\node[right] at (-0.1,1.85) {$\;\;\sigma(\tilde{g}(x))$};
		\node[tinycircle] (31) at (-1.5,2.75) {};
		\node[tinycircle] (32) at (1.5,2.75) {};
		\node[right] at (1.5,2.75) {$\sigma(\tilde{g}(x))$};
			\node[left] at (-1.5,2.75) {$\sigma(\tilde{g}(x))$};
		\filldraw[->,blue] (2)--(31) {};
				\filldraw[->,blue] (2)--(32) {};
	\node[tinycircle] (3) at (0,3.5) {};
		\filldraw[->,blue] (31)--(3) {};
			\filldraw[->,blue] (32)--(3) {};
		\node[right] at (0.2,3.5) {%
		{\fontsize{11}{11}\selectfont
		$\sigma(2\sigma(\tilde{g}(x))-1)$}};
	\node[bigbox] (xx) at (0,4.3) {};
	
		\node[myboxxx] (4) at(0,4.75) {\fontsize{9}{9}\selectfont{\begin{minipage}{232.56pt} 
					\centering
					$k$ layers sub-network  equipped with the architecture described in   Lemma C.5 of \cite{zhangzihan2023classification} \ and  representing the  function  $t\mapsto2^{k}\sigma(t)$
					\end{minipage}}};
	\node[mypt]  (x31) at (-1.25,4.225) {};
	\node[mypt]  (x32) at (1.25,4.225) {};
	\filldraw[->,blue] (3)--(x31) {};
	\filldraw[->,blue] (3)--(x32) {};
\node[tinycircle] (40) at (0,6) {};
\node[right] at (0.3,5.94) {\fontsize{11}{11}\selectfont
		$2^k\cdot\sigma(2\sigma(\tilde{g}(x))-1)=:\tilde g_0(x)$};
	\node[mypt]  (x33) at (-1.25,5.27) {};
	\node[mypt]  (x34) at (1.25,5.27) {};
\filldraw[->,blue] (x33)--(40) {};
\filldraw[->,blue] (x34)--(40) {};

\node[tinycircle] (51) at (-1.6,7.25) {};
\node[tinycircle] (52) at (-4.3,7.25) {};
\node[tinycircle] (53) at (1.6,7.25) {};
\node[tinycircle] (54) at (4.3,7.25) {};
\node[tinycircle] (55) at (0,7.25) {};
\node[tinycircle] (6) at (0,8.7) {};

\node[right]  at (1.6,7.25) {$\sigma(\tilde{g}_0(x))$};
\node[left]  at (-1.6,7.25) {$\sigma(\tilde{g}_0(x))$};
\node[left]  at (-0,7.25) {$1$};

\node[right]  at (4.3,7.25) {$\sigma(\tilde{g}_0(x)-1)$};
\node[left]  at (-4.3,7.25) {$\sigma(\tilde{g}_0(x)-1)$};

\node[right]  at (0.1,8.8) {$2\sigma(\tilde{g}_0(x)-1)-2\sigma(\tilde{g}_0(x)-1)-1=\tilde{\eta}(x)$};

\filldraw[->,blue] (40)--(51) {};
\filldraw[->,blue] (40)--(52) {};
\filldraw[->,blue] (40)--(54) {};
\filldraw[->,blue] (40)--(53) {};
\filldraw[->,blue] (40)--(55) {};

\filldraw[->,blue] (51)--(6) {};
\filldraw[->,blue] (52)--(6) {};
\filldraw[->,blue] (53)--(6) {};
\filldraw[->,blue] (54)--(6) {};
\filldraw[->,blue] (55)--(6) {};

\node[below] at (0,9.5) {Output};

		\eetikz
		\caption{The network representing the function $\tilde{\eta}$. }
		\label{fig241017035901}
	\end{figure}

We then use Theorem 2.31 of \cite{steinwart2008support} to obtain  by applying \eqref{231215060654} and \eqref{231215061726}, \eqref{231215061738} that
\[
&\inf\set{\mc{E}_P^{\phi_{\mathbf{h}}}({f})}{f\in\fdnn_d\ykh{\mr{c}_{13}\cdot\log\frac{1}{\delta},K\cdot\mr{c}_{13}\cdot\abs{\frac{1}{\delta}}^{\frac{d_*}{\beta\cdot(1\qx\beta)^q}},K\cdot\mr{c}_{13}\cdot\abs{\frac{1}{\delta}}^{\frac{d_*}{\beta\cdot(1\qx\beta)^q}}\cdot\log\frac{1}{\delta},1,1}}
\\&\leq\mc{E}_P^{\phi_{\mathbf{h}}}(\tilde{\eta})=\int_{[0,1]^d}\abs{\tilde{l}(\tilde{g}(x))-\mr{sgn}(2\cdot P(\hkh{1}|x)-1)}\cdot\abs{2\cdot P(\hkh{1}|x)-1}\mr{d}P_X(x)
\\&=
\int_{[0,1]^d}\abs{\tilde{l}(\tilde{g}(x))-\mr{sgn}(2\cdot g(x)-1)}\cdot\abs{2\cdot g(x)-1}\mr{d}P_X(x)
\\&=
\int_{[0,1]^d}\abs{\tilde{l}(\tilde{g}(x))-\mr{sgn}(2\cdot g(x)-1)}\cdot\abs{2\cdot g(x)-1}\cdot\idf_{[0,\delta]}(\abs{2\cdot g(x)-1})\mr{d}P_X(x)
\\&\leq
\int_{[0,1]^d}2\delta\cdot\idf_{[0,\delta]}(\abs{2\cdot g(x)-1})\mr{d}P_X(x)=2\delta\cdot P_X\ykh{\set{x\in[0,1]^d}{\abs{2\cdot P(\hkh{1}|x)-1}\leq\delta}}\\
&\leq 
2\delta\cdot\alpha\cdot\delta^s
=
2\alpha\cdot\delta^{s+1}. 
\] This completes the proof of this lemma. \end{proof}

We now give the proof of Theorem \ref{20231125000604}.

\begin{proof}[Proof of Theorem \ref{20231125000604}] It suffices to prove the claim 
 given at the beginning of Appendix  \ref{20250918073254}. 
According to Lemma \ref{231215142342}, there exists a constant $\mr{c}_{13}\in(2,\infty)$ only depending on $(d_\star,d_*,\beta,r,q)$, such that  \beq\label{231222055448}
&\inf\set{\mc{E}_P^{\phi_{\mathbf{h}}}({f})}{f\in\fdnn_d\ykh{\mr{c}_{13}\cdot\log\frac{1}{\delta},K\cdot\mr{c}_{13}\cdot\abs{\frac{1}{\delta}}^{\frac{d_*}{\beta\cdot(1\qx\beta)^q}},K\cdot\mr{c}_{13}\cdot\abs{\frac{1}{\delta}}^{\frac{d_*}{\beta\cdot(1\qx\beta)^q}}\cdot\log\frac{1}{\delta},1,1}}
\\&\leq
2\alpha\cdot\delta^{s+1},\;\forall\;\delta\in \left(0,{\textstyle\frac{1}{2}}\qx\tau\right],\;\forall\;P\in\mc{H}^{d,\beta,r}_{q,K,d_*,d_\star}\cap \mc{T}^{d,s}_{\alpha,\tau}. 
\eeq Take   
\[
&\csi:=\frac{\beta\cdot(1\qx\beta)^q}{d_*+2\cdot\beta\cdot(1\qx\beta)^q}\cdot \mr{c}_{13},\\
&\mr{c}_{14}:=\max\hkh{1,\;\ykh{\frac{K\cdot \mr{c}_{13}}{a}}^{\frac{\beta\cdot(1\qx\beta)^q}{d_*}}},\\
&\cyi:=\inf\set{z\in\mb N\cap [3,\infty)}{\mr{c}_{14}\cdot\ykh{\frac{(\log t)^3}{t}}^{\frac{\beta\cdot (1\qx\beta)^q}{d_*+(s+2)\cdot \beta\cdot(1\qx\beta)^q}}\leq \frac{1}{2}\qx\tau,\;\forall\;t\in[z,\infty)},\\
&\mr{c}_{15}:= 3+\nu+\log(2\cdot(b+1)^3), \\
&\mr{c}_{16}:=\mr{c}_{15}\cdot (1+b)\cdot(5+2b), 
\\
&\mr{c}_{17}:=\mr{c}_{16}+(d\log d)\cdot(b+1)\cdot(5+2b)+d\cdot\mr{c}_{16}+(\log d)\cdot(b+1)\cdot(5+2b),
\\&\cer:={4\cdot\alpha\cdot \bigabs{\mr{c}_{14}}^{s+1}+227\cdot\abs{\textstyle\alpha\qd 1\qd\frac{1}{\tau}}\cdot\mr{c}_{17}}. 
\] We then show that the constants $\cyi, \cer, \csi$ defined above have all the desired properties. Obviously,   $\cyi$  only depends on $(d_\star,d_*,\beta,r,q,K,\tau,s,a)$, $\cer$ only depends on $(d,d_\star,d_*,\beta,r,q,K,\tau,s,\alpha,\nu,a,b)$, $\csi$  only depends on   $(d_\star,d_*,\beta,r,q)$, $\cyi\in[3,\infty)$,  $\cer\in(0,\infty)$, and $\csi\in(0,\infty)$. From now on,  we assume that  \eqref{231220060903} holds. It remains to prove \eqref{231220061649}. Let $P$ be an arbitrary probability distribution in $\mc{H}^{d,\beta,r}_{q,K,d_*,d_\star}\cap \mc{T}^{d,s}_{\alpha,\tau}$.  Take  
\[
\zeta= \mr{c}_{14}\cdot\ykh{\frac{(\log n)^3}{n}}^{\frac{\beta\cdot (1\qx\beta)^q}{d_*+(s+2)\cdot \beta\cdot(1\qx\beta)^q}}. 
\]Since  $3\leq\cyi\leq n$, we have  that \beq\label{231220065526}
\frac{1}{n}<\ykh{\frac{1}{n}}^{\frac{\beta\cdot (1\qx\beta)^q}{d_*+(s+2)\cdot \beta\cdot(1\qx\beta)^q}}\leq \zeta\leq \frac{1}{2}\qx\tau<1\leq \frac{n}{(\log n)^3}.\eeq Consequently,  
\beq\label{231220065537}
&2\log2<\mr{c}_{13}\cdot\log\frac{1}{\zeta}\leq\mr{c}_{13}\cdot \frac{\beta\cdot (1\qx\beta)^q}{d_*+(s+2)\cdot \beta\cdot(1\qx\beta)^q}\cdot\log n\leq \mr{c}_{13}\cdot \frac{\beta\cdot (1\qx\beta)^q\cdot\log n}{d_*+2\cdot \beta\cdot(1\qx\beta)^q}=\csi\log n\leq G, 
\\&
K\mr{c}_{13}\abs{\frac{1}{\zeta}}^{\frac{d_*}{\beta\cdot(1\qx\beta)^q}}=K\mr{c}_{13}\abs{\frac{1}{\mr{c}_{14}}}^{\frac{d_*}{\beta\cdot(1\qx\beta)^q}}\cdot\abs{\frac{n}{(\log n)^3}}^{\frac{d_*}{d_*+(s+2)\cdot\beta\cdot(1\qx\beta)^q}}\leq a\cdot \abs{\frac{n}{(\log n)^3}}^{\frac{d_*}{d_*+(s+2)\cdot\beta\cdot(1\qx\beta)^q}}\leq N,
\eeq and 
\beq\label{231220065547}
&K\cdot\mr{c}_{13}\cdot\abs{\frac{1}{\zeta}}^{\frac{d_*}{\beta\cdot(1\qx\beta)^q}}\cdot\log\frac{1}{\zeta}\leq K\cdot\mr{c}_{13}\cdot\abs{\frac{1}{\zeta}}^{\frac{d_*}{\beta\cdot(1\qx\beta)^q}}\cdot\log n\\&=K\cdot\mr{c}_{13}\cdot\abs{\frac{1}{\mr{c}_{14}}}^{\frac{d_*}{\beta\cdot(1\qx\beta)^q}}\cdot\abs{\frac{n}{(\log n)^3}}^{\frac{d_*}{d_*+(s+2)\cdot\beta\cdot(1\qx\beta)^q}}\cdot\log{n}\leq a\cdot\abs{\frac{n}{(\log n)^3}}^{\frac{d_*}{d_*+(s+2)\cdot\beta\cdot(1\qx\beta)^q}}\cdot\log{n}\leq S. 
\eeq Combining \eqref{231220065526}, \eqref{231220065537}, \eqref{231220065547},  and \eqref{231222055448}, we obtain
\beq\label{231222021850}
&\inf\set{\mc{E}_P^{\phi_{\mathbf{h}}}({f})}{f\in\fdnn_d(G,N,S,B,\infty)}\\
&\leq\inf\set{\mc{E}_P^{\phi_{\mathbf{h}}}({f})}{f\in\fdnn_d\ykh{\mr{c}_{13}\cdot\log\frac{1}{\zeta},K\cdot\mr{c}_{13}\cdot\abs{\frac{1}{\zeta}}^{\frac{d_*}{\beta\cdot(1\qx\beta)^q}},K\cdot\mr{c}_{13}\cdot\abs{\frac{1}{\zeta}}^{\frac{d_*}{\beta\cdot(1\qx\beta)^q}}\cdot\log\frac{1}{\zeta},1,1}}
\\
&\leq2\alpha\cdot\zeta^{s+1}=2\alpha\cdot \bigabs{\mr{c}_{14}}^{s+1}\cdot\ykh{\frac{(\log n)^3}{n}}^{\frac{(s+1)\cdot\beta\cdot (1\qx\beta)^q}{d_*+(s+2)\cdot \beta\cdot(1\qx\beta)^q}}. 
\eeq Let 
\[
W=\max\hkh{3,\;\;\mc{N}\ykh{\setm{2.3}{f|_{[0,1]^d}}{f\in\fdnn_d(G,N,S,B,\infty)},1/n}}. 
\] Then it follows from Theorem A.1 of \cite{zhangzihan2023classification} that 
\[
&\log W\leq(S+1+Gd)\cdot(5+2G)\cdot\log\bigykh{(1+\max\hkh{N,d})\cdot(B\qd1)\cdot(2G+2)\cdot n}\\
&\leq 
|S+1+bd\log n|\cdot(5+2b\log n)\cdot\log\bigykh{(1+\max\hkh{bn,d})\cdot(1+b\cdot n^\nu)\cdot(2bn+2)\cdot n}\\
&\leq
\abs{b\cdot (\log n)\cdot \ykh{\frac{n}{(\log n)^3}}^{\frac{d_*}{d_*+(s+2)\cdot\beta\cdot(1\qx\beta)^q}}+1+bd\log n}\cdot(5+2b\log n)\cdot\log\bigykh{ 2d\cdot(b+1)^3\cdot n^{3+\nu}}
\\&\leq
(b+1)\cdot\abs{ (\log n)\cdot \ykh{\frac{n}{(\log n)^3}}^{\frac{d_*}{d_*+(s+2)\cdot\beta\cdot(1\qx\beta)^q}}+d\log n}\cdot(5+2b)\cdot(\log n)\cdot\bigabs{\log d+\mr{c}_{15}\cdot\log n}\\
&\leq\mr{c}_{16}\cdot (\log n)^3\cdot \ykh{\frac{n}{(\log n)^3}}^{\frac{d_*}{d_*+(s+2)\cdot\beta\cdot(1\qx\beta)^q}}+(d\log d)\cdot(b+1)\cdot(5+2b)\cdot(\log n)^2\\
&\;\;\;\;\;\;\;\;\;\;+d\cdot\mr{c}_{16}\cdot(\log n)^3+(\log d)\cdot(b+1)\cdot(5+2b)\cdot(\log n)^2\cdot \ykh{\frac{n}{(\log n)^3}}^{\frac{d_*}{d_*+(s+2)\cdot\beta\cdot(1\qx\beta)^q}}\\
&\leq
\mr{c}_{17}\cdot (\log n)^3\cdot \ykh{\frac{n}{(\log n)^3}}^{\frac{d_*}{d_*+(s+2)\cdot\beta\cdot(1\qx\beta)^q}},
\] meaning that 
\beq\label{231222021833}
\frac{\log W}{n}\leq 
\mr{c}_{17}\cdot \ykh{\frac{(\log n)^3}{n}}^{\frac{(s+2)\cdot\beta\cdot(1\qx\beta)^q}{d_*+(s+2)\cdot\beta\cdot(1\qx\beta)^q}}\leq
\mr{c}_{17}\cdot \ykh{\frac{(\log n)^3}{n}}^{\frac{(s+1)\cdot\beta\cdot(1\qx\beta)^q}{d_*+(s+2)\cdot\beta\cdot(1\qx\beta)^q}}
. 
\eeq Combining \eqref{231222021850}, \eqref{231222021833} and Theorem \ref{231126053801}, we obtain 
\[
&{\bm E}_{P^{\otimes n}}\zkh{\mc{E}_P(\hat{f}_n^{\FNN})}- 4\alpha\cdot \bigabs{\mr{c}_{14}}^{s+1}\cdot\ykh{\frac{(\log n)^3}{n}}^{\frac{(s+1)\cdot\beta\cdot (1\qx\beta)^q}{d_*+(s+2)\cdot \beta\cdot(1\qx\beta)^q}}
\\&\leq{\bm E}_{P^{\otimes n}}\zkh{\mc{E}_P(\hat{f}_n^{\FNN})}-(1+1)\cdot \inf\set{\mc{E}_P^{\phi_{\mathbf{h}}}({f})}{f\in\fdnn_d(G,N,S,B,\infty)}
\\&\leq
\abs{2+1}\cdot\frac{1}{n}+\frac{16 \cdot(1+1)}{n}\cdot\log W+8\cdot\abs{\frac{6\cdot\abs{\alpha\qd\frac{1}{\tau}}\cdot|1+1|^2}{n\cdot1}\cdot\log W}^{\frac{s+1}{s+2}}
\\&\leq
\frac{35}{n}\cdot\log W+192\cdot\abs{\textstyle\alpha\qd 1\qd\frac{1}{\tau}}\cdot\abs{\frac{1}{n}\cdot\log W}^{\frac{s+1}{s+2}}
\\
&\leq
\Bigykh{35\cdot\mr{c}_{17}+192\cdot\abs{\textstyle\alpha\qd 1\qd\frac{1}{\tau}}\cdot\mr{c}_{17}}\cdot \ykh{\frac{(\log n)^3}{n}}^{\frac{(s+1)\cdot\beta\cdot(1\qx\beta)^q}{d_*+(s+2)\cdot\beta\cdot(1\qx\beta)^q}}
\\&\leq
{227\cdot\abs{\textstyle\alpha\qd 1\qd\frac{1}{\tau}}\cdot\mr{c}_{17}}\cdot \ykh{\frac{(\log n)^3}{n}}^{\frac{(s+1)\cdot\beta\cdot(1\qx\beta)^q}{d_*+(s+2)\cdot\beta\cdot(1\qx\beta)^q}}, 
\] which yields
\[
{\bm E}_{P^{\otimes n}}\zkh{\mc{E}_P(\hat{f}_n^{\FNN})}&\leq
\ykh{4\alpha\cdot \bigabs{\mr{c}_{14}}^{s+1}+227\cdot\abs{\textstyle\alpha\qd 1\qd\frac{1}{\tau}}\cdot\mr{c}_{17}}\cdot \ykh{\frac{(\log n)^3}{n}}^{\frac{(s+1)\cdot\beta\cdot(1\qx\beta)^q}{d_*+(s+2)\cdot\beta\cdot(1\qx\beta)^q}}\\&=\cer\cdot \ykh{\frac{(\log n)^3}{n}}^{\frac{(s+1)\cdot\beta\cdot(1\qx\beta)^q}{d_*+(s+2)\cdot\beta\cdot(1\qx\beta)^q}}. 
\] Since $P$ is arbitrary, the desired inequality \eqref{231220061649} follows immediately. This completes the proof. 
\end{proof}

\subsection{Proof of Theorem \ref{231215231016}}\label{240104170001}\hypertarget{20250918113923}{} \label{20250918113923}

\begin{proof}[Proof of Theorem \ref{231215231016}]Let $b\in(0,\infty)$ and $\nu\in[0,\infty)$. To prove Theorem \ref{231215231016}, it suffices to show that  
 there exist constants  $\csan\in(0,\infty)$ depending only on $(d_\star,d_*,\beta,r,q,\tau)$ and $\cshi\in(0,\infty)$ depending only on $(d,\alpha,\tau, \nu, b)$ such that: 
\beq\label{231222054609}
\sup\setm{3}{{\bm E}_{P^{\otimes n}}\zkh{\mc{E}_P(\hat{f}_n^{\FNN})}}{P\in\mc{H}^{d,\beta,r}_{q,K,d_*,d_\star}\cap \mc{T}^{d,\infty}_{\alpha,\tau}}\leq \cshi\cdot\frac{\log n}{n}
\eeq 
provided that 
\beq\label{231222054545}
&2\leq n,\;\;\csan\leq G\leq b, \;\;K\cdot\csan\leq N\leq b,\\
&1\leq B\leq b \cdot n^\nu,\;\;K\cdot \csan\leq S. 
\eeq  According to Lemma \ref{231215142342}, there exists a constant $\mr{c}_{13}\in(2,\infty)$ only depending on $(d_\star,d_*,\beta,r,q)$, such that \beq \label{231222053857}
&\inf\set{\mc{E}_P^{\phi_{\mathbf{h}}}({f})}{f\in\fdnn_d\ykh{\mr{c}_{13}\cdot\log\frac{1}{\delta},K\cdot\mr{c}_{13}\cdot\abs{\frac{1}{\delta}}^{\frac{d_*}{\beta\cdot(1\qx\beta)^q}},K\cdot\mr{c}_{13}\cdot\abs{\frac{1}{\delta}}^{\frac{d_*}{\beta\cdot(1\qx\beta)^q}}\cdot\log\frac{1}{\delta},1,1}}
\\&\leq
0,\;\forall\;\delta\in \left(0,{\textstyle\frac{1}{2}}\qx\tau\right],\;P\in\mc{H}^{d,\beta,r}_{q,K,d_*,d_\star}\cap \mc{T}^{d,\infty}_{\alpha,\tau}. 
\eeq  Take
\[
&\csan:=\mr{c}_{13}\cdot\abs{\frac{3}{1\qx\tau}}^{\frac{d_*}{\beta\cdot(1\qx\beta)^q}}\cdot\log\frac{3}{1\qx\tau},\\
&\cshi:=227\cdot \abs{\alpha\qd1\qd\frac{1}{\tau}}\cdot d\cdot b^4\cdot 18\cdot(6\cdot\abs{\log b}+\nu+1+\log d). 
\] We then show that the constants $\csan$ and $\cshi$ defined above have all the desired properties. Obviously, $\csan$ only depends on $(d_\star,d_*,\beta,r,q,\tau)$, $\cshi$ only depends on $(d,\alpha,\tau, \nu, b)$, $\csan\in(0,\infty)$, and $\cshi\in(0,\infty)$. From now on, we assume that \eqref{231222054545} holds. It remains to prove \eqref{231222054609}. Let $P$ be an arbitrary probability distribution in $\mc{H}^{d,\beta,r}_{q,K,d_*,d_\star}\cap \mc{T}^{d,\infty}_{\alpha,\tau}$. Take $\zeta:=\frac{1\qx\tau}{3}$. Then we have
\[
&0<\xi<\frac{1}{2}\qx\tau<1,\;\;\\
&2\log 3\leq\mr{c}_{13}\cdot\log\frac{1}{\zeta}= \mr{c}_{13}\cdot\log\frac{3}{1\qx\tau}  
<
\mr{c}_{13}\cdot\abs{\frac{3}{1\qx\tau}}^{\frac{d_*}{\beta\cdot(1\qx\beta)^q}}\cdot\log\frac{3}{1\qx\tau} =
\csan\leq G\leq b,\\
&K\cdot\mr{c}_{13}\cdot\abs{\frac{1}{\zeta}}^{\frac{d_*}{\beta\cdot(1\qx\beta)^q}}=K\cdot\mr{c}_{13}\cdot\abs{\frac{3}{1\qx\tau}}^{\frac{d_*}{\beta\cdot(1\qx\beta)^q}}
<K\cdot\mr{c}_{13}\cdot\abs{\frac{3}{1\qx\tau}}^{\frac{d_*}{\beta\cdot(1\qx\beta)^q}}\cdot\log\frac{3}{1\qx\tau}
=K\cdot\csan\leq N, 
\\
&K\cdot\mr{c}_{13}\cdot\abs{\frac{1}{\zeta}}^{\frac{d_*}{\beta\cdot(1\qx\beta)^q}}\cdot\log\frac{1}{\zeta}=K\cdot\mr{c}_{13}\cdot\abs{\frac{3}{1\qx\tau}}^{\frac{d_*}{\beta\cdot(1\qx\beta)^q}}\cdot\log\frac{3}{1\qx\tau} =K\cdot\csan\leq S, 
\] which, together with  \eqref{231222053857}, yields\beq\label{231222100253}
&\inf\set{\mc{E}_P^{\phi_{\mathbf{h}}}({f})}{f\in\fdnn_d\ykh{G,N,S,B,\infty}}\\
&\leq\inf\set{\mc{E}_P^{\phi_{\mathbf{h}}}({f})}{f\in\fdnn_d\ykh{\mr{c}_{13}\cdot\log\frac{1}{\zeta},K\cdot\mr{c}_{13}\cdot\abs{\frac{1}{\zeta}}^{\frac{d_*}{\beta\cdot(1\qx\beta)^q}},K\cdot\mr{c}_{13}\cdot\abs{\frac{1}{\zeta}}^{\frac{d_*}{\beta\cdot(1\qx\beta)^q}}\cdot\log\frac{1}{\zeta},1,1}}
\leq 0.
\eeq Let 
\[
W=\max\hkh{3,\;\;\mc{N}\ykh{\setm{2.3}{f|_{[0,1]^d}}{f\in\fdnn_d(G,N,S,B,\infty)},1/n}}. 
\] Since any (real-valued) neural network defined on $\mbR^d$ of which the depth and the width are  less than or equal to $G$ and $N$ respectively has at most $(G\cdot N+N^3+d\cdot N)$ parameters,  we must have
\[
\fdnn_d(G,N,S,B,\infty)&\subset \fdnn_d(G,N,G\cdot N+N^3+d\cdot N,B,\infty)\subset \fdnn_d(b,b,2\cdot b^3+b\cdot d,B,\infty).
\]  Then it follows from Theorem A.1 of \cite{zhangzihan2023classification} and Lemma 10.6 of \cite{bartlett2009} that 
\[
\log W
&=
\max\hkh{\log3,\;\;\log\ykh{\mc{N}\ykh{\setm{2.3}{f|_{[0,1]^d}}{f\in\fdnn_d(G,N,S,B,\infty)},1/n}}}
\\&
\leq\max\hkh{\log3,\;\;\log\ykh{\mc{N}\ykh{\setm{2.3}{f|_{[0,1]^d}}{f\in\fdnn_d(b,b,2\cdot b^3+b\cdot d,B,\infty)},\frac{1}{2n}}}}
\\&
\leq(2\cdot b^3+b\cdot d+1+b\cdot d)\cdot(5+2b)\cdot\log\bigykh{(1+\max\hkh{b,d})\cdot(B\qd1)\cdot(2b+2)\cdot n}\\
&\leq d\cdot b^4\cdot 12\cdot\log\bigykh{{d}\cdot n^{1+\nu}\cdot b^6}
\leq d\cdot b^4\cdot 18\cdot(\log d+6\log b+\nu+1)\cdot\log n,
\] meaning that
\beq\label{231222100325}
&227\cdot \abs{\alpha\qd1\qd\frac{1}{\tau}}\cdot\frac{\log W}{n}\leq 227\cdot \abs{\alpha\qd1\qd\frac{1}{\tau}}\cdot d\cdot b^4\cdot 18\cdot(6\cdot\abs{\log b}+\nu+1+\log d)\cdot\frac{\log n}{n}\\&=\cshi\cdot\frac{\log n}{n}. 
\eeq Combining \eqref{231222100253}, \eqref{231222100325} and Theorem \ref{231126053801}, we obtain \[
&{{\bm E}_{P^{\otimes n}}\zkh{\mc{E}_P(\hat{f}_n^{\FNN})}}\leq {{\bm E}_{P^{\otimes n}}\zkh{\mc{E}_P(\hat{f}_n^{\FNN})}}-(1+1)\cdot\inf\set{\mc{E}_P^{\phi_{\mathbf{h}}}({f})}{f\in\fdnn_d\ykh{G,N,S,B,\infty}}
\\
&\leq 
\abs{2+1}\cdot\frac{1}{n}+\frac{16 \cdot(1+1)}{n}\cdot\log W+8\cdot\abs{\frac{6\cdot\abs{\alpha\qd\frac{1}{\tau}}\cdot|1+1|^2}{n\cdot1}\cdot\log W}
\\&\leq
227\cdot\abs{\frac{\abs{\alpha\qd1\qd\frac{1}{\tau}}}{n}\cdot\log W}\leq\cshi\cdot\frac{\log n}{n}. 
\] Since $P$ is arbitrary,  the desired inequality \eqref{231222054609} follows immediately. This completes the proof. \end{proof}

\subsection{Proof of Theorem \ref{20231204212113}}\hypertarget{20251011004301}{}  \label{20251011004301}

To prove Theorem \ref{20231204212113}, 
we first establish an upper bound for the covering number of the space $\mc G_d^{\mathbf{CHOM}}(q, K,d_\star, d_*, \beta,r)$.

\begin{lemma}\label{20231204220916}Let $d\in\mb  N$, $q\in\mb N\cup\hkh{0}$, $K\in\mb N$,  $d_\star\in\mb N$, $d_*\in\mb N$, $\beta\in(0,\infty)$, and $r\in(0,\infty)$. Suppose $d_*\leq d$ and $d_*\leq K$.   Then there exists a constant $\mr{c}_7\in(0,\infty)$ only depending on $(d,q,K,d_\star,d_*,\beta,r)$, such that \[&\log\ykh{\mc{N}(\mc G_d^{\mathbf{CHOM}}(q, K,d_\star, d_*, \beta,r),\gamma)}\leq \mr{c}_{7}\cdot \ykh{\frac{1}{\gamma}}^{\frac{d_*}{\beta\cdot(1\qx\beta)^q}},\;\forall\;\gamma\in(0,1]. \]
\end{lemma}
\begin{proof} According to Theorem \uppercase\expandafter{\romannumeral13\relax}  of  \cite{kolmogorov1961e} (see equation (65) therein), there exists a constant $A_1\in(0,\infty)$ only depending on $(d_*,\beta,r)$, such that 
\beq\label{231201145522}
\log\ykh{\mc{N}(\mc{B}^\beta_r([0,1]^{d_*}),t)}\leq  \frac{1}{A_1}\cdot \ykh{\frac{1}{t}}^{\frac{d_*}{\beta}}\leq \frac{1}{A_1}\cdot \ykh{\frac{2}{t}}^{\frac{d_*}{\beta}},\;\forall\;t\in(0,A_1],
\eeq which, together with Lemma 10.6 of \cite{bartlett2009}, gives 
\[
&\log\ykh{\mc{N}\ykh{\setm{2.3}{f\in\mc{B}^\beta_r([0,1]^{d_*})}{\mathbf{ran}(f)\subset[0,1]},t}}\leq \log\ykh{\mc{N}\ykh{{\mc{B}^\beta_r([0,1]^{d_*})},t/2}}\\&\leq  \frac{1}{A_1}\cdot \ykh{\frac{2}{t}}^{\frac{d_*}{\beta}},\;\forall\;t\in(0,A_1].
\] Consequently, for each $t\in(0,A_1]$, there exist sets $\mc{W}_{t,1}\subset \mc{B}^\beta_r([0,1]^{d_*})$ and \[\mc{W}_{t,0}\subset \setm{2.3}{f\in\mc{B}^\beta_r([0,1]^{d_*})}{\mathbf{ran}(f)\subset[0,1]}\] such that 
\[
&{\#(\mc{W}_{t,1})}={\mc{N}(\mc{B}^\beta_r([0,1]^{d_*}),t)}\leq \exp\Bigykh{ \frac{1}{A_1}\cdot \ykh{\frac{2}{t}}^{\frac{d_*}{\beta}}},\\
&{\#(\mc{W}_{t,0})}=\mc{N}\ykh{\setm{2.3}{f\in\mc{B}^\beta_r([0,1]^{d_*})}{\mathbf{ran}(f)\subset[0,1]},t}\leq \exp\Bigykh{ \frac{1}{A_1}\cdot \ykh{\frac{2}{t}}^{\frac{d_*}{\beta}}},\] and\beq\label{231203155214}
&\mc{B}^\beta_r([0,1]^{d_*})\subset\bigcup_{g\in\mc{W}_{t,1}}\set{u\in\mc{B}^\beta_r([0,1]^{d_*})}{\sup_{x\in[0,1]^{d_*}}\Bigabs{u(x)-g(x)}\leq t},\\
&\setm{2.3}{f\in\mc{B}^\beta_r([0,1]^{d_*})}{\mathbf{ran}(f)\subset[0,1]}\subset\bigcup_{g\in\mc{W}_{t,0}}\set{u\in\mc{B}^\beta_r([0,1]^{d_*})}{\sup_{x\in[0,1]^{d_*}}\Bigabs{u(x)-g(x)}\leq t}.
\eeq For each $i\in\hkh{0,1,\ldots, q}$, denote $d^{\mathbf{in}}_{i}:=d\cdot\idf_{\hkh{0}}(i)+K\cdot\idf_{\mb N}(i)$, $d^{\mathbf{out}}_{i}:=K+(1-K)\cdot\idf_{\hkh{q}}(i)$, $R_i:=\begin{cases}\mbR,&\text{ if }i=q,\\
[0,1],&\text{ if }i<q,\end{cases}$ $\iota_i:=\idf_{\hkh{q}}(i)$,  and 
\[
\mc{U}_i:=\set{h}{\begin{minipage}{285.6pt}$h$ is a function from $[0,1]^{d_i^{\mathbf{in}}}$ to $R_i^{d_i^{\mathbf{out}}}$, and for any $j\in\hkh{1,2,\ldots,d_i^{\mathbf{out}}}$, either the  function $[0,1]^{d_i^{\mathbf{in}}}\ni x\mapsto \big(h(x)\big)_j\in R_i$  belongs to $ \mc G_{d_i^{\mathbf{in}}}^{\mathbf{M}}(d_\star)$, or there exist a set $I\subset\hkh{1,2,\ldots,d_i^{\mathbf{in}}}$   and a function $g\in \mc{B}^\beta_r([0,1]^{d_*})$ such that $\#(I)=d_*$ and $\big(h(x)\big)_j=g((x)_I)\text{ for any }x\in[0,1]^{d_i^{\mathbf{in}}}$\end{minipage}}. 
\] Then for each $i\in\hkh{0,1,\ldots,q}$ and  $t\in(0,A_1]$, define \[
\mc{V}_{i,t}:=\set{h}{\begin{minipage}{285.6pt}$h$ is a function from $[0,1]^{d_i^{\mathbf{in}}}$ to $R_i^{d_i^{\mathbf{out}}}$, and for any $j\in\hkh{1,2,\ldots,d_i^{\mathbf{out}}}$, either the  function $[0,1]^{d_i^{\mathbf{in}}}\ni x\mapsto \big(h(x)\big)_j\in R_i$  belongs to $ \mc G_{d_i^{\mathbf{in}}}^{\mathbf{M}}(d_\star)$, or there exist a set $I\subset\hkh{1,2,\ldots,d_i^{\mathbf{in}}}$   and a function $g\in \mc{W}_{t,\iota_i}$ such that $\#(I)=d_*$ and $\big(h(x)\big)_j=g((x)_I)\text{ for any }x\in[0,1]^{d_i^{\mathbf{in}}}$\end{minipage}}. 
\] Next, for each $t\in(0, A_1]$, define
\[
\mc{H}_t:=\setm{2.3}{h_{q}\circ h_{q-1}\cdots\circ h_1\circ h_0}{h_i\in\mc{V}_{i,t},\;\forall\;i\in\hkh{0,1,\ldots,q}}. 
\] Obviously, 
\[
&\mc{H}_t\subset \setm{2.3}{h_{q}\circ h_{q-1}\cdots\circ h_1\circ h_0}{h_i\in\mc{U}_{i},\;\forall\;i\in\hkh{0,1,\ldots,q}}\\&=\mc G_d^{\mathbf{CHOM}}(q, K,d_\star, d_*, \beta,r),\;\forall\;t\in(0,A_1],
\] and
\[
&\#(\mc{H}_t)=\prod_{i=0}^q\#(\mc{V}_{i,t})\leq \prod_{i=0}^q\abs{\#(\mc G_{d_i^{\mathbf{in}}}^{\mathbf{M}}(d_\star))+{{d_i^{\mathbf{in}}}\choose{d_*}}\cdot\#(\mc{W}_{t,\iota_i})}^{d_i^{\mathbf{out}}}\\&=
\prod_{i=0}^q\abs{{{d_i^{\mathbf{in}}}\choose{d_*}}\cdot\#(\mc{W}_{t,\iota_i})+\sum_{j=1}^{d_\star}{d_i^{\mathbf{in}}\choose j}}^{d_i^{\mathbf{out}}}
\leq
\prod_{i=0}^q\abs{\exp\Bigykh{ \frac{1}{A_1}\cdot \ykh{\frac{2}{t}}^{\frac{d_*}{\beta}}}\cdot\Bigykh{{{d_i^{\mathbf{in}}}\choose{d_*}}+\sum_{j=1}^{d_\star}{d_i^{\mathbf{in}}\choose j}}}^{d_i^{\mathbf{out}}}\\
&=
\exp\Bigykh{\sum_{i=0}^q \frac{d_i^{\mathbf{out}}}{A_1}\cdot \ykh{\frac{2}{t}}^{\frac{d_*}{\beta}}}\cdot \prod_{i=0}^q\abs{\Bigykh{{{d_i^{\mathbf{in}}}\choose{d_*}}+\sum_{j=1}^{d_\star}{d_i^{\mathbf{in}}\choose j}}}^{d_i^{\mathbf{out}}}
=
\mr{c}_8\cdot \exp\Bigykh{\frac{Kq+1}{A_1}\cdot \ykh{\frac{2}{t}}^{\frac{d_*}{\beta}}},\;\forall\;t\in(0,A_1],
\]  where the constant 
\[
&\mr{c}_8:=\prod_{i=0}^q\abs{\Bigykh{{{d_i^{\mathbf{in}}}\choose{d_*}}+\sum_{j=1}^{d_\star}{d_i^{\mathbf{in}}\choose j}}}^{d_i^{\mathbf{out}}}\\&=\prod_{i=0}^q\abs{\Bigykh{{{d\cdot\idf_{\hkh{0}}(i)+K\cdot\idf_{\mb N}(i)}\choose{d_*}}+\sum_{j=1}^{d_\star}{d\cdot\idf_{\hkh{0}}(i)+K\cdot\idf_{\mb N}(i)\choose j}}}^{K+(1-K)\cdot\idf_{\hkh{q}}(i)}
\]only depends on $(d,q,K,d_*,d_\star)$.  Now we are ready to estimate the covering number of \[\mc G_d^{\mathbf{CHOM}}(q, K,d_\star, d_*, \beta,r).\] Let $\gamma$ be an arbitrary number in $(0,1]$. Take\[
\mr{c}_9:=(q+1)\cdot \Bigabs{r\cdot d_*^{1\qx\beta}}^{\sum\limits_{i=0}^{q-1}(1\qx\beta)^i},
\] which only depends on $(q,d_*,\beta,r)$,  and take \[
z:=\min\hkh{A_1,\;1,\;\abs{\frac{\gamma}{\mr{c}_9}}^{\frac{1}{(1\qx\beta)^q}}}.
\] Then we have $0<z\leq 1\qx A_1\leq A_1$. For any $f\in \mc G_d^{\mathbf{CHOM}}(q, K,d_\star, d_*, \beta,r)$, there exist functions $h_i\in\mc{U}_i\;(i=0,1,\ldots,q)$ such that $f=h_q\circ h_{q-1}\circ\cdots\circ h_1\circ h_0$. Then it follows from \eqref{231203155214} that there exist functions $\tilde{h}_i\in\mc{V}_{i,z} \;(i=0,1,\ldots,q)$, such that 
\beq\label{231203161055}
\sup_{x\in[0,1]^{d_i^{\mathbf{in}}}}\norm{h_i(x)-\tilde{h}_i(x)}_\infty\leq z\leq 1,\;\forall\;i\in\hkh{0,1,\ldots,q}. 
\eeq Take $\tilde f=\tilde{h}_q\circ\cdots\circ\tilde h_1\circ\tilde h_0$. Then we have $\tilde f\in\mc{H}_z$, and it follows from \eqref{231203161055} and Lemma C.11 of \cite{zhangzihan2023classification} that
\[
&\sup_{x\in[0,1]^d}\Bigabs{f(x)-\tilde{f}(x)}\leq \Bigabs{r\cdot d_*^{1\qx\beta}}^{\sum\limits_{i=0}^{q-1}(1\qx\beta)^i}\cdot\sum_{i=0}^q\abs{\sup_{x\in[0,1]^{d_i^{\mathbf{in}}}}\norm{h_i(x)-\tilde{h}_i(x)}_\infty}^{(1\qx\beta)^{q-i}}
\\&\leq
\Bigabs{r\cdot d_*^{1\qx\beta}}^{\sum\limits_{i=0}^{q-1}(1\qx\beta)^i}\cdot\sum_{i=0}^q\bigabs{z}^{(1\qx\beta)^{q-i}}\leq 
\Bigabs{r\cdot d_*^{1\qx\beta}}^{\sum\limits_{i=0}^{q-1}(1\qx\beta)^i}\cdot\sum_{i=0}^q\bigabs{z}^{(1\qx\beta)^{q}}
=\mr{c}_9\cdot \bigabs{z}^{(1\qx\beta)^{q}}\leq \gamma. 
\] Thus 
\[
f&\in\set{u\in\mc G_d^{\mathbf{CHOM}}(q, K,d_\star, d_*, \beta,r)}{\sup_{x\in[0,1]^d}\Bigabs{u(x)-\tilde{f}(x)}\leq\gamma}\\&\subset\bigcup_{g\in\mc{H}_z}\set{u\in\mc G_d^{\mathbf{CHOM}}(q, K,d_\star, d_*, \beta,r)}{\sup_{x\in[0,1]^d}\Bigabs{u(x)-g(x)}\leq\gamma}. 
\] Since $f$ is arbitrary, we deduce that
\[
\mc G_d^{\mathbf{CHOM}}(q, K,d_\star, d_*, \beta,r)\subset\bigcup_{g\in\mc{H}_z}\set{u\in\mc G_d^{\mathbf{CHOM}}(q, K,d_\star, d_*, \beta,r)}{\sup_{x\in[0,1]^d}\Bigabs{u(x)-g(x)}. \leq\gamma}.
\] Therefore, 
\beq\label{20231204154325}
&\mc{N}(\mc G_d^{\mathbf{CHOM}}(q, K,d_\star, d_*, \beta,r),\gamma)\leq \#(\mc{H}_z)\leq \mr{c}_8\cdot \exp\Bigykh{\frac{Kq+1}{A_1}\cdot \ykh{\frac{2}{z}}^{\frac{d_*}{\beta}}}
\\&=
\max\hkh{\mr{c}_8\cdot \exp\Bigykh{\frac{Kq+1}{A_1}\cdot \ykh{\frac{2}{1\qx A_1}}^{\frac{d_*}{\beta}}},\mr{c}_8\cdot \exp\Bigykh{\frac{Kq+1}{A_1}\cdot 2^{\frac{d_*}{\beta}}\cdot\ykh{\frac{\mr{c}_9}{\gamma}}^{\frac{d_*}{\beta\cdot(1\qx\beta)^q}}}}
\\&\leq \mr{c}_8\cdot\exp\ykh{\max\hkh{\frac{Kq+1}{A_1}\cdot \ykh{\frac{2}{1\qx A_1}}^{\frac{d_*}{\beta}},\;\frac{Kq+1}{A_1}\cdot 2^{\frac{d_*}{\beta}}\cdot\Bigabs{\mr{c}_9}^{\frac{d_*}{\beta\cdot(1\qx\beta)^q}}}\cdot \ykh{\frac{1}{\gamma}}^{\frac{d_*}{\beta\cdot(1\qx\beta)^q}}}.
\eeq Take
\[
\mr{c}_{10}:=\max\hkh{\frac{Kq+1}{A_1}\cdot \ykh{\frac{2}{1\qx A_1}}^{\frac{d_*}{\beta}},\;\frac{Kq+1}{A_1}\cdot 2^{\frac{d_*}{\beta}}\cdot\Bigabs{\mr{c}_9}^{\frac{d_*}{\beta\cdot(1\qx\beta)^q}}},
\] which only depends on $(q,K,d_*,\beta,r)$, and take 
\[
\mr{c}_{7}:=\bigabs{\log(\mr{c}_8)}+\mr{c}_{10}, 
\] which only depends on $(d,q,K,d_\star,d_*,\beta,r)$. Then it follows from \eqref{20231204154325} that 
\[
\log\ykh{\mc{N}(\mc G_d^{\mathbf{CHOM}}(q, K,d_\star, d_*, \beta,r),\gamma)}\leq \log(\mr{c}_8)+\mr{c}_{10}\cdot\ykh{\frac{1}{\gamma}}^{\frac{d_*}{\beta\cdot(1\qx\beta)^q}}\leq \mr{c}_{7}\cdot \ykh{\frac{1}{\gamma}}^{\frac{d_*}{\beta\cdot(1\qx\beta)^q}}.
\] This completes the proof. \end{proof}

We now prove Theorem \ref{20231204212113}.

\begin{proof}[Proof of Theorem \ref{20231204212113}]Let \[\xi:=\frac{\tau\qx 1}{3}\cdot\ykh{\frac{1}{n}}^{\frac{1}{s+2+\frac{d_*}{\beta\cdot(1\qx\beta)^q}}}. \] According to Lemma \ref{20231204220916}, there exist a constant $\mr{c}_7\in(0,\infty)$ only depending on $(d,q,K,d_\star,d_*,\beta,r)$ and a subset $\mc{W}$ of  ${\mc G_d^{\mathbf{CHOM}}(q, K,d_\star, d_*, \beta,r)}$ such that 
\beq\label{231208000749}
1\leq \#(\mc{W})={\mc{N}(\mc G_d^{\mathbf{CHOM}}(q, K,d_\star, d_*, \beta,r),\xi)}\leq\exp\Bigykh{\mr{c}_7\cdot {\ykh{\frac{1}{\xi}}}^{\frac{d_*}{\beta\cdot (1\qx\beta)^q}}}
\eeq and 
\beq\label{231208003632}
\mc G_d^{\mathbf{CHOM}}(q, K,d_\star, d_*, \beta,r)\subset\bigcup_{g\in\mc{W}}\set{u\in\mc G_d^{\mathbf{CHOM}}(q, K,d_\star, d_*, \beta,r)}{\sup_{x\in[0,1]^d}\Bigabs{u(x)-g(x)}\leq \xi}. 
\eeq Let $\phi_{\mathbf{h}}:\mbR\to[0,\infty),\;t\mapsto\max\hkh{0,1-t}$ be the hinge loss,  \beq\label{75eq240119025342}
\mc{F}:=\setm{2.3}{g:[0,1]^d\to\hkh{-1,1}}{\exists\;f\in\mc{W}\;\mr{s.t.}\; g(x)= \mr{sgn}(2\cdot f(x)-1)\text{ for all }x\in[0,1]^d},\eeq and $\hat{f}^{\lozenge}_n$ be the empirical hinge risk minimizer over $\mc{F}$, that is,  
\[
\hat{f}_n^{\lozenge}\in\mathop{\arg\min}_{f\in\mc{F}}\frac{1}{n}\sum_{i=1}^n \phi_{\mathbf{h}}(Y_i\cdot f(X_i)). 
\] Then it follows from \eqref{231208000749} that
\beq\label{231208033253}
&W:=\max\hkh{3,\;\mc{N}(\mc{F},1/n)}\leq \max\hkh{3,\;\#(\mc{F})}\leq\max\hkh{3,\;\#(\mc W)}\leq \exp\Bigykh{\bigabs{2\qd\mr{c}_7}\cdot {\ykh{\frac{1}{\xi}}}^{\frac{d_*}{\beta\cdot (1\qx\beta)^q}}}\\
&=
\exp\Bigykh{\bigabs{2\qd\mr{c}_7}\cdot {\ykh{\frac{3}{\tau\qx 1}}}^{\frac{d_*}{\beta\cdot (1\qx\beta)^q}}\cdot {\Bigabs{{n}}^{\frac{{d_*}{}}{(s+2)\cdot \beta\cdot (1\qx\beta)^q+{d_*}{}}}}}.
\eeq Take 
\[
\mr{c}_{11}:=\bigabs{2\qd\mr{c}_7}\cdot {\ykh{\frac{3}{\tau\qx 1}}}^{\frac{d_*}{\beta\cdot (1\qx\beta)^q}}
\] and 
\[
\cwu:=227\cdot\abs{1\qd\alpha\qd\frac{1}{\tau}}\cdot\mr{c}_{11}+4\cdot\alpha.
\] Obviously, $\cwu$ only depends on $(d,q,K,d_\star,d_*,\beta,r,\tau,\alpha)$ and belongs to $(0,\infty)$. We then show that  $\hat{f}_n^{\lozenge}$ and $\cwu$ defined above satisfy the desired inequality \eqref{231208034119}. 

Let $P$ be an arbitrary probability measure in $\mc{H}^{d,\beta,r}_{q,K,d_*,d_\star}\cap \mc{T}^{d,s}_{\alpha,\tau}$. According to Theorem \ref{231126053801}, 
\beq\label{231208030946}
&\bm{E}_{P^{\otimes n}}\zkh{\mc{E}_P\big(\hat{f}_n^\lozenge\big)}\\&\leq \abs{2+1}\cdot\frac{1}{n}+\frac{16 \cdot(1+1)}{n}\cdot\log W+8\cdot\abs{\frac{6\cdot\abs{\alpha\qd\frac{1}{\tau}}\cdot|1+1|^2}{n\cdot 1}\cdot\log W}^{\frac{1}{1+1/(s+1)}}+(1+1)\cdot\inf_{u\in\mc{F}}\mc{E}_P^{\phi_{\mathbf{h}}}(u)\\
&\leq \frac{35}{n}\cdot\log W+8\cdot\abs{\frac{24\cdot\abs{\alpha\qd\frac{1}{\tau}}}{n}\cdot\log W}^{\frac{1}{1+1/(s+1)}}+2\cdot\inf_{u\in\mc{F}}\mc{E}_P^{\phi_{\mathbf{h}}}(u).
\eeq  Since $P\in \mc{H}^{d,\beta,r}_{q,K,d_*,d_\star}$, there exists a function $g_*\in \mc G_d^{\mathbf{CHOM}}(q, K,d_\star, d_*, \beta,r)$ such that 
\beq\label{231208021446}
P_X\ykh{\setm{2.3}{x\in[0,1]^d}{P(\hkh{1}|x)=g_*(x)}}=1. 
\eeq We then obtain from \eqref{231208003632} that there exists a function $f_*\in\mc{W}$ such that 
\beq\label{231208021647}
\sup_{x\in[0,1]^d}\Bigabs{f_*(x)-g_*(x)}\leq \xi. 
\eeq Take \[v_*:[0,1]^d\to\hkh{-1,1},\;x\mapsto \mr{sgn}(2\cdot f_*(x)-1),\] which is a function in $\mc{F}$. Then it follows from   Theorem 2.31 of \cite{steinwart2008support} and \eqref{231208021446}, \eqref{231208021647} that 
\[
&\inf_{u\in\mc F}\mc{E}_P^{\phi_{\mathbf{h}}}(u)\leq \mc{E}_P^{\phi_{\mathbf{h}}}( v_*)=\int_{[0,1]^d}\abs{{v_*(x)-\mr{sgn}\big(2\cdot P(\hkh{1}|x)-1\big)}}\cdot\bigabs{2\cdot P(\hkh{1}|x)-1}\mr{d}P_X(x)\\
&=\int_{[0,1]^d}\abs{{\mr{sgn}\big(2\cdot f_*(x)-1\big)-\mr{sgn}\big(2\cdot g_*(x)-1\big)}}\cdot\bigabs{2\cdot g_*(x)-1}\mr{d}P_X(x)\\
&=
\int_{[0,1]^d}\abs{{\mr{sgn}\big(2\cdot f_*(x)-1\big)-\mr{sgn}\big(2\cdot g_*(x)-1\big)}}\cdot\bigabs{2\cdot g_*(x)-1}\cdot\idf_{[0,2\xi]}(\abs{2\cdot g_*(x)-1})\mr{d}P_X(x)
\\
&\leq
\int_{[0,1]^d}2\cdot2\xi\cdot\idf_{[0,2\xi]}(\abs{2\cdot g_*(x)-1})\mr{d}P_X(x)= 4\xi\cdot P_X\ykh{\set{x\in[0,1]^d}{\abs{2\cdot P(\hkh{1}|x)-1}\leq 2\xi}}
\\&\leq 
4\xi\cdot \alpha\cdot \bigabs{2\xi}^s=2\alpha\cdot\abs{\frac{2}{3}\cdot\ykh{\tau\qx 1}\cdot\ykh{\frac{1}{n}}^{\frac{1}{s+2+\frac{d_*}{\beta\cdot(1\qx\beta)^q}}}}^{s+1}
\leq 
2\alpha\cdot\ykh{\frac{1}{n}}^{\frac{ \beta\cdot(1\qx\beta)^q}{\ykh{1+\frac{1}{s+1}}\cdot \beta\cdot(1\qx\beta)^q+\frac{d_*}{s+1}}},
\] which, together with \eqref{231208033253} and \eqref{231208030946}, yields 
\[
&\bm{E}_{P^{\otimes n}}\zkh{\mc{E}_P\big(\hat{f}_n^\lozenge\big)}\leq\frac{35}{n}\cdot\log W+8\cdot\abs{\frac{24\cdot\abs{\alpha\qd\frac{1}{\tau}}}{n}\cdot\log W}^{\frac{1}{1+{1}/\ykh{s+1}}}+2\cdot\inf_{u\in\mc{F}}\mc{E}_P^{\phi_{\mathbf{h}}}(u)\\
&\leq \frac{35}{n}\cdot {\mr{c}_{11}\cdot {\Bigabs{{n}}^{\frac{{d_*}{}}{(s+2)\cdot \beta\cdot (1\qx\beta)^q+{d_*}{}}}}}+192\cdot\abs{\alpha\qd\frac{1}{\tau}}^{\frac{1}{1+1/(s+1)}}\cdot\abs{\frac{{\mr{c}_{11}\cdot {\Bigabs{{n}}^{\frac{{d_*}{}}{(s+2)\cdot \beta\cdot (1\qx\beta)^q+{d_*}{}}}}}}{n}}^{\frac{1}{1+1/(s+1)}}+2\cdot\inf_{u\in\mc{F}}\mc{E}_P^{\phi_{\mathbf{h}}}(u)
\\&\leq
(35+192)\cdot\abs{1\qd\alpha\qd\frac{1}{\tau}}\cdot\mr{c}_{11}\cdot\abs{\frac{{ {\Bigabs{{n}}^{\frac{{d_*}{}}{(s+2)\cdot \beta\cdot (1\qx\beta)^q+{d_*}{}}}}}}{n}}^{\frac{1}{1+1/(s+1)}}+2\cdot 2\alpha\cdot\ykh{\frac{1}{n}}^{\frac{ \beta\cdot(1\qx\beta)^q}{\ykh{1+\frac{1}{s+1}}\cdot \beta\cdot(1\qx\beta)^q+\frac{d_*}{s+1}}}\\
&=\cwu\cdot  \ykh{\frac{1}{n}}^{\frac{ \beta\cdot(1\qx\beta)^q}{\ykh{1+\frac{1}{s+1}}\cdot \beta\cdot(1\qx\beta)^q+\frac{d_*}{s+1}}}. 
\] Since $P$ is arbitrary, the desired inequality \eqref{231208034119} follows immediately. This completes the proof of Theorem \ref{20231204212113}. \end{proof}

\subsection{Proof of Theorem \ref{23102601}}\hypertarget{20251011004523}{}  \label{20251011004523}

The proof of Theorem \ref{23102601} is quite long. It requires several lemmas,  which we use to establish two theorems below, namely, Theorem \ref{23090801} and Theorem \ref{23101701}, corresponding to the cases $s < \infty$ and $s = \infty$ of Theorem \ref{23102601}, respectively.

The following lemma provides upper bounds for the H\"older-$\beta$ norm of a linear combination of disjointly supported translations $u(x - a)$, where $u : [0,1]^d \to \mbR$ is a compactly supported H\"older-$\beta$ smooth function.

\begin{lemma}\label{23092901}Let $d\in\mb N$, $\beta\in(0,\infty)$,  $\rho\in(0,\frac{1}{2})$, $Q\in\mb N$,  \[
	G_{Q,d}:=\setl{(\frac{k_1}{2Q},\ldots,\frac{k_d}{2Q})^\top}{k_1,\ldots,k_d\text{ are odd integers}}\cap[0,1]^d,
	\]    $u\in\mc{C}^{\ceil{\beta-1}}(\mbR^d)$,  $T$ be a map from  $G_{Q,d}$ to $
	\zkh{-1,1}$ , and 
\[
f:\mbR^d\to\mbR, x\mapsto \sum_{a\in G_{Q,d}}\frac{1}{Q^\beta}\cdot T(a)\cdot u(Q\cdot (x-a)). 
\]Suppose 
\[
u(z)=0,\;\forall\;z\in\hkh{x\in\mbR^d\Big|\norm{x}_\infty\geq\rho}. 
\] Then $f\in\mc{C}^{\ceil{\beta-1}}(\mbR^d)$ and $\norm{f}_{\mathbf{H}^\beta(\mbR^d)}\leq3 \cdot\norm{u}_{\mathbf{H}^\beta(\mbR^d)}$. 
\end{lemma}
\begin{proof}
Obviously, $f\in\mc{C}^{\ceil{\beta-1}}(\mbR^d)$. It remains to show $\norm{f}_{\mathbf{H}^\beta(\mbR^d)}\leq 3 \cdot\norm{u}_{\mathbf{H}^\beta(\mbR^d)}$. Let 
\[
u_a:\mbR^d\to\mbR,\;x\mapsto\frac{1}{Q^\beta}\cdot T(a)\cdot u(Q\cdot (x-a)) 
\] for $a\in G_{Q,d}$. Recall the definition of the infinity-norm ball $\mathscr{C}(\;\cdot\;,\;\cdot\;)$ given in \eqref{20250918033935}. Note \[\setl{x\in\mbR^d}{u_a(a)\neq 0}\subset\mathscr{C}(a,\frac{\rho}{Q}),\;\forall\;a\in G_{Q,d},\] and  
\[\mathscr{C}(a,\frac{\rho}{Q})\cap \mathscr{C}(a',\frac{\rho}{Q})=\varnothing,\;\forall\;G_{Q,d}\ni a\neq a'\in G_{Q,d}, 
\] meaning that the supports of the functions $u_a$   are mutually disjoint.  Therefore, 
\beq\label{23092803}
&\norm{\mr{D}^{\bm m}f}_{\mbR^d}=\norm{\sum_{a\in G_{Q,d}}\mr{D}^{\bm m} u_a}_{\mbR^d}\leq \max_{a\in G_{Q,d}}\norm{\mr{D}^{\bm m}u_a}_{\mbR^d}\\&=\max_{a\in G_{Q,d}}\;\sup_{x\in\mbR^d}\abs{\frac{Q^{\norm{\bm m}_1}\cdot T(a)\cdot \mr{D}^{\bm m}u(Q\cdot(x-a))}{Q^\beta}}
\\&=
\max_{a\in G_{Q,d}}\abs{\frac{Q^{\norm{\bm m}_1}\cdot T(a)}{Q^\beta}}\cdot \norm{\mr{D}^{\bm m}u}_{\mbR^d}
\leq
\norm{\mr{D}^{\bm m}u}_{\mbR^d},\;\forall\;\bm m\in\mb N_0^{d,\beta}. 
\eeq We next show
\beq\label{23092802}
\abs{\mr{D}^{\bm m}f(x)-\mr{D}^{\bm m}f(y)}\leq 2\cdot \norm{\raisebox{2ex}{}x-y}_2^{\beta+1-\ceil{\beta}}\cdot \norm{u}_{\mathbf{H}^\beta(\mbR^d)},\;\forall\;\bm m\in\mb N_0^{d,\beta}\setminus\mb N_0^{d,\beta-1},\;x\in\mbR^d,\;y\in\mbR^d. 
\eeq Let $\bm m\in\mb N_0^{d,\beta}\setminus\mb N_0^{d,\beta-1}$, $x\in\mbR^d$, $y\in\mbR^d$ be arbitrary. Since the supports of the functions $u_a$   are mutually disjoint,  we  have  that there exists a set  $I\subset G_{Q,d}$  such that 
$1\leq \#(I)\leq 2$ and 
\[
\mr{D}^{\bm m}u_a(x)=\mr{D}^{\bm m}u_a(y)=0,\;\forall\;a\in G_{Q,d}\setminus I. 
\] Therefore, 
\[
&\abs{\mr{D}^{\bm m}f(x)-\mr{D}^{\bm m}f(y)}
=
\abs{\sum_{a\in G_{Q,d}}\ykh{\mr{D}^{\bm m}u_a(x)-\mr{D}^{\bm m}u_a(y)}}
=
\abs{\sum_{a\in I}\ykh{\mr{D}^{\bm m}u_a(x)-\mr{D}^{\bm m}u_a(y)}}
\\&\leq
\sum_{a\in I}\abs{\mr{D}^{\bm m}u_a(x)-\mr{D}^{\bm m}u_a(y)}
=
\sum_{a\in I}\abs{\frac{Q^{\norm{\bm m}_1}\cdot T(a)\cdot \Bigykh{\mr{D}^{\bm m}u(Q\cdot(x-a))-\mr{D}^{\bm m}u(Q\cdot(y-a))}}{Q^\beta}}
\\&\leq 
\sum_{a\in I}\frac{Q^{\ceil{\beta-1}}}{Q^\beta}\cdot\Bigabs{{\mr{D}^{\bm m}u(Q\cdot(x-a))-\mr{D}^{\bm m}u(Q\cdot(y-a))}}
\\&\leq
\sum_{a\in I}\frac{Q^{\ceil{\beta-1}}}{Q^\beta}\cdot\norm{\raisebox{3ex}{}(Q\cdot(x-a))-(Q\cdot(y-a))}_2^{\beta+1-\ceil{\beta}}\cdot \norm{u}_{\mathbf{H}^\beta(\mbR^d)}
=
\sum_{a\in I} \norm{\raisebox{2ex}{}x-y}_2^{\beta+1-\ceil{\beta}}\cdot \norm{u}_{\mathbf{H}^\beta(\mbR^d)}
\\&=
\#(I)\cdot \norm{\raisebox{2ex}{}x-y}_2^{\beta+1-\ceil{\beta}}\cdot \norm{u}_{\mathbf{H}^\beta(\mbR^d)}
\leq
2\cdot \norm{\raisebox{2ex}{}x-y}_2^{\beta+1-\ceil{\beta}}\cdot \norm{u}_{\mathbf{H}^\beta(\mbR^d)}. 
\] This proves \eqref{23092802}. Combining \eqref{23092803} and \eqref{23092802}, we obtain that 
\[
&\norm{f}_{\mathbf{H}^\beta(\mbR^d)}=\max_{\bm m\in\mb N_0^{d,\beta}}\|\mr D^{\bm m} f\|_{{\mbR^d}}+\max_{\bm m\in \mb N_0^{d,\beta}\setminus \mb N_0^{d,\beta-1}}\abs{\sup_{\mbR^d\ni x\neq y\in\mbR^d}\frac{\abs{\mr D^{\bm m} f(x)-\mr D^{\bm m}f(y)}}{\norm{x-y}_2^{\beta+1-\ceil{\beta}}}}
\\&\leq
\max_{\bm m\in\mb N_0^{d,\beta}}\|\mr D^{\bm m} u\|_{{\mbR^d}}+2\cdot\norm{u}_{\mathbf{H}^\beta(\mbR^d)}
\leq 
\norm{u}_{\mathbf{H}^\beta(\mbR^d)}+2\cdot\norm{u}_{\mathbf{H}^\beta(\mbR^d)}
\leq
3 \cdot\norm{u}_{\mathbf{H}^\beta(\mbR^d)}. 
\] This completes the proof. 
\end{proof}

The next lemma computes the Radon-Nikodym derivative  between two probability measures defined on $[0,1]^d\times\hkh{-1,1}$ which have the same marginal distribution on $[0,1]^d$. 

\begin{lemma}\label{23090103}Let  $\eta_1:[0,1]^d\to[0,1]$ and $\eta_2:[0,1]^d\to[0,1]$ be two Borel measurable functions, and $\mathscr Q$ be a Borel probability measure on $[0,1]^d$. Suppose 
	\[
	\eta_2(x)\in(0,1),\;\forall\;x\in\setl{z\in[0,1]^d}{\eta_1(z)\neq \eta_2(z)}. 
	\]Then $P_{\eta_1,\mathscr Q}$ is absolutely continuous with respect to $P_{\eta_2,\mathscr Q}$, and the Radon-Nikodym derivative $\frac{\mr{d}P_{\eta_1,\mathscr Q}}{\mr{d}P_{\eta_2,\mathscr Q}}$ satisfies \[\frac{\mr{d}P_{\eta_1,\mathscr Q}}{\mr{d}P_{\eta_2,\mathscr Q}}(x,y)=\begin{cases}\frac{\eta_1(x)}{\eta_2(x)},&\;\text{if }\eta_1(x)\neq\eta_2(x)\text{ and }y=+1,\\
		\frac{1-\eta_1(x)}{1-\eta_2(x)},&\;\text{if }\eta_1(x)\neq\eta_2(x)\text{ and }y=-1,\\
		1,&\;\text{if }\eta_1(x)=\eta_2(x). \end{cases}\]\end{lemma}
\begin{proof}	Let $f:[0,1]^d\times\hkh{-1,1}\to[0,\infty),\;(x,y)\mapsto\begin{cases}\frac{\eta_1(x)}{\eta_2(x)},&\;\text{if }\eta_1(x)\neq\eta_2(x)\text{ and }y=+1,\\
		\frac{1-\eta_1(x)}{1-\eta_2(x)},&\;\text{if }\eta_1(x)\neq\eta_2(x)\text{ and }y=-1,\\
		1,&\;\text{if }\eta_1(x)=\eta_2(x). \end{cases}$  Then $f$ is well defined and Borel  measurable. 
	For any Borel subset $A$ of $[0,1]^d\times\hkh{-1,1}$, let \[
	&S_0:=\setl{x\in[0,1]^d}{\eta_1(x)=\eta_2(x)},\] and \[
	&S_1:=\setl{x\in[0,1]^d}{ \eta_1(x)\neq \eta_2(x)}. \]   Then we have that 
	\begin{align*}
		&\int_{A}f(x,y)\mr{d}P_{\eta_2,\mathscr Q}(x,y)\\&=\int_{S_1}\int_{\hkh{1}}f(x,y)\idf_A(x,y)\mr{d}\mathscr M_{\eta_2(x)}(y)\mr{d}{\mathscr Q}(x)+\int_{S_1}\int_{\hkh{-1}}f(x,y)\idf_A(x,y)\mr{d}\mathscr M_{\eta_2(x)}(y)\mr{d}{\mathscr Q}(x)\\
		&\;\;\;\;\;\;\;\;+\int_{S_0}\int_{\hkh{-1,1}}f(x,y)\idf_A(x,y)\mr{d}\mathscr M_{\eta_2(x)}(y)\mr{d}{\mathscr Q}(x)\\
		&=\int_{S_1}\eta_2(x)f(x,1)\idf_A(x,1)\mr{d}{\mathscr Q}(x)+\int_{S_1}(1-\eta_2(x))f(x,-1)\idf_A(x,-1)\mr{d}{\mathscr Q}(x)\\
		&\;\;\;\;\;\;\;\;+\int_{S_0}\int_{\hkh{-1,1}}1\cdot\idf_A(x,y)\mr{d}\mathscr M_{\eta_1(x)}(y)\mr{d}{\mathscr Q}(x)\\
		&=\int_{S_1}\eta_2(x)\cdot\frac{\eta_1(x)}{\eta_2(x)}\cdot\idf_A(x,1)\mr{d}{\mathscr Q}(x)+\int_{S_1}(1-\eta_2(x))\cdot\frac{1-\eta_1(x)}{1-\eta_2(x)}\cdot\idf_A(x,-1)\mr{d}{\mathscr Q}(x)\\
&\;\;\;\;\;\;\;\;+\int_{S_0}\int_{\hkh{-1,1}}\idf_A(x,y)\mr{d}\mathscr M_{\eta_1(x)}(y)\mr{d}{\mathscr Q}(x)\\
		&=\int_{S_1}\int_{\hkh{1}}\idf_{A}(x,y)\mr{d}\mathscr M_{\eta_1(x)}(y)\mr{d}{\mathscr Q}(x)+\int_{S_1}\int_{\hkh{-1}}\idf_A(x,y)\mr{d}\mathscr M_{\eta_1(x)}(y)\mr{d}{\mathscr Q}(x)\\
&\;\;\;\;\;\;\;\;+\int_{S_0}\int_{\hkh{-1,1}}\idf_A(x,y)\mr{d}\mathscr M_{\eta_1(x)}(y)\mr{d}{\mathscr Q}(x)\\		
&=\int_{S_0\cup S_1}\int_{\hkh{-1,1}}\idf_A(x,y)\mr{d}\mathscr M_{\eta_1(x)}(y)\mr{d}{\mathscr Q}(x)=\int_{[0,1]^d}\int_{\hkh{-1,1}}\idf_A(x,y)\mr{d}\mathscr M_{\eta_1(x)}(y)\mr{d}{\mathscr Q}(x)\\
&=\int_{[0,1]^d\times\hkh{-1,1}}\idf_A(x,y)\mr{d}P_{\eta_1,\mathscr Q}(x,y)=P_{\eta_1,\mathscr Q}(A). 
	\end{align*}Since $A$ is arbitrary, we deduce that $P_{\eta_1,\mathscr Q}<<P_{\eta_2,\mathscr Q}$, and $\frac{\mr{d}P_{\eta_1,\mathscr Q}}{\mr{d}P_{\eta_2,\mathscr Q}}=f$. This completes the proof. \end{proof}

\begin{lemma}\label{2301050215}
	Let $\e\in(0,1/8]$, $a_1\in[1/2-\e,1/2+\e]$, and  $a_2\in[1/2-\e,1/2+\e]$. Then
	\begin{align*}
		a_1\log\ykh{\frac{a_1}{a_2}}+(1-a_1)\log\ykh{\frac{1-a_1}{1-a_2}}\leq   18\e^2. 
	\end{align*}
\end{lemma}
\begin{proof}
	By using the elementary inequality
	\beq
	\log(t)\leq t-1,\;\forall\;t\in(0,\infty), 
	\eeq    we immediately obtain \begin{align*}
		&a_1\cdot\log\ykh{\frac{a_1}{a_2}}+(1-a_1)\cdot\log\ykh{\frac{1-a_1}{1-a_2}}
		\leq a_1\cdot\ykh{\frac{a_1}{a_2}-1}+(1-a_1)\cdot\ykh{\frac{1-a_1}{1-a_2}-1}\\
		&=(a_1-a_2)\cdot\ykh{\frac{a_1}{a_2}-\frac{1-a_1}{1-a_2}}=(a_1-a_2)^2\cdot\frac{1}{a_2\cdot(1-a_2)}
		\leq (2\e)^2\cdot\frac{1}{\frac{3}{8}\cdot(1-\frac{3}{8})}\leq \frac{256}{15}\cdot\e^2<18\e^2. 
	\end{align*}This completes the proof.  \end{proof}

The next lemma establishes upper bounds for the Kullback-Leibler divergence (KL divergence) between two probability distributions on $[0,1]^d\times\hkh{-1,1}$ of which the  marginal distributions on $[0,1]^d$ are the same.   Recall that for any two probability distributions with  $\mathscr P$ and $\mathscr Q$ with $\mathbf{dom}(\mathscr P)=\mathbf{dom}(\mathscr Q)$, the KL divergence from $\mathscr Q$ to $\mathscr P$ is defined as \[
\mr{KL}(\mathscr P||\mathscr Q):=\begin{cases}\int \log\ykh{\frac{\mr{d}\mathscr P}{\mr{d}\mathscr Q}}\mr{d}\mathscr P, &\text{if $\mathscr P$ is absolutely continuous with respect to $\mathscr Q$},\\
	+\infty,&\text{otherwise,}\end{cases}
\] where $\frac{\mr{d}\mathscr P}{\mr{d}\mathscr Q}$ denotes the Radon-Nikodym derivative. 

\begin{lemma}\label{23091523}Let $\e\in(0,\frac{1}{8}]$, $\mathscr Q$ be a Borel probability measure on $[0,1]^d$, and  $A$ be a Borel subset of $[0,1]^d$. Suppose Borel measurable functions $\eta_1:[0,1]^d\to [0,1]$ and  $\eta_2:[0,1]^d\to[0,1]$ satisfy 
\[
\eta_1(x)= \eta_2(x),\;\forall\;x\in[0,1]^d\setminus A,
\] and 
\[
\hkh{\eta_1(x),\eta_2(x)}\subset [1/2-\e,1/2+\e],\;\forall\;x\in A. 
\] Then there hold
	\[
	\mr{KL}(P_{\eta_1,\mathscr Q}||P_{\eta_2,\mathscr Q})\leq  18\e^2\cdot\mathscr Q(A).
	\]\end{lemma}
\begin{proof}Let 
	$S_0:=\setl{x\in[0,1]^d}{\eta_1(x)=\eta_2(x)}$ and 	$S_1:=\setl{x\in[0,1]^d}{\eta_1(x)\neq \eta_2(x)}$.  Obviously, $S_1\subset A$,  which implies 
	\[
	\eta_2(x)\in\hkh{\eta_1(x),\eta_2(x)}\subset[1/2-\e,1/2+\e]\subset(0,1),\;\forall\;x\in S_1. 
	\]   Thus we can use Lemma  \ref{23090103} and Lemma  \ref{2301050215} to obtain
	\begin{align*}
		&\mr{KL}(P_{\eta_1,\mathscr Q}||P_{\eta_2,\mathscr Q})\\&=\int_{S_1\times\hkh{-1,1}}\log\ykh{\frac{\eta_1(x)}{\eta_2(x)}\idf_{\hkh{1}}(y)+\frac{1-\eta_1(x)}{1-\eta_2(x)}\idf_{\hkh{-1}}(y)}\mr{d}P_{\eta_1,\mathscr Q}(x,y)+\int_{S_0\times\hkh{-1,1}}\log1 \mr{d}P_{\eta_1,\mathscr Q}(x,y)\\
		&=\int_{S_1}\ykh{\eta_1(x)\log\ykh{\frac{\eta_1(x)}{\eta_2(x)}}+(1-\eta_1(x))\log\ykh{\frac{1-\eta_1(x)}{1-\eta_2(x)}}}\mr{d}\mathscr Q(x)+0\\
		&\leq \int_{S_1}18\e^2\mr{d}\mathscr Q(x)=18\e^2\cdot\mathscr Q(S_1)\leq 18\e^2\cdot \mathscr Q(A). 
\end{align*}This proves the desired result. \end{proof}

The next lemma gives the so-called Varshamov-Gilbert bound. 

\begin{lemma}\label{vgbound} Let $m\in(1,\infty)\cap\mb{Z}$,   $\Omega$ be a set with $\#(\Omega)=m$, and $\hkh{0,1}^\Omega$ be the set of all functions from $\Omega$ to $\hkh{0,1}$.      Then there exists a set $E\subset\hkh{0,1}^\Omega$, such that $\#(E)\geq 1+2^{m/8}$, and 
	\[
	\#\ykh{\setl{x\in \Omega}{f(x)\neq g(x)}}\geq \frac{m}{8},\;\forall\;E\ni f\neq g\in E. 
	\]\end{lemma}
\begin{proof}
	If $m\leq 8$, then  $E=\hkh{0,1}^{\Omega}$ has all the desired properties.   The proof of the case $m>8$  can be found in Lemma 2.9 of \cite{tsybakov2009introduction}.\end{proof}

The next lemma, which is similar to Lemma C.22 of \cite{zhangzihan2023classification}, tells us how to establish lower bounds for the sum of  the excess misclassification errors of a function $f:[0,1]^d\to\mbR$ with respect to two probability measures on $[0,1]^d\times\hkh{-1,1}$ of which the marginal distributions on $[0,1]^d$ are the same.

\begin{lemma}\label{2301051433}
	Let
	\beq\label{2301050326}
	\mc J:[0,1]^2&\to\mbR,\\
	(x,y)&\mapsto\min\hkh{x+y,2-x-y}-\min\hkh{x,1-x}-\min\hkh{y,1-y},
	\eeq $\mathscr Q$ be a Borel probability measure on $[0,1]^d$, and    $\eta_1:[0,1]^d\to[0,1]$ and  $\eta_2:[0,1]^d\to[0,1]$ are two Borel measurable functions. Then there holds 
	\beq\label{082701230105}
	\mc{J}(x,y)=\mc{J}(y,x)\geq 0,\;\forall\;x\in[0,1],\;y\in[0,1],
	\eeq
	\beq\label{082602230105}
	{\textstyle\mc J(\frac{1}{2}-\e,\frac{1}{2}+\e)=\mc J(\frac{1}{2}+\e,\frac{1}{2}-\e)=2\e,\;\forall\;\e\in(0,\frac{1}{2}], }
	\eeq and
	\beq\label{082603230105}
	\int_{[0,1]^d}\mc J(\eta_1(x),\eta_2(x))\mr{d}\mathscr Q(x)\leq\inf_{f\in\mc F_d}\abs{\mc{E}_{P_{\eta_1,\mathscr Q}}(f)+\mc{E}_{P_{\eta_2,\mathscr Q}}(f)}. 
	\eeq
\end{lemma}
\begin{proof}
	For any $x\in[0,1]$ and any $y\in[0,1]$, we have \[x+y\geq \min\hkh{x,1-x}+\min\hkh{y,1-y}\] and \[2-x-y=(1-x)+(1-y)\geq \min\hkh{x,1-x}+\min\hkh{y,1-y},\] meaning that , 
	\begin{align*}
		\min\hkh{x+y,2-x-y}\geq\min\hkh{x,1-x}+\min\hkh{y,1-y}.
	\end{align*} This proves \eqref{082701230105}. For any $\e\in(0,1/2]$, we have
	\begin{align*}
		&{\textstyle\mc J(\frac{1}{2}-\e,\frac{1}{2}+\e)=\mc J(\frac{1}{2}+\e,\frac{1}{2}-\e)=\min\hkh{1,2-1}-\min\hkh{\frac{1}{2}-\e,\frac{1}{2}+\e}-\min\hkh{\frac{1}{2}+\e,\frac{1}{2}-\e}}\\
		&{\textstyle=1-\ykh{\frac{1}{2}-\e}-\ykh{\frac{1}{2}-\e}=2\e}, 
	\end{align*} which yields \eqref{082602230105}. For any Borel measurable function  $f:[0,1]^d\to\mbR$, by Example 2.4 of \cite{steinwart2008support}, we have 
	\begin{align*}
		&\abs{\mc{E}_{P_{\eta_1,\mathscr Q}}(f)+\mc{E}_{P_{\eta_2,\mathscr Q}}(f)}={\mc{E}_{P_{\eta_1,\mathscr Q}}(f)+\mc{E}_{P_{\eta_2,\mathscr Q}}(f)}=\sum_{i=1}^2\mc{E}_{P_{\eta_i,\mathscr Q}}(f)\\
		&=\sum_{i=1}^2\ykh{\mc{R}_{P_{\eta_i,\mathscr Q}}(f)-\int_{[0,1]^d}\min\hkh{\eta_i(x),1-\eta_i(x)}\mr{d}\mathscr Q(x)}\\
		&=\sum_{i=1}^2\int_{[0,1]^d}\ykh{\eta_i(x)\idf_{\hkh{-1}}(\mr{sgn}(f(x)))+(1-\eta_i(x))\idf_{\hkh{1}}(\mr{sgn}(f(x)))-\min\hkh{\eta_i(x),1-\eta_i(x)}}\mr{d}\mathscr Q(x)\\
		&=\int_{[0,1]^d}\sum_{i=1}^2\ykh{\eta_i(x)\idf_{\hkh{-1}}(\mr{sgn}(f(x)))+(1-\eta_i(x))\idf_{\hkh{1}}(\mr{sgn}(f(x)))-\min\hkh{\eta_i(x),1-\eta_i(x)}}\mr{d}\mathscr Q(x)\\
		&\geq \int_{[0,1]^d}\min_{a\in\hkh{-1,1}}\ykh{\sum_{i=1}^2\ykh{\eta_i(x)\idf_{\hkh{-1}}(a)+(1-\eta_i(x))\idf_{\hkh{1}}(a)-\min\hkh{\eta_i(x),1-\eta_i(x)}}}\mr{d}\mathscr Q(x)\\
		&=\int_{[0,1]^d}\mc J(\eta_1(x),\eta_2(x))\mr{d}\mathscr Q(x). 
	\end{align*}  Since $f$ is arbitrary,  \eqref{082603230105} follows immediately. \end{proof}

 The next Lemma \ref{23102003} provides the key inequality for obtaining minimax lower bounds for the excess misclassification error, which is analogous to the one established in Section 2.2 of \cite{tsybakov2009introduction}  for nonparametric  estimation in pseudometric spaces (cf. (2.9) of \cite{tsybakov2009introduction}), and is essentially the same as the one used in (C.144) of \cite{zhangzihan2023classification} (with the excess logistic risk replaced by the excess misclassification error). Note that Lemma \ref{23102003} cannot be derived directly from (2.9) of \cite{tsybakov2009introduction} due to  the absence of the pseudometric.  The original condition which is required to obtain (2.9) of \cite{tsybakov2009introduction}, that is, (2.7) of \cite{tsybakov2009introduction}, is replaced by condition \eqref{23102004} here.  
 
 Recall that 
 $$\mc F_d:=\hkh{\text{Borel measurable functions from $[0,1]^d$ to $\mbR$}}$$ and $\mc{X}_d^n:=\ykh{[0,1]^d\times\hkh{-1,1}}^n.$

\begin{lemma}\label{23102003} Let $d\in\mb N$, $v\in(0,\infty)$, $n\in\mb N$, and $\texttt M\in\mb N$. Suppose $P_0, P_1, \ldots, P_{\texttt M}$ are Borel  probability measures on $[0,1]^d\times\hkh{-1,1}$ such that \beq\label{23102004}
	\inf_{f\in\mc F_d}\ykh{\mc E_{P_j}(f)+\mc E_{P_i}(f)}\geq v,\;\;\;\;\forall\;\;i\neq j\in[0,\texttt M]\cap\mb Z. 
	\eeq  Then \beq\label{23102101}
& 
\frac{v}{2}\cdot\inf\setl{ \sup_{j\in[0,\texttt M]\cap\mb Z}{P_j^{\otimes n}}\ykh{\hkh{z\in \mc{X}_d^n \big|g(z)\neq j}}}{
	\text{$g$ is a   Borel measurable function  from $\mc{X}^n_d$ to $[0,\texttt M]\cap\mb Z$}}\\
	&\leq \inf_{\hat f_n}\sup_{j\in[0,\texttt M]\cap\mb Z}\bm{E}_{{P_j}^{\otimes n}}\zkh{\mc{E}_{P_j}(\hat{f}_n)}
	\eeq where the infimum  is taken over all estimators $\hat f_n:[0,1]^d\to\mbR$  based upon the i.i.d. sample $\hkh{(X_i,Y_i)}_{i=1}^n$ in ${[0,1]^d\times\hkh{-1,1}}$. \end{lemma} 
	\begin{proof} \newcommand{\xdn}{\mc{X}_d^n} Let $\hat f_n:[0,1]^d\to\mbR$ be an arbitrary estimator  based upon the i.i.d. sample $\hkh{(X_i,Y_i)}_{i=1}^n$ in ${[0,1]^d\times\hkh{-1,1}}$. Then there exists a map  ${T_1:\xdn\to\mc{F}_d}$ such that $\hat f_n=T_1(X_1,Y_1,X_2,Y_2,\ldots,X_n,Y_n)$. Define 
		\[
		T_0:\mc F_d\to[0,\texttt M]\cap\mb Z, \;\;f\mapsto \inf\mathop{\arg\min}_{j\in[0,\texttt M]\cap\mb Z}\mc E_{P_j}(f),
		\]  and  $h_*= T_0\circ T_1$. Note that, for each $j\in[0,\texttt M]\cap\mb Z$  and $f\in\mc F_d$,  by \eqref{23102004} we have	\[
		T_0(f)\neq j\Rightarrow \mc{E}_{P_{ T_0(f)}}({f})+\mc{E}_{P_{j}}({f})\geq v\Rightarrow \mc{E}_{P_{j}}({f})+\mc{E}_{P_{j}}({f})\geq v\Rightarrow \mc{E}_{P_{j}}({f})\geq\frac{v}{2}. 
		\] Therefore, 
		\beq
		&\idf_{\mbR\setminus\hkh{j}}(h_*(z))=\idf_{\mbR\setminus\hkh{j}}(T_0( T_1((z))))\\&\leq\idf_{[\frac{v}{2},\infty]}(\mc E_{P_j}(T_1(z))),\;\forall\;z\in\xdn,\;\forall\;j\in[0,\texttt M]\cap\mb Z. 
		\eeq  Consequently, 
		\[
		& \sup_{j\in\zkh{0,\texttt M}\cap\mb Z}\bm{E}_{{P_j}^{\otimes n}}\zkh{\mc{E}_{P_j}(\hat{f}_n)}\\&=\sup_{j\in\zkh{0,\texttt M}\cap\mb Z}\int_{\xdn}\mc{E}_{P_j}( T_1(z))\mr d P_j^{\otimes n}(z)\geq \sup_{j\in\zkh{0,\texttt M}\cap\mb Z}\int_{\xdn}\frac{\idf_{[\frac{v}{2},\infty]}(\mc{E}_{P_j}(T_1(z)))}{2/v}\mr d P_j^{\otimes n}(z)\\&\geq \sup_{j\in\zkh{0,\texttt M}\cap\mb Z}\int_{\xdn}\frac{\idf_{\mbR\setminus\hkh{j}}(h_*(z))}{2/v}\mr d P_j^{\otimes n}(z)=\sup_{j\in\zkh{0,\texttt M}\cap\mb Z}\frac{{P_j^{\otimes n}}\ykh{\hkh{z\in\xdn\big|h_*(z)\neq j}}}{2/v}\\
&\geq \frac{v}{2}\cdot\inf\setl{ \sup_{j\in[0,\texttt M]\cap\mb Z}{P_j^{\otimes n}}\ykh{\hkh{z\in\xdn\big|g(z)\neq j}}}{
	\text{$g$ is a Borel measurable map  from $\mc{X}^n_d$ to $[0,\texttt M]\cap\mb Z$}}.  
		\] Since $\hat f_n$ is arbitrary, the desired inequality \eqref{23102101}  follows immediately.\end{proof}

Here we would like to point out that Lemma \ref{23102003} actually  gives a similar but slightly stronger inequality than that derived in  Proposition 15.1 of \cite{wainwright2019high}. To see this, note that the value of the infimum on the right-hand side of  \eqref{23102101} satisfies 
\[
&\inf\setl{ \sup_{j\in[0,\texttt M]\cap\mb Z}{P_j^{\otimes n}}\ykh{\hkh{z\in\mc{X}_d^n\big|g(z)\neq j}}}{
	\text{$g$ is a Borel measurable function  from $\mc{X}^n_d$ to $[0,\texttt M]\cap\mb Z$}}
	\\&\geq
\inf\setl{ \frac{1}{\texttt M}\cdot\sum_{j=0}^{\texttt M}{P_j^{\otimes n}}\ykh{\hkh{z\in\mc{X}_d^n\big|g(z)\neq j}}}{
	\text{$g$ is a Borel measurable function  from $\mc{X}^n_d$ to $[0,\texttt M]\cap\mb Z$}}
\\&=
\inf\setm{2.3}{ \mb{P}(g(Z)\neq J)}{
	\text{$g$ is a Borel measurable function  from $\mc{X}^n_d$ to $[0,\texttt M]\cap\mb Z$}}
\] with random variables $Z$ and $J$ constructed by the same procedure as the one  described at the beginning of $\text{Section  15.1.2}$ of \cite{wainwright2019high}, that is: 
\begin{enumerate}[label={{ {\hfil(\arabic*)\hfil}} }]%
\item $J$ follows the uniform distribution on $\hkh{0,1,\ldots,\texttt M}$; 
\item Given $J=j$, the conditional distribution of $Z$ is $P_j^{\otimes n}$, for all $j=0,1,\ldots,\texttt M$. 
\end{enumerate} One may refer to Section 2.7.1 of \cite{tsybakov2009introduction} for some more discussions.

Lemma \ref{23091519} below demonstrates how to use Lemma \ref{23102003} to  derive  minimax lower bounds for the excess misclassification error of estimators  by constructing suitable probability measures $P_0,P_1,\ldots,P_{\texttt M}$ ($M>1$)  with the KL divergence between  any two of them being well controlled. For this purpose, we apply Proposition 2.3 of \cite{tsybakov2009introduction} to establish a lower bound for the right-hand side of \eqref{23102101}, which is essentially the same bound as obtained by using the standard Fano's method (see e.g. Proposition 15.12  of \cite{wainwright2019high}). 

\begin{lemma}\label{23091519} Let $d\in\mb N$, $\mc {G}\subset \hkh{\text{Borel probability  measures on $[0,1]^d\times\hkh{-1,1}$}}$,   $\texttt M\in\mb N\cap(1,\infty)$,  $v\in(0,\infty)$, $u\in(0,\infty)$, and  $n\in\mb N$.    Suppose $P_0, P_1, \ldots, P_{\texttt M}$ are probability  measures in $\mc{G}$ satisfying that  \beq\label{23091301}
\frac{1}{\texttt M}\cdot\sum_{j=1}^{\texttt M}\mr{KL}(P_j||P_0)\leq u
	\eeq and 
\beq\label{23091302}
\inf_{f\in\mc F_d}\ykh{\mc E_{P_j}(f)+\mc E_{P_i}(f)}\geq v,\;\;\;\;\forall\;\;\mb Z\cap[0,\texttt M]\ni i\neq j\in[0,\texttt M]\cap\mb Z. 
\eeq	 Then there holds 
 \[
&\inf_{\hat f_n}\sup_{P\in \mc{G}}\bm{E}_{P^{\otimes n}}\zkh{\mc{E}_P(\hat{f}_n)}\geq \frac{v}{4}\cdot \ykh{1-\frac{2\cdot n\cdot u+\sqrt{{{2}\cdot n\cdot u }}}{\log {{\texttt M}}}},
\] where the infimum  is taken over all estimators $\hat f_n:[0,1]^d\to\mbR$  based upon the i.i.d. sample $\hkh{(X_i,Y_i)}_{i=1}^n$ from $P$ on ${[0,1]^d\times\hkh{-1,1}}$.  
\end{lemma}
\begin{proof}  
	
	\newcommand{\xdn}{\mc{X}_d^n} Using Lemma \ref{23102003}, we obtain 
	\beq\label{23102103}
	&\inf_{\hat f_n}\sup_{P\in \mc{G}}\bm{E}_{P^{\otimes n}}\zkh{\mc{E}_P(\hat{f}_n)}\geq\inf_{\hat f_n}\sup_{j\in[0,\texttt M]\cap\mb Z}\bm{E}_{{P_j}^{\otimes n}}\zkh{\mc{E}_{P_j}(\hat{f}_n)}
	\\&\geq 
	\frac{v}{2}\cdot\inf\setl{ \sup_{j\in[0,\texttt M]\cap\mb Z}{P_j^{\otimes n}}\ykh{\hkh{z\in \mc{X}_d^n \big|g(z)\neq j}}}{
		\text{$g$ is a Borel  measurable map from $\mc{X}^n_d$ to $[0,\texttt M]\cap\mb Z$}}. 
	\eeq
	  We then  bound the right hand side of the above inequality.  Indeed, it follows from    \eqref{23091301} that 
\beq\label{23}
\frac{1}{\texttt M}\cdot\sum_{j=1}^{\texttt M}\mr{KL}(P_j^{\otimes n}||P_0^{\otimes n})=n\cdot \frac{1}{\texttt M}\cdot\sum_{j=1}^{\texttt M}\mr{KL}(P_j||P_0)\leq n\cdot u, 
\eeq which, together with  Proposition 2.3 of \cite{tsybakov2009introduction}, yields 
\beq\label{23102102}
& \frac{v}{2}\cdot\inf\setl{ \sup_{j\in[0,\texttt M]\cap\mb Z}{P_j^{\otimes n}}\ykh{\hkh{z\in\xdn\big|g(z)\neq j}}}{
	\text{$g$ is a Borel  measurable function  from $\mc{X}^n_d$ to $[0,\texttt M]\cap\mb Z$}}\\
&\geq\frac{v}{2}\cdot   \sup_{ t\in(0,1)}\ykh{\frac{t\cdot \texttt M}{1+t\cdot \texttt M}\cdot \ykh{\frac{n\cdot u+\sqrt{{\frac{1}{2}\cdot n\cdot u }}}{\log t}+1}}\geq  \frac{v}{2}\cdot   {\frac{\frac{1}{\sqrt{\texttt M}}\cdot \texttt M}{1+\frac{1}{\sqrt{\texttt M}}\cdot \texttt M}\cdot \ykh{\frac{n\cdot u+\sqrt{{\frac{1}{2}\cdot n\cdot u }}}{\log \frac{1}{\sqrt{\texttt M}}}+1}}\\
&=\frac{v}{2}\cdot   {\frac{{\sqrt{\texttt M}}}{1+{\sqrt{\texttt M}}}\cdot \ykh{1-\frac{2\cdot n\cdot u+\sqrt{{{2}\cdot n\cdot u }}}{\log {{\texttt M}}}}}.
\eeq Combining \eqref{23102102} and \eqref{23102103}, we deduce that 
\beq\label{23091402}
&\inf_{\hat f_n}\sup_{P\in \mc{G}}\bm{E}_{P^{\otimes n}}\zkh{\mc{E}_P(\hat{f}_n)}\geq \frac{v}{2}\cdot   \frac{{\sqrt{\texttt M}}}{1+{\sqrt{\texttt M}}}\cdot \max\hkh{1-\frac{2\cdot n\cdot u+\sqrt{{{2}\cdot n\cdot u }}}{\log {{\texttt M}}},0}\\
& \geq\frac{v}{4}\cdot \max\hkh{1-\frac{2\cdot n\cdot u+\sqrt{{{2}\cdot n\cdot u }}}{\log {{\texttt M}}},0}\geq \frac{v}{4}\cdot \ykh{1-\frac{2\cdot n\cdot u+\sqrt{{{2}\cdot n\cdot u }}}{\log {{\texttt M}}}}. 
\eeq This completes the proof.    \end{proof}

 Theorem \ref{23090801} proves the case of $s<\infty$ in Theorem \ref{23102601}.

\begin{theorem}\label{23090801}
	Let $q\in\mb N\cup\hkh{0}$,  $d\in\mb N$, $d_*\in\mb N$, $K\in\mb N$, $\alpha\in(0,\infty)$,  $\tau\in(0,\infty]$, $s\in[0,\infty)$, $\beta\in(0,\infty)$,  and  $\Lambda\in(1,\infty)$. Suppose $d_*\leq K$, $d_*\leq d$,  and  $\idf_{(0,1)}(\alpha)\cdot\idf_{[1,\infty]}(\tau)\neq 1$.   Then there exist  constants $\mr{c}_3\in(0,\infty)$  only depending on $(\beta,\Lambda)$ and  $\mr{c_4}\in(0,\infty)$ only depending on $(d_*,\beta,s,q)$ such that \[
	&\inf_{\hat{f}_n}\sup\setl{{\bm E}_{P^{\otimes n}}\zkh{\mc{E}_P(\hat{f}_n)}\vphantom{\Bigg|}}{P\in\mc{H}^{d,\beta,r,\Lambda}_{q,K,d_*}\cap\mc{T}^{d,s}_{\alpha,\tau}}
	\geq 
	\min\hkh{\alpha,\frac{\Lambda-1}{256\Lambda}}\cdot{\mr{c}_4}\cdot \ykh{\frac{1}{{n}}}^{\frac{(s+1)\cdot\beta\cdot(1\qx\beta)^q}{{d_*+(s+2)\cdot\beta\cdot(1\qx\beta)^q}}} 
	\] for any $r\in[\mr{c}_3,\infty)$ and any   integer  $n>\ykh{\max\hkh{\frac{9\Lambda}{\Lambda-1},\idf_{\ykh{0,\infty}}(s)\cdot\Bigabs{3^{d_*}}^{\frac{1}{ s\cdot\beta\cdot(1\qx\beta)^q}}}}^{d_*+(s+2)\cdot\beta\cdot(1\qx\beta)^q}$,  where the infimum  is taken over all estimators $\hat f_n:[0,1]^d\to\mbR$  based upon the i.i.d. sample $\hkh{(X_i,Y_i)}_{i=1}^n$ from $P$ on ${[0,1]^d\times\hkh{-1,1}}$. 
\end{theorem}

\begin{proof} 
We will prove this theorem by using Lemma \ref{23091519}. To this end, we need to construct suitable probability measures $P_0, P_1, P_2, \ldots P_{\texttt M}$ in $\mc{H}^{d,\beta,r,\Lambda}_{q,K,d_*}\cap\mc{T}^{d,s}_{\alpha,\tau}$  which satisfy the conditions of  Lemma \ref{23091519}. 

Let
\[
\mc{V}_{\beta,\Lambda}:=\setl{w\in\mc{C}^{\ceil{\beta-1}}(\mbR)}{\begin{minipage}{191.36pt}$\mathbf{ran}(w)\subset[0,1]$, and $w(x)=\idf_{[0,\frac{1}{\Lambda}]}(1-x)$ for any $x\in\zkh{0,\frac{\Lambda-1}{2\Lambda}}\cup\zkh{\frac{\Lambda-1}{\Lambda},1}$\end{minipage}}
\] and 
\beq\label{23101301}
\mr{c}_3:=1+\inf\setl{\norm{w}_{\mathbf{H}^{\beta}(\mbR)}}{w\in\mc{V}_{\beta,\Lambda}}. 
\eeq Then %
$\mr{c}_3$ is in $[3,\infty)$  and only depends on $(\beta,\Lambda)$. Let $r$ be an arbitrary number in $[\mr{c}_3,\infty)$ and    $n$ be an arbitrary  integer greater than $\ykh{\max\hkh{\frac{9\Lambda}{\Lambda-1},\idf_{\ykh{0,\infty}}(s)\cdot\Bigabs{3^{d_*}}^{\frac{1}{ s\cdot\beta\cdot(1\qx\beta)^q}}}}^{d_*+(s+2)\cdot\beta\cdot(1\qx\beta)^q}$.    Take ${Q:= \flr{n^{\frac{1}{d_*+(s+2)\cdot\beta\cdot(1\qx\beta)^q}}}+1}$ and  $\texttt M:=\ceil{2^{\ykh{Q^{d_*}\cdot\frac{\Lambda-1}{32\Lambda}}}}$.
Then we have $1/Q<\frac{\Lambda-1}{9\Lambda}$.   Let  
\[
&G_{Q,d_*}:=\setl{(\frac{k_1}{2Q},\ldots,\frac{k_{d_*}}{2Q})^\top}{k_1,\ldots,k_{d_*}\text{ are odd integers}}\cap\zkh{0,{1}}^{d_*},\\
&G_{Q,d_*,\Lambda}:=\setl{(\frac{k_1}{2Q},\ldots,\frac{k_{d_*}}{2Q})^\top}{k_1,\ldots,k_{d_*}\text{ are odd integers, and $\frac{k_1}{2Q}+\frac{1}{2Q}< \frac{\Lambda-1}{2\Lambda}$}}\cap\zkh{0,{1}}^{d_*},   \] and   $\mc J$ be the function defined in   \eqref{2301050326}. It is easy to verify that \[Q^{d_*}\cdot\frac{\Lambda-1}{2\Lambda}\geq \#(G_{Q,d_*,\Lambda})\geq Q^{d_*}\cdot\frac{\Lambda-1}{4\Lambda}\geq Q\cdot \frac{\Lambda-1}{4\Lambda}\geq \frac{9}{4}.\] Then it follows from Lemma  \ref{vgbound} that there exists  $\overline{T}_j:G_{Q,d_*,\Lambda}\to \hkh{0,1}$, $j=0,1,2,\ldots,\texttt M$, such that 
\beq
\#\ykh{\setl{a\in G_{Q,d_*,\Lambda}}{\overline T_j(a)\neq \overline T_l(a)}}\geq
\frac{\#(G_{Q,d_*,\Lambda})}{8}
\geq \frac{Q^{d_*}}{32}\cdot\frac{\Lambda-1}{\Lambda},\;\forall\;j\neq l\in[0,\texttt M]\cap\mb Z. 
\eeq For each $j\in\hkh{0,1,\ldots,\texttt M}$, define $T_j:G_{Q,d_*}\to\hkh{0,1}, a\mapsto \begin{cases}
\overline{T}_j(a),&\text{ if } a\in	G_{Q,d_*,\Lambda},\\
0,&\text{ if } a\in G_{Q,d_*}\setminus G_{Q,d_*,\Lambda}. 
	\end{cases}$

Let 
\[
\mc{U}_{d_*,\beta,s}:=\setr{w\in\mc{C}^{\ceil{\beta-1}}(\mbR^{d_*})}{\begin{minipage}{321.528pt}$\mathbf{ran}(w)\subset[0,1]$, and $w(x)=\idf_{[0,\infty]}(\frac{1}{2}-\frac{1/3}{1+d_*\cdot\idf_{\hkh{0}}(s)}-\norm{x}_\infty)$ for any $x\in\setr{z\in\mbR^{d_*}}{\norm{z}_\infty\in\mbR\setminus\ykh{\frac{1}{2}-\frac{1/3}{1+d_*\cdot\idf_{\hkh{0}}(s)}, \;\frac{1}{2}-\frac{1/4}{1+d_*\cdot\idf_{\hkh{0}}(s)}}}$  \end{minipage}}
\] and 
\[
\mr{c}_2:=1+\inf\setl{\norm{w}_{\mathbf{H}^\beta(\mbR^{d_*})}}{w\in\mc{U}_{d_*,\beta,s}}. 
\] Then $\mr{c}_2\in[2,\infty)$ and there exists a function $u\in\mc{U}_{d_*,\beta,s}$ such that \[
1\leq \norm{u}_{\mathbf{H}^\beta(\mbR^{d_*})}<\mr{c_2}. 
\] Take $\mr{c}_1:=\frac{1}{\mr{4\cdot c_2}}$, and 
\[
f_j:\mbR^{d_*}\to \mbR,\;x\mapsto \sum_{a\in G_{Q,d_*}}\frac{\mr{c}_1}{Q^\beta}\cdot T_j(a)\cdot u(Q\cdot (x-a))
\] for $j=0,1,\ldots,\texttt M$. Then $\mr{c}_1$ only depends on $(d_*,\beta,s)$ and belongs to $(0,1/8]$. 
   Recall the definition of the infinity-norm ball $\mathscr{C}(\;\cdot\;,\;\cdot\;)$ given in \eqref{20250918033935}. 
   Note that
\[
\setl{x\in\mbR^{d_*}}{u(Q\cdot(x-a))\neq 0}\subset \mathscr{C}\ykh{a,\frac{1}{Q}\cdot\Bigykh{ \frac{1}{2}-\frac{1/4}{1+d_*\cdot\idf_{\hkh{0}}(s)}}},\;\forall\;a\in G_{Q,d_*}, 
\]	 and 
\[
\mathscr{C}\ykh{a,\frac{1}{Q}\cdot\Bigykh{ \frac{1}{2}-\frac{1/4}{1+d_*\cdot\idf_{\hkh{0}}(s)}}}\cap \mathscr{C}\ykh{a',\frac{1}{Q}\cdot\Bigykh{ \frac{1}{2}-\frac{1/4}{1+d_*\cdot\idf_{\hkh{0}}(s)}}}=\varnothing,\;\forall\; a\neq a'\in G_{Q,d_*}, 
\] meaning that the supports of the functions $x\mapsto u(Q\cdot (x-a))$ are mutually disjoint. From this we can easily obtain   \beq\label{23090802}
	&0\leq f_j(x)\leq \frac{\mr c_1}{Q^\beta}\cdot\norm{u}_{\mbR^{d_*}}=\frac{\mr{c}_1}{Q^\beta}\leq\frac{1}{8},\;\forall\;x\in\mbR^{d_*},\;\forall\;j\in\hkh{0,1,\ldots,\texttt M}, \eeq\beq\label{23091506}
	&f_j(x)=\frac{\mr c_1}{Q^\beta}\cdot T_j(a),\;\forall\; a\in G_{Q,d_*},\;x\in \mathscr{C}\ykh{a,\frac{1}{Q}\cdot\Bigykh{ \frac{1}{2}-\frac{1/3}{1+d_*\cdot\idf_{\hkh{0}}(s)}}},\;\forall\;j\in\hkh{0,1,\ldots,\texttt M},\eeq
	 and 
	 \beq \label{23091001}
	&\setl{x\in\mbR^{d_*}}{f_j(x)\neq 0}\subset \bigcup_{\substack{a\in G_{Q,d_*},T_j(a)\neq 0}}\mathscr{C}\ykh{a,\frac{1}{Q}\cdot\Bigykh{ \frac{1}{2}-\frac{1/4}{1+d_*\cdot\idf_{\hkh{0}}(s)}}}\\&\subset\bigcup_{a\in G_{Q,d_*,\Lambda},\;\overline{T}_j(a)\neq 0}\mathscr{C}\ykh{a,\frac{1}{Q}\cdot\Bigykh{ \frac{1}{2}-\frac{1/4}{1+d_*}}}\subset\setr{x\in(0,1)^{d_*}}{(x)_1\leq \frac{\Lambda-1}{2\Lambda}},\;\forall\;j\in\hkh{0,1,\ldots,\texttt M}. 
	\eeq Besides, it follows from Lemma \ref{23092901} that, 
\beq
f_j\in\mc{C}^{\ceil{\beta-1}}(\mbR^{d_*}) \text{ and } \norm{f_j}_{\mathbf{H}^\beta(\mbR^{d_*})}\leq3\cdot\mr{c}_1 \cdot\norm{u}_{\mathbf{H}^\beta(\mbR^{d_*})}\leq 3\cdot \mr{c_1}\cdot\mr{c_2}=\frac{3}{4},\;\forall\;j\in\hkh{0,1,\ldots,\texttt M}. 
\eeq

According to \eqref{23101301},  there exists a function $v\in\mc{V}_{\beta,\Lambda}$ such that 
\[
1\leq \norm{v}_{\mathbf{H}^\beta(\mbR)}<\mr{c}_3. 
\] Define 
\[
g:\mbR^{d_*}\to\mbR,\;x\mapsto v((x)_1). 
\] Thus it follows from $v\in\mc{V}_{\beta,\Lambda}$ that 
	\beq\label{23100601}
0\leq 	\inf_{x\in\mbR^{d_*}}g(x)\leq \sup_{x\in\mbR^{d_*}}g(x)\leq 1\leq \norm{g}_{\mathbf H^\beta(\mbR^{d_*})}=\norm{v}_{\mathbf H^\beta(\mbR)}\leq \mr{c}_3
	\eeq and \beq\label{23091003}
	g(x)=\idf_{[\frac{\Lambda-1}{\Lambda},1]}\bigykh{(x)_1},\;\forall\;x\in\setr{z\in[0,1]^{d_*}}{(z)_1\in\zkh{0,\frac{\Lambda-1}{2\Lambda}}\cup\zkh{\frac{\Lambda-1}{\Lambda},1}}. 
	\eeq Then we have from \eqref{23091001} and \eqref{23091003}
	\beq\label{23091002}
	\setl{x\in[0,1]^{d_*}}{f_j(x)\neq 0}\cap \setl{x\in[0,1]^{d_*}}{g(x)\neq 0}=\varnothing,\;\forall\;j\in\hkh{0,1,\ldots,\texttt M}. 
	\eeq Define $\e:=\frac{1}{8}\cdot\ykh{\frac{\mr c_1}{Q^\beta}}^{(1\qx\beta)^q}\in(0,1/8)$. Then it follows from \eqref{23090802}, \eqref{23091001}, \eqref{23091002}, and \eqref{23100601} that 
	\beq\label{23091004}
	0&\leq f_j(x)\cdot \ykh{\frac{1}{4}}^{\ykh{\frac{1}{1\qx\beta}}^q}+g(x)\cdot \ykh{\frac{1}{2}+\e}^{\ykh{\frac{1}{1\qx\beta}}^q}\leq f_j(x)\cdot {\frac{1}{4}}+g(x)\cdot \ykh{\frac{1}{2}+\e} \\
	&\leq \max\hkh{\frac{1}{4}\cdot\frac{\mr{c_1}}{Q^\beta},\frac{1}{2}+\e}=\frac{1}{2}+\e=1-\abs{\frac{1}{2}-\e}<1,\;\forall\;x\in[0,1]^{d_*},\;j\in\hkh{0,1,\ldots,\texttt M}. 
	\eeq Next, for each $j\in\hkh{0,1,\ldots,\texttt M}$, we define 
	\[
	h_{0,j}:[0,1]^d\to \mbR,\; (x_1,\ldots,x_d)^\top\mapsto  f_j(x_1,\ldots,x_{d_*})\cdot \ykh{\frac{1}{4}}^{\ykh{\frac{1}{1\qx\beta}}^q}+g(x_1,\ldots,x_{d_*})\cdot \ykh{\frac{1}{2}+\e}^{\ykh{\frac{1}{1\qx\beta}}^q}, 
	\] if $q=0$; and define 	\[
	h_{0,j}:[0,1]^d&\to \mbR^K,\;\\ (x_1,\ldots,x_d)^\top&\mapsto \ykh{f_j(x_1,\ldots,x_{d_*})\cdot \ykh{\frac{1}{4}}^{\ykh{\frac{1}{1\qx\beta}}^q}+g(x_1,\ldots,x_{d_*})\cdot \ykh{\frac{1}{2}+\e}^{\ykh{\frac{1}{1\qx\beta}}^q},0,0,\ldots,0}^\top
	\] if $q>0$. From \eqref{23091004} we see 
	\[
	\mathbf{ran}(h_{0,j})\subset[0,1]^{K\cdot\idf_{\mb N}(q)+ \idf_{\hkh{0}}(q)},\;\forall\;j\in\hkh{0,1,\ldots,\texttt M}. 
	\] Define
	\[
	\mr u_0:[0,1]^{d_*}\to[0,1],\;x\mapsto \abs{(x)_1}^{(1\qx\beta)}, 
	\] and  \[
	\mr u_1:[0,1]^{d_*}\to\mbR,\;x\mapsto\frac{1}{2}-\e+ \abs{(x)_1}^{(1\qx\beta)}. 
	\] Then, for each $i\in\mb N$ and $j\in\hkh{0,1,\ldots,\texttt M}$,  define \[h_{i,j}:[0,1]^K\to[0,1],\;x\mapsto \mr u_0\bigykh{(x)_1,(x)_2,\ldots,(x)_{d_*}}\] if $i=q>0$; and define 
	\[
	h_{i,j}:[0,1]^K\to[0,1]^K,\;x\mapsto \ykh{\mr u_0\bigykh{(x)_1,(x)_2,\ldots,(x)_{d_*}},0,0,\ldots,0,0}^\top
	\] otherwise. Finally, for each $j\in\hkh{0,1,\ldots,\texttt M}$, define \[
	\eta_j:[0,1]^d\to\mbR,\;x\mapsto 1/2-\e+h_{q,j}\circ h_{q-1,j}\circ\cdots\circ h_{3,j}\circ h_{2,j}\circ h_{1,j}\circ h_{0,j}(x). 
	\] $\eta_j$ is well-defined because 
	\[
	\begin{minipage}{285.6pt}
	\begin{enumerate}[label={{ {\hfil(\roman*)\hfil}} }]%
		\item   ${\mathbf{{dom}}}(h_{i,j})=[0,1]^K$ if $0<i\leq q$,  and ${\mathbf{{dom}}}(h_{0,j})=[0,1]^d$;
		\item $\mathbf{ran}(h_{i,j})\subset[0,1]^K$ if $0 \leq i<q$, and  $\mathbf{ran}(h_{q,j})\subset\mbR$
		\end{enumerate}
	\end{minipage}
	\] for all $j\in\hkh{0,1,\ldots,\texttt M}$. Recall that $r\geq\mr{c}_3\geq 3$ and $0<\e<1/8$.  Thus it is easy to show
	\[
	\hkh{\mr u_0,\mr u_1}\subset \mc B^\beta_3([0,1]^{d_*})\subset \mc B^\beta_r([0,1]^{d_*}),
	\]and 
	\[
	&\Big\|f_j\Big\|_{\mathbf H^\beta([0,1]^{d_*})}\cdot \ykh{\frac{1}{4}}^{\ykh{\frac{1}{1\qx\beta}}^q}+\Big\|g\Big\|_{\mathbf H^\beta([0,1]^{d_*})}\cdot \ykh{\frac{1}{2}+\e}^{\ykh{\frac{1}{1\qx\beta}}^q}\\
	&\leq  \abs{\frac{1}{2}-\e}+\Big\|f_j\Big\|_{\mathbf H^\beta([0,1]^{d_*})}\cdot \ykh{\frac{1}{4}}^{\ykh{\frac{1}{1\qx\beta}}^q}+\Big\|g\Big\|_{\mathbf  H^\beta([0,1]^{d_*})}\cdot \ykh{\frac{1}{2}+\e}^{\ykh{\frac{1}{1\qx\beta}}^q}\\&\leq\abs{\frac{1}{2}-\e}+ \Big\|f_j\Big\|_{\mathbf H^\beta([0,1]^{d_*})}\cdot \ykh{\frac{1}{4}}^{\ykh{\frac{1}{1\qx\beta}}^q}+\mr{c}_3\cdot \ykh{\frac{1}{2}+\e}^{\ykh{\frac{1}{1\qx\beta}}^q}\\
	&\leq \abs{\frac{1}{2}-\e}+  \ykh{\frac{1}{4}}^{\ykh{\frac{1}{1\qx\beta}}^q}+\mr{c}_3\cdot \ykh{\frac{1}{2}+\e}^{\ykh{\frac{1}{1\qx\beta}}^q}\leq \abs{\frac{1}{2}-\e}+{\frac{1}{4}}+\mr{c}_3\cdot \ykh{\frac{1}{2}+\e}\leq\mr{c}_3\leq r,\;\forall\;j\in\hkh{0,1,\ldots,\texttt M}. 
	\] Therefore, we must have 
	\beq\label{23091504}
	\eta_j\in \mc G_d^{\mathbf{CH}}(q, K, d_*, \beta,r),\;\forall\;j\in\hkh{0,1,\ldots,\texttt M}. 
	\eeq An elementary calculation gives
\beq\label{23091201}
&0<\frac{1}{2}-\e\leq \eta_j(x)\\&=\frac{1}{2}-\e+\abs{f_j\bigykh{(x)_1,\ldots,(x)_{d_*}}\cdot \ykh{\frac{1}{4}}^{\ykh{\frac{1}{1\qx\beta}}^q}+g\big((x)_1,\ldots,(x)_{d_*}\big)\cdot \ykh{\frac{1}{2}+\e}^{\ykh{\frac{1}{1\qx\beta}}^q}}^{(1\qx\beta)^q}\\
&\xlongequal{\because\eqref{23091002}}\frac{1}{2}-\e+\frac{1}{4}\cdot\Bigabs{f_j\bigykh{(x)_1,\ldots,(x)_{d_*}}}^{(1\qx\beta)^q}+\ykh{\frac{1}{2}+\e}\cdot\Bigabs{g\big((x)_1,\ldots,(x)_{d_*}\big)}^{(1\qx\beta)^q}\\
&\leq \frac{1}{2}-\e+\sup_{z\in[0,1]^{d_*}}\ykh{\frac{1}{4}\cdot\bigabs{f_j\ykh{z}}^{(1\qx\beta)^q}+\ykh{\frac{1}{2}+\e}\cdot\bigabs{g(z)}^{(1\qx\beta)^q}}\\
&\xlongequal[\text{ and }{\mathbf{ran}(g)\subset[0,1]}]{\because\eqref{23091002},\eqref{23090802},\eqref{23091003}}\frac{1}{2}-\e+\ykh{\frac{1}{2}+\e}\cdot \bigabs{1}^{(1\qx\beta)^q}=\frac{1}{2}-\e+\ykh{\frac{1}{2}+\e}\\&=1,\;\forall\;x\in[0,1]^d,\;j\in\hkh{0,1,\ldots,\texttt M}.  
\eeq From this we can easily derive 
\beq\label{23091221}
&\eta_j(x)\xlongequal{\because\eqref{23091201}} \frac{1}{2}-\e+\frac{1}{4}\cdot\Bigabs{f_j\bigykh{(x)_1,\ldots,(x)_{d_*}}}^{(1\qx\beta)^q}+\ykh{\frac{1}{2}+\e}\cdot\Bigabs{g\big((x)_1,\ldots,(x)_{d_*}\big)}^{(1\qx\beta)^q}\\
&\xlongequal{\because\eqref{23091003}}\frac{1}{2}-\e+\frac{1}{4}\cdot\Bigabs{f_j\bigykh{(x)_1,\ldots,(x)_{d_*}}}^{(1\qx\beta)^q}\leq \frac{1}{2}-\e+\frac{1}{4}\cdot\abs{\frac{\mr c_1}{Q^\beta}}^{(1\qx\beta)^q}=\frac{1}{2}-\e+2\e\\
&=\frac{1}{2}+\e,\;\forall\;x\in\setr{z\in[0,1]^d}{(z)_1\in\zkh{0,\frac{\Lambda-1}{2\Lambda}}},\;j\in\hkh{0,1,\ldots,\texttt M},
\eeq and 
\beq\label{23091220}
&\eta_j(x)\xlongequal{\because\eqref{23091201}} \frac{1}{2}-\e+\frac{1}{4}\cdot\Bigabs{f_j\bigykh{(x)_1,\ldots,(x)_{d_*}}}^{(1\qx\beta)^q}+\ykh{\frac{1}{2}+\e}\cdot\Bigabs{g\big((x)_1,\ldots,(x)_{d_*}\big)}^{(1\qx\beta)^q}\\
&\xlongequal{\because\eqref{23091001}}\frac{1}{2}-\e+\ykh{\frac{1}{2}+\e}\cdot\Bigabs{g\big((x)_1,\ldots,(x)_{d_*}\big)}^{(1\qx\beta)^q}\\
&\xlongequal{\because\eqref{23091001}}  \frac{1}{2}-\e+\frac{1}{4}\cdot\Bigabs{f_l\bigykh{(x)_1,\ldots,(x)_{d_*}}}^{(1\qx\beta)^q}+\ykh{\frac{1}{2}+\e}\cdot\Bigabs{g\big((x)_1,\ldots,(x)_{d_*}\big)}^{(1\qx\beta)^q}\\
&\xlongequal{\because\eqref{23091201}}\eta_l(x),\;\forall\;x\in\setr{z\in[0,1]^d}{(z)_1\notin\zkh{0,\frac{\Lambda-1}{2\Lambda}}},\;j\in\hkh{0,1,\ldots,\texttt M},\;l\in\hkh{0,1,\ldots,\texttt M}. 
\eeq 
Let 
\[
D_a:=\setr{x\in[0,1]^d}{\bigykh{(x)_1,(x)_2,\ldots,(x)_{d_*}}^\top\in\mathscr{C}\big(a,\frac{1}{2Q}-\frac{1/3}{Q+Q\cdot d_*\cdot\idf_{\hkh{0}}(s)}\big)},\;\forall\;a\in \mbR^{d_*}, 
\] and 
\[
&A_1:=\setr{x\in[0,1]^d}{(x)_1\leq\frac{\Lambda-1}{2\Lambda}},\\
&A_2:=\setr{x\in[0,1]^d}{(x)_1\geq\frac{\Lambda-1}{\Lambda}},\\
&
A_0:=\setr{x\in[0,1]^d}{\bigykh{(x)_1,(x)_2,\ldots,(x)_{d_*}}^\top\in\bigcup\limits_{a\in G_{Q,d_*,\Lambda}}\mathscr{C}\big(a,\frac{1}{2Q}-\frac{1/3}{Q+Q\cdot d_*\cdot\idf_{\hkh{0}}(s)}\big)}.
\] 
Obviously, 
\beq\label{23091518}
D_a\cap D_b=\varnothing,\;\forall\;G_{Q,d_*,\Lambda}\ni a\neq b\in G_{Q,d_*,\Lambda}, 
\eeq and  \beq\label{23091501}
\bigcup\limits_{a\in G_{Q,d_*,\Lambda}}D_a=A_0\subset A_1\subset [0,1]^d\setminus A_2.
\eeq Moreover, we have by \eqref{23091201}, \eqref{23091003}, and \eqref{23091002} 
\beq\label{23091507}
&\eta_j(x)=
\frac{1}{2}-\e+\frac{1}{4}\cdot\Bigabs{f_j\bigykh{(x)_1,\ldots,(x)_{d_*}}}^{(1\qx\beta)^q}+\ykh{\frac{1}{2}+\e}\cdot\Bigabs{g\big((x)_1,\ldots,(x)_{d_*}\big)}^{(1\qx\beta)^q}
\\&=
\frac{1}{2}-\e+\frac{1}{4}\cdot\Bigabs{0}^{(1\qx\beta)^q}+\ykh{\frac{1}{2}+\e}\cdot\Bigabs{1}^{(1\qx\beta)^q}=1,\;\forall\;x\in A_2,\;j\in\hkh{0,1,\ldots,\texttt M}, 
\eeq and by \eqref{23091201}, \eqref{23091501}, \eqref{23091506}, and \eqref{23091003},
\beq\label{23091508}
&\eta_j(x)
=
\frac{1}{2}-\e+\frac{1}{4}\cdot\Bigabs{f_j\bigykh{(x)_1,\ldots,(x)_{d_*}}}^{(1\qx\beta)^q}+\ykh{\frac{1}{2}+\e}\cdot\Bigabs{g\big((x)_1,\ldots,(x)_{d_*}\big)}^{(1\qx\beta)^q}
\\
&=
\frac{1}{2}-\e+\frac{1}{4}\cdot\Bigabs{\frac{\mr c_1}{Q^\beta}\cdot T_j(a)}^{(1\qx\beta)^q}+\ykh{\frac{1}{2}+\e}\cdot\Bigabs{0}^{(1\qx\beta)^q}
\\
&=
\frac{1}{2}-\e+2\e\cdot \overline{T}_j(a),\;\forall\;j\in\hkh{0,1,\ldots,\texttt M},\;\forall\;a\in G_{Q,d_*,\Lambda}, \;\forall\;x\in D_a, 
\eeq which means that
\beq\label{23091509}
\abs{2\eta_j(x)-1}=2\e,\;\forall\;j\in\hkh{0,1,\ldots,\texttt M},\;\forall\;x\in  A_0. 
\eeq The functions $\eta_0,\eta_1,\ldots,\eta_{\texttt M}$ will  serve as the conditional class probability function of the probability measures $P_0, P_1,\ldots,P_{\texttt M}$.

We then construct a suitable probability measure $\mathscr Q_n$ which will serve as the marginal distribution of $P_j$ on $[0,1]^d$. Define  \beq\label{23091502}
V_n:[0,1]^d&\to[-\infty,\infty],\\z&\mapsto \left\{\ba
&\frac{\min\hkh{\alpha,\frac{\Lambda-1}{256\Lambda}}\cdot 2^s\cdot\e^s}{\int_{\mbR^d}\idf_{A_0}(x)\mr{d}x},&&\text{ if }z\in A_0,\\
&0,&&\text{ if }z\in[0,1]^d\setminus (A_0\cup A_2),\\
&\Lambda\cdot\ykh{1-\min\hkh{\alpha,\frac{\Lambda-1}{256\Lambda}}\cdot 2^s\cdot\e^s},&&\text{ otherwise}. 
\ea\right.
\eeq Recall that \[0&<\e=\frac{1}{8}\cdot\ykh{\frac{\mr c_1}{Q^\beta}}^{(1\qx\beta)^q}\leq \frac{1}{8}\cdot\ykh{\frac{1}{Q^\beta}}^{(1\qx\beta)^q}=\frac{1}{8}\cdot\ykh{\flr{n^{\frac{1}{d_*+(s+2)\cdot\beta\cdot(1\qx\beta)^q}}}+1}^{-\beta\cdot(1\qx\beta)^q}\\
&\leq \frac{1}{8}\cdot\ykh{{n^{\frac{1}{d_*+(s+2)\cdot\beta\cdot(1\qx\beta)^q}}}}^{-\beta\cdot(1\qx\beta)^q}=\frac{1}{8}\cdot{\ykh{\frac{1}{n}}^{\frac{\beta\cdot(1\qx\beta)^q}{d_*+(s+2)\cdot\beta\cdot(1\qx\beta)^q}}}\] and
\[
&n>\ykh{\max\hkh{\frac{9\Lambda}{\Lambda-1},\idf_{\ykh{0,\infty}}(s)\cdot\Bigabs{3^{d_*}}^{\frac{1}{ s\cdot\beta\cdot(1\qx\beta)^q}}}}^{d_*+(s+2)\cdot\beta\cdot(1\qx\beta)^q}
\\&\geq \idf_{(0,\infty)}(s)\cdot \Bigabs{3^{d_*}}^{\frac{d_*+(s+2)\cdot\beta\cdot(1\qx\beta)^q}{ s\cdot\beta\cdot(1\qx\beta)^q}}. 
\]Therefore, 
\[
0&<\min\hkh{\alpha,\frac{\Lambda-1}{256\Lambda}}\cdot 2^s\cdot\e^s
\leq \min\hkh{\alpha,\frac{\Lambda-1}{256\Lambda}}\cdot {\ykh{\frac{1}{n}}^{\frac{s\cdot\beta\cdot(1\qx\beta)^q}{d_*+(s+2)\cdot\beta\cdot(1\qx\beta)^q}}}\\
&\leq
\min\hkh{\alpha,\frac{\Lambda-1}{256\Lambda}}\cdot \max\hkh{\frac{1}{3^{d_*}},\idf_{\hkh{0}}(s)}
\leq  \frac{\Lambda-1}{4\Lambda}\cdot \max\hkh{\frac{1}{ 3^{d_*}},\frac{\idf_{\hkh{0}}(s)}{2}}
\\&= \frac{\Lambda-1}{4\Lambda}\cdot Q^{d_*}\cdot\frac{1}{Q^{d_*}}\cdot\max\hkh{\frac{1}{ 3^{d_*}},\frac{\idf_{\hkh{0}}(s)}{2}}
\leq 
\#(G_{Q,d_*,\Lambda})\cdot\frac{1}{Q^{d_*}}\cdot\max\hkh{\frac{1}{ 3^{d_*}},\frac{\idf_{\hkh{0}}(s)}{2}}
\\&= 
\sum_{a\in G_{Q,d_*,\Lambda}}\frac{1}{Q^{d_*}}\cdot\max\hkh{\frac{1}{ 3^{d_*}},\frac{\idf_{\hkh{0}}(s)}{2}}
 \leq
\sum_{a\in G_{Q,d_*,\Lambda}} \ykh{\frac{1}{Q}-\frac{2/3}{Q+Q\cdot d_*\cdot\idf_{\hkh{0}}(s)}}^{d_*}
\\&=
\sum_{a\in G_{Q,d_*,\Lambda}} \int_{\mbR^{d}}\idf_{D_a}(x)\mr{d}x
=
\int_{\mbR^d}\idf_{A_0}(x)\mr{d}x
\leq
\int_{\mbR^d}\idf_{[0,1]^d}(x)\mr{d}x
= 1. 
\]
 Thus \beq\label{23091503}
0\leq V_n(z)\leq \Lambda, \;\forall\;z\in[0,1]^d. \eeq Let $\mathscr Q_n$ be the Borel  measure on $[0,1]^d$  given by 
\[
\mathscr Q_n(\Omega):=\int_{[0,1]^d}\idf_\Omega(z)\cdot V_n(z)\mr{d}z,\;\forall\; \text{Borel measurable set    $\Omega\subset[0,1]^d$}.
\] Note from \eqref{23091501} and \eqref{23091502}  that 
\beq\label{23091510}
&\mathscr Q_n([0,1]^d\setminus A_2)=\int_{[0,1]^d} V_n(z)\cdot \idf_{A_0}(z)\mr{d}z+\int_{[0,1]^d}V_n(z)\cdot\idf_{[0,1]^d\setminus (A_0\cup A_2)}(z)\mr{d}z
\\
&=\int_{[0,1]^d} \frac{\min\hkh{\alpha,\frac{\Lambda-1}{256\Lambda}}\cdot 2^s\cdot\e^s}{\int_{\mbR^d}\idf_{A_0}(x)\mr{d}x}\cdot \idf_{A_0}(z)\mr{d}z+\int_{[0,1]^d}0\cdot\idf_{[0,1]^d\setminus (A_0\cup A_2)}(z)\mr{d}z\\
&=\min\hkh{\alpha,\frac{\Lambda-1}{256\Lambda}}\cdot 2^s\cdot\e^s=1-\int_{[0,1]^d}\Lambda\cdot\ykh{1-\min\hkh{\alpha,\frac{\Lambda-1}{256\Lambda}}\cdot 2^s\cdot\e^s}\cdot \idf_{A_2}(z)\mr{d}z\\
&=1-\int_{[0,1]^d}V_n(z)\cdot\idf_{A_2}(z)\mr{d}z=1-\mathscr Q_n(A_2),  
\eeq meaning that   $\mathscr Q_n$ is a probability measure. We then deduce from \eqref{23091503} that 
\beq\label{23091505}
\mathscr Q_n\in\mc{M}_{\Lambda,d}. 
\eeq

 Take  $P_j:=P_{\eta_j,\mathscr Q_n}$ for $j=0,1,\ldots,\texttt M$. We have to show $P_0,P_1,\ldots,P_{\texttt M}$ belong to $\mc{H}^{d,\beta,r,\Lambda}_{q,K,d_*}\cap\mc{T}^{d,s}_{\alpha,\tau}$.  It follows from \eqref{23091504},  \eqref{23091201} and \eqref{23091505} that 
 \beq\label{23091514}
 P_j\in \mc{H}^{d,\beta,r,\Lambda}_{q,K,d_*},\;\forall\;j\in\hkh{0,1,\ldots,\texttt M}. 
 \eeq We next show that  $P_0,P_1,\ldots,P_{\texttt M}$ belong to $\mc{T}^{d,s}_{\alpha,\tau}$. 
  Indeed, we have by \eqref{23091509} and \eqref{23091507}
 \beq\label{23091511}
  {\setr{x\in A_0}{\vphantom{\big|}\abs{2\eta_j(x)-1}\leq t}}=\varnothing,\;\forall\;t\in(-\infty,2\e),\;j\in\hkh{0,1,\ldots,\texttt M}, 
 \eeq and 
 \beq\label{23091512}
  {\setr{x\in A_2}{\vphantom{\big|}\abs{2\eta_j(x)-1}\leq t}}=\varnothing,\;\forall\;t\in(-\infty,1),\;j\in\hkh{0,1,\ldots,\texttt M}. 
 \eeq Let  $j\in\hkh{0,1,\ldots,\texttt M}$ and  $t\in(0,\tau]\cap\mbR$ be arbitrary. If $t<2\e<1$, then we have by \eqref{23091511}, \eqref{23091512} and \eqref{23091502} that \[
 &
 P_j\ykh{\setl{(x,y)\in[0,1]^d\times\hkh{-1,1}}{\abs{2P_j(\hkh{1}|x)-1}\leq t}}
 \\&=
 \mathscr Q_n \ykh{\setl{x\in[0,1]^d}{\abs{2\eta_j(x)-1}\leq t}}=
 \mathscr Q_n \ykh{\setl{x\in[0,1]^d\setminus(A_0\cup A_2)}{\abs{2\eta_j(x)-1}\leq t}}
 \\
 &\leq 
 \mathscr Q_n \ykh{\hkh{[0,1]^d\setminus(A_0\cup A_2)}}
 =
 \int_{[0,1]^d}V_n(x)\cdot\idf_{[0,1]^d\setminus(A_0\cup A_2)}(x)\mr{d}x
 \\
 &=
 \int_{[0,1]^d}0\cdot\idf_{[0,1]^d\setminus(A_0\cup A_2)}(x)\mr{d}x
 =
 0
 \leq
 \alpha\cdot t^s; 
 \] If $2\e\leq t<1$, then it follows from \eqref{23091512} and \eqref{23091510}  that 
\[
&
P_j\ykh{\setl{(x,y)\in[0,1]^d\times\hkh{-1,1}}{\abs{2P_j(\hkh{1}|x)-1}\leq t}}
\\&=
\mathscr Q_n \ykh{\setl{x\in[0,1]^d}{\abs{2\eta_j(x)-1}\leq t}}
=
\mathscr Q_n \ykh{\setl{x\in[0,1]^d\setminus A_2}{\abs{2\eta_j(x)-1}\leq t}}
\\
&\leq
\mathscr Q_n \ykh{{[0,1]^d\setminus A_2}}
=
\min\hkh{\alpha,\frac{\Lambda-1}{256\Lambda}}\cdot 2^s\cdot\e^s\leq \alpha\cdot (2\e)^s\leq \alpha\cdot t^s;
\]	If $t\geq 1$, then it follows from $t\in(0,\tau]\cap\mbR$ and $\idf_{(0,1)}(\alpha)\cdot\idf_{[1,\infty]}(\tau)\neq 1$ that $\tau\in[1,\infty]$ and  $\alpha\notin(0,1)$, meaning that  \[
&
P_j\ykh{\setl{(x,y)\in[0,1]^d\times\hkh{-1,1}}{\abs{2P_j(\hkh{1}|x)-1}\leq t}}
\leq 
1
\leq
\alpha
=
\alpha\cdot 1^s
\leq
\alpha\cdot t^s. \] In conclusion, we always have $P_j\ykh{\setl{(x,y)\in[0,1]^d\times\hkh{-1,1}}{\abs{2P_j(\hkh{1}|x)-1}\leq t}}\leq \alpha\cdot t^s$. Since $j$ and $t$ are arbitrary, we deduce that 
\beq\label{23091513}
P_j\in\mc{T}^{d,s}_{\alpha,\tau},\;\forall\;j\in\hkh{0,1,\ldots,\texttt M}. 
\eeq	Combining \eqref{23091514} and \eqref{23091513}, we obtain 
\beq\label{230915163}
P_j\in\mc{H}^{d,\beta,r,\Lambda}_{q,K,d_*}\cap\mc{T}^{d,s}_{\alpha,\tau},\;\forall\;j\in\hkh{0,1,\ldots,\texttt M}.
\eeq

We then establish the inequality of the form \eqref{23091302}. Indeed, it follows from Lemma \ref{2301051433} that 
\[
&\inf_{f\in\mc F_d}\ykh{\mc{E}_{P_{j}}(f)+\mc{E}_{P_{l}}(f)}
\geq
\int_{[0,1]^d}\mc J(\eta_j(x),\eta_l(x))\mr{d}\mathscr Q_n(x)
\geq 
\int_{[0,1]^d}\mc J(\eta_j(x),\eta_l(x))\idf_{A_0}(x)\mr{d}\mathscr Q_n(x)
\\
&\xlongequal{\because\eqref{23091501}\text{ and }\eqref{23091518}}
\sum_{a\in G_{Q,d_*,\Lambda}}\int_{[0,1]^d}\mc J(\eta_j(x),\eta_l(x))\idf_{D_a}(x)\mr{d}\mathscr Q_n(x)
\\&\xlongequal{\because\eqref{23091508}}
\sum_{a\in G_{Q,d_*,\Lambda}}\int_{[0,1]^d}\mc J\ykh{\frac{1}{2}-\e+2\e\cdot \overline{T}_j(a), \frac{1}{2}-\e+2\e\cdot \overline{T}_l(a)}\idf_{D_a}(x)\mr{d}\mathscr Q_n(x)
\\
&\geq
\sum_{\substack{a\in G_{Q,d_*,\Lambda}\\\overline{T}_j(a)\neq \overline{T}_l(a)}}\int_{[0,1]^d}\mc J\ykh{\frac{1}{2}-\e+2\e\cdot \overline{T}_j(a), \frac{1}{2}-\e+2\e\cdot \overline{T}_l(a)}\idf_{D_a}(x)\mr{d}\mathscr Q_n(x)
\\
&=
\sum_{\substack{a\in G_{Q,d_*,\Lambda}\\\overline{T}_j(a)\neq \overline{T}_l(a)}}\int_{[0,1]^d}\mc J\ykh{\frac{1}{2}-\e, \frac{1}{2}+\e}\idf_{D_a}(x)\mr{d}\mathscr Q_n(x)
=
\sum_{\substack{a\in G_{Q,d_*,\Lambda}\\\overline{T}_j(a)\neq \overline{T}_l(a)}}\int_{[0,1]^d}2\e\idf_{D_a}(x)V_n(x)\mr{d}x
\\
&\xlongequal{\because\eqref{23091501}\text{ and }\eqref{23091502}}
\sum_{\substack{a\in G_{Q,d_*,\Lambda}\\\overline{T}_j(a)\neq \overline{T}_l(a)}}\int_{[0,1]^d}2\e\idf_{D_a}(x)\cdot \frac{\min\hkh{\alpha,\frac{\Lambda-1}{256\Lambda}}\cdot 2^s\cdot\e^s}{\int_{\mbR^d}\idf_{A_0}(z)\mr{d}z}\mr{d}x
\\&\xlongequal{\because \eqref{23091501}\text{ and }\eqref{23091518}}
 \frac{\min\hkh{\alpha,\frac{\Lambda-1}{256\Lambda}}\cdot 2^{s+1}\cdot\e^{s+1}}{\sum\limits_{a\in G_{Q,d_*,\Lambda}}\int_{[0,1]^d}\idf_{D_a}(z)\mr{d}z}\cdot{\sum\limits_{\substack{a\in G_{Q,d_*,\Lambda}\\\overline{T}_j(a)\neq \overline{T}_l(a)}}\int_{[0,1]^d}\idf_{D_a}(x)\mr{d}x}
 \\&=
 \frac{\min\hkh{\alpha,\frac{\Lambda-1}{256\Lambda}}\cdot 2^{s+1}\cdot\e^{s+1}}{\#\ykh{G_{Q,d_*,\Lambda}}}\cdot\#\ykh{\setr{a\in G_{Q,d_*,\Lambda}}{\overline{T}_j(a)\neq \overline{T}_l(a)}}
 \\&\geq
 \frac{\min\hkh{\alpha,\frac{\Lambda-1}{256\Lambda}}\cdot 2^{s+1}\cdot\e^{s+1}}{\#\ykh{G_{Q,d_*,\Lambda}}}\cdot\frac{\#(G_{Q,d_*,\Lambda})}{8}
 \\&=
{\min\hkh{\alpha,\frac{\Lambda-1}{256\Lambda}}\cdot\e^{s+1}}\cdot\frac{2^{s}}{4},\;\forall\;\mb Z\cap[0,\texttt M]\ni j\neq l\in[0,\texttt M]\cap\mb Z. 
\] Thus we have 
\beq\label{230915161}
&\inf_{f\in\mc F_d}\ykh{\mc{E}_{P_{j}}(f)+\mc{E}_{P_{l}}(f)}
&\geq  {\min\hkh{\alpha,\frac{\Lambda-1}{256\Lambda}}\cdot\e^{s+1}}\cdot\frac{2^{s}}{4},\;\forall\;\mb Z\cap[0,\texttt M]\ni j\neq l\in[0,\texttt M]\cap\mb Z. 
\eeq

We then establish upper bounds for the average KL divergence $\frac{1}{\texttt M}\cdot\sum_{j=1}^{\texttt M}\mr{KL}(P_j||P_0)$. It follows from \eqref{23091201} and \eqref{23091221} that 
\beq\label{23091517}
\frac{1}{2}-\e\leq\eta_j(x)\leq\frac{1}{2}+\e,\;\forall\;x\in A_1,\;j\in\hkh{0,1,\ldots,\texttt M}. 
\eeq Besides, we can derive from \eqref{23091220} that 
\beq\label{23091515}
\eta_j(x)=\eta_l(x),\;\forall\;x\in[0,1]^d\setminus A_1,\;j\in\hkh{0,1,\ldots,\texttt M},\;l\in\hkh{0,1,\ldots,\texttt M}. 
\eeq Combining  \eqref{23091517}, \eqref{23091515}, and Lemma \ref{23091523}, we obtain by \eqref{23091510} 
\beq\label{230915162}
&\frac{1}{\texttt M}\cdot\sum_{j=1}^{\texttt M}\mr{KL}(P_j||P_0)
\leq
\frac{1}{\texttt M}\cdot\sum_{j=1}^{\texttt M} \Bigabs{18\e^2\cdot\mathscr Q_n(A_1)}
=
18\e^2\cdot \mathscr Q_n(A_1)
\leq 
18\e^2\cdot \mathscr Q_n([0,1]^d\setminus A_2)
\\&=
18\e^2\cdot \min\hkh{\alpha,\frac{\Lambda-1}{256\Lambda}}\cdot 2^s\cdot\e^s
\leq
18\cdot {\frac{\Lambda-1}{256\Lambda}}\cdot 2^s\cdot\e^{s+2}.
\eeq

Finally, we deduce from \eqref{230915163},  \eqref{230915161},  \eqref{230915162} and Lemma \ref{23091519} that  \[
&\inf_{\hat{f}_n}\sup\setl{{\bm E}_{P^{\otimes n}}\zkh{\mc{E}_P(\hat{f}_n)}\vphantom{\Bigg|}}{P\in\mc{H}^{d,\beta,r,\Lambda}_{q,K,d_*}\cap\mc{T}^{d,s}_{\alpha,\tau}}
\\&\geq {\min\hkh{\alpha,\frac{\Lambda-1}{256\Lambda}}\cdot\e^{s+1}}\cdot\frac{2^{s}}{16}\cdot\ykh{1-\frac{36n\cdot {\frac{\Lambda-1}{256\Lambda}}\cdot 2^s\cdot\e^{s+2}+\sqrt{36n\cdot {\frac{\Lambda-1}{256\Lambda}}\cdot 2^s\cdot\e^{s+2}}}{\log\texttt M}}
\\&\geq
{\min\hkh{\alpha,\frac{\Lambda-1}{256\Lambda}}\cdot\e^{s+1}}\cdot\frac{2^{s}}{16}\cdot\ykh{1-\frac{36n\cdot {\frac{\Lambda-1}{256\Lambda}}\cdot 2^s\cdot\e^{s+2}+{72n\cdot {\frac{\Lambda-1}{256\Lambda}}\cdot 2^s\cdot\e^{s+2}}+\frac{1}{8}}{\log\texttt M}}
\\&\geq
{\min\hkh{\alpha,\frac{\Lambda-1}{256\Lambda}}\cdot\e^{s+1}}\cdot\frac{2^{s}}{16}\cdot\ykh{1-\frac{1/8}{\log 2}-\frac{108n\cdot {\frac{\Lambda-1}{256\Lambda}}\cdot 2^s\cdot\e^{s+2}}{\log\texttt M}}
\\&\geq
{\min\hkh{\alpha,\frac{\Lambda-1}{256\Lambda}}\cdot\e^{s+1}}\cdot\frac{2^{s}}{16}\cdot\ykh{1-\frac{1/8}{\log 2}-\frac{108n\cdot {\frac{\Lambda-1}{256\Lambda}}\cdot 2^s\cdot\e^{s+2}}{\ykh{Q^{d_*}\cdot\frac{\Lambda-1}{32\Lambda}}\cdot \log{2}}}
\\&=
{\min\hkh{\alpha,\frac{\Lambda-1}{256\Lambda}}\cdot\e^{s+1}}\cdot\frac{2^{s}}{16}\cdot\ykh{1-\frac{1/8}{\log 2}-\frac{ \Bigabs{\mr c_1}^{(1\qx\beta)^q\cdot(s+2)}\cdot\abs{\frac{n^{\frac{1}{(s+2)\cdot\beta\cdot(1\qx\beta)^q+d_*}}}{\flr{n^{\frac{1}{d_*+(s+2)\cdot\beta\cdot(1\qx\beta)^q}}}+1}}^{(s+2)\cdot\beta\cdot(1\qx\beta)^q+d_*} }{\frac{1}{3456}\cdot4^s\cdot64\cdot 256\cdot \log{2}}}
\\&\geq 
{\min\hkh{\alpha,\frac{\Lambda-1}{256\Lambda}}\cdot\e^{s+1}}\cdot\frac{2^{s}}{16}\cdot\ykh{1-\frac{1/8}{\log 2}-\frac{3456}{64\cdot 256\cdot \log{2}}}
\geq
{\min\hkh{\alpha,\frac{\Lambda-1}{256\Lambda}}\cdot\e^{s+1}}\cdot\frac{2^{s}}{16}\cdot\frac{1}{2}
\\&=
{\min\hkh{\alpha,\frac{\Lambda-1}{256\Lambda}}\cdot\frac{1}{256}\cdot\frac{\bigabs{\mr c_1}^{(1\qx\beta)^q\cdot(s+1)}}{4^s}\cdot \ykh{\frac{1}{\flr{n^{\frac{1}{d_*+(s+2)\cdot\beta\cdot(1\qx\beta)^q}}}+1}}^{(s+1)\cdot\beta\cdot(1\qx\beta)^q}}
\\&\geq
{\min\hkh{\alpha,\frac{\Lambda-1}{256\Lambda}}\cdot\frac{1}{256}\cdot\frac{\bigabs{\mr c_1}^{(1\qx\beta)^q\cdot(s+1)}}{4^s}\cdot \ykh{\frac{1}{2\cdot{n^{\frac{1}{d_*+(s+2)\cdot\beta\cdot(1\qx\beta)^q}}}}}^{(s+1)\cdot\beta\cdot(1\qx\beta)^q}}
\\&\geq
\min\hkh{\alpha,\frac{\Lambda-1}{256\Lambda}}\cdot\frac{1}{256}\cdot\frac{1}{4^s}\cdot \abs{\frac{\mr c_1}{2^\beta}}^{(1\qx\beta)^q\cdot(s+1)}\cdot \ykh{\frac{1}{{n}}}^{\frac{(s+1)\cdot\beta\cdot(1\qx\beta)^q}{{d_*+(s+2)\cdot\beta\cdot(1\qx\beta)^q}}}. 
\] Take  $\mr{c}_4:=\frac{1}{256}\cdot\frac{1}{4^s}\cdot \abs{\frac{\mr c_1}{2^\beta}}^{(1\qx\beta)^q\cdot(s+1)}$, which only depending on $(d_*,\beta,s,q)$. Then we have 
\[
&\inf_{\hat{f}_n}\sup\setl{{\bm E}_{P^{\otimes n}}\zkh{\mc{E}_P(\hat{f}_n)}\vphantom{\Bigg|}}{P\in\mc{H}^{d,\beta,r,\Lambda}_{q,K,d_*}\cap\mc{T}^{d,s}_{\alpha,\tau}}
\geq 
\min\hkh{\alpha,\frac{\Lambda-1}{256\Lambda}}\cdot{\mr{c}_4}\cdot \ykh{\frac{1}{{n}}}^{\frac{(s+1)\cdot\beta\cdot(1\qx\beta)^q}{{d_*+(s+2)\cdot\beta\cdot(1\qx\beta)^q}}}. 
\]  Since $r$ and $n$ are arbitrary, we conclude that the constants $\mr c_3$ and $\mr{c}_4$ defined above have all the desired properties. Thus the proof of this theorem is completed. \end{proof}
	
	Now it remains to prove the $s=\infty$ case of Theorem \ref{23102601}. To this end, we need the following Lemma \ref{23102201}, which is derived by taking $\texttt M=1$ in Lemma \ref{23102003}. 
	 Recall that if a probability measure $P_0$ is absolutely continuous with respect to another probability measure $P_2$, then $\frac{\mr{d}P_0}{\mr{d}P_2}$ denote the Radon-Nikodym derivative. 
\begin{lemma}\label{23102201}Let $d\in\mb N$, $v\in(0,\infty)$, and $n\in\mb N$. Suppose $P_0$, $P_1$, $P_2$ are three Borel  probability measures on $[0,1]^d\times\hkh{-1,1}$ satisfying that\[
\text{$P_j$ is absolutely continuous
	with respect to $P_2$, }\;\forall\;j\in\hkh{0,1}, 	
	\] and \beq\label{23102506}
	\inf_{f\in\mc F_d}\ykh{\mc E_{P_0}(f)+\mc E_{P_1}(f)}\geq v. 
	\eeq Then there holds  \beq\label{23102507}
	&\inf_{\hat f_n}\sup_{j\in\hkh{0,1}}\bm{E}_{{{P_j}^{\otimes n}}}\zkh{\mc{E}_{P_j}(\hat{f}_n)}\geq \frac{v}{4}\cdot \ykh{1-\frac{1}{2}\cdot \int_{\mc{X}_d^n}\abs{\frac{\mr{d}P_1^{\otimes n}}{\mr{d}P_2^{\otimes n}}-\frac{\mr{d}P_0^{\otimes n}}{\mr{d}P_2^{\otimes n}}}\mr{d}P_2^{\otimes n}}, 
	\eeq where the infimum  is taken over all estimators $\hat f_n:[0,1]^d\to\mbR$  based upon the i.i.d. sample $\hkh{(X_i,Y_i)}_{i=1}^n$ in ${[0,1]^d\times\hkh{-1,1}}$.	
\end{lemma}

\begin{proof} \newcommand{\xdn}{\mc{X}_d^n}
For simplicity, in this proof, we denote ${f_0:=\frac{\mr{d}P_0^{\otimes n}}{\mr{d}P_2^{\otimes n}}}$,   ${f_1:=\frac{\mr{d}P_1^{\otimes n}}{\mr{d}P_2^{\otimes n}}}$, and 
\[
\Omega:=\hkh{\text{Borel measurable functions from $\mc{X}_d^n$ to $\hkh{0,1}$}}. 
\]
	
Then	it follows from Lemma \ref{23102003} that 
\[
&\inf_{\hat f_n}\sup_{j\in\hkh{0,1}}\bm{E}_{{P_j}^{\otimes n}}\zkh{\mc{E}_{P_j}(\hat{f}_n)}
\geq 
\frac{v}{2}\cdot\inf\setl{ \sup_{j\in\hkh{0,1}}{P_j^{\otimes n}}\ykh{\hkh{z\in \mc{X}_d^n \big|g(z)\neq j}}}{
	\text{$g\in\Omega$}}
\\&\geq 
\frac{v}{4}\cdot\inf\setl{ \sum_{j\in\hkh{0,1}}{P_j^{\otimes n}}\ykh{\hkh{z\in \mc{X}_d^n \big|g(z)\neq j}}}{
	\text{$g\in\Omega$}}
\\&=
\frac{v}{4}\cdot\inf\setl{ \int_{\xdn}\ykh{\idf_{\hkh{0}}(g(z))\cdot f_1(z)+\idf_{\hkh{1}}(g(z))\cdot f_0(z)}\mr{d}P_2^{\otimes n}(z)}{
	\text{$g\in\Omega$}}
\\&\geq
\frac{v}{4}\cdot\inf\setl{ \int_{\xdn}\min\hkh{f_0(z),f_1(z)}\mr{d}P_2^{\otimes n}(z)}{
	\text{$g\in\Omega$}}
\\&=
\frac{v}{4}\cdot{ \int_{\xdn}\frac{f_0(z)+f_1(z)-\abs{f_0(z)-f_1(z)}}{2}\mr{d}P_2^{\otimes n}(z)}
=
\frac{v}{4}\cdot \ykh{1-\frac{1}{2}\cdot \int_{\xdn}\abs{f_0(z)-f_1(z)}\mr{d}P_2^{\otimes n}(z)}. 
\] This completes the proof. \end{proof}

During the above proof, we actually establish the inequality 
\beq\label{240112043330}
\inf\setl{ \sup_{j\in\hkh{0,1}}{P_j^{\otimes n}}\ykh{\hkh{z\in \mc{X}_d^n \big|g(z)\neq j}}}{
	\text{$g\in\Omega$}}\geq \frac{1}{2}\cdot \ykh{1-\frac{1}{2}\cdot \int_{\mc{X}_d^n}\abs{\frac{\mr{d}P_1^{\otimes n}}{\mr{d}P_2^{\otimes n}}-\frac{\mr{d}P_0^{\otimes n}}{\mr{d}P_2^{\otimes n}}}\mr{d}P_2^{\otimes n}},
\eeq which can be directly obtained from part ({\romannumeral1}) of Theorem 2.2 in \cite{tsybakov2009introduction}.  Nevertheless,   we provide the detailed proof of \eqref{240112043330} for the convenience of readers. The technique of using \eqref{240112043330} to establish minimax lower bounds is   known as Le Cam's method (see for example, Section 15.2 of \cite{wainwright2019high}).

The next theorem lead to the $s=\infty$ case of Theorem \ref{23102601}. 
\begin{theorem}\label{23101701}
	Let $q\in\mb N\cup\hkh{0}$,  $d\in\mb N$, $d_*\in\mb N$, $K\in\mb N$, $\alpha\in(0,\infty)$,  $\tau\in(0,\infty]$,  $\beta\in(0,\infty)$,  and  $\Lambda\in(1,\infty)$. Suppose $d_*\leq K$, $d_*\leq d$,  and  $\idf_{(0,1)}(\alpha)\cdot\idf_{[1,\infty]}(\tau)\neq 1$.   Then there exists a constant  $\mr{c}_3\in(0,\infty)$  only depending on $(\beta,\Lambda)$ such that \[
	&\inf_{\hat{f}_n}\sup\setl{{\bm E}_{P^{\otimes n}}\zkh{\mc{E}_P(\hat{f}_n)}\vphantom{\Bigg|}}{P\in\mc{H}^{d,\beta,r,\Lambda}_{q,K,d_*}\cap\mc{T}^{d,\infty}_{\alpha,\tau}}
	\geq 
\frac{1}{32\cdot n}
	\] for any $r\in[\mr{c}_3,\infty)$ and any   positive integer  $n\geq\frac{1}{\Lambda-1}$,  where the infimum  is taken over all estimators $\hat f_n:[0,1]^d\to\mbR$  based upon the i.i.d. sample $\hkh{(X_i,Y_i)}_{i=1}^n$ from $P$ on ${[0,1]^d\times\hkh{-1,1}}$. 
\end{theorem}

\begin{proof} \newcommand{\xdn}{\mc{X}_d^n}
Let $\mr{c}_3$ be the constant defined in \eqref{23101301}. Then $\mr{c}_3$ belongs to $[3,\infty)$  and only depends on $(\beta,\Lambda)$.  Let $r\in[\mr{c}_3,\infty)$ and $n\in\mb N\cap[\frac{1}{\Lambda-1},\infty)$ be arbitrary numbers.  We then use Lemma \ref{23102201} to prove the desired inequality. To this end, we need to construct three probability measures $P_0$, $P_1$, $P_2$ with $\hkh{P_0,P_1}\subset \mc{H}^{d,\beta,r,\Lambda}_{q,K,d_*}\cap\mc{T}^{d,\infty}_{\alpha,\tau}$ which satisfy the conditions of Lemma \ref{23102201}.

We first construct the conditional class probability functions of $P_0$ and $P_1$, which we will denote by $\eta_0$ and $\eta_1$ respectively.  According to the definition of $\mr{c}_3$, there  exists a function $w\in\mc{C}^{\ceil{\beta-1}}(\mbR)$ such that 
\beq\label{23102202}
\textstyle w(y)=0\leq w(x)\leq 1=w(z),\;\forall\;x\in\mbR,\;\forall\;y\in[0,\frac{\Lambda-1}{2\Lambda}],\;\forall\;z\in[\frac{\Lambda-1}{\Lambda},1],
\eeq and 
\beq\label{23102001}
1\leq \norm{w}_{\mathbf{H}^\beta(\mbR)}<\mr{c}_3\leq r. 
\eeq Define 
\[h_{0,0}:[0,1]^d\to[0,1],\;x\mapsto w((x)_1)
\] if $q=0$; and
 \[h_{0,0}:[0,1]^d\to[0,1]^K,\;x\mapsto \bigykh{w((x)_1),0,\ldots,0}^\top
 \] if $q>0$. Then, for each $i\in\mb N$,  define \[h_{i,0}:[0,1]^K\to[0,1],\;x\mapsto (x)_1\] if $i=q>0$; and define
 \[
 h_{i,0}:[0,1]^K\to[0,1]^K,\;x\mapsto \bigykh{(x)_1,0,0,0,\ldots,0,0}^\top
 \] otherwise. Next, for each $i\in\mb N\cup\hkh{0}$, define
 \[
 h_{i,1}:[0,1]^{d\cdot\idf_{\hkh{0}}(i)+K\cdot\idf_{\mb N}(i)}\to [0,1]^{(1-K)\cdot\idf_{\hkh{q}}(i)+K},\;x\mapsto (1,1,1,\ldots,1,1,1)^\top.  
 \] It follows from \eqref{23102001} that 
 \[
 [0,1]^{d_*}\ni x\mapsto w((x)_1)\in[0,1] 
 \] belongs to $\mc B^\beta_r([0,1]^{d_*})$. Besides, it is easy to verify that both the functions 
 \[
 [0,1]^{d_*}\ni x\mapsto 1\in[0,1]
 \] and 
 \[
 [0,1]^{d_*}\ni x\mapsto (x)_1\in[0,1]
 \] belongs to $\mc B^\beta_2([0,1]^{d_*})$, which is contained in $\mc B^\beta_r([0,1]^{d_*})$. Consequently, 
 \[
 h_{q,j}\circ h_{q-1,j}\circ\cdots\circ  h_{1,j}\circ h_{0,j}\in \mc G_d^{\mathbf{CH}}(q, K, d_*, \beta,r),\;\forall\;j\in\hkh{0,1}.
 \] Define $\eta_j:=h_{q,j}\circ h_{q-1,j}\circ\cdots\circ  h_{1,j}\circ h_{0,j}$ for $j\in\hkh{0,1}$. Thus we have
 \beq\label{23102002}
 \eta_j\in \mc G_d^{\mathbf{CH}}(q, K, d_*, \beta,r),\;\forall\;j\in\hkh{0,1}.
 \eeq Moreover,  an elementary calculation gives
 \beq\label{23102203}
 \eta_1(x)=1,\;\forall\;x\in[0,1]^d,
 \eeq and 
 \beq\label{23102204}
 0\leq \eta_0(x)=w((x)_1)\leq 1,\;\forall\;x\in[0,1]^d. 
 \eeq

 We next construct the marginal distributions of $P_0$ and $P_1$ on $[0,1]^d$, which are actually the same one, denoted by $\mathscr Q_n$.   Let 
 \[
 V_n:[0,1]^d\to\mbR,\;x\mapsto \begin{cases}
 	\frac{\Lambda/n}{\Lambda-1}, &\text{ if }0\leq (x)_1\leq \frac{\Lambda-1}{2\Lambda}, \\
 	0,&\text{ if }\frac{\Lambda-1}{2\Lambda}<(x)_1<\frac{\Lambda-1}{\Lambda}, \\
 \Lambda\cdot\ykh{1-\frac{1}{2n}},&\text{ if }\frac{\Lambda-1}{\Lambda}\leq(x)_1\leq 1. 
 \end{cases}
 \] Obviously, $\mathbf{ran}(V_n)\subset[0,\Lambda]$, and $\int_{[0,1]^d}V_n(x)\mr{d}x=1$. Thus the measure $\mathscr Q_n$ given by
 \[
 \mathscr Q_n(A):=\int_{[0,1]^d}V_n(x)\cdot\idf_A(x)\mr{d}x,\;\;\forall\;\text{Borel set $A\subset[0,1]^d$}
 \] belongs to $\mc{M}_{\Lambda,d}$.

 Define $\eta_2:[0,1]^d\to(0,1),\;x\mapsto 1/2$, and $P_j:=P_{\eta_j,\mathscr Q_n}$ for $j=0,1,2$. We then show that $P_0$ and $P_1$ are both in $\mc{H}^{d,\beta,r,\Lambda}_{q,K,d_*}\cap\mc{T}^{d,\infty}_{\alpha,\tau}$.  It follows from $\mathscr Q_n\in \mc{M}_{\Lambda,d}$ and  \eqref{23102002} that 
 \beq\label{23102501}
 P_j\in\mc{H}^{d,\beta,r,\Lambda}_{q,K,d_*}, \;\forall\;j\in\hkh{0,1}. 
 \eeq Besides, it follows from \eqref{23102202} that
 \[
 &1=\mathscr Q_n(\setr{x\in[0,1]^d}{\textstyle(x)_1\notin(\frac{\Lambda-1}{2\Lambda},\frac{\Lambda-1}{\Lambda})})\leq \mathscr Q_n(\setl{x\in[0,1]^d}{w((x)_1)\in\hkh{0,1}})
 \\&\leq 
 P_j\ykh{\setl{(x,y)\in[0,1]^d\times\hkh{-1,1}}{w((x)_1)\in\hkh{0,1}}}
 \\&\xlongequal{\because\eqref{23102203}\text{ and }\eqref{23102204}}
  P_j\ykh{\setl{(x,y)\in[0,1]^d\times\hkh{-1,1}}{w((x)_1)\in\hkh{0,1}\text{ and }\abs{2\eta_j(x)-1}=1}}
  \\&\leq 
  P_j\ykh{\setl{(x,y)\in[0,1]^d\times\hkh{-1,1}}{\abs{2\eta_j(x)-1}=1}}\leq 1, \;\forall\;j\in\hkh{0,1}, 
 \] %
meaning that 
\[
P_j\ykh{\setl{(x,y)\in[0,1]^d\times\hkh{-1,1}}{\abs{2\eta_j(x)-1}= 1}}=1,\;\forall\;j\in\hkh{0,1}, 
\] and 
\[
P_j\ykh{\setl{(x,y)\in[0,1]^d\times\hkh{-1,1}}{\abs{2\eta_j(x)-1}\leq t}}\leq\idf_{[1,\infty]}(t)\leq \alpha\cdot t^\infty,\;\forall\;t\in(0,\tau]\cap\mbR,\;\forall\;j\in\hkh{0,1}. 
\]  Thus 
\beq\label{23102205}
&\mc{R}_{P_j}(f)-\mc{E}_{P_j}(f)=\inf_{g\in\mc{F}_d}\mc{R}_{P_j}(g)=P_{j}\ykh{\setl{(x,y)\in[0,1]^d\times\hkh{-1,1}}{\mr{sgn}(2\eta_j(x)-1)\neq y}}\\
&=0,\;\forall\;f\in\mc{F}_d,\;\forall\;j\in\hkh{0,1}, 
\eeq and 
\beq\label{23102503}
P_j\in \mc{T}^{d,\infty}_{\alpha,\tau},\;\forall\;j\in\hkh{0,1}. 
\eeq Combining \eqref{23102501} and \eqref{23102503}, we obtain 
\beq\label{23102505}
P_j\in\mc{H}^{d,\beta,r,\Lambda}_{q,K,d_*}\cap\mc{T}^{d,\infty}_{\alpha,\tau},\;\forall\;j\in\hkh{0,1}. 
\eeq

We then establish an inequality of the form \eqref{23102506}.  Indeed,   it follows from \eqref{23102205} that 
\[
&\mc{E}_{P_0}(f)+\mc{E}_{P_1}(f)
=
\mc{R}_{P_0}(f)+\mc{R}_{P_1}(f)\\&=
\int_{[0,1]^d}(2-\eta_0(x)-\eta_1(x))\cdot\idf_{\hkh{1}}(\mr{sgn}(f(x)))\mr{d}\mathscr Q_n(x)\\&\;\;\;\;\;\;\;\;\;\;\;\;\;\;\;\;\;+\int_{[0,1]^d}(\eta_0(x)+\eta_1(x))\cdot\idf_{\hkh{-1}}(\mr{sgn}(f(x)))\mr{d}\mathscr Q_n(x)
\\&=
\int_{[0,1]^d}(1-\eta_0(x))\cdot\idf_{\hkh{1}}(\mr{sgn}(f(x)))\mr{d}\mathscr Q_n(x)+\int_{[0,1]^d}(\eta_0(x)+1)\cdot\idf_{\hkh{-1}}(\mr{sgn}(f(x)))\mr{d}\mathscr Q_n(x)
\\&\geq
\int_{[0,1]^d}(1-\eta_0(x))\cdot\idf_{\hkh{1}}(\mr{sgn}(f(x)))\mr{d}\mathscr Q_n(x)+\int_{[0,1]^d}(1-\eta_0(x))\cdot\idf_{\hkh{-1}}(\mr{sgn}(f(x)))\mr{d}\mathscr Q_n(x)
\\&=
\int_{[0,1]^d}(1-\eta_0(x))\mr{d}\mathscr Q_n(x)
=
\int_{[0,1]^d}(1-\eta_0(x))V_n(x)\mr{d}x
\\&\geq
\int_{[0,1]^d}(1-\eta_0(x))V_n(x){\textstyle\idf_{[0,1]}\bigykh{{\frac{2\Lambda}{\Lambda-1}\cdot(x)_1}}}\mr{d}x
\\&=
\int_{[0,1]^d}\ykh{1-0}\cdot{\textstyle\frac{\Lambda/n}{\Lambda-1}\cdot \idf_{[0,1]}\bigykh{{\frac{2\Lambda}{\Lambda-1}\cdot(x)_1}}}\mr dx
=
{\textstyle\frac{\Lambda/n}{\Lambda-1}\cdot\frac{\Lambda-1}{2\Lambda}} 
=
\frac{1}{2n},\;\forall\;f\in\mc{F}_d, 
\] which means that 
\beq\label{23102504}
\inf_{f\in\mc{F}_d}(\mc{E}_{P_0}(f)+\mc{E}_{P_1}(f))\geq\frac{1}{2n}. 
\eeq

We then estimate the quantity $\ykh{1-\frac{1}{2}\cdot \int_{\mc{X}_d^n}\abs{\frac{\mr{d}P_1^{\otimes n}}{\mr{d}P_2^{\otimes n}}-\frac{\mr{d}P_0^{\otimes n}}{\mr{d}P_2^{\otimes n}}}\mr{d}P_2^{\otimes n}}$, which appears on the right hand side of \eqref{23102507}.  According to Lemma \ref{23090103},  $P_j$ is  absolutely continuous with respect to $P_2$, and 
\[
\frac{\mr{d}P_j}{\mr{d}P_2}(x,y)=\begin{cases}
2\eta_j(x),&\text{ if }y=+1,\\
2-2\eta_j(x),&\text{ if }y=-1,	
\end{cases},\;\forall\; (x,y)\in[0,1]^d\times\hkh{-1,1}
\]  for any $j\in\hkh{0,1}$. Thus  $P_0^{\otimes n}$ and  $P_1^{\otimes n}$ are both absolutely continuous with respect to $P_2^{\otimes n}$.  Denote $\psi_j:=\frac{\mr{d}{P_j}}{\mr{d}{P_2}}$ and  $f_j:=\frac{\mr{d}{P_j}^{\otimes n}}{\mr{d}{P_2}^{\otimes n}}$ for $j\in\hkh{0,1}$. Then we have 
\[
&0\leq f_j(x_1,y_1,x_2,y_2,\ldots,x_n,y_n)=\prod_{i=1}^n\psi_j(x_i,y_i)\\&\leq 2^n,\;\forall\;\hkh{(x_i,y_i)}_{i=1}^n\subset[0,1]^d\times\hkh{-1,1},\;\forall\;j\in\hkh{0,1},
\] which, together with \eqref{23102203}, gives
\[
f_1(x_1,y_1,\ldots,x_n,y_n)=2^n\cdot\prod_{i=1}^n\idf_{\hkh{1}}(y_i),\;\forall\;\hkh{(x_i,y_i)}_{i=1}^n\subset[0,1]^d\times\hkh{-1,1}. 
\] Consequently, 
\[
&\int_{\xdn}\abs{f_0(z)-f_1(z)}\mr{d}P_2^{\otimes n}(z)
\\&\leq 
\int_{\xdn}\abs{f_0(z)}\mr{d}P_2^{\otimes n}(z)+\int_{\xdn}\abs{f_0(x_1,y_1,\ldots,x_n,y_n)-2^n}\cdot\prod_{i=1}^n\idf_{\hkh{1}}(y_i)\mr{d}P_2^{\otimes n}(x_1,y_1,\ldots,x_n,y_n)
\\&=
1+\int_{\xdn}\ykh{2^n-\prod_{i=1}^n\psi_0(x_i,1)}\cdot\prod_{i=1}^n\idf_{\hkh{1}}(y_i)\mr{d}P_2^{\otimes n}(x_1,y_1,\ldots,x_n,y_n)
\\&=1+\ykh{\prod_{i=1}^n\int_{\mc{X}^1_d}2\cdot\idf_{\hkh{1}}(y_i)\mr{d}P_2(x_i,y_i)}-\ykh{\prod_{i=1}^n\int_{\mc{X}^1_d}\psi_0(x_i,1)\cdot\idf_{\hkh{1}}(y_i)\mr{d}P_2(x_i,y_i)}
\\&=
1+1-\ykh{\int_{\mc{X}^1_d}\psi_0(x,1)\cdot\idf_{\hkh{1}}(y)\mr{d}P_2(x,y)}^n
=
2-\ykh{\int_{[0,1]^d}\frac{1}{2}\cdot\psi_0(x,1)\mr{d}\mathscr Q_n(x)}^n
\\&=
2-\ykh{\int_{[0,1]^d}\eta_0(x)\mr{d}\mathscr Q_n(x)}^n
\xlongequal{\because\eqref{23102204}}
2-\ykh{\int_{[0,1]^d}w((x)_1)V_n(x)\mr{d}x}^n
\\&\leq 
2-\ykh{\int_{[0,1]^d}w((x)_1)V_n(x)\idf_{[0,1]}(\Lambda\cdot(x)_1-\Lambda+1)\mr{d}x}^n
\\&=
2-\ykh{\int_{[0,1]^d}\Lambda\cdot{\textstyle(1-\frac{1}{2n})}\cdot\idf_{[0,1]}(\Lambda\cdot(x)_1-\Lambda+1)\mr{d}x}^n
=
2-\ykh{{\textstyle1-\frac{1}{2n}}}^n\leq \frac{3}{2},
\] which means that 
\beq\label{23102502}
 1-\frac{1}{2}\cdot \int_{\mc{X}_d^n}\abs{\frac{\mr{d}P_1^{\otimes n}}{\mr{d}P_2^{\otimes n}}-\frac{\mr{d}P_0^{\otimes n}}{\mr{d}P_2^{\otimes n}}}\mr{d}P_2^{\otimes n}\geq\frac{1}{4}. 
\eeq

Finally, combining  \eqref{23102505}, \eqref{23102504},  \eqref{23102502}, and Lemma \ref{23102201}, we obtain that
\[
&\inf_{\hat{f}_n}\sup\setl{{\bm E}_{P^{\otimes n}}\zkh{\mc{E}_P(\hat{f}_n)}\vphantom{\Bigg|}}{P\in\mc{H}^{d,\beta,r,\Lambda}_{q,K,d_*}\cap\mc{T}^{d,\infty}_{\alpha,\tau}}
\\&\geq
\inf_{\hat{f}_n}\sup_{{j\in\hkh{0,1}}}{{\bm E}_{{P_j}^{\otimes n}}\zkh{\mc{E}_{P_j}(\hat{f}_n)}\vphantom{\Bigg|}}
\geq
\frac{1}{8n}\cdot \ykh{1-\frac{1}{2}\cdot \int_{\mc{X}_d^n}\abs{\frac{\mr{d}P_1^{\otimes n}}{\mr{d}P_2^{\otimes n}}-\frac{\mr{d}P_0^{\otimes n}}{\mr{d}P_2^{\otimes n}}}\mr{d}P_2^{\otimes n}}
\geq
\frac{1}{32\cdot n}.
\] Thus the constant $\mr c_3$ has all the desired properties. This completes the proof of this theorem. 
\end{proof}

We are now in a position to conclude the proof of Theorem \ref{23102601}. 

\begin{proof}[Proof of Theorem \ref{23102601}] Take \[&\cliu:=\text{the constant $\mr{c}_3$ in Theorem \ref{23090801} and Theorem \ref{23101701}, which is precisely defined in \eqref{23101301}},
\\&
\cba:=\frac{\idf_{\hkh{\infty}}(s)}{\Lambda-1}+\idf_{[0,\infty)}(s)\cdot\ykh{\max\hkh{\frac{9\Lambda}{\Lambda-1},\idf_{\ykh{0,\infty}}(s)\cdot\Bigabs{3^{d_*}}^{\frac{1}{ s\cdot\beta\cdot(1\qx\beta)^q}}}}^{d_*+(s+2)\cdot\beta\cdot(1\qx\beta)^q},
\\&
\cqi:=	\begin{cases}
1,&\text{ if }s=\infty,\\
\text{the constant $\mr{c}_4$ in Theorem \ref{23090801}},&\text{ if }s\in[0,\infty),
\end{cases}
\\&\cshisan:= \min\hkh{{\textstyle\alpha,\frac{\Lambda-1}{256\Lambda}}}\cdot{\cqi}. 
	 \] Then the desired result follows immediately from Theorem \ref{23090801} and Theorem \ref{23101701}. \end{proof}

\subsection{Proof of Theorem \ref{23102602}}\hypertarget{20251011010150}{}   \label{20251011010150}

In this part, we will prove  Theorem \ref{23102602}. Since all the preliminary results have already been established during the proof of Theorem \ref{23102601} in the previous subsection, we can start the proof of Theorem \ref{23102602} right now.

\begin{proof}[Proof of Theorem \ref{23102602}]Let $n$ be  an arbitrary positive integer. Take   ${Q\mkern-1mu:=\flr{n^{\frac{1}{d_*+2\beta\cdot(1\qx\beta)^q}}}+1}$,  ${\texttt M:=\ceil{2^{{Q^{d_*}}/{8}}}}$, and $r':=\frac{\min\hkh{1,r-1/2}}{32}$.   As in the proof of  Theorem \ref{23090801}, we will prove this theorem by using Lemma \ref{23091519}. Hence we need to construct suitable probability measures $P_0, P_1, P_2, \ldots P_{\texttt M}$ in $\mc{H}^{d,\beta,r,\Lambda}_{q,K,d_*}$  which satisfy the conditions of  Lemma \ref{23091519}.  Let \[G_{Q,d_*}:=\setl{(\frac{k_1}{2Q},\ldots,\frac{k_{d_*}}{2Q})^\top}{k_1,\ldots,k_{d_*}\text{ are odd integers}}\cap[0,1]^{d_*}. \]  Note that $\#(G_{Q,d_*})=Q^{d_*}$. Thus it follows from Lemma \ref{vgbound} that there exist maps  $T_j:G_{Q,d_*}\to\hkh{0,1}$, $j=0,1,\ldots,\texttt M$, such that \beq\label{231116232723}
	\#\ykh{\setl{a\in G_{Q,d_*}}{ T_j(a)\neq  T_l(a)}}\geq
	\frac{Q^{d_*}}{8},\;\forall\;\mb Z\cap[0,\texttt M]\ni j\neq l\in[0,\texttt M]\cap\mb Z. 
	\eeq	Let 
	\[
	\mc{U}_{d_*,\beta,0}:=\setr{w\in\mc{C}^{\ceil{\beta-1}}(\mbR^{d_*})}{\begin{minipage}{270.36pt}$\mathbf{ran}(w)\subset[0,1]$, and $w(x)=\idf_{[0,\infty]}(\frac{1}{2}-\frac{1/3}{1+d_*}-\norm{x}_\infty)$ for any $x\in\setr{z\in\mbR^{d_*}}{\norm{z}_\infty\in\mbR\setminus\ykh{\frac{1}{2}-\frac{1/3}{1+d_*}, \;\frac{1}{2}-\frac{1/4}{1+d_*}}}$  \end{minipage}}
	\] and 
	\[
	\mr{c}_5:=\frac{1}{2}+\inf\setl{\norm{w}_{\mathbf{H}^\beta(\mbR^{d_*})}}{w\in\mc{U}_{d_*,\beta,0}}. 
	\] Then $\mr{c}_5\in(1,\infty)$ and there exists a function $u\in\mc{U}_{d_*,\beta,0}$ such that \[
	1\leq \norm{u}_{\mathbf{H}^\beta(\mbR^{d_*})}<\mr{c_5}. 
	\] Take $\mr{c}_6:=\frac{1}{\mr{ c_5}}\cdot\frac{\min\hkh{1,r-1/2}}{32}$, and 
	\[
	f_j:\mbR^{d_*}\to \mbR,\;x\mapsto \sum_{a\in G_{Q,d_*}}\frac{\mr{c}_6}{Q^\beta}\cdot T_j(a)\cdot u(Q\cdot (x-a))
	\] for $j=0,1,\ldots,\texttt M$. Then $\mr{c}_6$ only depends on $(d_*,\beta,r)$ and belongs to $(0,\frac{1}{32}]$. 
	 Recall the definition of the infinity-norm ball $\mathscr{C}(\;\cdot\;,\;\cdot\;)$ given in \eqref{20250918033935}. 
	Note that
	\[
	\setl{x\in\mbR^{d_*}}{u(Q\cdot(x-a))\neq 0}\subset \mathscr{C}\ykh{a,\frac{1}{Q}\cdot\Bigykh{ \frac{1}{2}-\frac{1/4}{1+d_*}}},\;\forall\;a\in G_{Q,d_*}, 
	\]	 and 
	\[
	\mathscr{C}\ykh{a,\frac{1}{Q}\cdot\Bigykh{ \frac{1}{2}-\frac{1/4}{1+d_*}}}\cap \mathscr{C}\ykh{a',\frac{1}{Q}\cdot\Bigykh{ \frac{1}{2}-\frac{1/4}{1+d_*}}}=\varnothing,\;\forall\;G_{Q,d_*}\ni a\neq a'\in G_{Q,d_*}, 
	\] meaning that the supports of the functions $x\mapsto u(Q\cdot (x-a))$ are mutually disjoint. From this we can easily obtain   \beq\label{23102702}
	&0\leq f_j(x)\leq \frac{\mr c_6}{Q^\beta}\cdot\norm{u}_{\mbR^{d_*}}=\frac{\mr{c}_6}{Q^\beta}\leq \mr{c}_6\leq\frac{1}{32},\;\forall\;x\in\mbR^{d_*},\;\forall\;j\in\hkh{0,1,\ldots,\texttt M}, \eeq and \beq\label{231116220304}
	&f_j(x)=\frac{\mr c_6}{Q^\beta}\cdot T_j(a),\;\forall\; a\in G_{Q,d_*},\;x\in \mathscr{C}\ykh{a,\frac{1}{Q}\cdot\Bigykh{ \frac{1}{2}-\frac{1/3}{1+d_*}}},\;\forall\;j\in\hkh{0,1,\ldots,\texttt M}. \eeq Besides, it follows from Lemma \ref{23092901} that 	\beq\label{23102709}
		&f_j\in\mc{C}^{\ceil{\beta-1}}(\mbR^{d_*}),\;\forall\; j\in\hkh{0,1,\ldots,\texttt M}, 
		\eeq and 
	\beq\label{23102701}
	& \norm{f_j}_{\mathbf{H}^\beta(\mbR^{d_*})}\leq3\mr{c}_6 \norm{u}_{\mathbf{H}^\beta(\mbR^{d_*})}\leq 3\mr{c_6}\mr{c_5}=3 r'={\textstyle\frac{\min\hkh{1,r-1/2}}{32/3}}<r-\frac{1}{2},\;\forall\;j\in\hkh{0,1,\ldots,\texttt M}. 
	\eeq  Define \[{\e:=\frac{1}{2}\cdot \abs{\frac{\mr c_6}{Q^\beta}}^{(1\qx\beta)^q}\cdot\bigabs{r'}^{\sum_{k=0}^{q-1}(1\qx\beta)^k}}.\] Then we have that 
	\[
	0<\e\leq r'\leq \frac{1}{32}. 
	\] Define
	\[
	\mr u_0:[0,1]^{d_*}\to\mbR,\;x\mapsto r'\cdot \abs{(x)_1}^{(1\qx\beta)}, 
	\] and  \[
	\mr u_1:[0,1]^{d_*}\to\mbR,\;x\mapsto\frac{1}{2}-\e+r'\cdot  \abs{(x)_1}^{(1\qx\beta)}. 
	\] It is easy to verify that 
	\beq\label{23102703}
	\mathbf{ran}(\mr u_0)\cup \mathbf{ran}(\mr u_1)\subset[0,1], 
	\eeq and 
	\beq\label{23102704}
	\hkh{\mr u_0,\mr u_1}\subset\mc{C}^{\ceil{\beta-1}}([0,1]^{d_*})\text{ and }0<\norm{\mr u_0}_{\mathbf{H}^\beta([0,1]^{d_*})}\leq \norm{\mr u_1}_{\mathbf{H}^\beta([0,1]^{d_*})}\leq \frac{1}{2}-\e+2\cdot r'\leq r. 
	\eeq Next, for each $i\in\hkh{0,1,\ldots,q}$, define $d^{\mathbf{in}}_{i}:=d\cdot\idf_{\hkh{0}}(i)+K\cdot\idf_{\mb N}(i)$ and $d^{\mathbf{out}}_{i}:=K+(1-K)\cdot\idf_{\hkh{q}}(i)$. Then for each $i\in\hkh{0,1,\ldots,q}$ and $j\in\hkh{0,1,\ldots,\texttt M}$, define
\[
h_{i,j}:&[0,1]^{d^{\mathbf{in}}_{i}}\to\mbR^{d^{\mathbf{out}}_{i}},\;\\
&\;\;\;x\mapsto \begin{pmatrix}1\\0\\0\\\vdots\\0\\0\end{pmatrix}\cdot \begin{pmatrix}{f_j((x)_1,\ldots, (x)_{d_*})\cdot\idf_{\hkh{0}}(i)+\mr{u}_0((x)_1,\ldots,(x)_{d_*})\cdot\idf_{\mb N}(i)+\abs{{\textstyle\frac{1}{2}-\e}}\cdot\idf_{\hkh{q}}(i)}\end{pmatrix},
\] that is,  $h_{i,j}(x)$ is a  ${d_i^{\mathbf{out}}}$-dimensional real vector of which the first component is \[{f_j((x)_1,\ldots, (x)_{d_*})\cdot\idf_{\hkh{0}}(i)+\mr{u}_0((x)_1,\ldots,(x)_{d_*})\cdot\idf_{\mb N}(i)+\abs{{\textstyle\frac{1}{2}-\e}}\cdot\idf_{\hkh{q}}(i)},\] and the other components  are all zero. From \eqref{23102702} and \eqref{23102703} we see that \beq\label{23102706}
\mathbf{ran}(h_{i,j})\subset[0,1]^{d^{\mathbf{out}}_i},\;\;\forall\;i\in\hkh{0,1,\ldots,q},\;\forall\;j\in\hkh{0,1,\ldots,\texttt M}.\eeq Thus we can well define
\[
\eta_j:=h_{q,j}\circ h_{q-1,j}\circ\cdots h_{1,j}\circ h_{0,j}
\]	 for each $j\in\hkh{0,1,\ldots,\texttt{M}}$.  Then it follows from \eqref{23102709}, \eqref{23102701}, \eqref{23102704} and \eqref{23102706} that 	\beq\label{23102705}
\mathbf{ran}(\eta_j)\subset[0,1]\;\;\text{ and }\;\;\eta_j\in \mc G_d^{\mathbf{CH}}(q, K, d_*, \beta,r),\;\forall\;j\in\hkh{0,1,\ldots,\texttt M}. 
\eeq   Let $\mathscr Q$ be the Lebesgue measure on $[0,1]^d$ and define $P_j:=P_{\eta_j,\mathscr Q}$ for $j=0,1,\ldots,\texttt M$. Then we obtain from \eqref{23102705} that 
\beq\label{231117030859}
P_j\in\mc{H}^{d,\beta,r,1}_{q,K,d_*}\subset \mc{H}^{d,\beta,r,\Lambda}_{q,K,d_*},\;\forall\;j\in\hkh{0,1,\ldots,\texttt M}. 
\eeq%

We then establish an inequality of the same form as \eqref{23091302}. Indeed, it follows from the definitions of $h_{i,j}$ and $\eta_j$ ($j\in
\{0,1,\ldots,\texttt M\}$, $i\in\{0,1,\ldots,q\}$)  that 
\beq\label{2311152210}
\eta_j(x)=\frac{1}{2}-\e+\Bigabs{f_j\big((x)_1,(x)_2,\ldots,(x)_{d_*}\big)}^{(1\qx\beta)^q}\cdot\bigabs{r'}^{\sum_{k=0}^{q-1}(1\qx\beta)^k},\;\forall\;x\in[0,1]^d,\;\forall\;j\in\hkh{0,1,\ldots,\texttt M}. 
\eeq Let \[D_a:=\setr{x\in[0,1]^d}{\bigykh{(x)_1,(x)_2,\ldots,(x)_{d_*}}^\top\in\mathscr{C}\Big(a,\frac{1}{2Q}-\frac{1/3}{Q+Q\cdot d_*}\Big)},\;\forall\;a\in \mbR^{d_*}, 
\] $\displaystyle A_0:=\setl{x\in[0,1]^d}{\exists \;a\in G_{Q,d_*}\text{ s.t. }x\in D_a}$, and $\mc{J}$ be the function defined by \eqref{2301050326}. Obviously, 
\beq\label{231116230556}
D_a\cap D_b=\varnothing,\;\forall\;G_{Q,d_*}\ni a\neq b\in G_{Q,d_*}. 
\eeq It follows from \eqref{231116220304} and \eqref{2311152210} that
\beq\label{231117025106}
&\eta_j(x)=\frac{1}{2}-\e+\abs{\frac{\mr c_6}{Q^\beta}\cdot T_j(a)}^{(1\qx\beta)^q}\cdot\bigabs{r'}^{\sum_{k=0}^{q-1}(1\qx\beta)^k}
\\&=
\frac{1}{2}-\e+T_j(a)\cdot2\cdot\e\in[{1}/{2}-\e,{1}/{2}+\e]\subset(0,1),\;\forall\;j\in\hkh{0,1,\ldots,\texttt{M}},\;\forall\;a\in G_{Q,d_*},\;\forall\;x\in D_a. 
\eeq We then use \eqref{231116232723} and Lemma \ref{2301051433} to obtain that
\[
&\inf_{f\in\mc F_d}\ykh{\mc{E}_{P_{j}}(f)+\mc{E}_{P_{l}}(f)}
\geq
\int_{[0,1]^d}\mc J(\eta_j(x),\eta_l(x))\mr{d}\mathscr Q(x)
\geq 
\int_{[0,1]^d}\mc J(\eta_j(x),\eta_l(x))\cdot\idf_{A_0}(x)\mr{d}\mathscr Q(x)\\
&\xlongequal{\because\eqref{231116230556}}
\sum_{a\in G_{Q,d_*}}\int_{[0,1]^d}\mc J(\eta_j(x),\eta_l(x))\cdot\idf_{D_a}(x)\mr{d}\mathscr Q(x)
\\&\geq
\sum_{\substack{a\in G_{Q,d_*}\\T_j(a)\neq T_l(a)}}\int_{[0,1]^d}\mc J(\eta_j(x),\eta_l(x))\cdot\idf_{D_a}(x)\mr{d}\mathscr Q(x)
\\&=
\sum_{\substack{a\in G_{Q,d_*}\\T_j(a)\neq T_l(a)}}\int_{[0,1]^d}\mc J\ykh{\frac{1}{2}-\e+T_j(a)\cdot2\cdot\e,\frac{1}{2}-\e+T_l(a)\cdot2\cdot\e}\cdot\idf_{D_a}(x)\mr{d}\mathscr Q(x)
\\&=
\sum_{\substack{a\in G_{Q,d_*}\\T_j(a)\neq T_l(a)}}\int_{[0,1]^d}\mc J\ykh{\frac{1}{2}-\e+0\cdot2\cdot\e,\frac{1}{2}-\e+1\cdot2\cdot\e}\cdot\idf_{D_a}(x)\mr{d}\mathscr Q(x)
\\&=
\sum_{\substack{a\in G_{Q,d_*}\\T_j(a)\neq T_l(a)}}\int_{[0,1]^d}\mc J\ykh{\frac{1}{2}-\e,\frac{1}{2}+\e}\cdot\idf_{D_a}(x)\mr{d}x
=
\sum_{\substack{a\in G_{Q,d_*}\\T_j(a)\neq T_l(a)}}\int_{[0,1]^d}2\e\cdot\idf_{D_a}(x)\mr{d}x
\\&=
\#\ykh{\setr{a\in G_{Q,d_*}}{\vphantom{2^2}T_j(a)\neq T_l(a)}}\cdot 2\e\cdot \abs{\frac{1}{Q}-\frac{2/3}{Q+Q\cdot d_*}}^{d_*}
\geq
\frac{Q^{d_*}}{8}\cdot 2\e\cdot \abs{\frac{1}{Q}-\frac{2/3}{Q+Q\cdot d_*}}^{d_*}
\\&=
\frac{\e}{4}\cdot \abs{{1}-\frac{2/3}{1+ d_*}}^{d_*}\geq\frac{\e}{8},\;\forall\;\mb{Z}\cap[0,\texttt M]\ni j\neq l\in[0,\texttt M]\cap\mb{Z}.
\] Thus we have that
\beq\label{231117030752}
\inf_{f\in\mc F_d}\ykh{\mc{E}_{P_{j}}(f)+\mc{E}_{P_{l}}(f)}\geq
\frac{\e}{8}, \;\forall\;\mb{Z}\cap[0,\texttt M]\ni j\neq l\in[0,\texttt M]\cap\mb{Z}.
\eeq

We then establish upper bounds for $\frac{1}{\texttt M}\cdot\sum_{j=1}^{\texttt M}\mr{KL}(P_j||P_0)$. Indeed, it follows from \eqref{23102702} and \eqref{2311152210} that 
\[
&0<\frac{1}{2}-\frac{1}{32}\leq\frac{1}{2}-\e\leq\eta_j(x)=\frac{1}{2}-\e+\Bigabs{f_j\big((x)_1,(x)_2,\ldots,(x)_{d_*}\big)}^{(1\qx\beta)^q}\cdot\bigabs{r'}^{\sum_{k=0}^{q-1}(1\qx\beta)^k}\\&\leq
\frac{1}{2}-\e+\abs{\frac{\mr c_6}{Q^\beta}}^{(1\qx\beta)^q}\cdot\bigabs{r'}^{\sum_{k=0}^{q-1}(1\qx\beta)^k}=\frac{1}{2}+\e\leq\frac{1}{2}+\frac{1}{32}<1
,\;\forall\;x\in[0,1]^d,\;\forall\;j\in\hkh{0,1,\ldots,\texttt M}, 
\] which, together with Lemma \ref{23091523}, yields that 
\beq\label{231117030732}
\frac{1}{\texttt M}\cdot\sum_{j=1}^{\texttt M}\mr{KL}(P_j||P_0)
\leq
\frac{1}{\texttt M}\cdot\sum_{j=1}^{\texttt M}18\e^2=18\e^2. 
\eeq

Finally, we deduce from \eqref{231117030859}, \eqref{231117030752}, \eqref{231117030732}, and Lemma \ref{23091519} that
\[
&\inf_{\hat{f}_n}\sup\setl{{\bm E}_{P^{\otimes n}}\zkh{\mc{E}_P(\hat{f}_n)}\vphantom{\Bigg|}}{P\in\mc{H}^{d,\beta,r,\Lambda}_{q,K,d_*}}
\geq\frac{\e}{32}\cdot\ykh{1-\frac{2\cdot n\cdot18\e^2+\sqrt{2\cdot n\cdot 18\e^2}}{\log\texttt M}}
\\&\geq
\frac{\e}{32}\cdot\ykh{1-\frac{36n\e^2+54n\e^2+1/6}{\log\texttt M}}
\geq
\frac{\e}{32}\cdot\ykh{1-\frac{36n\e^2+54n\e^2}{\log\ykh{2^{Q^{d_*}/8}}}-\frac{1/6}{\log 2}}
\\&=
\frac{\e}{32}\cdot\ykh{1-\frac{180n}{Q^{d_*}\cdot\log\ykh{2}}\cdot \abs{ \abs{\frac{\mr c_6}{Q^\beta}}^{(1\qx\beta)^q}\cdot\bigabs{r'}^{\sum_{k=0}^{q-1}(1\qx\beta)^k}}^2-\frac{1/6}{\log 2}}
\\&\geq
\frac{\e}{32}\cdot\ykh{1-\frac{180n}{Q^{d_*}\cdot\log\ykh{2}}\cdot \abs{ \abs{\frac{1/32}{Q^\beta}}^{(1\qx\beta)^q}\cdot\bigabs{1/32}^{\sum_{k=0}^{q-1}(1\qx\beta)^k}}^2-\frac{1/6}{\log 2}}
\\&\geq
\frac{\e}{32}\cdot\ykh{1-\frac{180n}{Q^{d_*}\cdot\log\ykh{2}}\cdot \abs{ \abs{\frac{1}{Q^\beta}}^{(1\qx\beta)^q}\cdot\bigabs{1/32}}^2-\frac{1/6}{\log 2}}
\\&\geq
\frac{\e}{32}\cdot\ykh{1-\frac{180}{\log\ykh{2}}\cdot \abs{ \frac{1}{32}}^2-\frac{1/6}{\log 2}}\geq\frac{\e}{64}
\\&=
\frac{1}{128}\cdot\abs{\frac{1}{\mr{c}_5\cdot2^\beta}}^{(1\qx\beta)^q}\cdot\abs{\frac{2}{Q}}^{\beta\cdot(1\qx\beta)^q}\cdot\abs{\frac{\min\hkh{1,r-\frac{1}{2}}}{32}}^{\sum\limits_{k=0}^{q}(1\qx\beta)^k}
\\&\geq
\frac{1}{128}\cdot\abs{\frac{1}{\mr{c}_5\cdot2^\beta}}^{(1\qx\beta)^q}\cdot\abs{\frac{1}{n}}^{\frac{\beta\cdot(1\qx\beta)^q}{d_*+2\beta\cdot(1\qx\beta)^q}}\cdot\abs{\frac{\min\hkh{1,r-\frac{1}{2}}}{32}}^{\sum\limits_{k=0}^{q}(1\qx\beta)^k}
=
\cjiu\cdot \ykh{\frac{1}{n}}^{\frac{\beta\cdot(1\qx\beta)^q}{d_*+2\beta\cdot(1\qx\beta)^q}}
\] with \[\cjiu:=\frac{1}{128}\cdot\abs{\frac{1}{\mr{c}_5\cdot2^\beta}}^{(1\qx\beta)^q}\cdot\abs{\frac{\min\hkh{1,r-\frac{1}{2}}}{32}}^{\sum\limits_{k=0}^{q}(1\qx\beta)^k}\in(0,\infty)\] only depending on $(d_*,\beta,q,r)$. This completes the proof of Theorem \ref{23102602}.  \end{proof}

\subsection{Proof of \eqref{231118003959}}\hypertarget{20251011004753}{}    \label{20251011004753}

In this part, we will prove the following Lemma \ref{231118004959}, which leads to \eqref{231118003959}.  For $q\in\mb N\cup\hkh{0}$, $d\in\mb N$, $d_*\in\mb N$, $K\in\mb N$,  $\beta\in(0,\infty)$, and $r\in(0,\infty)$, define
 \[
\mc{H}^{d,\beta,r}_{q,K,d_*}:=\setr{P_{\eta,\mathscr Q}}{\begin{minipage}{207.48pt}  $\eta\in\mc G_d^{\mathbf{CH}}(q, K, d_*, \beta,r)$, $\mathbf{ran}(\eta)\subset[0,1]$, and $\mathscr Q$ is a Borel probability measure on $[0,1]^d$\end{minipage}}.
\] 

\begin{lemma}\label{231118004959}Let $q\in\mb N\cup\hkh{0}$,  $d\in\mb N$, $d_*\in\mb N$, $K\in\mb N$,     $\beta\in(0,\infty)$, $r\in(0,\frac{1}{2}]$, $n\in\mb N$, and  $\Lambda\in[1,\infty)$. Suppose $d_*\leq K$, and $d_*\leq d$. Then 
\[
\inf_{\hat{f}_n}\sup\setl{{\bm E}_{P^{\otimes n}}\zkh{\mc{E}_P(\hat{f}_n)}\vphantom{\Bigg|}}{P\in\mc{H}^{d,\beta,r,\Lambda}_{q,K,d_*}}=\inf_{\hat{f}_n}\sup\setl{{\bm E}_{P^{\otimes n}}\zkh{\mc{E}_P(\hat{f}_n)}\vphantom{\Bigg|}}{P\in\mc{H}^{d,\beta,r}_{q,K,d_*}}=0,
\] where the infimum  is taken over all estimators $\hat f_n:[0,1]^d\to\mbR$  based upon the i.i.d. sample $\hkh{(X_i,Y_i)}_{i=1}^n$ in ${[0,1]^d\times\hkh{-1,1}}$.\end{lemma}

\begin{proof} Obviously, 
\[
\varnothing\subsetneqq \mc{H}^{d,\beta,r,\Lambda}_{q,K,d_*}\subset\mc{H}^{d,\beta,r}_{q,K,d_*}, 
\] meaning that
\beq\label{20231120014815}
0\leq \inf_{\hat{f}_n}\sup\setl{{\bm E}_{P^{\otimes n}}\zkh{\mc{E}_P(\hat{f}_n)}\vphantom{\Bigg|}}{P\in\mc{H}^{d,\beta,r,\Lambda}_{q,K,d_*}}\leq\inf_{\hat{f}_n}\sup\setl{{\bm E}_{P^{\otimes n}}\zkh{\mc{E}_P(\hat{f}_n)}\vphantom{\Bigg|}}{P\in\mc{H}^{d,\beta,r}_{q,K,d_*}}. 
\eeq Let $\hat{f}_n^{\diamond}$ be the estimator which always takes the constant function $-1$, that is, \[\hat f_n^{\diamond}(x)=-1,\;\forall\;x\in[0,1]^d\]for any sample $\hkh{(X_i,Y_i)}_{i=1}^n$ in $[0,1]^d\times\hkh{-1,1}$. Let $P$ be an arbitrary probability measure in $\mc{H}^{d,\beta,r}_{q,K,d_*}$. Then there exists a function $\eta\in\mc G_d^{\mathbf{CH}}(q, K, d_*, \beta,r)$ and a probability measure $\mathscr Q$ on $[0,1]^d$ such that $\mathbf{ran}(\eta)\subset[0,1]$ and $P=P_{\eta,\mathscr Q}$.  Take $u:=d\cdot\idf_{\hkh{0}}(q)+K\cdot\idf_{\mb N}(q)$.  By the definition of  $\mc G_d^{\mathbf{CH}}(q, K, d_*, \beta,r)$, there exist functions $h_0, h_1,\ldots, h_q$ such that $\eta=h_q\circ h_{q-1}\circ\cdots h_1\circ h_0$ and $h_q\in \mc G_u^{\mathbf{H}}(d_*, \beta,r)$. Thus there exist a function $g\in\mc{B}_r^\beta([0,1]^{d_*})$ and a set $I\subset\hkh{1,2,\ldots,u}$ such that $\#(I)=d_*$ and 
\[
h_q(x)=g((x)_I),\;\forall\;x\in\mathbf{dom}(h_q)=[0,1]^u,
\] meaning that
\beq\label{231120015045}
&\eta(z)=h_q\circ\cdots h_0(z)\leq \sup\setl{h_q(x)\raisebox{1em}{}}{x\in\mathbf{dom}(h_q)}=\sup\setl{h_q(x)\raisebox{1em}{}}{x\in[0,1]^u}\\&=\sup\setl{g((x)_I)\raisebox{1em}{}}{x\in[0,1]^u}\leq \norm{g}_{[0,1]^{d_*}}\leq\norm{g}_{\mathbf{H}^\beta([0,1]^{d_*})}\leq r\leq\frac{1}{2},\;\forall \;z\in[0,1]^d.  
\eeq Consequently, 
\[
&0\leq{\bm E}_{P^{\otimes n}}\zkh{\mc{E}_P(\hat{f}^\diamond_n)}={\bm E}_{P^{\otimes n}}\zkh{\mc{R}_P(\hat{f}^\diamond_n)}-\inf_{f\in\mc{F}_d}\mc{R}_P(f)\\&=P\ykh{\hkh{(x,y)\in[0,1]^d\times\hkh{-1,1}\Big|y\neq -1}}-\inf_{f\in\mc{F}_d}P\ykh{\hkh{(x,y)\in[0,1]^d\times\hkh{-1,1}\Big|\mr{sgn}(f(x))\neq y}}\\
&=
\int_{[0,1]^d}\eta(x)\mr{d}\mathscr Q(x)-\inf_{f\in\mc{F}_d}\int_{[0,1]^d}\Bigykh{\eta(x)\cdot\idf_{\hkh{-1}}(\mr{sgn}(f(x)))+(1-\eta(x))\cdot\idf_{\hkh{1}}(\mr{sgn}(f(x)))}\mr{d}\mathscr Q(x)
\\&\leq
\int_{[0,1]^d}\eta(x)\mr{d}\mathscr Q(x)-\int_{[0,1]^d}\inf\hkh{\eta(x)\cdot\idf_{\hkh{-1}}(z)+(1-\eta(x))\cdot\idf_{\hkh{1}}(z)\Big|z\in\hkh{-1,1}}\mr{d}\mathscr Q(x)\\
&=\int_{[0,1]^d}\eta(x)\mr{d}\mathscr Q(x)-\int_{[0,1]^d}\inf\big\{\eta(x),1-\eta(x)\big\}\mr{d}\mathscr Q(x)\\&\xlongequal{\because\eqref{231120015045}}
\int_{[0,1]^d}\eta(x)\mr{d}\mathscr Q(x)-\int_{[0,1]^d}\eta(x)\mr{d}\mathscr Q(x)=0.
\] Thus ${\bm E}_{P^{\otimes n}}\zkh{\mc{E}_P(\hat{f}^\diamond_n)}=0$. Since $P$ is arbitrary, we must have 
\beq
\sup\setl{{\bm E}_{P^{\otimes n}}\zkh{\mc{E}_P(\hat{f}^\diamond_n)}\vphantom{\Bigg|}}{P\in\mc{H}^{d,\beta,r}_{q,K,d_*}}=\sup\setr{0}{P\in\mc{H}^{d,\beta,r}_{q,K,d_*}}=0. 
\eeq Therefore, 
\beq\label{231121000305}
\inf_{\hat{f}_n}\sup\setl{{\bm E}_{P^{\otimes n}}\zkh{\mc{E}_P(\hat{f}_n)}\vphantom{\Bigg|}}{P\in\mc{H}^{d,\beta,r}_{q,K,d_*}}\leq \sup\setl{{\bm E}_{P^{\otimes n}}\zkh{\mc{E}_P(\hat{f}^\diamond_n)}\vphantom{\Bigg|}}{P\in\mc{H}^{d,\beta,r}_{q,K,d_*}}=0. 
\eeq Combining \eqref{20231120014815} and \eqref{231121000305}, we obtain
\[
0\leq \inf_{\hat{f}_n}\sup\setl{{\bm E}_{P^{\otimes n}}\zkh{\mc{E}_P(\hat{f}_n)}\vphantom{\Bigg|}}{P\in\mc{H}^{d,\beta,r,\Lambda}_{q,K,d_*}}\leq\inf_{\hat{f}_n}\sup\setl{{\bm E}_{P^{\otimes n}}\zkh{\mc{E}_P(\hat{f}_n)}\vphantom{\Bigg|}}{P\in\mc{H}^{d,\beta,r}_{q,K,d_*}}\leq 0,
\] which proves the desired result. \end{proof}

\section{Further Discussions}

\setcounter{figure}{0}

\hypertarget{20250928194706}{}    \label{20250928194706}

 In  Appendix  \ref{20250928194706}, we provide       
 further discussions of a few claims mentioned in the main body of this paper. 

\subsection{Comparison with the Setting of \cite{meyer2023optimal}}\label{20250919034320}\hypertarget{20250919034321}{}    \label{20250919034321}

In \cite{meyer2023optimal}, optimal convergence rates of ReLU DNNs in classification  are also obtained. However, the setting in \cite{meyer2023optimal} is quite different from that in the present paper. 
In what follows, we will point out the differences in detail.

The first difference is the regularity assumption on the CCP function $\eta_P(\cdot)=P(\hkh{1}|\;\cdot\;)$ of the data distribution $P$. 
In this paper, we assume that $\eta_P$ is a  compositional Hölder smooth function on $[0,1]^d$. 
In \cite{meyer2023optimal}, the author 
assumes that  $\eta_P$ has a  smooth decision boundary (that is, the boundary  of the Bayes decision set  $\set{x\in[0,1]^d}{\eta_P(x)\geq \textstyle\frac12}$ in  $[0,1]^d$), where  the smoothness is directly characterized  by the  error rates  of approximation by ReLU DNNs. Specifically, the author assumes that  
the Bayes decision set $\set{x\in[0,1]^d}{\eta_P(x)\geq \frac12}$ can be expressed as $\set{x\in[0,1]^d}{\eta_P(x)\geq \frac12}=\bigcup_{k=1}^uH_k$,  %
and there exist hyperrectangles $D_k=\prod\limits_{j=1}^d [a_{k,j},b_{k,j}]\subset[0,1]^d$ ($k=1,\ldots, u$) whose interior regions are disjoint such that for any $k\in\hkh{1,2,\ldots,u}$ there holds 
\beq\label{20250926163808}
H_k=D_k\cap\setl{x\in[0,1]^{d}}{\iota_k\cdot (x)_{i_k}\leq g_k((x)_{\hkh{1,\ldots,d}\setminus \hkh{i_k}})}
\eeq  for some $\iota_k\in\hkh{-1,1}$, $i_k\in\hkh{1,2,\ldots,d}$, $g_k\in\mc{G}$, where  $\mc{G}$ is a set of functions from $[0,1]^{d-1}$ to $\mbR$ such that for any $f\in \mc{G}$ and any $\e\in(0,\e_0)$ there exists a ReLU DNN \[\tilde{f}\in\fdnn_{d-1}(C_1\ceil{\log(1/\e)},\infty,  C_2\e^{-\rho}\log (1/\e),\infty,\infty)\] of which all the parameters take values in 
\[
[-1,1]\cap \set{\frac{k}{2^{C_3+C_4\ceil{\log(1/\e)}}}}{ k\in\mb{Z}}
\] such that $\norm{f-\tilde{f}}_{[0,1]^{d-1}}\leq \e$, where $\e_0\in(0,\infty)$, $C_1\in(0,\infty)$, $C_2\in(0,\infty)$, $C_3\in\mb{N}$, $C_4\in\mb{N}$ are constants independent of $(f,\e)$. This  assumption says that the part of the decision boundary of $P$ covered by $D_k$ is exactly part of the graph of some function $g_k\in\mc{F}$. Note that  functions in $\mc{F}$ must be  continuous because they can be approximated by DNNs in the uniform norm, and 
recall that the $d$-dimensional Lebesgue  measure of the  graph of a continuous function on $[0,1]^{d-1}$  must be zero. Therefore, \beq\label{20250926003155}
\text{under the above assumption, the Lebesgue measure of the decision boundary of $P$ must be $0$. }\eeq
Superficially, the assumption in this paper seems more restrictive because we require that $\eta_P$ has some smoothness on the whole input space $[0,1]^d$ while the assumption of \cite{meyer2023optimal} described above only requires the  smoothness of the boundary  of the superlevel set  $\set{x\in[0,1]^d}{\eta_P(x)\geq 1/2}$  of $\eta_P$ in $[0,1]^d$ without  any constrains on the smoothness of $\eta_P$ at points $x$  not on  this boundary.  In particular, one might think that the assumption of \cite{meyer2023optimal} described above covers our assumption because, intuitively,  if $\eta_P$  has  some smoothness,  then  the boundary of  $\set{x\in[0,1]^d}{\eta_P(x)\geq 1/2}$ in $[0,1]^d$ should inherit some smoothness as well. 
However, this is not correct. Our assumption also contains cases which are not covered by the assumption of \cite{meyer2023optimal} described above. Indeed, contrary to intuition, the boundary  of the Bayes decision set  $\set{x\in[0,1]^d}{\eta_P(x)\geq 1/2}$ in $[0,1]^d$ can be  very irregular  even if $\eta_P$ belongs to the compositional H\"older smooth function space  $\mc{G}_d^{\mathbf{CH}}(q, K, d_*, \beta,r)$. 
 To see this, let us consider the 
following example. 
Let $d\in\mb  N\cap[2,\infty)$, $q\in\mb N\cup\hkh{0}$, $K\in\mb N$,  $d_*\in\mb N$, $\beta\in(0,\infty)$,  $r\in[2,\infty)$ be arbitrary. 
Define
\[
\varphi:\mbR\to[0,\infty),\;x\mapsto\begin{cases}
0,&\text{ if }\abs{x}\geq 1,\\
\me^{\frac{1}{x^2-1}},&\text{ if }\abs{x}<1. 
\end{cases}
\] Then $\varphi$ is infinitely differentiable and supported in the $(-1,1)$. Let $\hkh{a_1,a_2,a_3,\ldots}$ be the set of all rational numbers in $(0,1/2)$. 
Take $m=\ceil{\beta+2}$ and $M={\sup\setm{3}{\abs{\mr{D}^j\varphi(x)}}{x\in\mbR,\;j\in\hkh{0,1,\ldots,\ceil{\beta+2}}}}$. Then, for
each $k\in\mb{N}$, take $\rho_k=\min\hkh{\frac{1}{6^k},  \frac{a_k}{2}, \frac{1/2-a_k}{2}}$ and  $b_k= \frac{1}{2^k}\cdot {\rho_k^m}\cdot \frac{1}{M}$. 
Then define 
\[
g:\mbR\to\mbR, \;x\mapsto \bigabs{\max\hkh{x-1/2,0}}^{m+1}-\sum_{k=1}^\infty b_k\cdot\varphi\ykh{\textstyle\frac{x-a_k}{\rho_k}}. 
\] Note that for any $j\in\hkh{0,1,\ldots,m}$,  the $j$-th derivative of the terms in the series $\sum\limits_{k=1}^\infty b_k\cdot\varphi\ykh{\textstyle\frac{x-a_k}{\rho_k}}$ converge uniformly, since 
\[
\sum_{k=1}^\infty\sup_{x\in\mbR}\abs{\frac{\mr{d}^j}{\mr{d}x^j}\ykh{b_k\cdot\varphi\ykh{\textstyle \frac{x-a_k}{\rho_k}}}}\leq 
\sum_{k=1}^\infty\frac{b_k}{\rho_k^j}\cdot\norm{\mr{D}^j\varphi}_{\mbR}
\leq \sum_{k=1}^\infty\frac{b_k}{\rho_k^m}\cdot M=\sum_{k=1}^\infty\frac{1}{2^k}\leq 1<\infty. 
\] Therefore, $\sum\limits_{k=1}^\infty b_k\cdot\varphi\ykh{\textstyle\frac{x-a_k}{\rho_k}}$ converges to a function that is $m$ times differentiable on $\mbR$, meaning that $g$ is H\"older-$\beta$ smooth  on $(-2,2)$. Define 
\[
\mr{u}:[0,1]^d\to[0,1],\;x\mapsto \frac{g((x)_1)}{3}\cdot \frac{1}{\norm{g}_{\mathbf{H}^{\beta}((-2,2))}}+\frac{1}{2}. 
\]Obviously, \beq\label{20250925131941}
&\norm{\mr{u}}_{\mathbf{H}^\beta([0,1]^d)}\leq \norm{\frac{1}{2}}_{\mathbf{H}^\beta([0,1]^d)}+\frac{\norm{g}_{\mathbf{H}^{\beta}([0,1])}}{3}\cdot \frac{1}{\norm{g}_{\mathbf{H}^{\beta}((-2,2))}}
\\&\leq \frac{1}{2}+\frac{1}{3}< 1\leq \norm{x\mapsto (x)_1}_{\mathbf{H}^\beta([0,1]^d)}=\norm{x\mapsto (x)_1}_{\mathbf{H}^\beta([0,1]^K)}\leq2\leq r. 
\eeq
Define 
\[
h_i:[0,1]^K\to[0,1]^{\iffun(i<q;K;1)},\;x\mapsto ((x)_1,0,0,\ldots,0)
\] for $i=1,2,\ldots,q$; and define  $h_0:[0,1]^d\to[0,1]^{\iffun(q>0;K;1)},\;x\mapsto(\mr{u}(x),0,0,\ldots,0)$. From \eqref{20250925131941} we see 
\beq h_q\circ h_{q-1}\circ\cdots\circ h_1\circ h_0\in \mc G_d^{\mathbf{CH}}(q, K, 1, \beta,2)\in \mc G_d^{\mathbf{CH}}(q, K, d_*, \beta,r). \eeq
Let $P$ be the probability measure on $[0,1]^d\times\hkh{-1,1}$ such that $P_X$ is the Lebesgue measure on $[0,1]^d$ and $\eta_P\xlongequal{P_X\text{-a.s.}}h_q\circ h_{q-1}\circ\cdots\circ h_1\circ h_0$. Then $P\in \mc{H}^{d,\beta,r,\Lambda}_{q,K,d_*}\subset \mc{H}^{d,\beta,r}_{q,K,d_*,d_\star}$ for any $\Lambda\in(1,\infty)$ and $d_\star\in\mb{N}$, meaning that this $P$ is within the scope of the assumption of this paper. However, this $P$ does not  satisfy the assumption of \cite{meyer2023optimal} described above. To show this, we compute the Lebesgue measure of the decision boundary of $P$.  Hereinafter, for  any measurable set $S\subset \mbR^d$, we use $\partial S$,  $\overline{S}$, and $\texttt{v}(S)$ to denote its boundary,  closure, and  Lebesgue measure in $\mbR^d$ respectively. Furthermore, for any measurable set $D\subset[0,1]^d$, we use ${\partial}_{\circ} D$ to denote its boundary in $[0,1]^d$, that is, \[{\partial}_\circ D:=\overline{D}\cap\overline{[0,1]^d\setminus D}. \]
Denote 
\[
&E=\set{x\in[0,1]^d}{h_q\circ h_{q-1}\circ\cdots\circ h_1\circ h_0(x)\geq\frac{1}{2}}, 
\\& F=\set{x\in[0,1]^d}{\eta_P(x)\geq\frac{1}{2}}, 
\\&A=\set{x\in[0,1]}{g(x)\geq 0\vphantom{\frac12}}, 
\\&U=\mbR^d\setminus E, \\
& V=\mbR^d\setminus F, \\
&N=(E\setminus F)\cup (F\setminus E). 
\] Then  it follows from ${ \eta_P\xlongequal{P_X\text{-a.s.}}h_q\circ h_{q-1}\circ\cdots\circ h_1\circ h_0}$ and $U\setminus N\subset V$ that \beq\label{20250925195512}
\texttt{v}(N)=P_X(N)=0\eeq  and 
\[
\overline{U}&=\set{x\in\mbR^d}{W\cap U\neq \varnothing\text{ for any open set $W$ that contains $x$}\vphantom{\frac12}}
\\&\xlongequal[\text{set in $\mbR^d$ must be positive, and $U$ is open}]{\text{$\because$ the Lebesgue measure of any nonempty open}}\set{x\in\mbR^d}{\texttt{v}(W\cap U)>0 \text{ for any open set $W$ that contains $x$}\vphantom{\frac12}}
\\&\xlongequal{\because \eqref{20250925195512}}\set{x\in\mbR^d}{\texttt{v}(W\cap U\setminus N)>0 \text{ for any open set $W$ that contains $x$}\vphantom{\frac12}}
\\&\subset\set{x\in\mbR^d}{\texttt{v}(W\cap V)>0 \text{ for any open set $W$ that contains $x$}\vphantom{\frac12}}\\
&
\subset\set{x\in\mbR^d}{W\cap V\neq\varnothing\text{ for any open set $W$ that contains $x$}\vphantom{\frac12}}=\overline{V}. 
\] Thus 
\[
&\partial E=\overline{E}\cap\overline{U}\xlongequal{\because \text{$E$ is closed}}E\cap\overline{U}\subset E\cap\overline{V}\subset(E\setminus F)\cup (F\cap \overline{V})\subset N\cup (\overline{F}\cap \overline{V})\\&=N\cup \partial F\subset N\cup \partial_{\circ} F\cup \partial [0,1]^d. 
\] Consequently, 
\[
&\texttt{v}\ykh{\partial_{\circ}\setl{x\in[0,1]^d}{\eta_P(x)\geq 1/2}}=\texttt{v}(\partial_\circ F)\xlongequal{\because\eqref{20250925195512}}\texttt{v}(N\cup\partial_\circ F)=
\texttt{v}(N\cup\partial_\circ F\cup\partial[0,1]^d)
\\&\geq\texttt{v}(\partial E)=\texttt{v}\ykh{\partial\setl{x\in[0,1]^d}{h_q\circ h_{q-1}\circ\cdots\circ h_1\circ h_0(x)\geq{1}/{2}}}
\\&=\texttt{v}\ykh{\partial\setl{x\in[0,1]^d}{\mr{u}(x)\geq{1}/{2}}}
=
\texttt{v}\ykh{\partial\set{x\in[0,1]^d}{\textstyle\frac{g((x)_1)}{3}\cdot \frac{1}{\norm{g}_{\mathbf{H}^{\beta}((-2,2))}}+\frac{1}{2}\geq{1}/{2}}}
\\&=\texttt{v}\ykh{\partial\setl{x\in[0,1]^d}{{g((x)_1)}{}\geq 0}}=\texttt{v}\ykh{\partial\ykh{A\times [0,1]^{d-1}}}\\&=\texttt{v}\ykh{\partial\ykh{\ykh{[0,1]\setminus\ykh{\textstyle\bigcup_{k=1}^\infty (a_k-\rho_k,a_k+\rho_k)}}\times[0,1]^{d-1}}}\\&\geq\texttt{v}\ykh{\overline{{\ykh{\textstyle\bigcup_{k=1}^\infty (a_k-\rho_k,a_k+\rho_k)}}\times[0,1]^{d-1}}\cap\ykh{\ykh{[0,1]\setminus\ykh{\textstyle\bigcup_{k=1}^\infty (a_k-\rho_k,a_k+\rho_k)}}\times[0,1]^{d-1}}}
\\&\xlongequal[\text{dense in $[0,1/2]$}]{\because\;\hkh{a_1,a_2,a_3,\ldots} \text{ is}}\texttt{v}\ykh{\ykh{{{[0,1/2]}}\times[0,1]^{d-1}}\cap\ykh{\ykh{[0,1]\setminus\ykh{\textstyle\bigcup_{k=1}^\infty (a_k-\rho_k,a_k+\rho_k)}}\times[0,1]^{d-1}}}
\\&=
\texttt{v}\ykh{\ykh{\ykh{[0,1/2]\setminus\ykh{\textstyle\bigcup_{k=1}^\infty (a_k-\rho_k,a_k+\rho_k)}}\times[0,1]^{d-1}}}
\\&\geq \texttt{v}\ykh{[0,1/2]\times[0,1]^{d-1}}-\sum_{k=1}^\infty\texttt{v}\ykh{(a_k-\rho_k,a_k+\rho_k)\times[0,1]^{d-1}}
\\&=\frac{1}{2}-\sum_{k=1}^\infty 2\rho_k\geq\frac{1}{2}-\sum_{k=1}^\infty \frac{2}{6^k}=\frac{1}{2}-\frac{2}{5}>0. 
\]
Therefore, the Lebesgue measure of the  decision boundary of $P$ is greater than zero. This means that $P$ does not satisfy the assumption of \cite{meyer2023optimal} described above because, as  mentioned in \eqref{20250926003155},  the Lebesgue measure of the decision boundary of the distributions that satisfy the assumption of $\text{\cite{meyer2023optimal}}$ described above must be zero. From this example we see the decision boundary of a distribution $P$ that has a compositional H\"older smooth CCP function $\eta_P$ (that is, satisfies the regularity assumption in this paper) can be highly non-smooth, possibly even having  positive Lebesgue measure,  and thus cannot satisfy the regularity assumption in \cite{meyer2023optimal} described above. As a result, the regularity assumption in this paper is not more restrictive than that in \cite{meyer2023optimal} described above.

The second difference is the noise condition. In this paper, we consider the noise condition \eqref{23080401} to characterize the behavior of the data points $x$ near the decision boundary. In \cite{meyer2023optimal}, the 
author considers the assumption where the following two conditions are required to hold simultaneously:

	\begin{enumerate}[label={{ {\hfil(\roman*)\hfil}} }]
\item  
for any set $G$ of the form 
$\set{x\in[0,1]^d}{\tilde{f}(x)=1}$,  where $\tilde{f}$ is some ReLU neural network whose parameters take values in $[-1,1]\cap \set{\frac{k}{2^{l}}}{ l\in\mb{N}, k\in\mb{Z}}$, there holds \beq\label{250703032425}
D_1(G,G^*_P)\geq c\cdot D_2(G,G^*_P)^{1+1/s}
\eeq for 
 some constant $c\in(0,\infty)$, where 
\[
&D_1(G_1, G_2):=\int_{(G_1\setminus G_2)\cup(G_2\setminus G_1)}\Bigabs{2 \eta_P(x)-1}\mr{d}P_X(x),\\
&D_2(G_1, G_2):=P_X\ykh{(G_1\setminus G_2)\cup(G_2\setminus G_1)}, 
\]  and 
\[
G_P^*:=\set{x\in[0,1]^d}{\eta_P(x)\geq 1/2}
\] is the Bayes decision set; 
\item for any $x$ in the decision boundary $\partial G^*_P$ with $x\in D_k$, there exists a H\"older-$\frac1s$ function $g_{k, x}$ with
\[
\mr{D}^\alpha g_{k,x}(0)=0,\;\forall\;\alpha\in \mb{N}_0^{d,1/s}
\] and $\e_1\in(0,\infty)$ such that 
\beq\label{20250926173458}
&\abs{2\eta_P\ykh{((x)_1,(x)_2,\ldots, (x)_{i_k-1}, t, (x)_{i_k+1}, (x)_{i_k+2},\ldots ,(x)_d)}-1}
\\&\leq g_{k,x}\ykh{\abs{t-(x)_{i_k}}},\;\forall\;t\in[\max\hkh{0,(x)_{i_k}-\e_1},\min\hkh{1,(x)_{i_k}+\e_1}]
\eeq
where recall that $D_k$ satisfies \eqref{20250926163808} and $G^*_P=\bigcup_{i=1}^uH_i$. 
 \end{enumerate}

It follows from Proposition 1 of  \cite{tsybakov2004optimal} that the condition \eqref{23080401} with $s>0$ is stronger than \eqref{250703032425}. Indeed, \eqref{23080401} with $s>0$  implies that \eqref{250703032425} holds for some $c>0$ and   any measurable set $G\subset[0,1]^d$. However, the condition  \eqref{20250926173458} is a fairly strong condition and it cannot be derived from the assumptions in our setting. Furthermore, the above conditions  \eqref{250703032425} and  \eqref{20250926173458}   considered in \cite{meyer2023optimal} exclude the case  $s=0$, that is, the case without the noise condition,  which is incorporated in our setting. 

The third difference is the quantity for which the convergence rates  are established. 
 In this paper, we establish convergence rates for the expected excess 0-1 risk 
\[
\bm{E}\zkh{\mc{E}_P(\hat{f}_n)}
\] to measure the performance of the estimator  $\hat{f}_n$. In \cite{meyer2023optimal}, the author considers  quantities
\beq\label{20250926220045}
\bm{E}\zkh{\ykh{\mc{E}_P(\hat{f}_n)}^p}
\eeq or 
\beq\label{20250926224818}
\bm{E}\zkh{{P_X\ykh{\setl{x\in[0,1]^d}{\mr{sgn}(f(x))\neq \mr{sgn}(\hat{f}_n(x))}}}^p}, 
\eeq where $p\geq 1$. Therefore, compared with this paper, a broader class of quantities characterizing the performance of  estimators is investigated in \cite{meyer2023optimal}. It is interesting to study 
whether we can use the results developed in this paper to establish convergence rates for the quantities in \eqref{20250926220045} and \eqref{20250926224818}  
in the future. 

The fourth  difference is the approach to obtain the neural network estimators.  In this paper, the neural network estimators for which we establish convergence rates are obtained from empirical hinge risk minimization,  while in \cite{meyer2023optimal}  the neural network estimators  are obtained from empirical 0-1 risk minimization.

From the above discussion we see that the setting of this paper and that of \cite{meyer2023optimal} are entirely different, and neither covers the other.

\subsection{On the  Inefficiency of the Tsybakov Noise Condition in Improving Convergence of  Logistic Risk }\label{20250919034414}\hypertarget{20250919034415}{}    \label{20250919034415}

In this part,  we will explain why  in general the Tsybakov noise condition $P\in\mc{T}^{d,s}_{\alpha,\tau}$ cannot help improve the optimal convergence rates of the expected excess logistic risk of estimators. 
  Indeed, for any $s\in[0,\infty]$, $\tau\in(0,\infty]$, and $\alpha\in(0,\infty)$ with 
\beq\label{241202002615}
\text{$\tau<1$ or $\idf_{[1,\infty]}(\tau)\cdot\idf_{[0,\infty)}(s)\cdot\idf_{(1,\infty)}(\alpha)=1$ or $\idf_{[1,\infty]}(\tau)\cdot \idf_{\hkh{0}}(s)\cdot\idf_{\hkh{1}}(\alpha)=1$,}\eeq  
it is easy to show that there exists a constant  $A\in[0,1)$, sufficiently close to $1$, such that 
\beq\label{241129203630}
\overline{\mc{H}}^{d,\beta,r}_{A,q,K,d_*}:=\set{P_g}{\begin{minipage}{193.2pt} $g\in \mc G_d^{\mathbf{CH}}(q, K, d_*, \beta,r)$,  $\ran(g)\subset[0,1]$, and $\int_{[0,1]^d}\idf_{[0,A]}(\abs{2g(x)-1})\mr{d}x=0$\end{minipage}} 
\subset \mc{H}^{d,\beta,r}_{q,K,d_*,d_\star}\cap \mc{T}^{d,s}_{\alpha,\tau},
\eeq
where  the space $\mc G_d^{\mathbf{CH}}(q, K, d_*, \beta,r)$ will be  defined in \eqref{241129192804} below, and $P_g$ is the Borel probability measure on $[0,1]^d\times\hkh{-1,1}$ such that $\eta_{P_g}=g$ and the marginal distribution of $P_g$ on $[0,1]^d$ is the Lebesgue measure, that is, 
\[
&P_{g}:\hkh{\text{all Borel subsets of $[0,1]^d\times\hkh{-1,1}$}}\to [0,1],\\
	&\;\;\;\;\;\;S\mapsto\int_{[0,1]^d}\Bigabs{g(x)\cdot\idf_{S}(x,1)+(1-g(x))\cdot\idf_{S}(x,-1)}\mr{d}x. 
\] 
To show that \eqref{241129203630} holds for $A\in[0,1)$ sufficiently close to $1$,  we first note the obvious fact  that 
\beq\label{241201234339}
\mc G_d^{\mathbf{CH}}(q, K, d_*, \beta,r)\subset \mc G_d^{\mathbf{CHOM}}(q, K,d_\star, d_*, \beta,r), 
\eeq 
which follows directly from the definition. Besides, it is easy to verify that   
\beq \label{241201234342}
 &\set{P_g}{\textstyle\int_{[0,1]^d}\idf_{[0,A]}(\abs{2g(x)-1})\mr{d}x=0}
 \\&
 \subset  \mc{T}^{d,s}_{\alpha,\tau},\;\forall\;
 A\in\begin{cases}
 (\tau,1),&\text{ if }\tau<1,\\
 \ykh{(1/\alpha)^{1/s},1},&\text{ if }\idf_{[1,\infty]}(\tau)\cdot\idf_{[0,\infty)}(s)\cdot\idf_{(1,\infty)}(\alpha)=1,\\
 (0,1),&\text{ if }\idf_{[1,\infty]}(\tau)\cdot \idf_{\hkh{0}}(s)\cdot\idf_{\hkh{1}}(\alpha)=1.
 \end{cases}
 \eeq 
 Combining \eqref{241201234339} and \eqref{241201234342}, we deduce that 
 \[
 \overline{\mc{H}}^{d,\beta,r}_{A,q,K,d_*}&:=\set{P_g}{\begin{minipage}{193.2pt} $g\in \mc G_d^{\mathbf{CH}}(q, K, d_*, \beta,r)$,  $\ran(g)\subset[0,1]$, and $\int_{[0,1]^d}\idf_{[0,A]}(\abs{2g(x)-1})\mr{d}x=0$\end{minipage}} 
 \\&\subset 
 \set{P_g}{\begin{minipage}{221.76pt} $g\in  \mc G_d^{\mathbf{CHOM}}(q, K,d_\star, d_*, \beta,r)$,  $\ran(g)\subset[0,1]$, and $\int_{[0,1]^d}\idf_{[0,A]}(\abs{2g(x)-1})\mr{d}x=0$\end{minipage}} 
 \\&\subset
 \set{P_g}{\begin{minipage}{229.92pt} $g\in  \mc G_d^{\mathbf{CHOM}}(q, K,d_\star, d_*, \beta,r)$ and  $\ran(g)\subset[0,1]$\end{minipage}}\cap \mc{T}^{d,s}_{\alpha,\tau}
 \subset \mc{H}^{d,\beta,r}_{q,K,d_*,d_\star}\cap \mc{T}^{d,s}_{\alpha,\tau} 
 \] 
 for any 
 $$A\in\begin{cases}
  (\tau,1),&\text{ if }\tau<1,\\
  \ykh{(1/\alpha)^{1/s},1},&\text{ if }\idf_{[1,\infty]}(\tau)\cdot\idf_{[0,\infty)}(s)\cdot\idf_{(1,\infty)}(\alpha)=1,\\
   (0,1),&\text{ if }\idf_{[1,\infty]}(\tau)\cdot \idf_{\hkh{0}}(s)\cdot\idf_{\hkh{1}}(\alpha)=1.
  \end{cases}$$ 
 This proves that  \eqref{241129203630} holds for $A\in[0,1)$ sufficiently close to $1$. On the other hand, Theorem 2.6 of \cite{zhangzihan2023classification} yields that 
  \beq\label{241202002200}
  \inf_{\hat{f}_n} \sup\setm{3}{{\bm E}_{P^{\otimes n}}\zkh{\mc{E}_P^{\phi_{\mathbf{l}}}(\hat{f}_n)}}{P\in \overline{\mc{H}}^{d,\beta,r}_{A,q,K,d_*}}\gtrsim n^{-\frac{\beta\cdot(1\qx\beta)^q}{d_*+\beta\cdot(1\qx\beta)^q}}. 
  \eeq 
Combining  \eqref{241121010851}, \eqref{241129203630}, and \eqref{241202002200}, we deduce that 
  \[
 & n^{-\frac{\beta\cdot(1\qx\beta)^q}{d_*+\beta\cdot(1\qx\beta)^q}} 
 \lesssim 
  \inf_{\hat{f}_n} \sup\setm{3}{{\bm E}_{P^{\otimes n}}\zkh{\mc{E}_P^{\phi_{\mathbf{l}}}(\hat{f}_n)}}{P\in \overline{\mc{H}}^{d,\beta,r}_{A,q,K,d_*}}
  \\&\leq 
  \inf_{\hat{f}_n} \sup\setm{3}{{\bm E}_{P^{\otimes n}}\zkh{\mc{E}_P^{\phi_{\mathbf{l}}}(\hat{f}_n)}}{P\in \mc{H}^{d,\beta,r}_{q,K,d_*,d_\star}\cap \mc{T}^{d,s}_{\alpha,\tau}}
  \\&\leq 
    \inf_{\hat{f}_n} \sup\setm{3}{{\bm E}_{P^{\otimes n}}\zkh{\mc{E}_P^{\phi_{\mathbf{l}}}(\hat{f}_n)}}{P\in \mc{H}^{d,\beta,r}_{q,K,d_*,d_\star}}
  \\&  \leq \sup\setm{3}{{\bm E}_{P^{\otimes n}}\zkh{\mc{E}_P^{\phi_{\mathbf{l}}}(\tilde{f}_n^{\FNN})}}{P\in\mc{H}^{d,\beta,r}_{q,K,d_*,d_\star}}
    \lesssim \ykh{\frac{(\log n)^5}{n}}^{{\frac{\beta\cdot(1\qx\beta)^q}{{d_*+\beta\cdot(1\qx\beta)^q}}}}, 
  \] 
provided that \eqref{241202002615} holds. Consequently, in the case where \eqref{241202002615} holds, the noise condition \eqref{23080401} can hardly help improve the convergence rate of the \zaverage excess logistic risk of estimators under the compositional assumption $P\in\mc{H}^{d,\beta,r}_{q,K,d_*,d_\star}$. Any improvement due to this condition is, at most, a negligible logarithmic factor $(\log n)^{{\frac{5\cdot\beta\cdot(1\qx\beta)^q}{{d_*+\beta\cdot(1\qx\beta)^q}}}}$. However, if \eqref{241202002615} does not hold, the situation changes. Note that the negation of \eqref{241202002615} is equivalent to one of the foloowing conditions:
  \beq\label{241202024026}
  \tau\geq 1\text{ and }s=\infty, 
  \eeq or 
  \beq\label{241202024027}
  \tau\geq 1\text{ and }0\leq s<\infty\text{ and }\alpha<1,
  \eeq or
  \beq\label{241202024028}
  \tau\geq 1\text{ and }0<s<\infty\text{ and }\alpha=1.
  \eeq
According to \eqref{23081002}, the condition \eqref{241202024027} is impossible. Furthermore, note that \eqref{241202024026} is equivalent to  
\beq\label{241204204619}
P_X\ykh{\set{x \in [0,1]^d}{\eta_P(x) \in \hkh{0,1}}} = 1,  
\eeq  
which is exactly the so-called noise-free case. In the noise-free scenario described by \eqref{241204204619}, it follows from our forthcoming work \cite{zhang2025classification} that both the \zaverage excess 0-1 risk and the \zaverage excess logistic risk of ReLU DNN estimators can achieve the optimal convergence rate of order ${\frac{\log^\theta n}{n}}$, up to a logarithmic factor $\log^\theta n$. This rate is significantly better than the rate ${n^{-\frac{\beta\cdot(1\qx\beta)^q}{d_*+\beta\cdot(1\qx\beta)^q}}}$, which holds under condition \eqref{241202002615}.  However, the case described by \eqref{241202024028} remains quite subtle. To the best of our knowledge, the optimal convergence rate for the \zaverage excess logistic risk of estimators under the assumption  $P\in \mc{H}^{d,\beta,r}_{q,K,d_*,d_\star}\cap \mc{T}^{d,s}_{\alpha,\tau}$ with \eqref{241202024028} remains unknown.
  
We can provide an intuitive explanation for why the noise condition \eqref{23080401} can help improve the convergence rate of the excess 0-1 risk of estimators, but generally does not enhance the convergence rate of the excess logistic risk. Recall that the noise condition \eqref{23080401} characterizes the amount of input data $x$ for which $\eta_P(x)$ is close to $1/2$: the stronger the condition \eqref{23080401}, the less frequently $\eta_P(x)$ is near $1/2$. Note that the target function (i.e., the minimizer) of the excess 0-1 risk $\mc{E}_P(\cdot)$ is $\mr{sgn}(2\eta_P(\cdot)-1)$. The function $\eta \mapsto \mr{sgn}(2\eta - 1)$ is discontinuous at $1/2$, but it is constant (and thus behaves well) on $[0, 1/2)$ and $(1/2, 1]$. Therefore, when fewer input data points $x$ yield $\eta_P(x)$ near $1/2$, the target function $\mr{sgn}(2\eta_P(\cdot)-1)$ becomes more regular, making the associated learning task easier. Consequently, the noise condition \eqref{23080401} can lead to improved convergence rates for the excess 0-1 risk of estimators. In contrast, the target function (i.e., the minimizer) of the excess logistic risk $\mc{E}_P^{\phi_{\mathbf{l}}}(\cdot)$ is $\log\frac{\eta_P(\cdot)}{1-\eta_P(\cdot)}$, which is smooth with respect to $\eta_P$ except at the endpoints $\eta_P=0$ and $1$. However, the noise condition \eqref{23080401} does not impose any restriction on the behavior of the input data $x$ for which $\eta_P(x)$ is close to $0$ or $1$—with the exception of condition \eqref{241202024026}, which implies that almost all input data points $x$ satisfy $\eta_P(x) \in \{0,1\}$, and condition \eqref{241202024028}, which bounds the probability that the CCP function $\eta_P$ takes values near $0$ or $1$ by $\delta$, i.e., $P_X\ykh{\set{x\in[0,1]^d}{\eta_P(x)\in[0,\delta)\cup(1-\delta,1]}}$, converges to zero at a rate no slower than 
\[
  & P_X\ykh{\set{x\in[0,1]^d}{\eta_P(x)\in[0,\delta)\cup(1-\delta,1]}}=1-P_X\ykh{\set{x\in[0,1]^d}{\abs{2\eta_P(x)-1}\leq 1-2\delta}}
  \\&\geq 1-\alpha\cdot (1-2\delta)^s=1-(1-2\delta)^s\gtrsim 2\cdot s\cdot \delta
\]
as $\delta\to 0+$. As a result, in general, the noise condition \eqref{23080401} does not contribute to improving the regularity of the target function $\log\frac{\eta_P(\cdot)}{1-\eta_P(\cdot)}$ associated with the excess logistic risk $\mc{E}_P^{\phi_{\mathbf{l}}}(\cdot)$. Therefore, it does not lead to improved convergence rates for the excess logistic risk of estimators.

\bibliographystyle{plain}

\bibliography{ref}
\end{document}